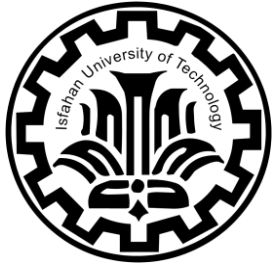

# Isfahan University of Technology

Department of Mechanical Engineering

# Stability Control of Walking Biped Robots based on Total Momentum

A Thesis

Submitted in partial fulfillment of the requirements

For the degree of Master of Science

**By**

**Mostafa Ghobadi Shahreza**

Evaluated and Approved by the Thesis Committee, on April 18, 2012

1. M. Keshmiri, Associate Prof. (Supervisor)

2. M.J. Sadigh, Associate Prof. (Advisor)

3. M. Danesh, Assistant Prof. (Examiner)

4. S. Hadian Jazi, Assistant Prof. (Examiner)

M.R. Salimpour, Assistant Prof. (Department Graduate Coordinator)

# Stability Control of Walking Biped Robots based on Total Momentum

## Mostafa Ghobadi Shahreza

m.ghobadishahreza@me.iut.ac.ir

04/18/2012

### Department of Mechanical Engineering

### Isfahan University of Technology, Isfahan 84156-83111, Iran

**Degree:** M.Sc.                                                        **Language:** Farsi

**Dr. Mehdi Keshmiri,** mehdik@cc.iut.ac.ir

## Abstract

With the development of robotics science, lots of research has been conducted in the area of biped robots and many industries have been attracted to the application of these robots. Because of the non-continuous nature of legged robots and bipeds in particular, not facing problems that wheeled or snake-like robots do in moving on stairs and ditches, their industrial and practical applications has been considered seriously. That is why a great number of humanoid robots have been developed all over the world and in some countries such as Japan, development of Humanoid Robots is considered as a national project to the extent that it is considered as an index of the science and technology development in the country. As the humanoid robots have the same structure as humans, many efforts are put to reach their level of ability to handle many tasks which have been normally assigned to humankind. Among the subjects which have been studied on biped motion, naming motion stability analysis, motion pattern generation, optimization, and control, the stability and control of these systems can be considered as one of the most important issues. The concept of stability of the biped motions, its criteria and how such stability can be guaranteed is of the greatest importance for researchers and at the same time is the most complicated dilemma not completely resolved yet. Most of the researches on stability and control of these systems have been based on the Postural Stability which regards the control of walkers' stability as the motion tracking of their joints and support limbs on some desired trajectories having postural stability. However, recently some scholars developed the concept of the stability of walkers correctly and equate it with not going to the states which leads to fall down or tip over. The current research is based on this new intuition of stability and unlike the traditional analyses which follows stability and control of the walker in *joint space* as internal or detailed quantities, a walker is considered as a total system so that the problem is followed in *whole-body momentum space*. Such an approach gives an exhaustive solution which enables its implementation on any biped walkers considering the existence of a map between these two spaces. Firstly, Principle Equation of Motion (for walkers) is derived that later results in introducing two piecewise-continuous dynamical systems namely *Simplified Walking Model (SWM)* and *Complete Walking Model (CWM)* which both describe the behavior of walker with emphasis on the motion in horizontal plane. Furthermore, by making some realistic assumptions based on human natural walking, a simplified equation of motion named Step-to-Step Equation of Walking is formulated. To calculate the solution of steady walking motions, the repetition condition is exerted on this equation that yields a significant finding named *Simple and Compound Motion Cycles* as general solutions of steady walking. Among all possible motion cycles, *Simple Forward Motion Cycle* is representing the normal walking pattern. These cycles have marginal stability that in practice cause the motion to diverge exponentially even under slight disturbance. By defining the stabilization of walking as the guidance of the motion initiated from arbitrary initial states to a desired motion cycle and controlling the motion about it, two major strategies are presented for stability control of the walkers: 1) Continuous altering of Center of Pressure (CoP) within support polygon, and 2) Continual planning of the step length and duration. Using these two strategies and based on Simplified Walking Model (SWM), four methods of stability control named generally as *Motion Cycle Stabilizers* are proposed and their theoretical aspects are inspected. Then, to examine the capability and detect deficiencies of the proposed stabilizers on Complete Model of Walking (CWM), some simulations are performed on a physical model with realistic constraints. Finally, to overcome the deficiencies of the Stabilizers, method of *Optimal Stability Control* is proposed to complete the solution. The numerical simulations show that the proposed approach to the stability and control of biped walkers provides us with a more complete and accurate solution compared with traditional approaches and guarantee the stability of walkers in a maximum sense.

## Keywords

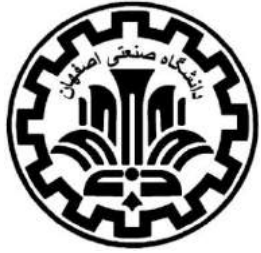

**دانشگاه صنعتی اصفهان**

دانشکده مهندسی مکانیک

# کنترل پایداری رباتهای راهرونده دوپا بر اساس اندازه حرکت کلی

پایان نامه کارشناسی ارشد مهندسی مکانیک

گرایش طراحی کاربردی

**مصطفی قبادی شهرضا**

استاد راهنما

**دکتر مهدی کشمیری**

**فروردین‌ماه ۱۳۹۱**

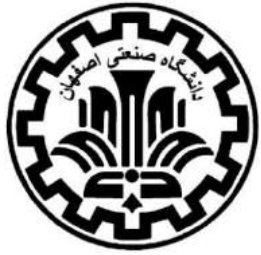

**دانشگاه صنعتی اصفهان**
دانشکده مهندسی مکانیک

پایان نامه کارشناسی ارشد رشته طراحی کاربردی آقای مصطفی قبادی شهرضا
تحت عنوان

# کنترل پایداری رباتهای راهرونده دوپا بر اساس اندازه حرکت کلی

در تاریخ ۱۳۹۱/۱/۳۰ توسط کمیته تخصصی زیر مورد بررسی و تصویب نهایی قرار گرفت.

| | | |
|---|---|---|
| ۱. | استاد راهنمای پایان نامه | دکتر مهدی کشمیری |
| ۲. | استاد مشاور پایان نامه | دکتر محمدجعفر صدیق |
| ۳. | استاد داور | دکتر محمد دانش |
| ۴. | استاد داور | دکتر شهرام هادیان جزی |
| | سرپرست تحصیلات تکمیلی دانشکده | دکتر محمدرضا سلیم‌پور |

خداوند یگانه را سپاسگذارم بابت عنایت و لطف بیکرانش در همه لحظات زندگی و خاصه به جهت توفیق در گذراندن دوره پر ثمر تحصیلات تکمیلی. جا دارد از زحمات و فداکاری های بی دریغ پدر و مادر گرانقدرم که حامی و مشوق بنده بوده اند، کمال تشکر و سپاسگذاری را داشته باشم.

همچنین سپاس فراوان دارم از زحمات و راهنمایی های استاد گرانقدر جناب آقای دکتر کشمیری که فراتر از آموزگاری فرزانه، همواره راهنمایی دلسوز، اسوه ای اخلاقی و الگویی انسانی برای بنده بوده اند. و نیز، از دوستان همراه و گوهران گرانتر از جان بی نهایت سپاسگذارم.

کلیه حقوق مادی مترتب بر نتایج مطالعات، ابتکارات و نوآوری‌های ناشی از تحقیق موضوع این پایان‌نامه (رساله) متعلق به دانشگاه صنعتی اصفهان است.

تقدیم به پدر و مادر عزیزم

که همواره مرا به علم آموزی و خرد اندوزی رهنما بوده‌اند

# فهرست مطالب

**فصل دوم: مدل دینامیکی حرکت راه‌رفتن**

**فصل سوم: کنترل پایداری**

## فصل چهارم: کنترل‌کننده پایداری بهینه

## فصل پنجم: جمع بندی و پیشنهادات

## چکیده

با گسترش و پیشرفت علم رباتیک، مطالعه و تحقیق بر روی رباتهای دارای پا و به خصوص ربات دوپا افزایش یافته است و توجه بسیاری از مراکز تحقیقاتی و صنعتی را به خود جلب کرده است. در سالهای اخیر، رباتهای انسان نمای متعددی در سراسر دنیا ساخته شده است و در برخی کشورها مانند ژاپن ساخت ربات انساننما به عنوان پروژه ملی مطرح است به نحوی که نمادی از پیشرفت تکنولوژی محسوب میشود. رباتهای دوپا به این دلیل که از نظر ساختاری شبیه انسان هستند، توانایی انجام بسیاری از فعالیتهای انسان را دارند. همچنین به علت ماهیت ناپیوسته حرکت ربات دوپا، بسیاری از مشکلاتی که رباتهای چرخدار و مارسان در عبور از سطوح ناهموار، پلهها، موانع و شکافها دارند برای این دسته از رباتها وجود ندارد. از این میان تحقیقات زیادی که بر روی رباتهای دوپا انجام شده میتوان به بررسی پایداری، طراحی مسیر، بهینه سازی راه رفتن و کنترل آنها اشاره نمود. از این میان، مهمترین و پیچیدهترین مسئله ای که هنوز به طور کامل حل نشده است، مفهوم پایداری راهروندهها و معیارها و روشهای کنترل آن میباشد. اگرچه تاکنون پایداری وضعی، که حفظ پایداری راهرونده را برابر با کنترل وضعیت قرارگیری تکیهگاهها و مفاصل بر روی یک مسیر از پیشطراحی-شده دارای پایداری وضعی میگیرد، مبنایی برای کنترل پایداری آنها به حساب میآمده است، برخی از محققین مفهوم پایداری این سیستمها را بسط دادهاند و آن را معادل با نرفتن به وضعیتی میدانند که منجر به زمین خوردن(افتادن) میشود که برداشت اخیر از پایداری و کنترل آن، مورد نظر تحقیق حاضر است. همچنین، برخلاف روشهای متداول که حرکت راهرونده را در فضای مفاصل(کمیتهای جزئی داخلی) پیگیری میکنند، تحلیلهای پایداری و کنترل پایداری ارائه شده در این پژوهش، به راهرونده به صورت یک سیستم کلی نگاه میکند و روشی جامع در فضای اندازه حرکت(کمیتهای کلی) ارائه میدهد که با توجه به وجود یک نگاشت بین این دوفضا، پیادهسازی آن را بر روی هر راهروندهای، ممکن میسازد. بر این اساس، ابتدا معادله حرکت پایه برای راهرونده در فضای اندازهحرکت کلی بدن استخراج شده است که در نهایت به ارائه دو مدل ریاضی(سیستم دینامیکی تکهتکه پیوسته) به نامهای مدل سادهشده راهرفتن(SWM) و مدل کامل راهرفتن(CWM) میانجامد. همچنین، با انجام فرضیاتی نزدیک به واقعیت بر روی معادله حرکت پایه، معادله حرکتی سادهشده، به نام معادله حرکت گام بهگام بهدست میآید. برای محاسبه حل حرکت دائمی سیستم، با اعمال شرط تکرار بر روی این معادله، سیکلهای حرکتی ساده و مرکب استخراج خواهند شد که ازاین میان سیکل ساده پیشرونده، الگویی برای حرکت راهرفتن معمولی ارائه میدهد. این سیکلها دارای پایداری مرزی هستند و در عمل به واسطه کوچکترین اختلالی با رشدی نمایی واگرا خواهند شد. با تعریف پایدارسازی حرکت راهرفتن به صورت هدایت حرکت از یک شرایط اولیه دلخواه به سمت یک سیکل حرکتی مطلوب و کنترل حرکت حول آن، دو راهبرد کنترل پایداری، ۱) تغییر پیوسته مرکز فشار در ناحیه تکیهگاهی و ۲) تغییر پیدرپی طول و زمان فرودگام، ارائه شده است. بر اساس این دو راهبرد و بر مبنا قراردادن مدل سادهشده راهرفتن(SWM)، چهار کنترل-کننده پایداری با نام کلی کنترل سیکل حرکتی ارائه شده و توانایی پایدارسازی هر یک از آنها تجزیه و تحلیل شده است. سپس برای بررسی توانایی پایدارسازها بر روی مدل کامل راهرونده(CWM)، تعدادی شبیهسازی بر روی یک مدل فیزیکی دارای محدودیت-های واقعی انجام شده و توانایی و نارسایی هر یک از کنترلکنندههای پایداری بررسی شده است. در پایان، برای فائق آمدن بر نارسایی-های پایدارسازهای سیکل حرکتی، روشی به نام کنترل پایداری بهینه به عنوان راهحلی کامل برای مسئله ارائه گردیده و عملکرد آن با انجام شبیهسازی بر روی مدل فیزیکی، بررسی شده است. در این پژوهش سعی شده است با نگاهی جامعتر به دو مقوله پایداری و کنترل پایداری در مقایسه با نگاه متداول، مسئله پایداری و کنترل پایداری راهرونده با رویکردی حداکثری دنبال گردد.

## کلمات کلیدی

پایداری راهروندهها، رباتهای راه رونده دوپا، مدل راه رفتن در فضای اندازه حرکت، سیکلهای حرکتی راه رفتن، پایدارسازی راه رفتن، کنترل گام بهکام راه رفتن

**فصل اول**

مقدمه

در این فصل ابتدا به تاریخچه و مفاهیم اولیه ربات‌های راه‌رونده و به خصوص ربات‌های دوپا می‌پردازیم، سپس با بررسی مفاهیمی همچون پایداری و کنترل پایداری، به طرح و بسط ایده پایداری و توصیف صورت مسئله اصلی پایان‌نامه خواهیم پرداخت تا در فصل‌های آتی با استخراج مدل دینامیکی راه‌رونده، به تعریف پایداری حرکت راه-رفتن و کنترل آن بپردازیم.

## ۱-۱  تاریخچه ربات‌های دارای پا

ربات‌های دارای پا[1] و به خصوص نوع ربات دوپای[2] آن که می‌تواند حرکت انسان را شبیه‌سازی کند، یکی از انواع ربات‌های جابجایی‌پذیر[3] محسوب می‌شوند. این ربات‌ها در مقایسه با دیگر انواع زمینی آن‌ها از جمله ربات-های چرخ‌دار[4] یا ربات‌های خزنده مار و کرم[5] دارای مزایای حرکتی زیادی ازجمله قابلیت حرکت در محیط‌های غیر مسطح و دارای موانع حرکتی در مقایسه با دسته اول و در عین حال دارای سرعت پیشروی بالاتری نسبت به دسته دوم هستند.

---

[1] Legged Robots
[2] Biped Robots
[3] Mobile Robots
[4] Wheeled/Belt Robots
[5] Crawling Robots e.g.SsnakeWworm

پیشینه ساخت ربات‌های دوپا شاید در سالیان بسیار دور از زمانی که بشر به دنبال ساخت موجودات شبیه به خود بود برمی‌گردد ولی اولین کارهای علمی قابل توجه در این زمینه پس از جنگ جهانی دوم و به طور مشخص به اواخر دهه ۶۰ و شروع دهه ۷۰ میلادی بازمی‌گردد. در سال ۱۹۶۹ محققین دانشگاه واسدا[1] در ژاپن موفق به ساخت ربات انسان نمایی شدند که با تعادل استاتیکی مرکز ثقل[2] و با سرعت بسیار کمی قادر به حرکت بر روی دوپای خود بود. پس از استخراج معادلات حرکت دینامیکی حاکم بر ربات‌های بازویی و با تعریف یک معیار پایداری دینامیکی به نام نقطه گشتاور صفر[3] در سال ۱۹۷۰ توسط ووکوبراتویچ[4]، نقطه عطفی جهت انجام تحقیقات بیشتر در زمینه‌های دینامیک، پایداری و کنترل انواع این ربات‌ها پدیدآمد.

شاید اواخر دهه ۹۰ و دهه آغازین قرن بیست و یکم، نقطه شکوفایی تحقیقات در زمینه ربات‌های دارای پا و به طور خاص نوع دوپای آن به‌حساب آید. در طی این سال‌ها، اهمیت و توجه به این پروژه در برخی از کشورها مانند ژاپن تا حدی بالا رفت که ساخت بهترین ربات انسان نما پروژه ملی این کشور بوده است. در حال حاضر شرکت های هوندا[5]، تویوتا[6] و سونی[7] که برخی، کار بر روی این پروژه را از ۲۵ سال پیش آغاز کرده‌بودند، پیشتازان این نوع تکنولوژی رباتیک در دنیا هستند به نحوی که ربات‌های ساخته شده توسط این شرکت‌ها (به ترتیب آسیمو[8]، ربات همیار[9] و کیوریو[10]) نشانه های پیشرفت و اقتدار ملی کشور ژاپن محسوب شده و در برخی از سفرهای خارجی دولتمردان این کشور را همراهی می‌کنند. در امریکا نیز در سال ۱۹۸۰، آقای مارک ریبرت[11] با تاسیس آزمایشگاه ربات‌های دارای پا در دانشگاه ام‌آی‌تی[12] و سپس ساخت انواع مختلفی از این ربات‌ها از جمله یک ربات تک‌پای جهش کننده دارای پایداری، سهم بالایی را در توسعه دانش در این زمینه ایفا کرد. وی بعدها در سال ۱۹۸۶ کتاب "Legged Robots That Balance" را منتشر کرد که یک منبع اصلی در این زمینه به‌حساب می‌آید. در ژاپن بعدها موسسات تحقیقاتی دیگری چون موسسه ملی تحقیقات پیشرفته علوم صنعتی[13] نیز به این جریان پیوستند. تاکنون در نقاط دیگری از دنیا نیز، تلاش‌هایی مجزا آغازشده‌است.  در زمینه انوع دیگر ربات‌های دارای پا نیز، شرکت بوستون داینامیکز[14] که توسط آقای ریبرت اداره‌می‌شود، توانسته‌است ربات چهارپایی بسازد که می‌تواند با پایداری دینامیکی فوق‌العاده بالایی در برابر اغتشاشات محیطی، بر روی سطوح غیر مسطح حرکت کند، با سرعت

---

[1] Waseda
[2] CoG: Center of Gravity
[3] ZMP: Zero Moment Point
[4] Vukobratović
[5] Honda
[6] Toyota
[7] Sony
[8] Asimo
[9] Partner Robot
[10] Qurio
[11] Marc Raibert
[12] MIT: Massachusetts Institute of Technology
[13] AIST: national institute of Advanced Industrial Science and Technology
[14] Boston Dynamics

بالایی بدود، از سراشیبی بالا برود و مانورهای حرکتی زیادی را به نمایش بگذارد که کاربرد آن حمل تجهیزات و پشتیبانی نیروها در مناطق ناهموارکوهستانی است.

## ۱-۲ اهداف و کاربردها

اگر بخواهیم به اهداف و کاربردهای این دسته از رباتها بپردازیم، بهتر است توصیف دقیقتری از مزایای اصلی آنها در مقایسه با دیگر رباتها از جمله مزیتهای حرکتی و همچنین مشابهتهای رفتاری و ارتباطی آن با انسان ارائهدهیم. اگر رباتهای دارای پا را برحسب تعداد پا به رباتهای تکپای جهشکننده[1]، رباتهای دوپا[2]، ربات-های سهپا[3]، چهارپا[4] و ششپا[5] و رباتهای چندپا[6] که داری بیش از شش پا هستند تقسیمکنیم، به غیر از مزایای حرکتی ذکر شده نسبت به دیگر رباتهای متحرک، این رباتها و به خصوص نوع پیشرفتهتر این رباتها یعنی ربات دوپای انساننما که دارای بیشترین قدرت مانور در میان رباتهای دارای پا میباشد، میتواند مانورهای حرکتی متنوعی را به صورت مواردی که در ذیل ذکر شده انجام دهد:

- راه رفتن، دویدن و پریدن از روی موانع

- حرکت بر روی سطوح شیبدار ، پله ، نردبان و انواع ناهمواری ها

- حرکت بر روی مسیرهای باریک، از میان شکافها و حفرهها

- بالا رفتن و پایین آمدن از صخرهها و گذر کردن از روی شکافهای عمیق

انواع مانورهای حرکتی ممکن برای ربات انساننما در شکل ۱-۱، نمایشدادهشدهاست.

---

[1] Monopod Hopping Robots
[2] Biped Robots
[3] Tripod Robots
[4] Quadruped Robots
[5] Hexapod Robots
[6] Multipeds Robots

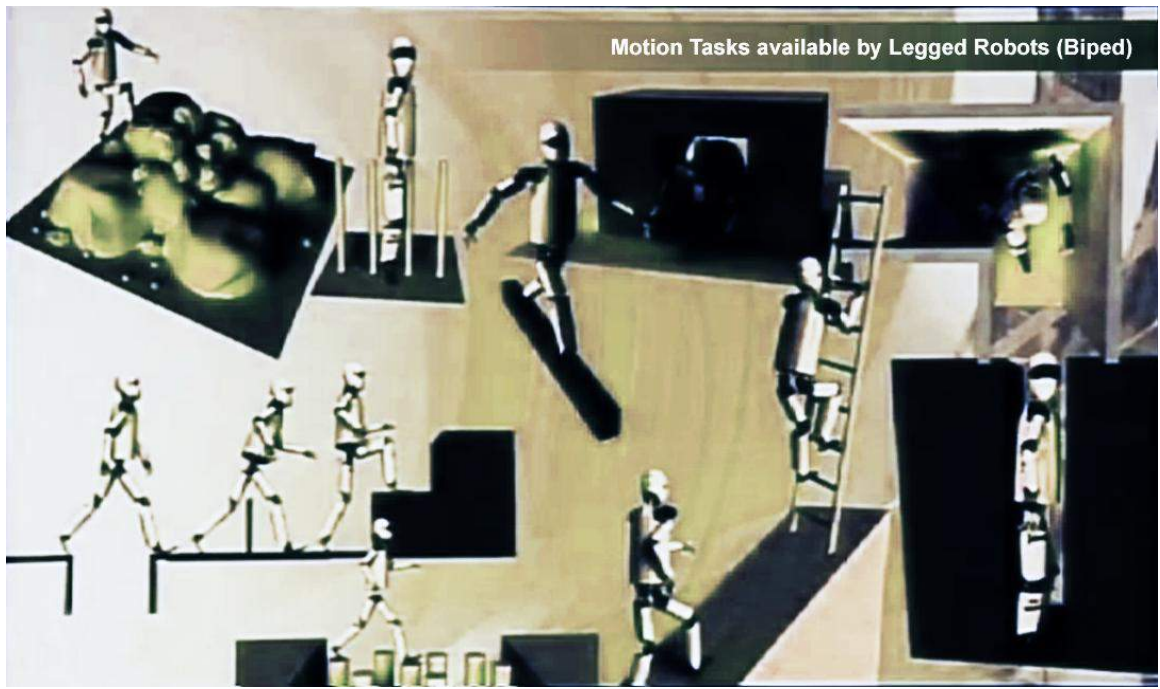

شکل ۱-۱. انواع مانورهای حرکتی ممکن ربات‌های انسان‌نما

مزیت دیگر ربات دوپای انسان‌نما سازگاری بالا با انسان و محیط زندگی و کار وی است که در موارد زیر قابل بیان است.

- ربات انسان‌نما به علت شباهت، بهترین شریک در میان ربات‌ها برای انسان محسوب می‌شود(می‌تواند با انسان‌ها از هر سنی ارتباط برقرار کند: از بازی با کودکان گرفته تا کمک به بزرگسالان)

- می‌تواند به هر مکانی که انسان‌ها می‌توانند بروند، حرکت کند.

- می‌تواند تقریبا تمامی وظایفی که یک انسان به عهده می‌گیرد را به عهده بگیرد.

با توجه به این ویژگی‌های منحصربفرد ربات دوپای انسان‌نما و همچنین مزایای نسبی دیگر انواع ربات‌های دارای پا به نظر می‌رسد کاربردهای زیر در نگاهی به آینده برای این ربات‌ها قابل پیش‌بینی باشد(شکل ۱-۲).

## ۱-۲-۱ ربات‌های شهری [1]

ربات‌های انسان‌نما می‌توانند در آینده وظیفه کمک به افراد و انجام بسیاری از کارها را در محیط کار و زندگی انسان به عهده بگیرند. این پروژه با هدف افزایش بهره‌وری در محیط‌های شهری در کشور ژاپن به عنوان یک پروژه

---

[1] Urban Robots

ملی در حال پیگیری است. در این زمینه، شرکت هوندا با معرفی ربات آسیمو تاکنون کارهایی از جمله ارائه سرویس در غذاخوری و درمانگاه‌ها و ارائه آموزش‌های برنامه‌ریزی‌شده به افراد انجام داده است. ربات کیوریو ساخت شرکت سونی می‌تواند فعالیت‌هایی از قبیل بازی با کودکان و ارتباط با افرادی از سنین مختلف از جمله آشنایی واظهار دوستی با اشخاص به‌وسیله به حافظه سپردن شکل و صدای آن‌ها و یادآوری آن در هنگام بازدید را انجام دهد. شرکت توبوتا نیز به تازگی با ساخت ربات همیار کارهایی ازجمله نواختن آلات موسیقی و انجام مراسم تشریفاتی را توسط این ربات ارائه‌کرده‌است. این کاربرد از ربات انساننما یعنی ربات شهری به علت امکان برخورد مداوم ربات با افراد و محیط‌های ناشناخته نیاز فوق‌العاده‌ای را در زمینه پایدارسازی حرکتی به وجود آورده است که تاکنون کارهایی در این زمینه انجام‌شده‌است و موضوع این تحقیق نیز می‌باشد.

## ۲-۲-۱ ربات پشتیبان[1]

کاربرد دیگر ربات‌های دارای پا، حمل تجهیزات برای نیروها در مناطق کوهستانی و صعب‌العبور است که به راحتی توسط این نوع ربات قابل انجام می‌باشد. شرکت بوستون داینامیکز[2] هم‌اکنون به سفارش دولت، ربات چهارپایی را به نام سگ عظیم‌الجثه[3] ساخته است که با تعادل بسیار بالا قادر به انجام این کار می‌باشد. با توجه به راهبردی بودن این وظیفه، کاربرد اخیر می‌تواند به یکی از مهمترین کاربردهای این ربات در آینده تبدیل شود.

## ۳-۲-۱ وسایل نقلیه جدید[4]

یکی دیگر از کاربردهای این ربات‌ها می‌تواند ساخت وسایل نقلیه حرکتی جدید برای گشت و گذار در مناطق کوهستانی و دشت‌ها برای تفریح و یا حرکت از نقطه‌ای به نقطه دیگر باشد. هم‌اکنون وسیله نقلیه‌ای به صورت یک ربات تک‌پای جهش‌کننده ساخته‌شده است که می‌تواند یک انسان را حمل کند. افرادی که این پروژه را انجام-داده‌اند هم‌اکنون در شرکت بوستون داینامیکز مشغول به کار هستند.

## ۴-۲-۱ کاربردهای جانبی[5]

کاربردهای جانبی که در عین حال یکی از مهمترین کاربردهای تحقیقات در این زمینه است، به استفاده از نتایج بخش‌های مختلف آن در دانش بیومکانیک جهت ساخت اعضای مصنوعی فعال[6] مانند پاهای مصنوعی متحرک، پوشش خارجی حرکتی[1]، رفع نواقص حرکتی[2] و حرکت درمانی[3] می‌پردازد.

---

[1] Military Logistics Robot
[2] Boston Dynamics
[3] Big Dog
[4] Legged Vehicles
[5] Side Benefits
[6] Active/Powered Limb Prosthesis

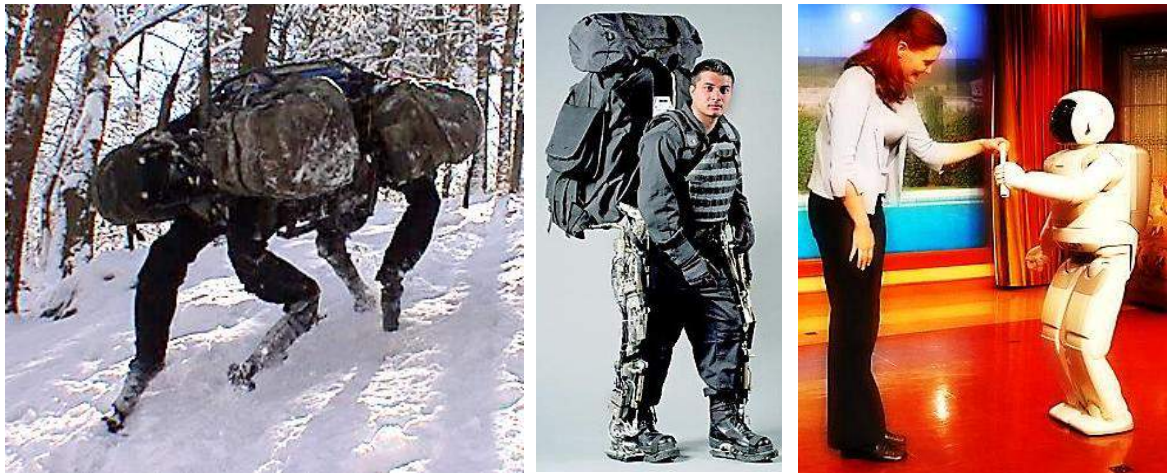

شکل ۲-۱. برخی کاربردهای ربات‌های دارای پا

## ۱-۳ مفاهیم اولیه ربات دوپا

### ۱-۳-۱ تقسیم بندی ربات‌های دوپا

- **انواع ربات‌های دوپا بر اساس نوع تحرک**

ربات‌های دوپا بر اساس داشتن و یا نداشتن تحریک خارجی به دو دسته کلی ربات‌های دوپا با مفاصل غیرفعال[4] و ربات‌های دوپا با مفاصل فعال[5] تقسیم‌بندی می‌شوند.

ربات دوپا با مفاصل غیر فعال رباتی است که مفاصل آن بدون تحریک خارجی است و ربات تنها بر اثر نیروی جاذبه زمین بر روی سطح شیب‌دار به پایین حرکت می‌کند. مفهوم ربات دوپای غیر فعال اولین بار توسط مک‌گیـر[6] در سال ۱۹۹۰ ارائه شد. وی استفاده از خمیدگی در کف پا برای راه رفتن ربات دوپای غیر فعال بر روی سـطحی بـا شیب کم را پیشنهاد نمود[۱]. همچنین، گوسوامی[7] و همکارانش تعریفی برای پایداری یک ربات دوپـای غیـر فعـال دو عضوی بیان کردند[۲]. از نقاط ضعف ربات دوپای غیرفعال می‌تـوان بـه وابسـتگی شـدید بـه شـیب سـطح، عـدم امکان حرکت در زمین‌های مسطح و شیب‌های رو به بالا و عدم تغییر سرعت و طول گام در شیب ثابت اشاره نمود.

---

[1] Exoskeleton
[2] Locomotory Rehabilitation
[3] Kinesiotherapy
[4] Passive Robot
[5] Active Robot
[6] McGeer
[7] Goswami

در ربات‌های دوپا با مفاصل فعال در هر یک از مفاصل عملگری وجود دارد. از این‌رو حرکت ربات قابل کنترل است و ربات توانایی حرکت بر روی مسیرهای مختلف را دارد. همین امر موجب گستردگی استفاده از ربات‌های دوپا با مفاصل فعال شده است.

- **انواع مختلف ربات دوپا از لحاظ تعداد عضوها**

مدل‌های متفاوتی برای ربات دوپا در نظر گرفته شده است. مدل سه لینکی، پنج لینکی، هفت لینکی و نه لینکی نمونه‌هایی از آن‌ها هستند. از آنجایی که در مدل‌های سه لینکی و پنج لینکی به ترتیب زانو و کف پا در نظر گرفته نمی‌شود، این دو مدل نمی‌توانند به خوبی گویای سینماتیک راه رفتن انسان باشند. از طرفی در مدل نه لینکی نسبت به مدل هفت لینکی پنجه در نظر گرفته می‌شود و همین امر سبب پیچیدگی سینماتیک حرکت ربات می‌گردد. مدل هفت لینکی به دور از پیچیدگی‌های مذکور، ویژگی‌های مهم راه رفتن را در نظر می‌گیرد. از این‌رو استفاده از این مدل کاربرد بیشتری نسبت به سایر مدل‌ها دارد. در شکل ۳-۱ مدل دوپای هفت لینکی نشان داده شده است.

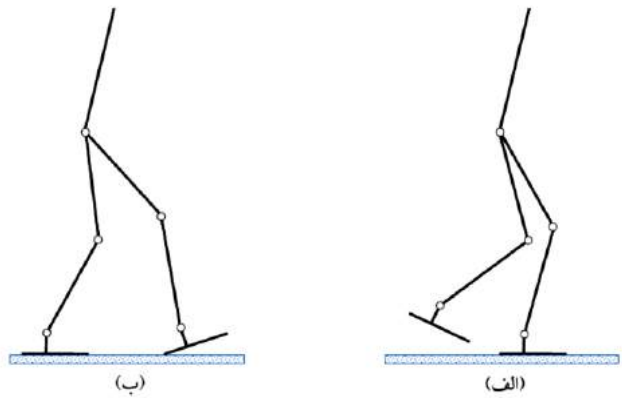

شکل ۳-۱ فازهای راه رفتن انسان: الف) فاز تک تکیه‌گاهی و ب) فاز دو تکیه‌گاهی

- **انواع مختلف ربات دوپا از منظر عضو کف پا**

ربات‌های دوپا از جهت نوع تماس کف پا با زمین به دو نوع ربات‌های ۱) دارای کف پای نقطه‌ای و ۲) ربات-های دارای کف پای مسطح یا چندتکه تقسیم می‌شوندکه نوع دوم، نوع متداول این ربات‌ها می‌باشد. ربات‌های دارای کف پای نقطه‌ای، دوپاهای کم‌عملگر[1] نامیده می‌شوند، زیرا تعداد محرک‌های این ربات‌ها همواره کمتر از تعداد درجات آزادی‌شان خواهد بود، به همین دلیل توانایی کنترل دقیق حرکت آن‌ها غیـرممکن اسـت و تنها مـی-توان در مورد پایداری آن‌ها حول سیکل‌های حدی سخن گفت.

---

[1] Underactuated Biped

### ۱-۳-۲    صفحات حرکت انسان

حرکات بدن انسان در سه صفحه طولی[1]، عرضی[2] و برشی[3] بیان می‌شود. هر یک از این صفحات در شکل ۱-۴
نشان داده شده است. اگرچه حرکت در صفحه عرضی نقش مهمی در پایداری ربات دوپا دارد ولی برای کاستن از
پیچیدگی تحلیل ربات، معمولا تنها حرکت ربات در صفحه طولی در نظر گرفته می‌شود. به این‌گونه مدل‌ها،
مدل‌های صفحه‌ای گفته می‌شود که مدل درنظرگرفته‌شده در این تحقیق نیز مدلی صفحه‌ای می‌باشد.

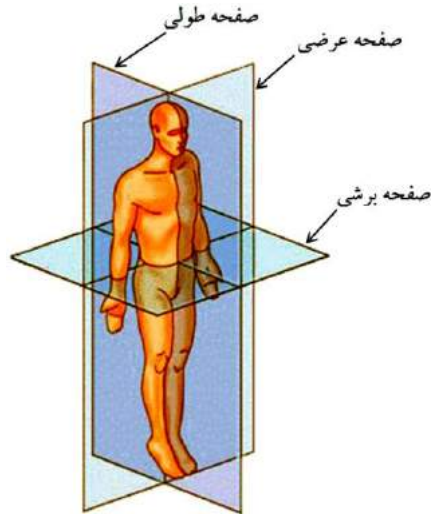

شکل ۱-۴ صفحه‌های حرکت انسان

### ۱-۳-۳    فازهای راه رفتن انسان

بر اساس تماس پاها با سطح زمین، راه رفتن انسان به دو فاز تک تکیه‌گاهی[4] و دو تکیه‌گاهی[5] تقسیم می‌شود. در
فاز تک تکیه‌گاهی پای تکیه‌گاهی[6] با زمین در تماس است و پای متحرک[7] به طرف جلو حرکت می‌کند. فاز تک
تکیه‌گاهی با بلند شدن پنجه پای متحرک از سطح زمین آغاز شده و با برخورد پاشنه آن با زمین خاتمه می‌یابد. در
این فاز یک زنجیره سینماتیکی باز به وجود می‌آید. در فاز دو تکیه‌گاهی هر دوپا با زمین در تماس هستند و کل
بدن به سمت جلو حرکت می‌کند. فاز دو تکیه‌گاهی با برخورد پاشنه پای متحرک با زمین آغاز شده و با بلند شدن

---

[1] Sagittal Plane
[2] Lateral Plane
[3] Transverse Plane
[4] Single Support
[5] Double Support
[6] Stance Leg
[7] Swing Leg

پنجه پای تکیه‌گاهی از سطح زمین خاتمه می‌یابد. در این فاز یک زنجیره سینماتیکی بسته[1] به وجود می‌آیـد. در شکل ۱-۳، فازهای حرکت یک ربات دوپای هفت عضوی نشان داده شده است.

### ۴-۳-۱ پایداری ربات دوپا

یکی از مهمترین مسائلی که در مورد ربات‌های دارای پا مطرح می‌شود، حفظ تعادل ربات در هنگـام راه رفتن است. در ربات‌های راه رونده نظیر ربات‌های هشت‌پا، شش‌پا، دوپا و یک‌پا بسته به نوع ربات، تمایل به ناپایداری متفاوت است. در ربات‌هایی که تعداد پاهای آن‌ها زیاد است، تمایل به ناپایداری بـه مراتـب کمتـر از ربات‌هـایی بـا تعداد پای کمتر است. در بخش بعدی به لزوم پرداختن به بحث پایداری، معیارهای پایداری[2] و بسط مفهوم پایـداری ربات‌های دارای پا خواهیم پرداخت تا به یک صورت مسئله کلی برای کنترل پایداری دست پیدا کنیم.

### ۴-۱ پایداری

### ۱-۴-۱ لزوم پرداختن به مسئله پایداری

یکی از کلیدی‌ترین نکات در تاریخچه تحقیقاتی ربات‌های دارای پا، گره‌خوردن مبحث پایداری در این ربات‌ها با روش‌های کنترلی آن و حتی مقدم‌بودن پایداری نسبت به کنترل است که نشان از اهمیت بالای مبحـث پایـداری دارد. ارائه معیارهای پایداری دینامیکی متعدد دیگری غیر از نقطه گشتاور صفر[3] از جمله معیـار شـاخص دوران پـا[4]، معیار نرخ اندازه‌حرکت زاویه‌ای صفر[5]، نشان از میزان اشتغال فکری محققان به این زمینه تحقیقـاتی اسـت کـه هنـوز به نتیجه‌ی نهایی نیز نرسیده است. لزوم پرداختن به مبحث پایداری و کنترل پایداری در موارد زیر قابل بیان است.

- در حرکت این ربات‌ها نیازی به پایداری ذاتی وجود دارد که به علت وجود کنش و واکنش نیرویی مداوم و درعین حال قطع و وصل شونده بین ربات و محیط می‌باشد (همواره پاها با زمین در حال برخورد و جداشدن هستند). (شکل ۱-۵- الف)

- حرکت در زمین‌های ناشناخته، ناهموار و یا دارای شیب، و جلوگیری از تاثیر اغتشاشات داخلی، نیاز به کنترل پایداری به صورت به‌هنگام[6] دارد. (شکل ۱-۵ – ب)

---

[1] Close-Chain Configuration
[2] Stability Criteria
[3] ZMP: Zero Moment Point
[4] FRI: Foot Rotation Index
[5] ZRAM: Zero Rate of Angular Momentum
[6] online

• هل داده شدن، وارد شدن ضربه در اثر برخورد با موانع و به وجود آمدن اختلالات حرکتی داخلی یا خارجی نیاز به طراحی کنترل‌کننده‌هایی با مقاومت بالا برای پایدارسازی مجدد خواهد داشت.(شکل ۱-۶)

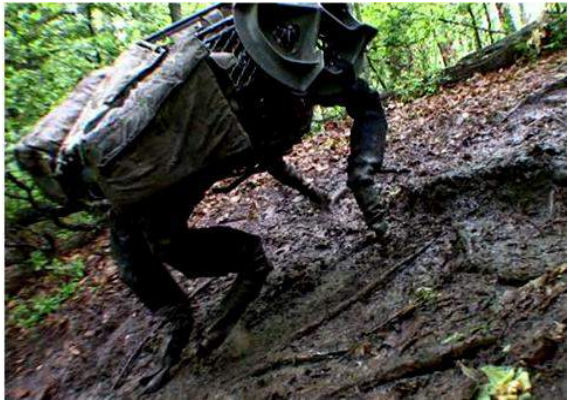 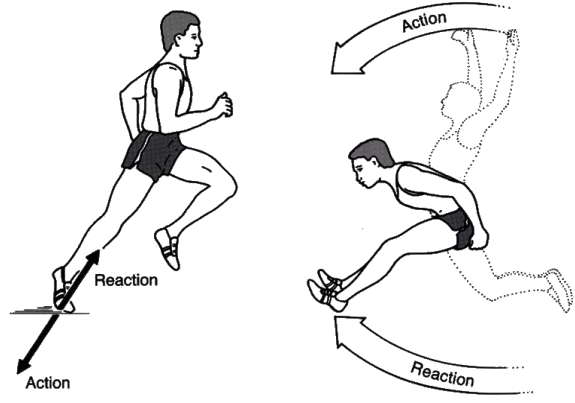

ب– حرکت در زمین‌های ناشناخته یا ناهموار      الف – کنش و واکنش نیرویی قطع و وصل شونده بین بدن و محیط

شکل ۱-۵. نیازهای موجود در پرداختن به پایداری و کنترل پایداری در ربات‌های دارای پا

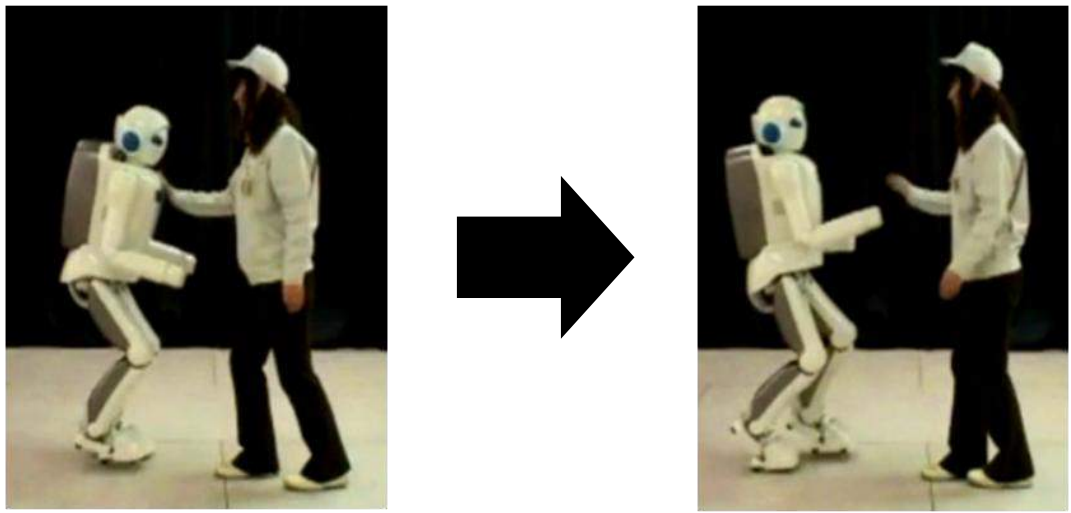

شکل ۱-۶. نیاز به پایدارسازی مجدد پس از دریافت ضربه

### ۲-۴-۱ معیارهای پایداری

**● پایداری وضعی**

همان‌طور که ذکر شد، پایداری یکی از مسئله‌های باقیمانـده در ربات‌هـای دارای پـا مـی باشـد. تـاکنون نـزد همـه محققان به غیر از مولفان مراجع [۳] و [۴]، پایداری این ربات‌ها برابر با پایداری وضعی [1] تعریف‌شده‌است. پایـداری وضعی، ادامه حرکت ربات در یک وضعیت ثابت و یا بر روی یک مسیر از پیش طراحی شده را تضمین می‌کند کـه به دو حالت پایداری استاتیکی و دینامیکی تقسیم‌بندی می‌شود.

برای معرفی انواع معیارهای پایداری وضعی نیاز به تعـاریف جدیـد خـواهیم داشـت: ناحیـه تکیـه‌گـاهی [2]، ناحیـه محدب متشکل از تمامی نقاط تماسی بین ربات و زمین است و نیـروی عکـس‌العمـل زمـین(GRF) [3]، نیـروی برآینـد تمام نیروها و گشتاورهای وارده از طرف سطح زمین بـه ربـات مـی‌باشـد و هـر لحظـه در یـک موقعیـت و راسـتای مشخص قرار دارد(شکل ۱-۷).

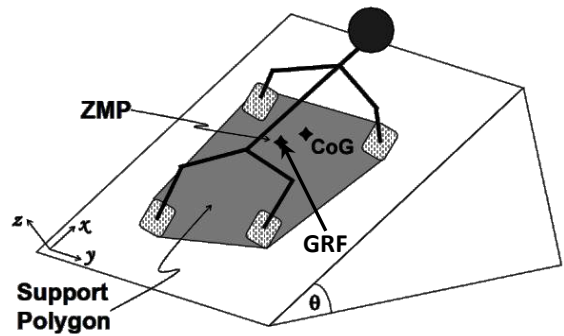

شکل ۱-۷. ناحیه تکیه‌گاهی(Support Polygon)، نیروی عکس‌العمل زمین(GRF)، تصویر مرکز ثقل(CoG) و نقطه گشتاور صفر(ZMP)

به طور کلی می‌توان گفت، تا زمانی که برآیند نیروهای واکنشی وارده از سطح زمین(GRF) داخل ناحیـه تکیـه‌گاهی قرار دارد، پایداری وضعی تضمین می‌شود ولی اگر بر روی مرز این ناحیه قرارگرفت، این پایـداری تضمین نمی‌شـود و بـا داشـتن ایـن اطلاعـات نمـی‌تـوان در مـورد پایـداری وضـعی سـخن بیشـتری گفـت. معیارهـای مرکزثقل(CoG) [4] و نقطه گشتاور صفر(ZMP) [5] بر این واقعیت استوار هستند. در ادامه معیار پایداری شـاخص دوران پا(FRI) [6] معرفی‌شده‌است که وضعیت پایداری وضعی را در ایـن حالـت بـا درنظرگـرفتن اطلاعـات بیشـتری ماننـد نیروهای وارد بر هر پا و گشتاور مچ پا مشخص می‌کند. نقطه صفرکننده نرخ اندازه حرکـت زاویـه‌ای(ZRAM) [7] نیـز

---

[1] Postural Stability
[2] Support Polygon
[3] GRF: Ground Reaction Force
[4] CoG: Center of Gravity
[5] ZMP: Zero Moment Point
[6] FRI: Foot Rotation Indicator
[7] ZRAM: Zero Rate of Angular Momentum

یک معیار پایداری محسوب می‌شود که میزان پایداری دورانی ربات[1] را بررسی می‌کند ولی هرگز یک شرط لازم یا کافی نه برای پایداری وضعی و نه برای معیار پایداری کلی به حساب نمی‌آید.

## ۱. معیار پایداری استاتیکی مرکز ثقل(CoG)

این معیار بیان می‌کند که ربات دارای پا، مادامی که تصویر مرکز ثقل آن در امتداد شتاب جاذبه از ناحیـه تکیـه‌گاهی بگذرد، دارای پایداری استاتیکی است که از نوشتن معادلات تعادل نیرو و گشتاور نتیجه می‌گردد.

## ۲. معیار پایداری دینامیکی نقطه گشتاور صفر(ZMP) یا مرکز فشار(CoP)[2]

نقطه گشتاور صفر(ZMP) که اول بار توسط ووکوبراتویچ[3] در سال ۱۹۷۲ معرفی‌شد، طبق تعریف اول اصلاح‌شده توسط وی در سال ۲۰۰۴، نقطه‌ای بر روی سطح زمین است کـه گشتاور ایجادشـده توسـط مجموعـه نیروهـای گرانشی و اینرسی ربات، در راستای محورهای واقع در سطح افق، در آن نقطه صفر می‌شود. معیار ZMP بیـان مـی‌کند که اگر نقطه ZMP در ناحیه محدب متشکل از همه نقاط تکیه‌گاهی قراربگیرد، ربات پایداری خواهـد داشـت ولی در حالتی که بر روی مرز این ناحیه بیافتد، تضمینی جهت حفظ پایداری وجود نخواهد داشت. ایـن معیـار تنهـا شرطی کافی برای پایداری بیان‌می‌کند. با نوشتن معادلات تعادل نیرو و گشتاور مـی‌تـوان نشـان داد نقطـه ZMP بـر مرکز فشار (CoP) نیروی عکس‌العمل زمین(GRF) منطبق است و بنابراین هردو تعریفی معادل را ارائه می‌کنند.

## ۳. معیار پایداری شاخص دوران پا (FRI)

گوسوامی[4] درسال ۱۹۹۹، نقطه‌ای با عنوان شاخص دوران پا (FRI) معرفی کرد. وی این نقطه را، مکانی بـر روی صفحه تماسی کف پا با زمین معرفی‌می‌کند که اگر بردار مجمـوع نیروهـا و گشـتاورهای واردشـده از طـرف سـطح زمین بر عضو کف پا در این نقطه عمل‌کند، از دوران کف پا جلوگیری‌می‌کند[۵]. بدیهی است این نقطه مـی‌توانـد خارج محوطه تکیه‌گاهی نیز قرار گیرد که در این صورت به دلیل عدم امکان تامین گشتاور لازم، کف پا حول پنجـه یا پاشنه دوران‌می‌کند و ناپایداری وضعی به وجود می‌آید. هر چه فاصله این نقطـه از مـرز ناحیـه تکیـه‌گـاهی دورتـر باشد، ناپایداری بیشتر خواهدشد و مزیتش نسبت به ZMP، ارائه معیاری لازم و کافی برای پایداری حرکـت ربـات براساس وضعیت کف پا خواهدبود. کاربرد این معیـار در انـدازه‌گیـری میـزان پایـداری بـه هنگـام کنتـرل بـه‌هنگـام پایداری وضعی است.

---

[1] Tip Over Stability
[2] CoP: Center of Pressure
[3] Vukobratović
[4] Goswami

## ۴. معیار نرخ اندازه‌حرکت زاویه‌ای صفر (ZRAM)

گوسوامی سپس درسال ۲۰۰۴ نقطه‌ای با عنوان نقطه نرخ اندازه‌حرکت زاویه‌ای صفر (ZRAM) مطرح کرد که حتی در حالتی که مرکز فشار(CoP) داخل ناحیه تکیه‌گاهی(نرسیده به مرز ناحیه) حفظ می‌شود، معیاری برای میزان ناپایداری ربات برحسب فاصله‌گرفتن از نرخ اندازه‌حرکت‌زاویه‌ای صفر به‌دست‌می‌دهد[۶]. نقطه ZRAM مکانی بر روی سطح تکیه‌گاهی است که اگر بردار مجموع نیروهای تماسی در آن نقطه عمل کند، از مرکز ثقل ربات می‌گذرد و نرخ اندازه‌حرکت زاویه‌ای کل ربات را صفرمی‌کند. البته در پایان می‌شود غیر صفر بودن نرخ اندازه‌حرکت‌زاویه‌ای، تنها دلیل ناپایداری نیست و غیر صفر بودن خود اندازه‌حرکت‌زاویه‌ای نیز منجر به ناپایداری می‌گردد(شکل ۱-۸).

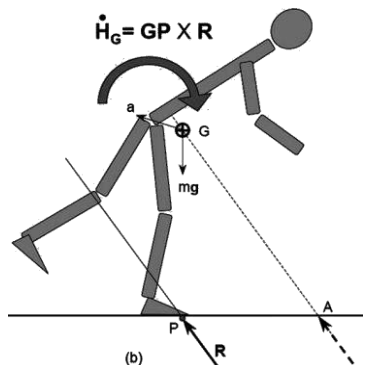

شکل ۱-۸  R نیروی GRF است که در P یا نقطه CoP (مرکز فشار) عمل می‌کند و A نقطه ZRAM را نشان‌می‌دهد.

• **پایداری کلی**

تلاشی که محققان برای رسیدن به یک معیار پایداری تعمیم‌یافته‌تر نسبت به معیارهای پایداری وضعی انجام‌داده‌اند تا کاستی‌های معیارهای موجود را برطرف بسازد، به خصوص کار مولفان [۳] و [۴] ما را تشویق کرد تا به بررسی مفهوم پایداری بپردازیم. توسعه این مفهوم در ادامه به ما کمک خواهد کرد تا با درنظرداشتن آن در این پژوهش، به یک روش جدید کنترل پایداری دست‌یابیم که به مسئله پایداری به عنوان حفظ متغیرهای سیستم درون یک ناحیه زیست‌پذیر که تنها خصوصیت آن جلوگیری از افتادن است، نگاه می‌کند. اگرچه در این روش کنترلی، معیاری جایگزین معیارهای وضعی مطرح نمی‌شود، اما، یک سطح پایداری بالاتری تعریف می‌گردد که محدوده وسیع‌تری از فضای حالت را پوشش می‌دهد. این روش کنترل پایداری با تعریف سیکل‌های حرکتی، بر پایه هدایت و کنترل حرکت به سمت این سیکل استوار خواهد بود و خود را محدود به طراحی یک مسیر پایدار برای زوایای مفاصل و کنترل حرکت مفاصل حول آن مسیر نخواهد کرد. این روش نیازی به بررسی پایداری خود با معیارهای پایداری وضعی ندارد چراکه دیگر توجهی به دوری یا نزدیکی حرکت مفاصل خود به یک مسیر حرکتی از‌پیش-طراحی‌شده نخواهد کرد.

### ۱-۴-۳ بررسی درستی معیارهای متداول پایداری و نقص‌های موجود در آن‌ها

معیارهای مطرح شده اگرچه هریک دارای مزایایی می‌باشند، نقص‌هایی نیز به همراه دارنـد. اگـر از معیار CoG که تنها حالت استاتیکی را توصیف می‌کند آغاز کنیم، همان‌طور که در شکل ۹-۱ نمایش‌داده‌شـده‌اسـت، ایـن معیار در حالت کلی نه شرط لازم و نه کافی برای پایداری است و تنها شرطی کافی برای حرکت‌های آرام بر روی زمین‌های مسطح با فرض غیردقیق تامین اصطکاک موردنیاز می‌باشد، ولی بـرای زمیـن‌هـای ناهمـوار بررسـی شـرط عـدم لغزش برای بررسی پایداری حتما باید انجام‌شود. همچنین درحالتی که تکیه‌گاهی در ارتفاعی بـالاتر از مرکز ثقل ربات قرار گیرد، این معیار شرطی لازم برای پایداری نیست.

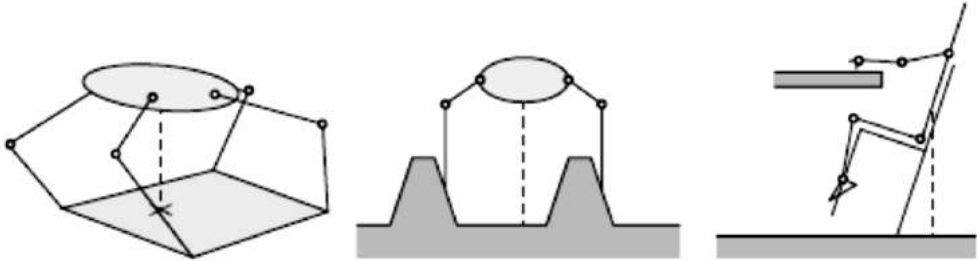

شکل ۹-۱. به طور کلی یک ربات دارای پا(Legged Robot) مادامی که امتداد مرکز ثقل آن از ناحیه محدب متشکل از نقاط تکیـه- گاهی بگذرد، دارای پایداری استاتیکی است(چپ) ولی گاهی این اتفاق می‌افتد و ربات پایدار نیست(وسط) و گاهی نیز مرکز ثقل این شرط را برآورده‌نمی‌کند ولی ربات تعادل پایدار دارد(راست).

معیار ZMP نیز دقیقا دارای همین اشکالات است به عـلاوه اینکـه ایـن معیـار تنهـا شـرطی کـافی بـرای پایداری وضعیت تکیه‌گاه ربات ارائه‌می‌دهد. در حالتی که ZMP در مرز ناحیه تکیه‌گـاهی قرارمـی‌گیـرد، وضعیـت پایـداری نامشخص است. اگرچه برای استفاده از معیارهای دینامیکی بـر روی سطوح غیریکنواخت نیـز تـلاش‌هـایی شـده‌ است[۶]، ولی معیارهای حاصل، معیارهایی غیردقیق به حساب می‌آیند.

معیار FRI نیز اگرچه شرطی لازم و کافی برای پایداری مطرح می‌کنـد ولـی نکتـه اینجـا اسـت کـه در همـه ایـن معیارهای متداول پایداری که تا اینجا بررسی شد، **پایداری وضعیت تکیه‌گاه‌ها** معیاری برای **پایداری کلـی راه-رونده** تلقی شده است که همان‌طور که در ادامه با پرداختن به مفهوم مسئله پایـداری، نشـان خـواهیم داد، **الزامـی برای معادل گرفتن این دو پایداری وجود ندارد.**

معیار ZRAM نیز هرچند از معادل‌دانستن پایداری ربات بـا پایـداری وضعیت تکیـه‌گاه‌هـا فراتـر رفتـه، ولـی بـه اعتراف خود همه علت‌ها یا به‌طور صریح‌تر علت اصلی ناپایداری در ایـن معیـار درنظرگرفتـه‌نشـده‌اسـت و بنـابراین نمی‌توان از آن به عنوان یک معیار پایداری جامع یادکرد.

### ۱-۴-۴ مفهوم پایداری ربات‌های دارای پا و ارائه تعریفی جدید از پایداری کلی

برای شروع، با یک مثال مسئله پایداری را بررسی می‌کنیم. اگر به یک ربات دوپـا کـه در حالـت ایسـتاده قرار دارد، اغتشاشی وارد شود ربات می‌تواند حالت خود را حفظ کند و یا به یکی از سه حالت (روی پاشنه، روی پنجه و

معلق در هوا)، نمایش داده‌شده در شکل ۱-۱۰ درآید ولی مادامی که این ربات بتواند دوباره به حالت ایستاده بازگردد، از آن به عنوان حرکتی پایدار یاد می‌کنیم. بدیهی است در صورت ناپایدارشدن، ربات به زمین خواهد افتاد[۴].

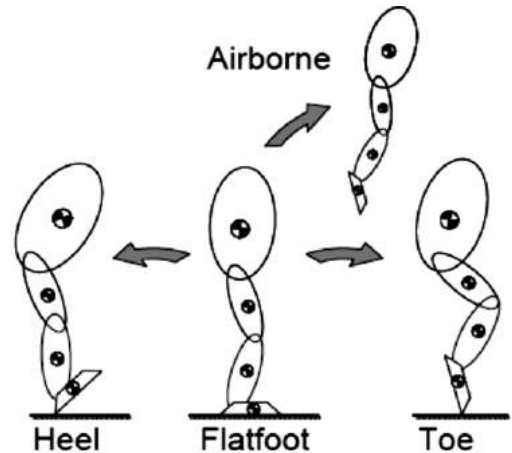

شکل ۱-۱۰. وضعیت ربات ایستاده پس از وارد کردن یک اغتشاش [۴]

با توجه به این مثال می‌توان نتیجه گرفت، مفهوم پایداری در ربات‌های دارای پا دقیقا به معنی پایداری حول یک نقطه تعادل یا مسیر مرجع و یا حرکت به سمت سیکل‌های حدی مانند دیگر سیستم‌های مکانیکی نمی‌باشد، بلکه مفهومی معادل با ادامه وظیفه حرکتی بدون واژگون شدن را یادآوری می‌کند. این نوع تعریف از پایداری با تئوری زیست‌پذیری[1] قابل بیان است. اگر مجموعه حالاتی که به طور حتم منجر به افتادن یا واژگون شدن ربات می‌شود را در $\mathcal{F}$ در نظربگیریم، یک حالت $(\theta, \dot{\theta})$ زیست‌پذیر است اگر و تنها اگر سیستم قادر باشد یک حرکت $(\theta(t))$ را از این حالت آغاز کند و هرگز وارد $\mathcal{F}$ نشود. در واقع مجموع تمام حالاتی از سیستم که جلوگیری از واژگون‌شدن در آن ممکن است و ربات می‌تواند به هدف حرکتی اصلی خود بازگردد، مجموعه‌ای از نقاط در فضای حالت را تعریف می‌کنند که از آن به محدوده یا هسته زیست‌پذیر[2] تعبیر می‌شود. در صورت استفاده همیشگی از حداکثر امکان برای جلوگیری از واژگونی، هسته زیست‌پذیر یک مجموعه نامتغیر[3] نیز محسوب می‌شود(شکل ۱-۱۱).

---

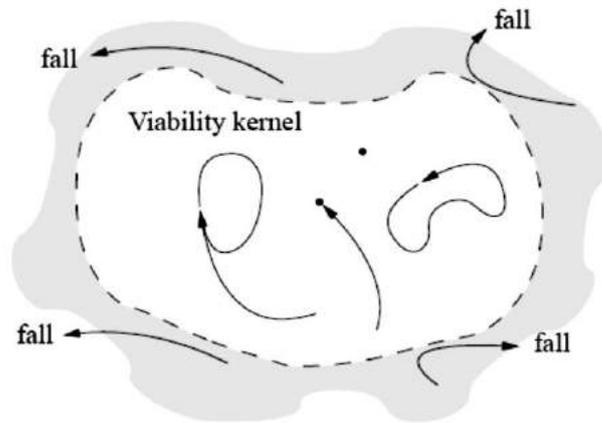

شکل ۱-۱۱. هسته زیست‌پذیر شامل همه حالاتی است که می‌توان از آن از زمین افتادن ربات جلوگیری کرد. برای مثال نقاط تعادل و سیکل‌های حدی جز این هسته اند ولی خارج بودن از این هسته ( قرارگرفتن در $F: Fall\ States$ ) حتما منجر به افتادن خواهدشد. [۳]

با تعریف این مجموعه نامتغیر، مرز پایداری و میزان پایداری قابل تعریف‌خواهدبود ولی نکتـه ایـن اسـت کـه تعریف ریاضی هسته زیست‌پذیر به دلیل وجود پیچیدگی بالای دینامیک ربات‌های راه‌رونده تقریبا ناممکن است. با این‌حال، با تعیین یک کنترل‌کننده مشخص، می‌توان به محـدوده‌ای کنترل‌پـذیر بـه صـورت یـک مجموعـه نـامتغیر دست‌یافت که این هسته زیست‌پذیر می‌باشد. در واقع مجموع تمام محدوده‌هـای کنترل‌پـذیر کـه توسط انواع کنترل‌کننده‌ها قابل دسترسی است، نزدیک‌ترین تعریف بـه هسته زیست‌پـذیر را ارائه‌خواهـد کـرد ولـی رسیدن به چنین تعریفی نیاز به طراحی کنترل‌کننده‌های متنوع و محاسبات عـددی بسـیار طـولانی خواهدداشـت کـه انجام آن قابل توجیه و احتمالا قابل انجام نیز نمی‌باشد.

شاید رسیدن به یک تخمین ضعیف‌تر از مرز این مجموعه نامتغیر با استفاده از پایداری لیاپانوف یا شبه‌لیاپـانوف، ممکن باشد. فرض کنیم تابع اسکالر مثبت غیـرمعین $V(\theta, \dot\theta)$ بـا مشتق پیوسته قابل تعریـف اسـت کـه مقـدار آن مادامی که کنترل‌کننده فعال می‌شود(و پایداری حفظ می‌گردد)، از یک حد مجاز $\alpha$ افزایش‌نمی‌یابد و بـالعکس در صورتی که $V(\theta, \dot\theta) > \alpha$ باشد، فعالیت کنترل‌کننده می‌تواند پایداری را حفظ کند. مـی‌تـوان نتیجـه‌گرفـت کـه مجموعه $V_\alpha = \left\{(\theta, \dot\theta) \mid V(\theta, \dot\theta) < \alpha\right\}$ مادامی که کنترل‌کننده فعال است، یـک مجموعـه نـامتغیر خواهـد بود. حال مجموع تمام مجموعه‌های $V_\alpha$ که در حالت فعال بودن کنترل‌کننـده بـا $F$ تـداخل ندارنـد، یـک مجموعـه نامتغیر را تشکیل می‌دهند که اگر بزرگترین آن‌ها را $V_\Omega$ فرض کنیم، فاصـله نقـاط حالـت بـا مرزهـای ایـن مجموعـه میزان پایداری را مشخص می‌کند و می‌توان از رابطه $\Omega - V(\theta, \dot\theta)$ ، به عنوان معیاری برای نزدیک‌شـدن بـه مـرز ناپایداری و یا میزان پایداری استفاده نمود. برای حالتی که کنترل‌کننده تابعی از زمـان باشـد نیـز مـی‌تـوان بـا انـدکی تغییرات تعریف مشابهی ارائه‌داد با این تفاوت که $\Omega(t)$ و $V(t, \theta, \dot\theta)$ تـوابعی از زمـان هسـتند و پایـداری در هـر لحظه با یک مجموعه نامتغیر قابل بیان است که در مرجع [۳] شرح داده شده است.

## ۱–۵ کنترل پایداری (روش‌های متداول)

کنترل‌کننده‌های متداول برای حفظ پایداری ربات‌های دارای پا به دو شیوه کلی تقسیم‌بندی می‌شوند که یکی کنترل غیربه‌هنگام و دیگری کنترل به‌هنگام پایداری است. مبنای هر دو روش حرکت بر روی یک مسیر از پیش‌طراحی‌شده است که پایداری آن‌ها با توجه به معیارهـای پایداری وضعی و به خصوص معیار گشتاور صفر مجازی (FZMP) تضمین شده باشد. به همین جهت در ابتدای این بخش، ابتدا به مبحث طراحی مسیر پایدار می‌پردازیم تا در معرفی دو شیوه کنترل‌کننده‌های پایداری به چگونگی به‌کارگیری آن در کنترل پایداری بپردازیم.

### ۱–۵–۱ طراحی مسیر پایدار[1]

معیار نقطه گشتاور صفر(ZMP) رابطه‌ای کلی برای طراحی مسیر حرکت این ربات‌ها است که پایداری وضعی آن مسیر را تضمین می‌کند. این معیار هنگام طراحی مسیر، یک رابطه مجازی است و با موقعیت و شتابی که ربات در عمل خواهد داشت، متفاوت خواهد بود(زیرا در عمل همواره نقطه گشتاور صفر (ZMP) منطبق بر نقطه مرکز فشار(CoP) است که هیچگاه از ناحیه تکیه‌گاهی خارج نمی‌شود)، به همین جهت این معیار را در زمان طراحی مسیر، نقطه گشتاور صفر مجازی(FZMP)[2] می‌نامیم. استفاده از این قاعده با رابطه (۱–۱) در راستای افقی $X$ و $Y$ زمین قابل بیان است. شرط کافی برای حفظ پایداری وضعی بدین صورت خواهد بود که FZMP محاسبه شده برای یک مسیر از طریق این فرمول باید داخل ناحیه تکیه‌گاهی قراربگیرد. اگرچه شرط عدم لغزش و عدم جداشدن از زمین نیز، شروطی لازم برای پایداری محسوب می‌شوند ولی معمولا شرایط ضعیف‌تری هستند و همواره برآورده می‌شوند.

$$x_{ZMP} = \frac{\sum_{i=1}^{n} x_{G_i} \times m_i(\ddot{z}_{G_i} + g) - \sum_{i=1}^{n} z_{G_i} \times m_i \ddot{x}_{G_i} - \sum_{i=1}^{n} \bar{I}_{y_i} \ddot{q}_{y_i}}{\sum_{i=1}^{n} m_i(\ddot{z}_{G_i} + g)}$$

$$(۱–۱)$$

$$y_{ZMP} = \frac{\sum_{i=1}^{n} y_{G_i} \times m_i(\ddot{z}_{G_i} + g) - \sum_{i=1}^{n} z_{G_i} \times m_i \ddot{y}_{G_i} - \sum_{i=1}^{n} \bar{I}_{x_i} \ddot{q}_{x_i}}{\sum_{i=1}^{n} m_i(\ddot{z}_{G_i} + g)}$$

دو شیوه کلی برای تولید مسیر حرکتی این نوع ربات‌ها وجود دارد که در ادامه توضیح داده خواهد شد. همچنین استفاده از مسیرهای حرکتی ضبط شده توسط انسان یا دیگر موجودات دارای پا یک راه‌حل دیگر برای این مسئله است که به علت محدودیت‌های آن کمتر مورد استفاده قرارمی‌گیرد.

---

[1] Walking Pattern Generation
[2] FZMP: Fictitious Zero Moment Point

## ۱) روش طراحی مسیر پارامتری

در این روش ابتدا یک مسیر حرکتی پارامتری برای نقاط کلیدی مانند پاها و بالاتنه درنظرگرفته‌می‌شود به نحوی که با مشخص بودن این مسیر، مسیر حرکت مفاصل قابل محاسبه باشد. سپس مقادیر عددی پارامترها را به نحوی می‌یابیم تا حرکتی را تولیدکنند که نقطه گشتاور صفر مجازی ناشی از آن همواره داخل ناحیه داخل تکیه‌گاهی قرار بگیرد(شکل ۱-۱۲).

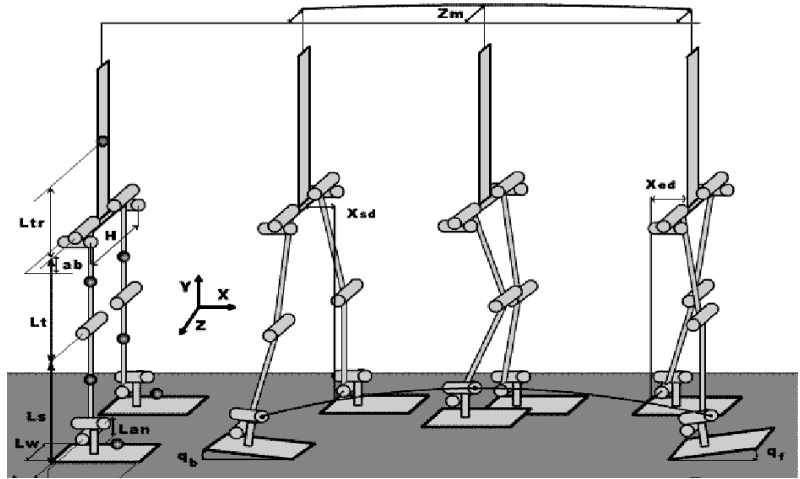

شکل ۱-۱۲. طراحی مسیر پارامتری حرکت ربات دوپا

## ۲) روش طراحی مسیر بر اساس مدل حرکتی

در این روش ابتدا یک مدل ساده برای حرکت ربات در نظرمی‌گیریم، به عنوان مثال یک مدل پاندول معکوس خطی برای حرکت بالاتنه درنظرگرفته‌می‌شود(شکل ۱-۱۳). سپس با درنظرگرفتن یک مسیر مطلوب برای نقطه گشتاور صفر مجازی، مسیر حرکت مرکز جرم، طوری طراحی می‌شود که نقطه گشتاور صفر مجازی حاصل از مجموع نیروهای وزن و اینرسی آن بر روی این مسیر قراربگیرد. در پایان، این مسیر طراحی شده برای حرکت مرکز جرم، توسط یک نگاشت بر روی مدل کامل یک ربات منطبق می‌شود و مسیر حرکت مفاصل به دست می‌آید.

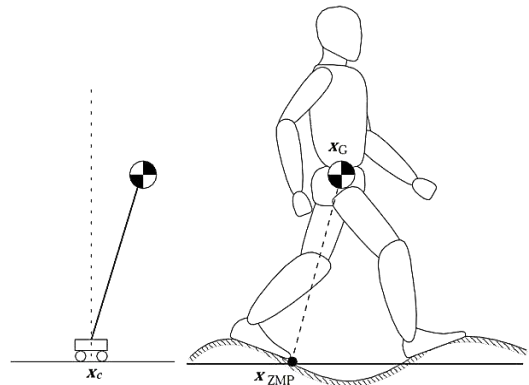

شکل ۱-۱۳. مدل ساده پاندول معکوس برای حرکت انسان

## ۳) نگاشت از فضای اندازه حرکت(موقعیت مرکز جرم و دوران) به فضای سرعت مفاصل(زوایای مفاصل)

علاوه بر داشتن مسیر حرکت مطلوب مرکز جرم بر حسب زمان، نیاز بـه تبـدیل آن بـه صـورت زوایـای مفاصل برای استفاده درکنترل‌کننده، خواهیم داشت. همواره رابطه مشخصی بین موقعیت مرکز جرم و زوایای مفاصل وجـود دارد. با درنظرگرفتن مشتق زمانی این روابط موقعیت، روابط سرعت، رابطـه‌ی جـامع بـرای تبـدیل انـدازه‌حرکـت کلی(خطی و زاویه‌ای) به سرعت مفاصل، توسط کاجیتا[1] در مرجع [۷] پیشنهاد شـده اسـت کـه در ایـن قسـمت بـه توصیف این کار می‌پردازیم. اگر ربات انسان‌نمایی با پیکـره‌بنـدی شـکل ۱-۱۴ را درنظربگیـریم، حرکـت ربـات بـه وسیله زوایای مفاصل ربات به همراه سرعت آن‌ها و سه مولفه بردار سرعت به همراه سه مولفه بردار سـرعت‌زاویـه‌ای یک عضو مبنای B (در اینجا عضو لگَن) قابل تعریف است. رابطه ورودی – خروجی این کمیت‌های حرکتـی بـا کمیت‌های حرکتی کلی که شامل سه مولفه اندازه حرکت خطی و سه مولفه اندازه حرکت زاویه‌ای می‌شـود توسـط رابطه (۱-۲) قابل بیان است. استخراج این رابطه می‌تواند به صورت بازگشتی از عضوهای انتهایی تا عضو مبنا و یـا از طریق گرفتن ژاکوبین رابطه اندازه حرکت انجام‌شود[۷]. در این حرکت هیچ قید حرکتی لحاظ‌نشده و بنابراین ربات در حالت پروازی(جدای از زمین) فرض‌شده‌است.

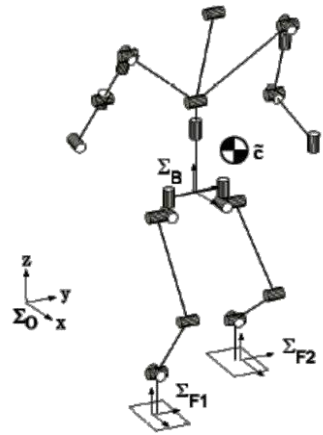

شکل ۱-۱۴. پیکره‌بندی کلی یک ربات انسان‌نما [۷]

$$\begin{bmatrix} L \\ H \end{bmatrix} = J(\theta) \begin{bmatrix} v_B \\ \omega_B \\ \dot{\theta} \end{bmatrix} = J_M(\theta) \begin{bmatrix} v_B \\ \omega_B \\ \dot{\theta}_{free} \end{bmatrix} + \sum_{i=1}^{2} J_{F,i}(\theta) \dot{\theta}_{leg,i} \qquad (۲-۱)$$

در این رابطه L اندازه‌حرکت خطی، H اندازه‌حرکت زاویه‌ای، $v_B$ و $\omega_B$ بردارهای سرعت و سـرعت زاویـه‌ای عضو مبنا و $\dot{\theta}$ بردار سرعت زاویه‌ای تمامی مفاصل ربات است. همچنین $J_M(\theta)$، یک ماتریس ژاکوبین است.

حال اگر بخواهیم معادلات قید حرکت (مسیر مطلوب برای سرعت‌های خطی و زاویه‌ای دو عضو کف پا) که به صورت رابطه (۳-۱) قابل بیان است، را وارد رابطه (۲-۱) کنیم،

---

[1] Kajita

$$\begin{bmatrix} v_{F_i} \\ \omega_{F_i} \end{bmatrix} = T_{F_i}(\theta) \begin{bmatrix} v_B \\ \omega_B \end{bmatrix} + J_{leg,i}(\theta)\, \dot\theta_{leg,i} , \qquad i = 1,2 \tag{۳-۱}$$

با جداسازی بردار سرعت مفاصل به صورت $\dot\theta = \begin{bmatrix} \dot\theta_{free} & \dot\theta_{leg,1} & \dot\theta_{leg,2} \end{bmatrix}$، که در آن $\dot\theta_{free}$ مربوط بـه مفاصل خارج از پاها (به غیر از عضو مبنا) هستند، و محاسبه $\dot\theta_{leg,i}$ با وارون‌گیری از رابطه (۳-۱)، به رابطه‌ای بـا فرم رابطه (۴-۱) خواهیم رسید:

$$\begin{bmatrix} L \\ H \end{bmatrix} = \tilde{J}_M(\theta) \begin{bmatrix} v_B \\ \omega_B \\ \dot\theta_{free} \end{bmatrix} + \sum_{i=1}^{2} \tilde{J}_{F,i}(\theta) \begin{bmatrix} v_{F_i} \\ \omega_{F_i} \end{bmatrix} \tag{۴-۱}$$

در نهایت با شبه‌وارون‌گرفتن از این رابطه به رابطه (۵-۱) خواهیم رسید کـه در صـورت افزونگـی درجـات آزادی، بینهایت جواب دقیق به ما می‌دهد. از تغییر بردار $z$ می‌توان برای بهینه‌سازی اهداف مختلفی از جمله نزدیـک کـردن مسیر حرکت مفاصل به یک مسیر دلخواه، استفاده‌نمود. این رابطه، $v_B$، $\omega_B$ و $\dot\theta_{free}$ را بـازای ورودی مطلـوب برای اندازه‌حرکت و حرکت کف پاها محاسبه می‌کنـد کـه بـا جایگـذاری در وارون رابطـه (۳-۱)، $\dot\theta_{leg,i}$ هـا نیـز محاسبه می‌شوند.

$$\begin{bmatrix} v_B \\ \omega_B \\ \dot\theta_{free} \end{bmatrix} = \tilde{J}_M^{\#} \left( \begin{bmatrix} L^{ref} \\ H^{ref} \end{bmatrix} - \sum_{i=1}^{2} \tilde{J}_{F,i} \begin{bmatrix} v_{F_i}{}^{ref} \\ \omega_{F_i}{}^{ref} \end{bmatrix} \right) + (E - \tilde{J}_M^{\#} \tilde{J}_M) z \tag{۵-۱}$$

## ۱-۵-۲ کنترل غیربه‌هنگام [1]

شیوه کنترل غیربه‌هنگام، در واقع، اساسا بازخورد [2] پایداری را در زمان حرکت در نظر نمی‌گیرد و بر این واقعیت استوار است که اگر مفاصل با دقت خوبی مسیر پایدار از پیش طراحی‌شده‌ای را (که حاشیه امنی را برای پایداری تضمین می‌کند،) دنبال کنند، متعاقبا ZMP ، رابطه (۱-۱)، نیز با دقت خوبی داخل ناحیه تکیه‌گاهی می‌افتد و پایداری حفظ خواهد شد.

برای تعقیب حرکت مفاصل دو روش کنترلی مفاصل مجزا (SJ) [3] و روش گشتاور محاسبه‌شده (CTM) [4] استفاده‌می‌شود. روش CTM با توجه به خطای ناچیز در تعقیب مسیر و مقاومت نسبتا بالا در برابر اختلالات ، روشی مناسب برای این شیوه از کنترل به حساب می‌آید ولی استفاده از روش SJ با توجه به سادگی در این شیوه کنترلی پرکاربردتر و متداول‌تر است که با توجه به زیاد بودن خطای تعقیب مسیر در آن ، تضمین ضعیفی را برای حفظ پایداری به دنبال دارد.

---

[1] Offline
[2] Feedback
[3] SJ: Separated Joint
[4] CTM: Computed Torque Method

## ۳-۵-۱ کنترل بههنگام[1]

این شیوه در واقع مبین کنترلکننده پایداری وضعی است که پیشخوراند از خطای پایداری در آن وجود دارد. معمولا از خطای موقعیت ZMP ، FRI و یا نیروهای تماسی برای کنترل پایداری ربات استفاده میشود. دو روش جبران کننده خطای نقطه گشتاور صفر[2] که با تغییر در مسیر حرکتی یک یا چند مفصل(مفاصل ران) انجام میشود و روش کنترل نیروهای تماسی[3] از عمدهترین روشها در این شیوه کنترلی است. یکی از کنترلکنندههای روش جبران کننده خطای نقطه گشتاور صفر که بسیار متداول و در عین حال ساده است، روش کنترل زوایه مطلق بالاتنه است. این روش بر این فرض مبتنی است که انحراف نقطه گشتاور صفر واقعی در مسیر مورد انتظار خود، بیش از هر چیز ناشی از حرکت بالاتنه است، زیرا این بخش از بدن، بیشترین جرم را داشته و همچنین در بالاترین ارتفاع قرارگرفته که با توجه به مدل پاندول معکوس، این دو ویژگی بیشترین تاثیر را در ناپایدار کردن حرکت خواهند داشت. بنابراین با کنترل بههنگام زاویه مطلق بالاتنه، میتوان تا حد زیادی، اختلاف بین نقطه گشتاور صفر واقعی(ZMP) با نقطه گشتاور صفر مجازی(FZMP) مورد انتظار را کم نمود و بنابراین پایداری را تا حد زیادی بهبود بخشید. اگرچه مبنای تئوری این روش، کامل و قدرتمند نیست ولی استفاده از آن در عمل بسیار موثر است.

## ۶-۱  مروری بر کارهای انجام شده مرتبط با موضوع پایاننامه

مطالعه در ارتباط با سه موضوع پایداری، طراحی مسیر و کنترل پایداری رباتهای راهرونده انجامشد که در این قسمت به بررسی مطالب اصلی ابتدا در مقالههای مرتبط با پایداری و سپس در مقالههای مرتبط با طراحی مسیر و کنترل پایداری خواهیم پرداخت.

مرجع [3] ابتدا به بررسی معیارهای متداول پایداری وضعی[4] و محدودیتها و نقصهای مفهومی آنها می-پردازد. برای این کار، معیار استاتیکی مرکز ثقل(CoG) و معیار دینامیکی نقطه گشتاور صفر(ZMP) را به عنوان دو معیار اصلی بررسی میکند. البته معیارهای دیگری چون شاخص دوران کف پا (FRI) که در منبع [5] معرفیشده-است و یا معیار نرخ اندازهحرکت صفر(ZRAM) که در منبع [6] معرفیشده است نیز میتوان در گروه همین معیارهای متداول پایداری وضعی از آنها نام برد ولی به دلیل خلاصهگویی در این مقاله بررسی نشدهاست. این نوشته پس از بررسی معیارهای متداول، به مسئله حاشیه پایداری رباتهای راهرونده میپردازد و با بهرهگرفتن از رابطه قیدهای تماسی شامل نیروی عمودی و نیروی اصطکاک در معادلات حرکت این نوع رباتها، به یک شرط کلی برای پایداری ربات میرسد. این شرط ساده در واقع همان قرارگیری بردار نیروهای تماسی در یک ناحیه مخروطی یا به طور ضعیفتر یک چندوجهی مخروطی است. سپس با بازنگری در مفهوم پایداری سیستمهای

---

[1] Online
[2] ZMP Compensator
[3] Foot/Contact Force Control
[4] Postural Stability

دارای پا، حاشیه این پایداری و استفاده از تئوری زیست‌پذیری، جایگزینی جدید برای معیار پایداری و حاشیه پایداری ارائه‌می‌دهد. همچنین در همین زمینه، نویسنده مرجع [۸] نیز با بررسی معیارهای پایداری و شرایط مورد نیاز برای یک معیار پایداری برای راهروونده‌ها، سعی می‌کند تا با استفاده از دیدگاه تئوری زیست‌پذیری، مرزی بر روی سرعت برای پایداری حرکت راه‌رفتن سریع پیدا کند. نویسنده این مرجع در کار دیگری در مقاله [۹]، نقطه یا ناحیه‌ای به نام نقطه یا ناحیه تسخیر[1] معرفی می‌کند که راه‌رونده با گذاشتن قدم در این ناحیه می‌تواند سرعت حرکت خود را به آرامی صفر کند و بایستد.

در مراجع [۱۰] و [۵]، مولفان برای اولین بار دو معیار پایداری وضعی ZMP و FRI را تعریف می‌کنند. این تعاریف در بخش‌های قبلی شرح داده شد. مرجع [۱۱] معیار ZMP را برای حالتی که ربات بر روی سطوح غیر مسطح حرکت‌می‌کند، تعمیم می‌دهد.

مرجع [۶]، ابتدا به تعریف معیاری جدید برای پایداری وضعی ربات انسان‌نما به نام ZRAM برحسب نرخ اندازه حرکت زاویه‌ای کل ربات و بررسی مزیت‌های آن نسبت به معیارهای ZMP و FRI می‌پردازد و سپس سه راه‌کار برای کنترل پایداری ربات با هدف صفرنگه‌داشتن نرخ اندازه حرکت زاویه‌ای کل ربات پیشنهاد‌می‌کند. اولین راه‌کار، تغییر در نحوه فرود آوردن پا برای افزایش سطح تکیه‌گاهی جهت دربرگرفتن نقطه ZRAM است. راه کار دوم، جابجایی مرکز جرم با توجه به مرکز فشار نیروی تکیه‌گاهی است و سومین راه‌کار پیشنهادی تغییر جهت نیروی تکیه‌گاهی است که با تغییر در بردار شتاب عضوهای ربات این‌کار انجام می‌شود. این روش با مشاهدات مرجع [۱۲] مطابقت دارد. مرجع [۱۳] نیز توجه ویژه‌ای به اندازه‌حرکت زاویه‌ای به عنوان معیاری برای پایداری دارد و مقدار اندازه حرکت زاویه‌ای در لحظه شروع فاز پروازی در حرکت دویدن را، معیاری مهم برای پایداری و پایدارسازی حرکت دویدن معرفی می‌کند.

مرجع [۱۲] که در واقع یک تحلیل حرکت‌شناسی بر روی اطلاعات ثبت‌شده از انسان به حساب‌می‌آید، یافته‌هایی جدید و در عین حال کاربردی ارائه‌می‌کند که نشان‌می‌دهد انسان حین حرکت راه رفتن مستقیم، اندازه‌حرکت زاویه‌ای‌اش را حول سه محور فضایی نزدیک به صفر تنظیم می‌کند. نکته دیگر آنکه امتداد نیروهای تماسی کف‌پا گذرنده از مرکز جرم، همواره از داخل ناحیه تکیه‌گاهی با زمین می‌گذرد. این روش به عنوان یک راهبرد کنترلی در مرجع [۶] برای کنترل نرخ اندازه حرکت زاویه‌ای پیشنهاد شده است. تحقیقات نویسنده مرجع [۱۲]، به وسیله نتایج به‌دست آمده توسط محققان دیگری در [۱۴] و در بعضی دیگر از تحقیقات که در ادامه از نتایج کلی آن‌ها استفاده خواهیم کرد، قابل تایید و استناد می‌باشد.

مرجع [۷] اگرچه در عنوان آن واژه کنترل دیده می‌شود ولی در واقع تنها یک فرمول‌بندی برای به دست آوردن سرعت مفاصل به عنوان خروجی برحسب اندازه‌حرکت خطی و زاویه‌ای مطلوب کل ربات و مسیرهای از پیش تعیین‌شده کف پاها(قیود هندسی) به عنوان ورودی ارائه می‌کند. این کار منجر به تولید مسیرهایی برای مفاصل

---

[1] Capture Point/Region

ربات بر حسب زمان به منظور تامین اندازه حرکت کلی ربات بر روی یک منحنی مطلوب می‌شود. در هر حال این مقاله به بررسی شرایطی که این منحنی مطلوب باید دارا باشد تا محدودیت‌های نیروهای تماسی ربات در شرایط واقعی تامین شود، نمی‌پردازد. همچنین کنترل پایداری ربات از نوع غیربه‌هنگام [1] است و به صورت کنترل‌کننده مجزا تنها وظیفه پیمودن مسیرهای محاسبه‌شده برای مفاصل را به عهده دارد و کنترل پایداری به صورت به‌هنگام [2] در آن دیده نمی‌شود. کار نسبتا مشابه دیگری که نگاشتی بین اندازه حرکت کلی و سرعت مفاصل پیدا می‌کند، کار محققان در مرجع [15] است که روش مذکور با تعریف ماتریس اندازه‌حرکت مرکزی انجام شده است. همچنین مرجع [16]، وظایف سینماتیکی و دینامیکی محول به یک ربات انسان‌نمای دارای افزونگی درجه آزادی را اولویت‌بندی می‌کند و بر اساس اولویت بندی، این وظایف را که بعضی در فضای مرکز جرم(اندازه حرکت) تعریف شده‌اند، در فضای مفاصل تصویر می‌کند که کاری متفاوت در زمینه نگاشت بین دو فضا محسوب می‌شود.

مرجع [4] به بررسی شیوه حفظ پایداری انسان در حالت ایستاده در مقابل اغتشاش ناشی از نیروی ضربه‌ای خارجی می‌پردازد و یک روش کنترل دو مرحله‌ای برای حفظ پایداری با توجه به مشاهدات واقعی از عکس‌العمل انسان (مبنی بر تلاش انسان درجهت افزایش اندازه حرکت زاویه‌ای در راستای ضربه ورودی برخلاف پیش بینی قابل تصور برای ما) ارائه‌می‌کند. این مقاله ابتدا با به نقدکشیدن معیارهای پایداری متداول، تعبیری همانند مقاله [3] برای پایداری ارائه می‌دهد که در قسمت‌های قبلی به آن اشاره شد. مدل حرکتی ربات ایستاده به صورت یک مجموعه چهار عضوی شامل کف پا، ساق پا، ران پا و بالاتنه فرض‌شده‌است. این روش، در مرحله اول، با استفاده از روش کنترل اندازه حرکت، مرکز جرم را به سمت مرکز فشار حرکت می‌دهد و پس از دفع ضربه در مرحله دوم، وضعیت ربات به آرامی به حالت ایستاده برمی‌گردد. مرحله دوم از طریق دو روش در جهت افزایش انرژی پتانسیل مکانیکی ربات انجام‌می‌شود.

نویسنده مراجع [17] و [18]، در هر دو کار خود، برای اولین بار مدل پاندول معکوس خطی را به ترتیب در حرکت صفحه‌ای و فضایی ربات راه‌رونده، به عنوان مبنایی برای طراحی مسیر و کنترل ربات قرار داده است. همچنین نویسنده [19]، مدلی نسبتا مبسوط برای ربات راه‌رونده به نام مدل پاندول معکوس با جرم انعکاسی [3] معرفی می‌کند که مدعی است رفتار راه‌رونده را بهتر نمایندگی می‌کند. این مدل از تعدادی جرم نقطه‌ای پراکنده در بالاتنه تشکیل شده است.

در همین زمینه، مرجع [20]، گزارش منتشرشده‌ای از کار محققان شرکت هوندا [4]، در سال ۲۰۰۹ است که نحوه تولید مسیر برای حرکت صفحه‌ای ربات آسیمو [5] را بررسی می‌کند. ابتدا یک مدل ساده برای حرکت ربات انسان نما در صفحه، به صورت یک سیستم سه جرمی ( بدنه و دو کف پا به صورت سه جرم نقطه‌ای) از یکدیگر مستقل

---

(فرض عدم وجود محدودیت بین حرکت این سه عضو) و فرض حرکت بدنه با مدل یک پاندول معکوس خطی، معرفی می‌گردد. سپس، معادله حرکت بدنه به صورت تابعی از مسیر مطلوب نقطه گشتاور صفر($ZMP_{desired}$) و حرکت کف پای متحرک در دستگاه چسبیده به کف پای تکیه‌گاهی نوشته‌می‌شود به طوری که با داشتن $ZMP_{desired}$ و حرکت کف پای متحرک بتوان مسیر حرکت بدنه ربات را محاسبه‌کرد. پس از آن با درنظرگرفتن یک حرکت ساده برای کف پای متحرک، یک پیشنهاد اولیه به صورت مسیری تکه‌تکه خطی و پیوسته برای $ZMP_{desired}$ ارائه‌می‌کند به نحوی که $ZMP_{desired}$ تقریبا همواره در وسط محدوده تکیه‌گاهی ربات قرارگیرد. در نهایت با مجزا کردن حرکت پاندولی بدنه به دو جزء همگرا و واگرا نسبت به زمان و اعمال روشی برای تغییر در منحنی $ZMP_{desired}$ جهت کنترل جزء واگرای حرکت بدنه به نحوی که نقطه انتهایی حرکت بدنه در هر سیکل بر نقطه ابتدایی حرکت بدنه در سیکل بعدی منطبق‌گردد، مسیر نهایی بالاتنه و به تبع آن مسیر مفاصل ربات برای حفظ نقطه ZMP بر روی مسیر اصلاح‌شده $ZMP_{desired}$ به‌دست‌می‌آید.

کار مولفان [۲۱] نیز، به عنوان نمونه‌ای از کنترل‌کننده بهینه پایداری(با استفاده از روش‌های عددی) یک دوپا در حالت ایستاده قابل بررسی است که سعی می‌کند همزمان با رعایت قید پایداری وضعی (محدوده مرکز فشار و مخروط اصطکاک)، اندازه‌حرکت کلی یک دوپا را صفر کند و یک مسیر برای نقاط انتهایی بدن را در حالت ایستاده به صورت حدودی تعقیب کند. با این‌حال، این روش تنها برای حرکت ایستاده و برای حفظ تعادل در هنگام حرکت اعضای غیرتکیه‌گاهی مانند سر و دست‌ها و یا در هنگام دریافت ضربه به‌کارگرفته شده است و برای حرکت‌هایی دینامیکی مانند راه‌رفتن راه‌کاری ارائه نمی‌دهد.

## ۷-۱  تعریف مسئله

در اکثر کارهای انجام‌شده تاکنون، طراحی کنترل‌کننده برپایه کنترل مسیر مفاصل بر روی یک مسیر حرکتی از پیش طراحی شده دارای پایداری وضعی و نهایتا جبران به‌هنگام پایداری وضعی حول این مسیر، استوار بوده‌است.

رسیدن به یک مفهوم تعمیم‌یافته و در عین‌حال ملموس برای پایداری کلی ربات‌های دارای پا به منظور ارائه روشی برای کنترل پایداری با میزان کنترل‌پذیری بالا برای هر شرایط اولیه دلخواه و مقاومت بالا در مقابل اختلالات داخلی و خارجی از جمله ضربه، ایده کلی این پایان‌نامه است. این پژوهش براساس قوانین کلی حاکم بر حرکت راه‌رفتن موجودات دوپا به خصوص انسان، مبنی بر نزدیک به صفر ماندن اندازه‌حرکت زاویه‌ای و نرخ آن بر حسب زمان که از اندازه‌گیری و ضبط داده‌های حرکت واقعی به دست آمده است، پایه‌ریزی خواهد شد. شیوه کنترلی موردنظر بر خلاف شیوه‌های معمول کنترل این ربات‌ها، مقید به الزام کنترل مفاصل(کمیتهای جزئی) بر روی یک مسیر از پیش‌طراحی‌شده در طول هرگام و تکرارپذیری دقیق حرکت‌ها نیست و در عوض کنترل کمیت‌های کلی از جمله اندازه‌حرکت خطی، اندازه حرکت زاویه‌ای و نرخ این دو را در کنار سازگاری هندسی قید(تماس پا یا پاها با زمین) نشانه‌گرفته است. این تحقیق هر سه چالش پایداری، مسیرحرکتی پایدار(سیکل‌های حرکتی) و

کنترل پایداری را دربرمی‌گیرد ولی در عین حال روش برخورد با این مسائل را از طریق کمیـت‌هـای کلـی ربـات پیگیری خواهد کرد.

## ۸-۱ روند تدوین پایان نامه

در فصل اول بعد از بیان مختصری از تاریخچه ربـات‌هـای دارای پـا، برخـی از کاربردهـای ایـن ربـات‌هـا و بـه‌خصوص ربات دوپا نام برده شد. به منظور آشنایی با ربات دوپا ابتدا مفـاهیم اولیـه آن بیـان گردیـد و پـس از آن بـه برخی از کارهای نظری و عملی انجام شده در زمینه پایداری و کنترل پایداری اشاره شد و بـا بسـط مفهـوم پایـداری کلی، چالش‌های اصلی این شاخه از رباتیک (ازجمله ارائه معیاری کلی برای پایـداری و کنتـرل پایـداری کلـی ایـن ربات‌ها)، مشخص گردید. سپس با ذکر برخی از تحقیقات مهم انجام شده تلاش‌هـایی کـه محققـان تـاکنون در ایـن زمینه انجام داده‌اند، بررسی شد. در پایان مساله‌ای که در این تحقیق به آن پرداختـه مـی‌شـود و روش حـل مـورد نظـر معرفی شد.

پس از این مقدمه، در ابتدای فصل دوم، معادله‌ای برای حرکت راهرفتن واقعی در فضای انـدازه‌حرکـت بـه نـام معادله حرکت(راهرفتن) پایه، استخراج می‌شود. در ادامه با ساده‌سازی آن به وسیله انجام فرضیاتی واقعـی، بـه معادلـه حرکتی ساده‌شده برای بررسی حرکت طبیعی(دینامیک صفر) راهرفتن بـه نـام معادلـه حرکـت گـام بـه‌گـام دسـت خواهیم یافت. معادله حرکت گام به‌گام، مبنایی برای استخراج انواع سیکل‌های حرکتی ساده و مرکب در ادامه ایـن بحث خواهد بود. سپس به بررسی پایداری حرکت حول سیکل‌هـای حرکتـی مـی‌پـردازیم و در پایـان بـا تحلیـل معادلات کامل حرکت راهرفتن و مقایسه آن با معادلات ساده‌شده، نگاهی سیستمی به معادلات حرکـت مـی‌انـدازیم که نتیجه آن جمع‌بندی معادلات ساده‌شده وکامل حرکت در قالب دو مدل ریاضی جامع با نام‌های مـدل سـاده‌شـده راه رفتن(SWM: Simplified Walking Model) و مـدل کامـل راهرفتـن(CWM: Complete Walking Model) خواهد بود.

در فصل سوم، توانایی هدایت حرکت و چگونگی هـدایت حرکـت راهرونـده بـه سـمت یـک سـیکل حرکتـی مطلوب و کنترل حول آن، مبنایی برای پایداری و طراحی کنترل‌کننده پایداری قرارمی‌گیرد. سپس با بررسی مـدل‌های ساده(SWM) و کامل(CWM) راهرفتن، با توجه به سـادگی و کـاهش‌یافتـه‌بـودن مـدل SWM و در عین حـال نزدیکی آن به واقعیت حرکت، این مدل مبنایی برای طراحی کنترل‌کننده‌های پایداری قرار می‌گیرد کـه از آن‌هـا بـا عنوان پایدارسازهای سیکل حرکتی نام می‌بریم. در ادامه فصل سوم، چهار پایدارساز معرفی خواهند شـد و پایـداری هر یک اثبات می‌شود، همچنین، توانایی‌های هر کدام از لحاظ تئوری بررسی خواهـد شـد. در پایان این فصـل بـا معرفی یک مدل فیزیکی کامل و قیود آن، به انجام شبیه‌سازی بر روی این مدل فیزیکی که معرف مـدل کامـل راه‌رفتن است خواهیم پرداخت و توانایی و نارسایی کنترل‌کننده‌هـا را در پایدارسـازی شـرایط اولیـه دور از سـیکل و همچنین در پایدارسازی مجدد پس از دریافت ضربه، بررسی خواهیم کرد.

در فصل چهارم، به معرفی روشی برای کنترل پایداری با استفاده از برنامه‌ریزی غیرخطی خواهیم پرداخت کـه بـا توجه به عملکرد کنترلی پایدارسازهای سیکل حرکتی در فصل سوم، عملکردی مشابه را برای محاسـبه ورودی‌هـای کنترلی با کمینه‌کردن شاخص پایداری، پایه‌ریزی می‌کند. این روش علاوه بر پایدارسازی، قیود واقعی مسـئله را نیـز در نظرمی‌گیرد. در پایان این فصل با انجام شبیه‌سازی بر روی مدل کامل مشاهده خواهد شد که این روش مـی‌توانـد پایداری را با قدرت بالایی حفظ کند و در عمل همه قیود مسئله را نیـز رعایـت کنـد. ایـن روش، راه حلـی کامـل و عملی برای مسئله کنترل پایداری ارائه می‌دهد که دیگر هیچ‌یک از نارسـایی‌هـای پایدارسـازهای سـیکل حرکتـی را نخواهد داشت.

نهایتا در فصل پنجم به جمع‌بندی و نتیجه‌گیـری از تحقیـق پرداختـه مـی‌شـود و نـوآوری‌هـا و دسـتاوردهای ایـن پژوهش بررسی می‌شود و در پایان پیشنهاداتی جهت ادامه و توسعه کار ارائه می‌گردد.

**فصل دوم**
**مدل دینامیکی حرکت راهرفتن**

تاکنون مدلهای مختلفی برای حرکت راهرونده دوپا پیشنهاد شده است کـه برخـی بـه طـور کامـل و برخـی بـه صورت سادهشده، حرکت این سیستم را بیان میکنند. در این تحقیق راهرونده بـه صـورت یـک سیسـتم کلـی دیـده میشود تا در نهایت بتوانیم به تحلیلی جامع از حرکت راهرونده بدون ورود به بحث در مورد کمیتهـای داخلـی آن مانند متغیرهای حرکت مفاصل برسیم.

از جمله روشهای پایدارسازی حرکت رباتهای راهرونده و کنترل پایداری این دسته از رباتها، کنترل انـدازه-حرکتهای خطی و زاویهای آنها است که تاکنون به صورت جامع به آن پرداخته نشده است. در این پژوهش بررسی پایداری و کنترل پایداری رباتهای راهرونده دوپا، با تکیه بر تحلیل رفتار این رباتها در فضای اندازه حرکت کلی[1] انجام میگیرد. در این تحلیل از پایدارسازی ربات، بیش از آنکه و پیش از آنکه توجه به حرکت راه-رونده در فضای مفاصل باشد، حرکت راهرونده در فضای اندازهحرکت کلی آن، مورد توجه قرار میگیرد و در حقیقت حرکت پایدار ربات از طریق کنترل اندازهحرکتهای خطی و زاویهای انجام میگیرد. در فصل حاضر، مدلسازی رفتار دینامیکی ربات در این فضا و در فصل بعد، راهبردهای کنترلی برای کنترل پایداری راهرونده از طریق کنترل اندازه حرکتهای کلی آن و متغیرهای موثر بر آن مورد بررسی قرار میگیرد.

---

[1] Total Momentun Space

در این فصل ابتدا معادله حرکت حاکم بر مدل دوبعدی رباتهای دارای پا را استخراج می‌کنیم. سپس با درنظرگرفتن چند فرض ساده کننده برای حرکت یک راه رونده دوپا، معادله حرکت آن را برای یک گام محاسبه خواهیم کرد و در نهایت معادلات حرکت گام به گام را بدست می‌آوریم و به طرح ایده سیکل‌های حرکتی و بررسی انواع سیکلهای حرکتی ساده و مرکب، محدودیت‌ها و میزان پایداری آنها با استفاده از تحلیل ریاضی و بررسی رفتاری آنها در صفحه فازی خواهیم پرداخت. همچنین در بخش آخر برای دست یافتن به دیدگاهی کامل‌تر به مقایسه نتایج معادله حرکت ساده شده با نتایج معادله حرکت کامل برای راه رونده دوپا می‌پردازیم.

در پایان این فصل، با جمع‌بندی تحلیل‌های حرکت راه‌رفتن و تجزیه و تحلیل رفتاری و مقایسه آنها با یکدیگر، به ارائه یک مدل ریاضی کامل از حرکت راه‌رفتن ( CWM : Complete Walking Model ) و یک مدل ریاضی ساده شده از حرکت راه‌رفتن ( SWM : Simplified Walking Model ) برای راه‌رونده دو پا می‌پردازیم که از آن به عنوان مبنایی برای طراحی کنترل‌کننده در فصل بعدی استفاده خواهیم کرد.

## ۱-۲  استخراج معادلات حرکت راه‌رفتن در فضای اندازه حرکت

اگر دو دسته حرکت دینامیکی کلی به نامهای حرکت راه‌رفتن و حرکت دویدن(حرکت جهش یا پریدن نیز می‌تواند در این دسته قراربگیرد.)، برای یک دوپا فرض کنیم، یک فرض اساسی در حرکت راه‌رفتن، وجود دائمی تماس حداقل یک نقطه بین کف پا و زمین می‌باشد، در صورتی که برای دویدن این فرض در بخشی از حرکت معتبر است. برای استخراج معادلات حرکت راه‌رفتن با درنظر گرفتن این فرض اساسی ، مدلی شبیه شکل ۲-۱ درنظرمی گیریم که نمادهای آن طبق جدول ۲-۱ تعریف شده باشند.

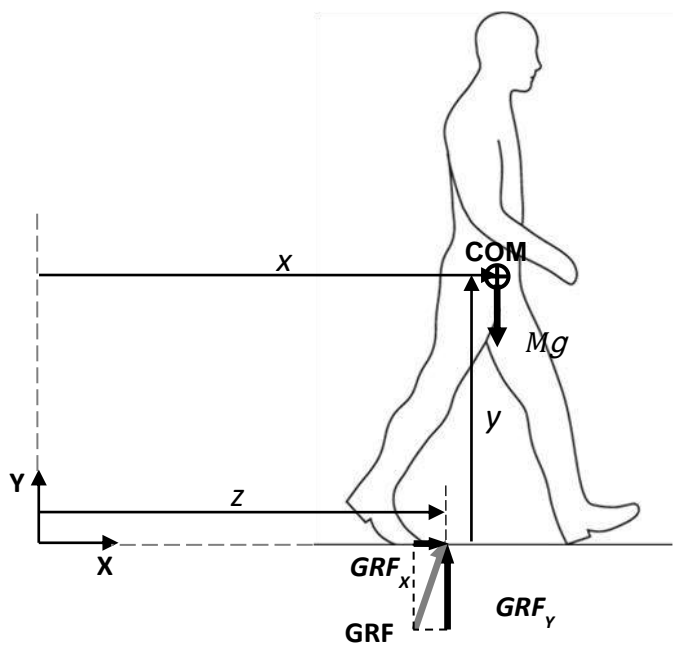

شکل ۲-۱. مدل حرکتی ربات دارای پا

جدول ۲-۱. تعاریف مربوط به مدل حرکتی ربات

| | |
|---|---|
| $CoM(Center\ of\ Mass)$ | مرکز جرم کل بدن |
| $GRF \triangleq Ground\ Reaction\ Force$ | برآیند نیروهای خارجی وارد بر کل بدن از طرف سطح زمین |
| $CoP(Center\ of\ Pressure)$ $or$ $ZMP(Zero\ Moment\ Point)$ | مرکز فشار نیروهای خارجی وارد بر کل بدن از طرف سطح زمین یا نقطه گشتاور صفر واقعی که $GRF$ در آن عمل می‌کند |
| $x \triangleq x_{CoM}$ | موقعیت مرکز جرم کل بدن در راستای محور X |
| $y \triangleq y_{CoM}$ | موقعیت مرکز جرم کل بدن در راستای محور Y |
| $z \triangleq x_{CoP} \equiv x_{ZMP}$ | موقعیت نقطه گشتاور صفر یا مرکز فشار در راستای محور X |
| $M$ | جرم کل بدن |
| $g$ | شتاب جاذبه زمین |

با نوشتن معادلات نیوتن به روابط (۲-۱) خواهیم رسید که معادله حرکت (۲-۲) در فضای اندازه حرکت و یا معادله (۲-۳) در فضای اندازه حرکت بدون جرم (یا فضای مرکزجرم) را نتیجه می‌دهد و کمیتهای آن طبق جدول ۲-۱ تعریف می‌شوند. این رابطه را معادله حرکت پایه (برای حرکت راه‌رفتن) نامگذاری می‌کنیم که برای همه راه‌روندها از جمله دوپای راه‌رونده شکل ۲-۱ معتبر است. باید درنظر داشت که علاوه بر حرکت راه‌رفتن، این معادله حرکت برای بخش‌هایی از حرکت دویدن که پای دونده با زمین در تماس است نیز، معتبر می‌باشد.

$$\begin{cases} \sum F_X = GRF_X = \dot{L}_X = M\ddot{x} \\ \sum F_Y = GRF_Y - Mg = \dot{L}_Y = M\ddot{y} \\ \sum T_{CoM} = y * GRF_X - (x - z) * GRF_Y = \dot{H} \end{cases} \qquad (۲-۱)$$

$$z = x + \frac{\dot{H} - y\dot{L}_X}{Mg + \dot{L}_Y} \qquad (۲-۲)$$

$Principal\ Equation$
$of$ $\quad : \quad z = x + \dfrac{\dot{H}/M - y\ddot{x}}{g + \ddot{y}} \qquad (۲-۳)$
$(Walking)\ Motion$

معادله حرکت پایه رابطه (۲-۳) به خوبی ارتباط بین مرکز فشار، موقعیت و شتاب مرکز جرم کل بدن و همچنین اندازه‌حرکت زاویه‌ای (بدون جرم‌شده) حول مرکز جرم را نشان می‌دهد به این معنی که اگر موقعیت مرکز فشار به عنوان ورودی این معادله حرکت مشخص باشد، موقعیت مرکز جرم و گردش کلی ربات با این رابطه نسبت به هم تغییر می‌کنند. بطور معکوس، اگر بخواهیم موقعیت مرکز جرم و گردش کلی ربات مسیرهای مشخصی را طی کنند، مسیر مشخصی برای مرکز فشار به دست‌می‌آید که برای انجام‌پذیر بودن این حرکت، ملزم به فراهم کردن تماس یا

حداقل یک نقطه از ربات(پای ربات) با سطح زمین بر روی این مسیر هستیم که این رابطه معکوس برای بحث طراحی مسیر پایدار کاربرد دارد و استفاده از آن در این تحقیق موضوعیتی ندارد.

جدول ۲-۲. تعریف کمیت‌های معادلات حرکت

| $\dot{L}_X$ | اندازه حرکت خطی کل بدن در راستای محور X |
|---|---|
| $\dot{L}_Y$ | اندازه حرکت خطی کل بدن در راستای محور Y |
| $\dot{H}$ | اندازه حرکت زاویه‌ای کل بدن حول مرکز جرم |
| $\sum F_X$ | برآیند کل نیروهای خارجی وارد بر بدن در راستای محور X |
| $\sum F_Y$ | برآیند کل نیروهای خارجی وارد بر بدن در راستای محور Y |
| $\sum T_{CoM}$ | برآیند کل گشتاورهای حاصل از نیروهای خارجی وارد بر بدن در نقطه مرکز جرم |

## ۲-۲  الگوی گام برداشتن و محدوده مرکز فشار

از آنجا که گام برداشتن مهم‌ترین بخش از حرکت یک راه‌رونده و زیربنای مباحث بعدی در بیان مـدل حرکتـی راه‌رفتن، مانند پایداری و کنترل پایداری است، در این قسمت سعی می‌کنیم تا بـه مبحـث گـام‌برداشتن و چگـونگی ارتباط آن با مباحث بعدی بپردازیم.

با دقت در فرایند راه‌رفتن موجودات زنده دیده می‌شود این راه‌رونده است که تصمیم می‌گیرد با برداشتن گـام‌های متوالی از طریق دو روش، ۱) برنامه‌ریزی طول گام و زمان مناسب فـرود آمـدن پـای متحـرک، ۲) نحـوه فـرود آوردن پای متحرک، نحوه برداشتن پای تکیه‌گاهی و برهم‌زمانی این دو، ناحیه تکیه‌گاهی حرکت خود یـا محـدوده مرکز فشار را چگونه تغییر دهد تا حرکتش را تحت کنترل درآورد و علاوه بر این دو، با روش ۳) جابجایـی مرکز فشار در ناحیه تکیه‌گاهی ایجاد شده، کنترل دقیقتری بر روی حرکت خود انجام‌دهد. از این میان، روش(۲) کـه نـوع برخورد و جدا شدن پاها از زمین را نمایندگی می‌کند، نقشی ناچیز در تصمیم‌گیـری راه‌رونـده دارد و بیشـتر بـه نـوع تماس پا با سطح زمین و ساختار کف پا که با مشخصات تغییرناپذیر یک راه‌رونده است، بستگی دارد و از این‌رو در کنترل پایداری کمتر مورد توجه است. از منظر ساختار کف پا که بر نحوه گام‌برداشتن تاثیر می‌گذارد، می‌تـوان راه‌روندها را به سه دسته کلی تقسیم‌بندی کرد و ناحیه تکیه‌گاهی قابل دسترس برای مرکز فشار آنها را بررسی نمود:

دسته اول – راه روندهای بدون کف پا که دارای تماس نقطه‌ای با زمین هستند. این راه روندهـا بـا نـام دوپـای کم‌عملگر[1] شناخته می‌شوند که به دلیل کم‌عملگربودن، هیچ کنترل‌کننده‌ای نمی‌تواند پایداری وضعی آنها را بـرای تعقیب یک مسیر از پیش‌طراحی شده، به طور کامل تضمین کند و تنها می‌توان از همگرایی حرکت آنهـا بـه سـمت

---

[1] Underactuated Biped

سیکل‌های حدی سخن گفت که زمینه‌ای برای محققان سیستم‌هـای غیرخطـی اسـت. شـکل ۲-۲، نمـودار محـدوده مرکز فشار قابل دسترس را برای بازه زمانی دو گام نمایش می‌دهد که حرکت راه‌رونده برای بازه زمانی گام اول آن به تصویر کشیده شده است.

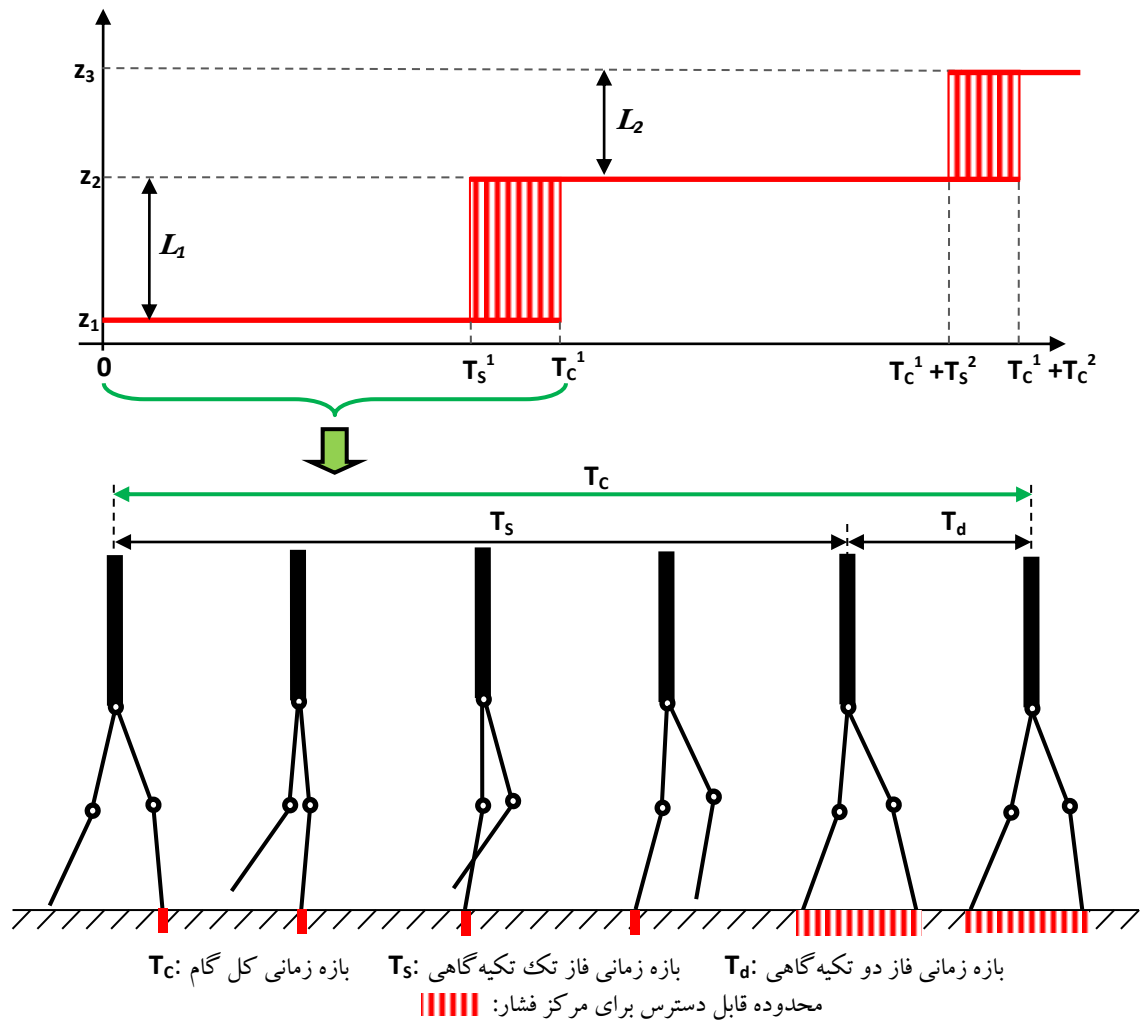

بازه زمانی فاز دو تکیه‌گاهی $T_d$:     بازه زمانی فاز تک تکیه‌گاهی $T_s$:     بازه زمانی کل گام $T_c$:

محدوده قابل دسترس برای مرکز فشار: ⦀

شکل ۲-۲. حرکت راه روندهای دوپای کم‌عملگر

دسته دوم — راه روندهها دارای کف پای مسطح هستند که می‌توانند تماس نقطه‌ای یا سطحی با زمین داشته-باشند. این راه روندهها می‌توانند یا بدون استفاده از حرکتی دورانی عضـو کـف پـا، نسـبتا شـبیه بـه دسـته اول حرکـت کنند (رباتهای دارای پاهای مجهز به مکانیزم موازی دومیله‌ای دارای چنین رفتاری هسـتند) و یـا اینکـه هماننـد دسته سوم مانند انسان از حرکت چرخشی کف پا در هر گام استفاده‌کننـد(رباتهای دارای پاهـای مجهـز بـه مفاصـل مجـزا دارای چنین رفتاری هستند) که بسته به نحوه گام برداشتن آنها، ناحیه تکیه‌گاهی متفاوتی را ایجاد مـی‌کننـد. تفـاوت نحوه گام برداشتن در این دو حالت در شکل ۲-۳ و شکل ۲-۴ به طور جداگانه بررسی شده است. در هر یک از این شکل‌ها، محدوده مرکز فشار قابل دسترس برای بازه زمانی دو گام و حرکت راه‌رونده برای بازه زمانی گـام اول آن به تصویر کشیده شده است.

۳۳

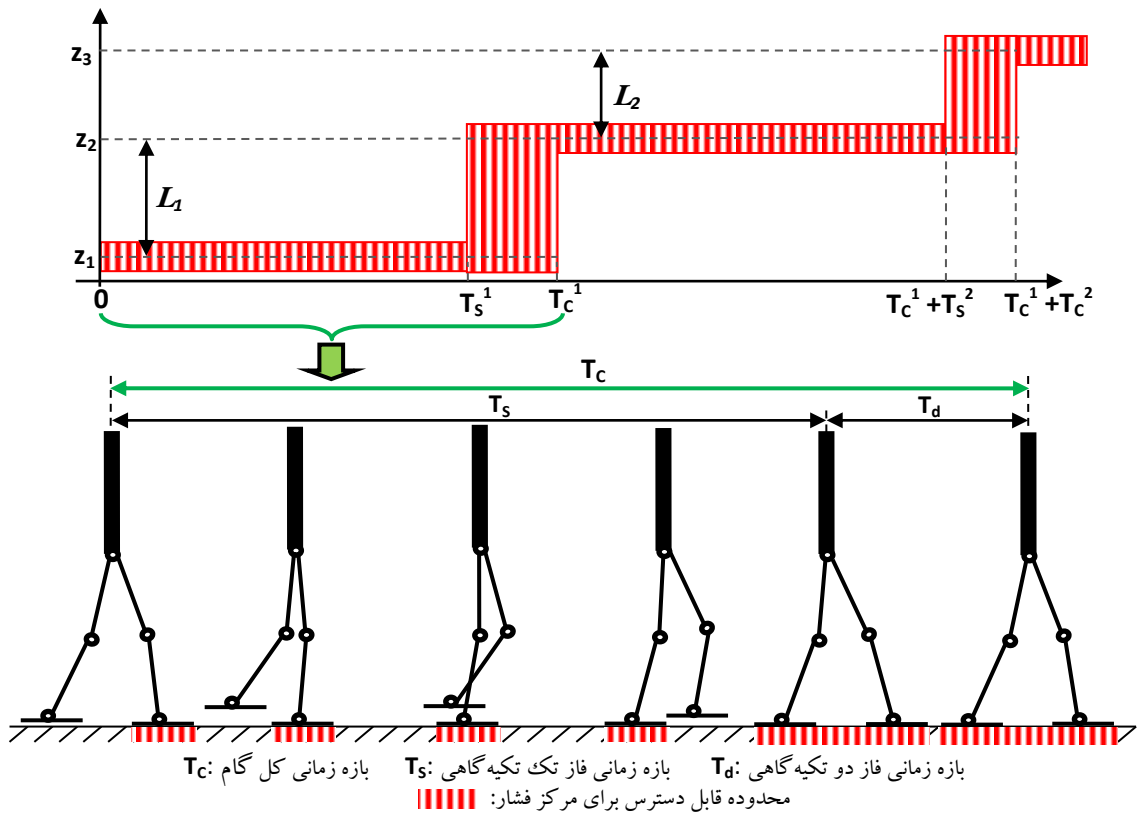

شکل ۳-۲. حرکت راه رونده دوپا با کف پای مسطح بدون چرخش کف پا

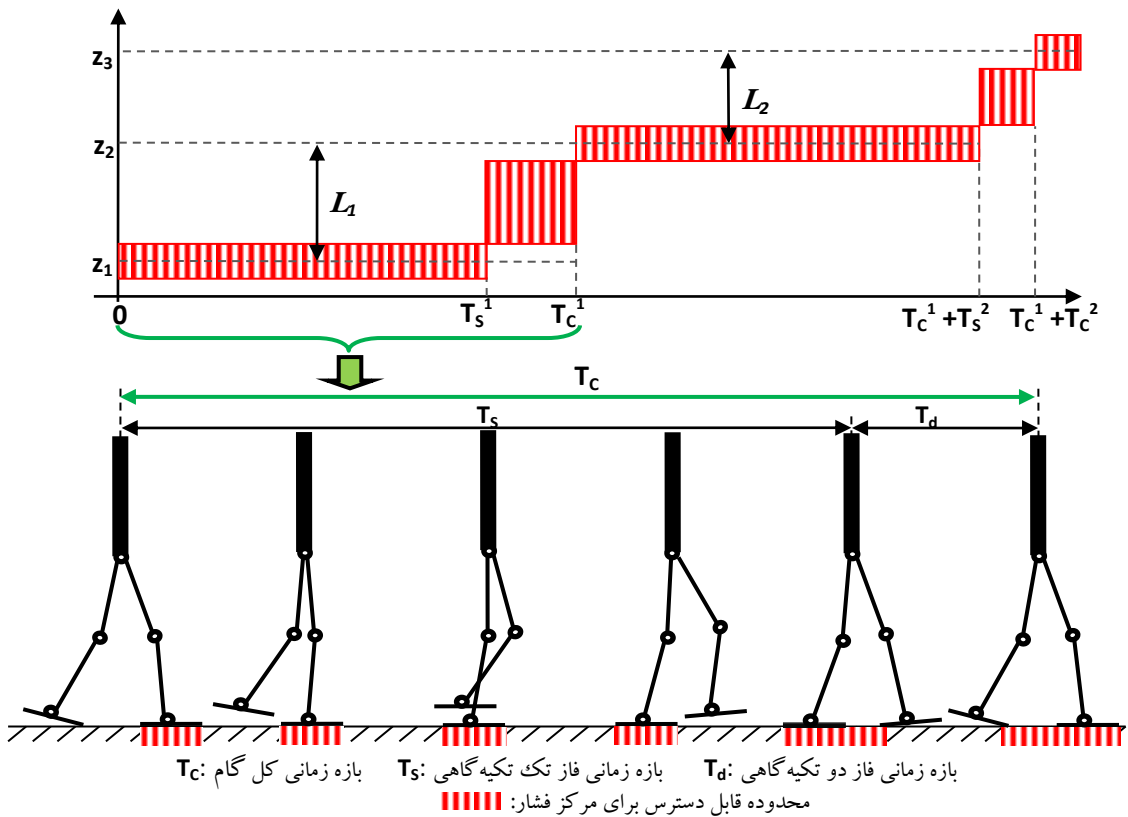

شکل ۴-۲. حرکت راه رونده دوپا با کف پای مسطح با چرخش کف پا

دسته سوم — راه رونده های دارای کف پای دوقسمتی دارای پنجه و پاشـنه هستند کـه بهترین نمونه آن انسـان است. شکل ۵-۲، نمودار محدوده مرکز فشار قابل دسترس را برای بازه زمانی دو گام نمـایش مـی‌دهـد کـه حرکت راه‌رونده برای بازه زمانی گام اول آن به تصویر کشیده شده است.

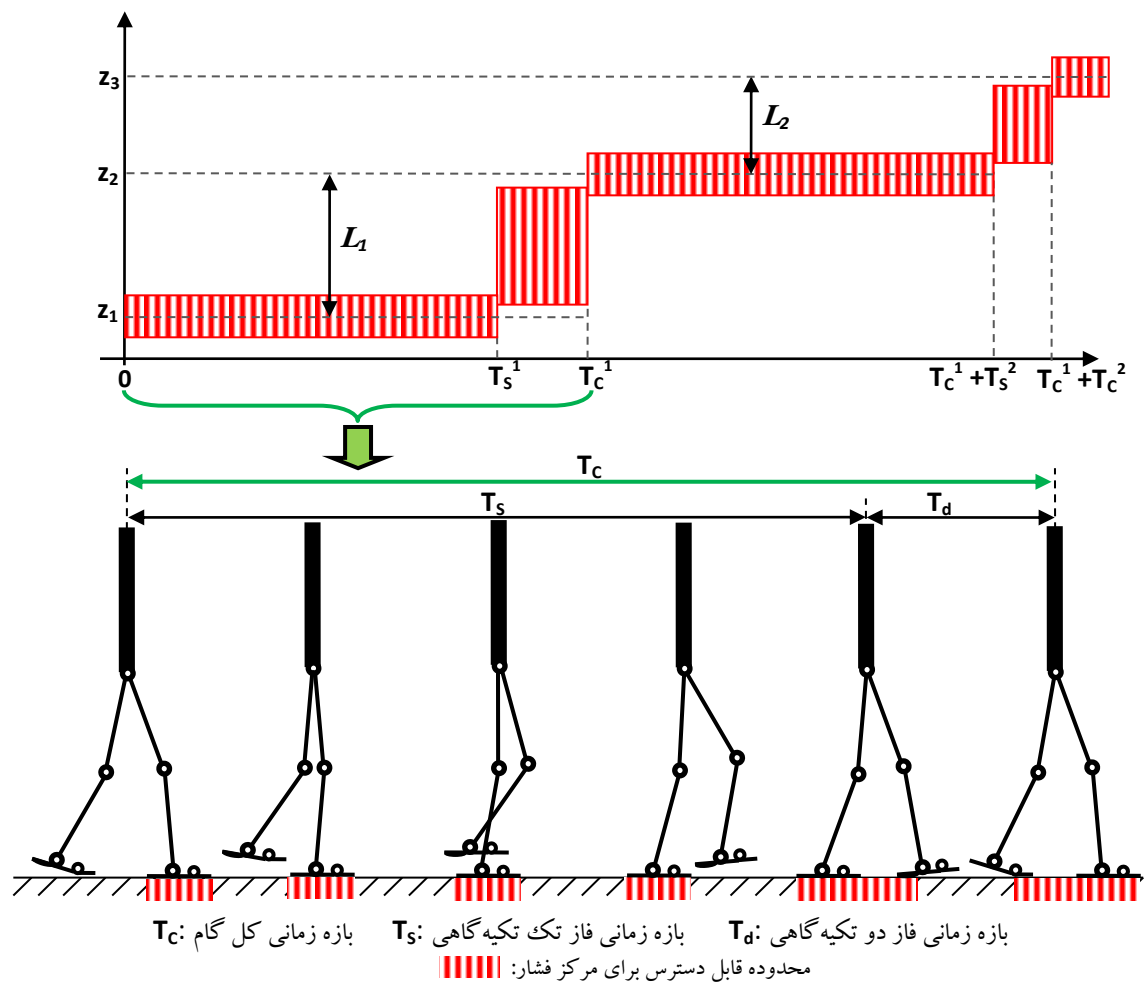

بازه زمانی فاز دو تکیه‌گاهی T_d:    بازه زمانی فاز تک تکیه‌گاهی T_s:    بازه زمانی کل گام T_c:
محدوده قابل دسترس برای مرکز فشار:

شکل ۵-۲. حرکت راه رونده دوپا با کف پای دوقسمتی دارای پنجه و پاشنه

از آنجا که در بازه زمانی فاز تک تکیه‌گاهی که حدود ۸۰٪ کل بازه زمانی یک گام را شـامل مـی‌شـود، قابلیـت تغییر موقعیت مرکز فشار تنها به کف یک پا محدود می‌شود، استفاده از روش (۳) برای کنترل حرکت ربات در ایـن بازه دارای محدودیت بسیار است. همچنین در بازه زمانی فاز دو تکیه‌گاهی که تنها ۲۰٪ کـل بـازه زمـانی یـک گـام است نیز، توانایی کنترل حرکت بر اساس تغییر مرکز فشار محدود است. در این بازه بـا وجـود اینکـه ناحیـه قابـل دسترس برای مرکز فشار محدوده بزرگتری را شامل می‌شود، بازه زمانی کوتـاه اسـت و توانـایی جبـران پایـداری از دست رفته که احتمالا در فاز قبلی اتفاق افتاده، محدود است. به همین علت، روش (۱) که اساسا به جای تغییر مرکـز فشار در ناحیه تکیه گاهی، به دنبال جابجایی خود ناحیه تکیـه گـاهی اسـت ، گزینـه مـؤثرتری بـرای کنتـرل پایـداری محسوب می‌شود. روش (۱) می‌تواند آثار عوامل ناپایدارکننده‌ای را که در بازه زمانی یک گام ایجاد شده است، در لحظه فرود آمدن پا جبران کند و یک راهبرد برتر برای کنترل پایداری محسوب می‌شود.

با این حال کارهایی که تاکنون برای طراحی مسیر پایدار طراحی انجام شده است و برخی در فصل دوم توضیح داده شد، بیشتر بر مبنای روش (۳) شکل گرفته است به این معنی که طول گام و زمان گام برداشتن، بدون توجه به تأثیر آن در پایداری و تنها با توجه به مشخصههای حرکتی مطلوب نظیر سرعت مطلوب و طول گام مطلوب درنظرگرفته می‌شود و سپس مسیر حرکت بدن طوری طراحی میشود که مسیر حرکت مرکز فشار یا داخل محدوده مجاز قرارگیرد [۲۲] و [۲۳] و یا دقیقا بر روی یک مسیر مطلوب داخل محدوده مجاز حرکت کند[۲۴]، [۲۵]، [۲۶] و[۲۷]. اخیـرا کارهایی نیز انجام شده است که با وجود محور قراردادن (۳) نگاهی نیـز بـه روش (۱) دارد. بـه عنـوان مثال [ ۲۰] و [۲۸] به دو شیوه کنترل عددی متفاوت طول گام را به گونهای طراحی میکننـد کـه مسـیر حرکتـی شـده در نهایت پایدار باشد.

## ۲-۳  مدل حرکتی ساده شده برای راهرفتن

در این قسمت فرض چهار فرض برای سادهسازی معادله حرکت پایه رابطه (۲-۳) در حالت راه رفتن درنظر می‌گیریم تا معادلات حاکم بر حرکت راه رفتن را به نحوی بدستآوریم که بتوان در تحلیلهای آتی از آن استفاده نمود.

**فرض اول**- با توجه به حرکت انسان و دیگر راه روندهها، ارتفاع مرکز جرم در حرکت راهرفتن با تقریب بسیار خوبی ثابت است. بنابراین موقعیت عمودی مرکز جرم در حرکت راه رفتن را طبق رابطه (۲-۴) ثابت میگیریم.

*Assumption* 1: $y = cte = h$
<div align="left">(۲-۴)</div>

**فرض دوم**- با توجه به فرض اول، ارتفاع مرکز جرم با تقریب بسیار خوبی ثابت است. همچنین در حرکت راه رفتن با توجه به سرعت کم نسبت به حرکت دویدن، با توجه به دامنه ناچیز تغییرات جزئی ارتفاع و کوچک بودن فرکانس تناوب، میتوان از شتاب عمودی مرکز جرم در حرکت راه رفتن در نسبت با شتاب جاذبه زمین طبق رابطه (۲-۵) صرفنظر کرد. این فرض با توجه به اندازهگیریهای انجامشده از حرکت انسان در [۲۹] و [۱۲] نیز قابل استناد است.

*Assumption* 2: $\ddot{y} \approx 0 \Rightarrow z = x - \dfrac{h}{g}\ddot{x} + \dfrac{\dot{H}}{Mg}$
<div align="left">(۲-۵)</div>

**فرض سوم**- با توجه به محدودیتهای ساختاری راهروندهها برای کنترل اندازهحرکت زاویهای، همواره اندازه حرکت زاویهای و نرخ آن نزدیک به صفر کنترل میشوند. همچنین با توجه به اندازهگیریهای انجام شده بر روی حرکت انسان نرخ اندازهحرکت زاویهای در حرکت راهرفتن انسان، همواره نزدیک به صفر است[۱۲] ، [۳۰] ، [۳۱] و [۳۲]. بنابراین میتوان با این فرض رابطه (۲-۶) را بهدست آورد.

*Assumption* 3: $\dot{H} \approx 0 \Rightarrow z = x - \dfrac{h}{g}\ddot{x}$
<div align="left">(۲-۶)</div>

**فرض چهارم**- با توجه به اینکه مرکز فشار می‌تواند در هر نقطه‌ای از محدوده مجاز قرار بگیرد و همچنین با توجه به اینکه در این تحقیق محدوده بیشتر از راهبرد برتر تغییر طول و زمان فرود گام‌های متوالی به جای راهبرد تغییر دادن مرکز فشار داخل محدوده مجاز استفاده می‌شود، فرض می‌کنیم در فاز تک تک تکیه‌گاهی مرکز فشار، $z$، در مرکز هندسی کف پای تکیه‌گاهی، $z_i$، قرارگیرد و در فاز دوتکیه‌گاهی که بلافاصله بعد از آن آغاز خواهد شد، مرکز هندسی پای دیگر(پای متحرک)، به عنوان مرکز فشار در نظر گرفته‌شود(در واقع وزن بدن در همان لحظه فرود پای متحرک بر روی آن منتقل شود)، آنگاه همواره به اندازه طول زمانی یک گام، مرکز فشار ثابت خواهد بود به این معنی که برای هر گام فرضی که از ابتدای فاز دو تکیه‌گاهی شروع می‌شود و تا پایان فاز تک تک تکیه‌گاهی بعد از آن ادامه می‌یابد مرکز هندسی یک کف پا، $z_i$، مرکز فشار، $z$، خواهد بود و بنابراین رابطه (۲-۷) را می‌توان فرض کرد که در آن $i$ اندیس گام و مربوط به گام $i$ اُم است.

*Assumption 4:* $\quad z = z_i = cte \quad \Rightarrow \quad \dot{z}_i = 0, \ddot{z}_i = 0$ $\qquad\qquad$ (۲-۷)

اگرچه برای راهروونده‌های داری کف پای غیرنقطه‌ای، درعمل می‌توان جابه‌جایی به اندازه $\Delta z_i$، بین مرکز فشار، $z$، و مرکز هندسی کف پا، $z_i$، درنظرگرفت که در آن صورت، $z = z_i + \Delta z_i$ در هرگام خواهد بود، ولی برای ساده‌سازی راهروونده‌های دارای کف پای غیرنقطه‌ای، برای آن‌ها نیز، فعلا فرض می‌کنیم $\Delta z_i = 0$ باشد.

با تعریف پارامتر $\omega$ و تعریف متغیر $s_i$ که فاصله مرکز جرم، $x$، تا مرکز تکیه‌گاه، $z_i$، (که در اینجا همان مرکز فشار فرض شده است) را نمایندگی می‌کند به صورت زیر:

$$\begin{cases} (1) \ \omega \triangleq \sqrt{g/h} & \ddot{x} - \omega^2(x - z_i) = 0 \\ & \qquad\qquad \downarrow \\ (2) \ s_i \triangleq x - z_i & \ddot{s}_i - \omega^2 s_i = 0 \ , \ \dot{s}_i \triangleq \dot{x} \ , \ \ddot{s}_i \triangleq \ddot{x} \end{cases} \Rightarrow \qquad (۲-۸)$$

معادله حرکت و بیان حالتِ آن به صورت روابط (۲-۹) قابل بازنویسی است.

$$\ddot{s}_i - \omega^2 s_i = 0 \xRightarrow{\substack{STATE \\ SPACE}} \begin{bmatrix} \dot{s}_i \\ \ddot{s}_i \end{bmatrix} = \begin{bmatrix} 0 & 1 \\ \omega^2 & 0 \end{bmatrix} \begin{bmatrix} s_i \\ \dot{s}_i \end{bmatrix} \qquad\qquad (۲-۹)$$

حال اگر دو متغیر سیستم را با تعریف نگاشت تبدیل (۲-۱۰) قطری‌سازی کنیم به رابطه نهایی(۲-۱۱) خواهیم رسید.

$$\begin{bmatrix} p_i \\ q_i \end{bmatrix} \triangleq T \begin{bmatrix} s_i \\ \dot{s}_i \end{bmatrix} \ , \ T = \begin{bmatrix} 1 & -1/\omega \\ 1 & 1/\omega \end{bmatrix} \ , \ T^{-1} = \begin{bmatrix} 1/2 & 1/2 \\ -\omega/2 & \omega/2 \end{bmatrix} \qquad (۲-۱۰)$$

$$\begin{bmatrix} \dot{p}_i \\ \dot{q}_i \end{bmatrix} = \begin{bmatrix} -\omega & 0 \\ 0 & \omega \end{bmatrix} \begin{bmatrix} p_i \\ q_i \end{bmatrix} \quad \Rightarrow \quad \begin{cases} \dot{p}_i + \omega \, p_i = 0 \\ \dot{q}_i - \omega \, q_i = 0 \end{cases} \qquad (۲-۱۱)$$

حل سیستم خطی مجزاشده (۲–۱۱) به صورت روابط(۲–۱۲) خواهد بود کـه معادله حرکـت پیوسـته راهرونده برای مدل سادهشده را در یک بازه زمانی یک گام نشان میدهد. در این روابط، مؤلفه $p_i$، یک تابع نمـایی بـا رشـد منفی است و به همین جهت آنرا مؤلفه همگرا مینامیم و مؤلفه $q_i$، یک تابع نمایی با رشد مثبت اسـت و بـه همـین جهت آنرا مؤلفه واگرا مینامیم.

$$\begin{matrix} Continuous\ Part \\ of \\ Motion\ Equation \\ for\ Simple\ Model \end{matrix} : \begin{cases} p_i = p_{i,0}\, e^{-\omega t_i} \\ q_i = q_{i,0}\, e^{\omega t_i} \end{cases}$$

$$(۲-۱۲)$$

## ۲-۳-۱ خواص معادله حرکت مدل سادهشده راه رونده در طول بازه زمانی یک گام

با مروری بر معادله حرکت رابطه (۲–۱۲) میتوان به خواص زیر در این معادله اشاره کرد.

۱-  همواره ضرب مؤلفه واگرا در مؤلفه همگرا در طول یک قدم طبق رابطه (۲–۱۳) ثابت است که میتوان توسط شکل ۲-۶، نمودار های صفحه فاز آن را نمایش داد.

$$p_i q_i = p_{i,0} q_{i,0} = k_i \qquad\qquad (۲-۱۳)$$

۲-  همواره علامت مؤلفه واگرا و مؤلفه همگرا در طول یک قدم حفظ میشود، به این معنی که هم مؤلفه واگرا و هم مؤلفه همگرا در بازه زمانی یک قدم همواره به طور جداگانه یا مثبت و یا منفی هستند.

۳-  مؤلفه واگرا و مؤلفه همگرا به طور هندسی بر اساس زمان افزایش و کاهش می یابند. به این معنی که مقدار جدید هر یک از مؤلفهها پس از گذشت یک بازه زمانی ثابت، نسبت به مقدار قبلی خود با ضریب ثابت افزایش یا کاهش پیدا میکنند(رابطه (۲–۱۴)).

$$\begin{cases} p_i(t + \Delta t) = p_i(t) e^{-\omega \Delta t} \\ q_i(t + \Delta t) = q_i(t) e^{\omega \Delta t} \end{cases}$$

$$(۲-۱۴)$$

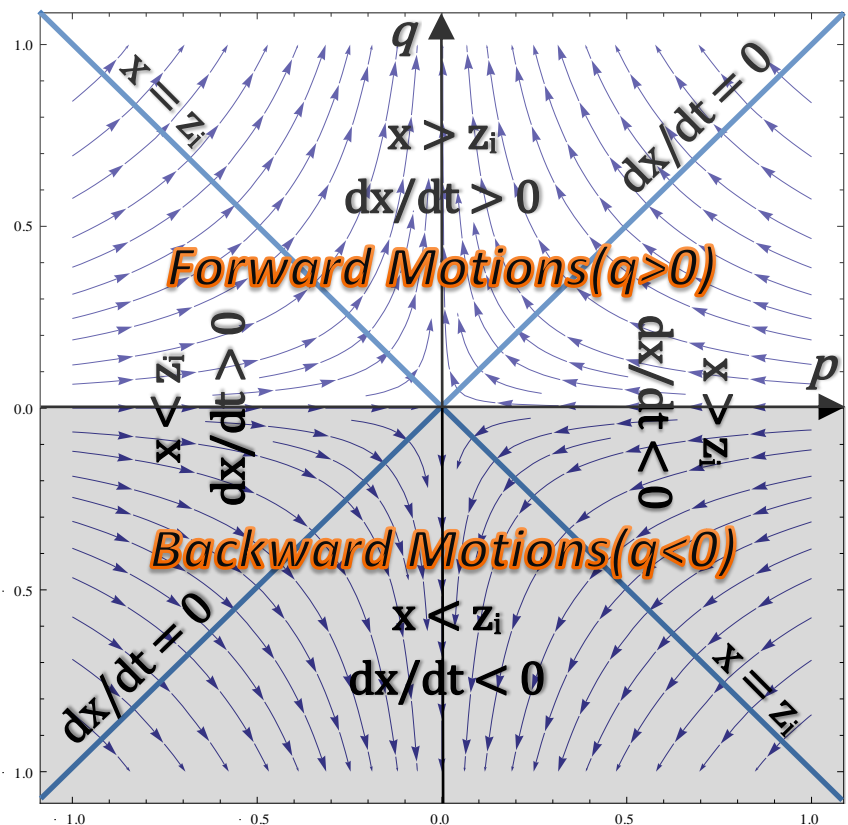

شکل ۲-۶. صفحه فاز مسیرهای حرکت مولفه واگرا نسبت به مولفه همگرا

(جهت زمانی حرکت در هر مسیر به صورت خطوط جریان رسم شده‌است.)

## ۲-۳-۲ حرکت پیش‌رونده[1] و پس‌رونده[2]

حرکتی را که یا سرعت مرکز جرم همواره در آن مثبت باشد یا در نهایت به سمت مثبت شدن پیش برود، حرکت پیش‌رونده یا روبه‌جلو و برخلاف آن حرکتی را که در آن یا سرعت مرکز جرم منفی باشد یا در نهایت به سمت منفی شدن پیش برود، حرکت پس‌رونده یا روبه‌عقب می‌نامیم. می‌توان اثبات کرد که حرکت‌های پیش‌رونده همواره دارای مولفه واگرای مثبت($q > 0$) هستند و در نمودار صفحه فازشکل ۲-۶ مسیرهای نیم صفحه بالایی را در ربع اول و بخصوص ربع دوم - به دلایلی که در ادامه توضیح داده خواهد شد- شامل می‌شوند، و بالعکس، حرکت‌های پس‌رونده همواره دارای مولفه واگرای منفی($q < 0$) هستند و در نمودار صفحه فاز شکل ۲-۶ مسیرهای نیم صفحه پایینی را در ربع سوم و بخصوص ربع چهارم شامل می‌شوند. آنچه در تحلیل حرکت پیش‌رونده و پس‌رونده حائز اهمیت بسیار خواهد بود و در فصل کنترل پایداری به آن خواهیم پرداخت، الزام به گام

---

[1] Forward Motion
[2] Backward Motion

برداشتن رو به جلو برای حفظ پایداری در حرکت‌های پیش‌رونده(شرایط اولیه پیش‌رونده) و الزام به گام برداشتن رو به عقب برای حفظ پایداری در حرکت‌های پس‌رونده(شرایط اولیه پس‌رونده) است.

همچنین در تحلیل‌های آتی همواره باید توجه داشت که هرگاه بحث از مؤلفه واگرا و یا همگرا می‌شود، از لحاظ مفهومی همواره روابط(۲-۱۵) بین این دو مؤلفه و دو کمیت فیزیکی قابل مشاهده یعنی سرعت مرکز جرم و فاصله مرکز جرم تا مرکز هندسی پای تکیه‌گاهی(که فعلا مرکز فشار را منطبق بر آن فرض کردیم)، حاکم است.

$$\begin{cases} (x - z_i) = \dfrac{q_i + p_i}{2} \\ \dot{x} = \omega \dfrac{q_i - p_i}{2} \end{cases} \quad and \quad \begin{cases} p_i = (x - z_i) - \dfrac{\dot{x}}{\omega} \\ q_i = (x - z_i) + \dfrac{\dot{x}}{\omega} \end{cases} \tag{۲-۱۵}$$

### ۲-۴  استخراج معادله حرکت گام به گام

با توجه به پیوسته بودن حرکت، اگر فرض کنیم در عمل گام برداشتن، فرودآمدن پای متحرک بـدون ضـربه وارد کردن به زمین انجام شود، همواره شرط پیوستگی موقعیت و سرعت مرکز جرم در لحظه انتهایی گـام قبلـی و ابتدای گام جدید حفظ خواهد شد و رابطه نهایی(۲-۱۶)همواره برقرار خواهد بود.

$$\begin{cases} t_i = T_i \Rightarrow p_{i,T_i} = (x - z_i) + \dfrac{\dot{x}}{\omega} \\ t_{i+1} = 0 \Rightarrow p_{i+1,0} = (x - z_{i+1}) + \dfrac{\dot{x}}{\omega} \end{cases} \Rightarrow p_{i+1,0} = p_{i,T_i} - (z_{i+1} - z_i) \tag{۲-۱۶}$$

این رابطه برای مؤلفه همگرا نوشته شده است و در آن، $T_i$، مدت زمان کل گـام $i$اُم را نمایش مـی‌دهـد. بـا نوشتن رابطه‌ای مشابه برای مؤلفه واگرا، در نهایت (۲-۱۷) به دست می‌آید که بخش گسسته معادله حرکـت را در لحظه فرود گام جدید نمایش می‌دهد.

$$\begin{matrix} Discrete\ Part \\ of \\ Motion\ Equation \end{matrix} : \begin{cases} p_{i+1,0} = p_{i,T_i} - L_i \\ q_{i+1,0} = q_{i,T_i} - L_i \end{cases} , \quad L_i \triangleq z_{i+1} - z_i \tag{۲-۱۷}$$

در این رابطه، $L_i$، به عنوان طول گام جدید تعریف می‌شود که برای حرکت با قـدم‌های روبه‌جلو مقـداری مثبـت و برای حرکت با قدم‌های روبه عقب مقداری منفی اختیار می‌کند. باید توجه کرد کـه ایـن پـارامتر در واقـع میـزان جابه‌جایی مرکز فشار یا فاصله مرکز پای تکیه‌گاهی فعلی تا مرکز پای تکیه‌گاهی بعدی در مـدت زمـان یک گام است و با تعریف متعارف از طول گام که منظور از آن فاصله طی شده توسط یک پا در مـدت زمـان یـک گام است، متفاوت می‌باشد.

با محاسبه معادله حرکت پیوسته ساده شده گام $i$اُم(رابطه (۲-۱۲)) برای لحظه نهایی این گام، $T_i$، که همان لحظه فرود پا است، (۲-۱۸) را خواهیم داشت.

$$t_i = T_i \quad \Rightarrow \quad \begin{cases} p_{i,T_i} = p_{i,0}e^{-\omega T_i} \\ q_{i,T_i} = q_{i,0}e^{\omega T_i} \end{cases} \tag{۲-۱۸}$$

با جایگذاری رابطه (۲-۱۸) در رابطه (۲-۱۷)، در نهایت به رابطه نهایی (۲-۱۹) خواهیم رسید:

$$\begin{matrix} Step - to - Step \;\; Equation \\ \textit{(Discrete+Contiuous Parts of} \\ \textit{Motion Equations for Simple Model )} \end{matrix} : \quad \begin{cases} p_{i+1,0} = p_{i,0}e^{-\omega T_i} - L_i \\ q_{i+1,0} = q_{i,0}e^{\omega T_i} \;\; - L_i \end{cases} \tag{۲-۱۹}$$

این رابطه گسسته را که در واقع مجموع دو بخش پیوسته و گسسته یک گام برای مدل ساده‌شده حرکت می‌باشد، معادله حرکت گام به‌گام می‌نامیم و دارای دو ورودی $L_i$ و $T_i$ است که شرایط اولیه گام $i$ اُم را به شرایط اولیه گام $i + 1$ اُم انتقال می‌دهد. به بیان دیگر این معادله، شرایط اولیه گام بعدی را با توجه به شرایط اولیه گام فعلی و طول و زمان فرود گام فعلی(گام جدید) تخمین می‌زند. برای نمایش این معادله حرکت گسسته نیز همانند نمایش معادله حرکت پیوسته برای یک گام، از نمودار صفحه فاز با اندکی تفاوت استفاده می‌کنیم. شمای کلی یک گام و انواع گام‌های قابل تصور در شکل ۲-۷ نمایش داده‌شده است. نمادهای شکلی استفاده شده در این شکل‌ها و شکل‌های بعدی که در این فصل و فصل‌های بعدی از آن استفاده خواهیم کرد، در جدول ۲-۳ توضیح داده شده‌است.

جدول ۲-۳. تعریف نمادهای موجود در شکل‌های صفحه فاز معادله حرکت فاز گام‌به‌گام

| | |
|---|---|
| 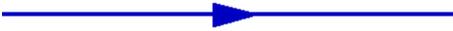 | مسیر حرکتی طی شده در بازه زمانی یک گام که از شرایط اولیه یک گام شروع و به شرایط نهایی همان گام ختم می‌شود. این مسیر بخش پیوسته معادله حرکت را نمایش می‌دهد. |
| 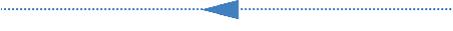 | خط انتقال که میزان انتقال شرایط نهایی یک گام به شرایط اولیه گام بعدی را (که بر اثر فرود پا در یک لحظه با مدت زمانی نزدیک به صفر اتفاق می‌افتد) نمایش می‌دهد. این خط بخش گسسته معادله حرکت را نمایش می‌دهد. |
|  | نقطه شرایط اولیه مولفه همگرا و واگرا در گام اول |
|  | نقطه شرایط اولیه مولفه همگرا و واگرا پس از برداشتن گام آخر |

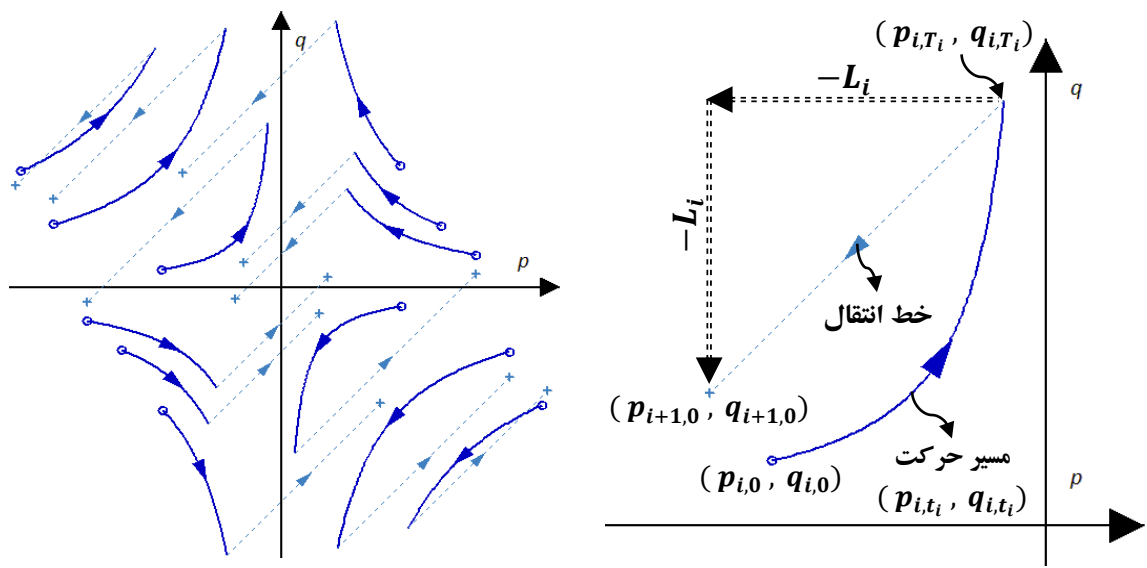

الف- شمای یک گام برای معادله حرکت گام‌به‌گام

ب- انواع نمونه گام‌های قابل تصور برای معادله حرکت گام‌به‌گام

شکل ۲-۷. صفحه فاز مسیرهای حرکت مولفه واگرا نسبت به مولفه همگرا برای نمایش مدل گسسته حرکت گام‌به‌گام

برای نمونه شکل ۲-۸، شکل ۲-۹ و شکل ۲-۱۰ چهارگام پی‌درپی از مدل حرکتی گام به‌گام را در صـفحه فـاز برای حرکت‌های مختلف (مقادیر مختلف طول گام و مدت زمان یک گام) به ترتیب پیش‌رونده، پس‌رونده و جلو و عقب رونده(یا راست وچپ رونده) نمایش می‌دهند.

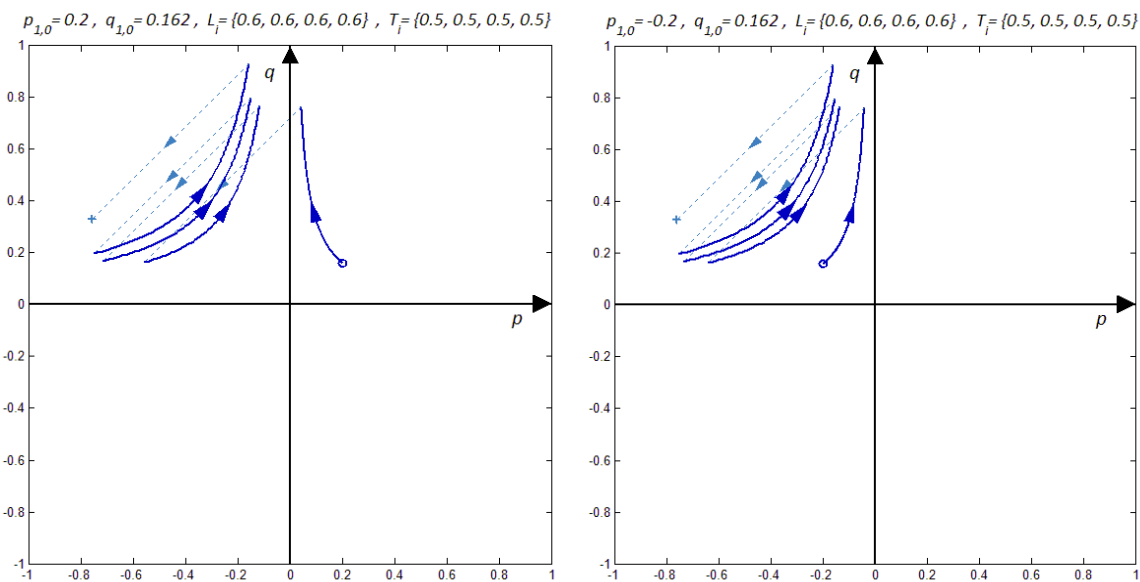

شکل ۲-۸ صفحه فاز مسیرهای حرکت مولفه واگرا نسبت به مولفه همگرا برای چهار گام پیش‌رونده پی‌درپی با شرایط اولیه متفاوت

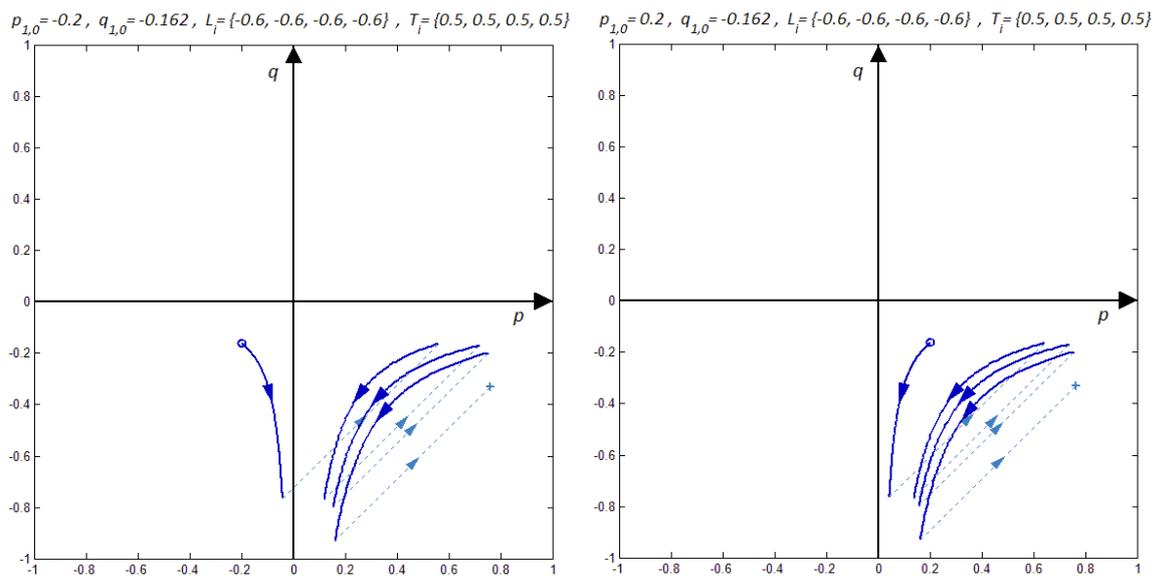

شکل ۲-۹. صفحه فاز مسیرهای حرکت مؤلفه واگرا نسبت به مؤلفه همگرا برای چهار گام پس‌رونده پی‌درپی با شرایط اولیه متفاوت

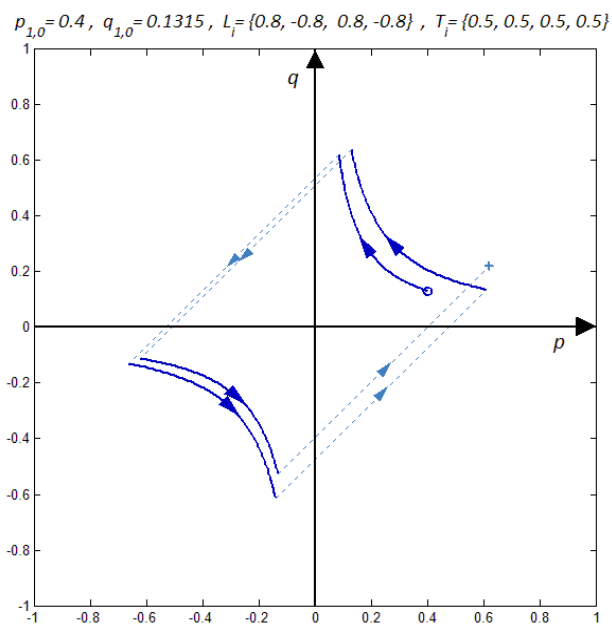

شکل ۲-۱۰. صفحه فاز مسیرهای حرکت مؤلفه واگرا نسبت به مؤلفه همگرا برای چهار گام جلو و عقب یا چپ و راست پی‌درپی

برای کنترل پایداری در فصل بعدی، از یک رابطه تخمین آنی که دقیق‌تر از (۲-۱۹) است، استفاده خواهیم کرد. رابطه مذکور شرایط اولیه گام بعدی را بر اساس شرایط آنی در گام فعلی و نه شرایط اولیه گام فعلی تخمین می‌زند. برای به‌دست‌آوردن رابطه مورد نظر، می‌توان رابطه‌ای به صورت (۲-۲۰) با توجه به خاصیت (۳) معادله حرکت پیوسته نوشت.

$$\begin{cases} p_{i+1,0} = p_{i,t_i} e^{-\omega(T_i - t_i)} - L_i \\ q_{i+1,0} = q_{i,t_i} e^{\omega(T_i - t_i)} - L_i \end{cases} \qquad (\text{۲۰}-\text{۲})$$

با تعریف شرایط اولیه تخمینی گام فعلی $i$ اُم، $p_{i,0}^E$ و $q_{i,0}^E$، که از روی شرایط آنی این گام توسط رابطه (۲۱-۲) محاسبه می‌شوند می‌توان به رابطه‌ای مشابه با (۱۹-۲) به صورت رابطه (۲۲-۲) رسید.

$$\begin{cases} p_{i,0}^E(t_i) = p_{i,t_i} e^{\omega t_i} \\ q_{i,0}^E(t_i) = q_{i,t_i} e^{-\omega t_i} \end{cases} \qquad (\text{۲۱}-\text{۲})$$

$$\begin{cases} p_{i+1,0} = p_{i,0}^E(t_i) e^{-\omega T_i} - L_i \\ q_{i+1,0} = q_{i,0}^E(t_i) e^{\omega T_i} - L_i \end{cases} \qquad (\text{۲۲}-\text{۲})$$

اگرچه این رابطه تخمینی تابعی از متغیر زمان در گام فعلی است و هیچگاه جایگزین رابطه اصلی معادله حرکت گسسته گام‌به‌گام (۱۹-۲) در تحلیل‌های مربوط به طراحی کنترل‌کننده و بررسی پایداری نخواهد شد، ولی در کنترل به‌هنگام پایداری برای محاسبه دقیق‌تر بازخورد[1] در طول بازه زمانی گام، قابل استفاده خواهد بود. این رابطه به علت در نظر گرفتن اثر عواملی که معادله حرکت پیوسته را از مسیر ایده‌آل خود در مدل ساده‌شده تا لحظه کنونی گام فعلی ($t_i$) منحرف کرده‌اند، به کار گرفته می‌شود تا پیش‌بینی دقیق‌تری را از شرایط اولیه گام بعدی (در صورت برداشتن گامی با مشخصات $L_i$ و $T_i$) در اختیارمان بگذارد(در صورت صفر بودن اختلالات، رابطه (۲۲-۲) همواره برابر با (۱۹-۲) است که در طراحی کنترل‌کننده معمولا به دلیل نامشخص بودن تابع اختلال، اثر آن را صفر فرض می‌کنیم ولی در زمان کنترل، اثر آن را در بازخورد اندازه‌گیری می‌کنیم).

## ۲-۵ سیکل‌های حرکتی

اگر بخواهیم با برداشتن گام‌هایی متوالی با طول مشخص و در زمان مناسب به حرکتی دایمی برسیم با دو سئوال مهم روبرو هستیم:
۱-   چه رابطه‌ای باید بین این کمیت‌ها به عنوان ورودی سیستم و شرایط اولیه حاکم باشد؟
۲-   آیا برای گام‌هایی متوالی با طول و زمان فرود مشخص و ثابت، این حرکت دایمی پایدار خواهد بود یا خیر؟

---

[1] Feedback

سعی می‌کنیم برای جواب به این دو سوال، ابتدا در این قسمت به طرح مفهومی به نام سیکل حرکتی و انواع آن بپردازیم و در قسمت بعدی پایداری آن را بررسی کنیم. حرکتی را سیکل می‌نامیم که حرکت در آن از یک شرایط اولیه دلخواه شروع شود و پس از طی یک گام یا چند گام متفاوت با طول و زمان مشخص به همین شرایط اولیه دلخواه بازگردد.

## ۲-۵-۱ سیکل‌های حرکتی ساده(تک گامی) پیش‌رونده[۱] و پس‌رونده[۲]

سیکل حرکتی ساده را سیکلی تعریف می‌کنیم که دارای تناوب یک گام است. طبق این تعریف اگر فرض کنیم برای شرایط اولیه یک گام مقادیر مولفه همگرا $p_c$ و مولفه واگرا $q_c$ باشد، آنگاه برای شرایط اولیه‌ای که پس از برداشتن گامی با اندازه طول $L_c$ و در مدت زمان $T_c$ ایجاد می‌شود نیز همین مقادیر باید تکرار گردد. با درنظرگرفتن این شرط بر روی رابطه (۲-۱۹) می‌توان روابط (۲-۲۳) را نوشت که رابطه‌ای بین شرایط اولیه مولفه همگرا و واگرا با طول و زمان گام در یک سیکل حرکتی ساده را توصیف می‌کنند.

$$\begin{cases} p_c = p_c e^{-\omega T_c} - L_c \\ q_c = q_c e^{\omega T_c} - L_c \end{cases} \Rightarrow \begin{cases} p_c = -\dfrac{L_c}{1 - e^{-\omega T_c}} \\ q_c = \dfrac{L_c}{e^{\omega T_c} - 1} \end{cases} \quad or \quad \begin{cases} L_c = -(p_c + q_c) \\ T_c = \dfrac{1}{\omega} Ln\left(-\dfrac{p_c}{q_c}\right) \end{cases} \tag{۲-۲۳}$$

سرعت متوسط سیکل حرکتی ساده از رابطه (۲-۲۴) بدست می‌آید که می‌توان از آن بجای پارامتر زمان سیکل استفاده نمود.

$$V_c = \frac{L_c}{T_c} = \frac{-\omega(p_c + q_c)}{Ln\left(-\dfrac{p_c}{q_c}\right)} , \quad (V_c, L_c) \Rightarrow (p_c, q_c) , \quad \begin{cases} p_c = -\dfrac{L_c}{1 - e^{-\frac{\omega L_c}{V_c}}} \\ q_c = \dfrac{L_c}{e^{\frac{\omega L_c}{V_c}} - 1} \end{cases} \tag{۲-۲۴}$$

ازآنجایی که $T_c$ همواره باید مقداری مثبت داشته باشد، مقدار ورودی تابع لگاریتم نمایی همواره باید مقداری بزرگتر از واحد اختیار کند که می‌توان روابط (۲-۲۵) را به عنوان شروط لازم از آن نتیجه‌گیری کرد.

$$-\frac{p_c}{q_c} > 1 \Rightarrow \begin{cases} -\dfrac{p_c}{q_c} > 0 \Rightarrow p_c q_c < 0 \\ 1 + \dfrac{p_c}{q_c} < 0 \end{cases} \Rightarrow \begin{matrix} Forward\ Cycles & \begin{cases} p_c < 0 \\ q_c > 0 \\ p_c + q_c < 0 \equiv L_c > 0 \end{cases} \\ or & \\ Backward\ Cycles & \begin{cases} p_c > 0 \\ q_c < 0 \\ p_c + q_c > 0 \equiv L_c < 0 \end{cases} \end{matrix} \tag{۲-۲۵}$$

---

[۱] Forward Cycle
[۲] Backward Cycle

این شروط با توجه به مثبت یا منفی بودن طول گام ( $L_c = -(p_c + q_c)$ ) ، دو نوع سیکل ساده حرکتی پیش‌رونده یا پس‌رونده را تعریف می‌کند. این دو نوع سیکل حرکتی ساده به ترتیب در ربع دوم و چهارم صفحه فاز مطابق شکل ۲-۱۱ قرار می‌گیرد. سیکل حرکتی ساده پیش‌رونده در واقع حرکت راه رفتن یکنواخت رو به جلو است که مهم‌ترین بخش از راه رفتن را توصیف می‌کند.

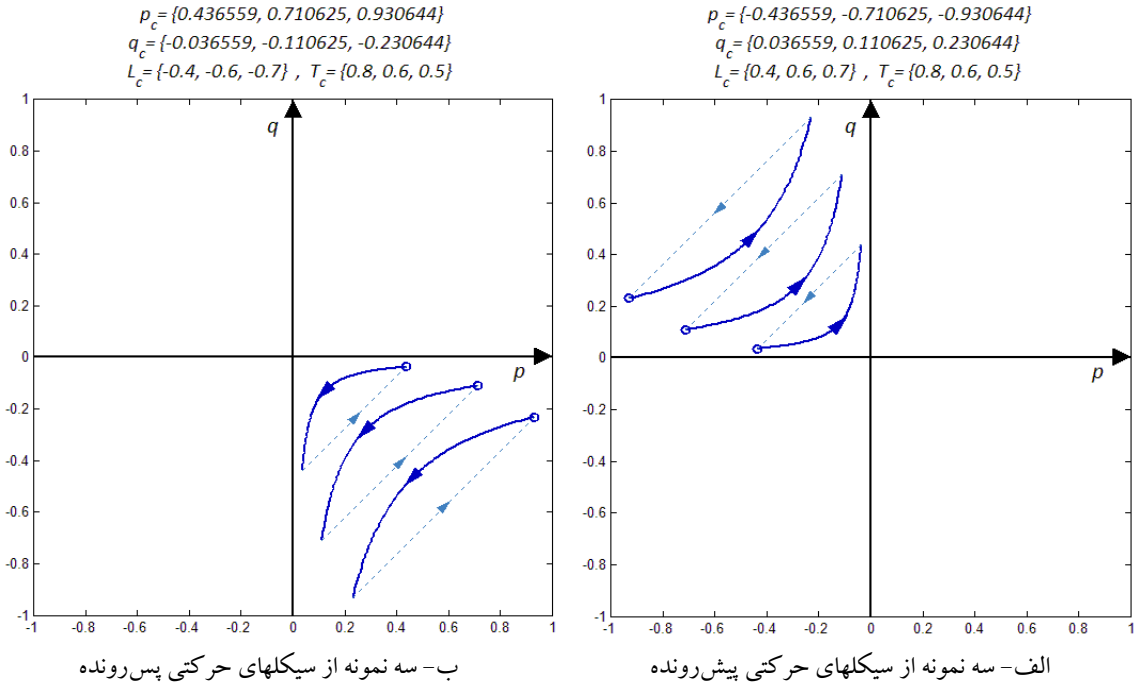

$p_c = \{0.436559, 0.710625, 0.930644\}$
$q_c = \{-0.036559, -0.110625, -0.230644\}$
$L_c = \{-0.4, -0.6, -0.7\}$ , $T_c = \{0.8, 0.6, 0.5\}$

$p_c = \{-0.436559, -0.710625, -0.930644\}$
$q_c = \{0.036559, 0.110625, 0.230644\}$
$L_c = \{0.4, 0.6, 0.7\}$ , $T_c = \{0.8, 0.6, 0.5\}$

ب- سه نمونه از سیکلهای حرکتی پس‌رونده

الف- سه نمونه از سیکلهای حرکتی پیش‌رونده

شکل ۲-۱۱. صفحه فاز مسیرهای حرکت مؤلفه واگرا نسبت به مؤلفه همگرا برای سیکل‌های حرکتی ساده

• محدودیت‌های سیکل‌های حرکتی ساده

با توجه به روابط (۲-۲۵)، محدوده قابل تعریف برای شرایط اولیه سیکلهای حرکتی ساده پیش‌رونده و پس‌رونده در شکل ۲-۱۲-الف به تصویر کشیده شده است. ولی با توجه به دیگر محدودیت‌های ربات از جمله حداکثر فاصله مجاز بین دو پا، $L_{max}$ ، یا حداقل زمان مورد نیاز برای گامی به طول $L_c$، $T_{min}$ ، تنها بخشی از این محدوده، محدوده مجاز یا قابل دسترس برای سیکل‌های حرکتی ساده محسوب می‌شود که در شکل ۲-۱۲-ب به تصویر کشیده شده است.

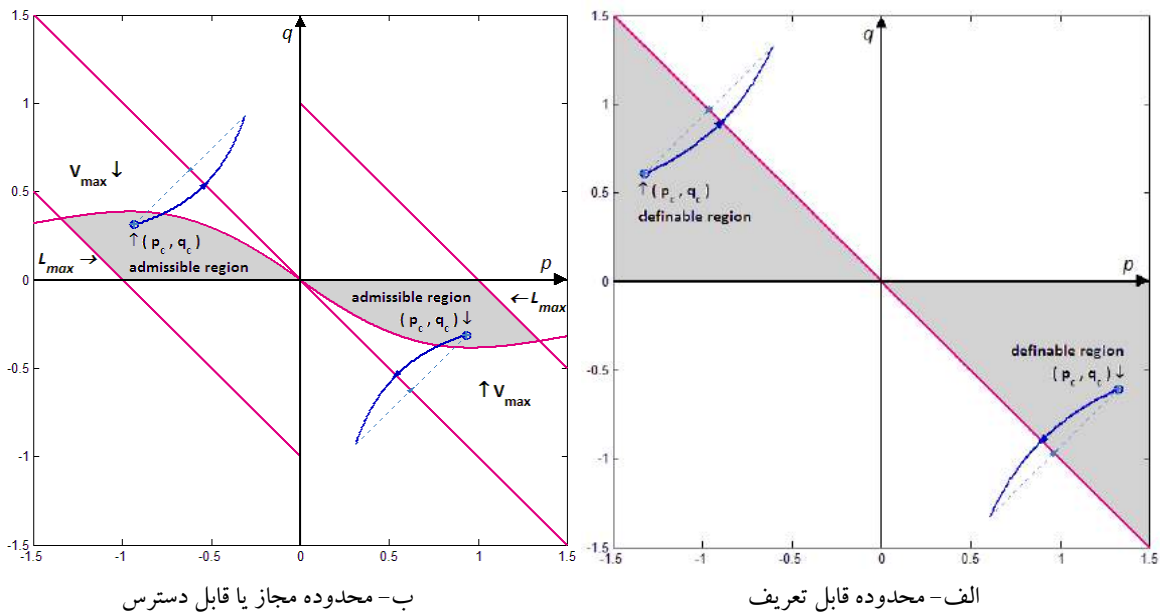

شکل ۲-۱۲. محدوده شرایط اولیه مولفه واگرا نسبت به مولفه همگرا در صفحه فاز برای سیکل‌های حرکتی ساده (ناحیه رنگ شده)

برای به دست آوردن محدودیت حداکثر طول بین دوپا، روابط (۲-۲۶) را می‌توان نوشت.

$$|L_c| < L_{max} \Rightarrow |p_c + q_c| < L_{max} \Rightarrow \begin{cases} Forward\ Cycles\ :\ q_c > -L_{max} - p_c \\ Backward\ Cycles:\ q_c < L_{max} - p_c \end{cases} \tag{۲-۲۶}$$

همچنین برای محدودیت حداقل زمان گام برداشتن اگر برای شتاب‌گیری و ترمز پا مدت زمان $T_0$ و برای حداکثر اندازه سرعت نسبی حرکت پا را به بدن مقدار $V_{max}$ را در نظر بگیریم، روابط (۲-۲۷) به‌دست می‌آیند که $W_0$ در آن، تابع اُمگای لمبرت[1] نام دارد و توسط رابطه (۲-۲۸) تعریف و به وسیله شکل ۲-۱۳ توصیف می‌شود.

$$T_c > T_{min}(L_c) \Rightarrow \frac{1}{\omega} Ln\left(-\frac{p_c}{q_c}\right) < \frac{|L_c|}{V_{max}} + T_0 \Rightarrow \frac{1}{\omega} Ln\left(-\frac{p_c}{q_c}\right) < \frac{|p_c + q_c|}{V_{max}} + T_0$$

$$\Rightarrow \begin{cases} Forward\ Cycles\ :\ q_c < -\frac{V_{max}}{\omega} W_0\left(\frac{\omega\, p_c}{V_{max}} e^{\omega\left(\frac{p_c}{V_{max}} - T_0\right)}\right) \\ Backward\ Cycles:\ q_c > \frac{V_{max}}{\omega} W_0\left(-\frac{\omega\, p_c}{V_{max}} e^{-\omega\left(\frac{p_c}{V_{max}} + T_0\right)}\right) \end{cases} \tag{۲-۲۷}$$

---

[1] Lambert-W function / Omega Function / Product Logarithm

$$Lambert\ W\ definition: \quad y = \boldsymbol{W_0}(x) \quad \Leftrightarrow \quad x = ye^y \ , \ y > -1$$

$$Taylor\ series\ of\ \boldsymbol{W_0}(x) = \sum_{n=1}^{\infty} \frac{(-n)^{n-1}}{n!} x^n = x - x^2 + \frac{3}{2}x^3 - \frac{8}{3}x^4 + \frac{125}{24}x^5 + \cdots \qquad (2-28)$$

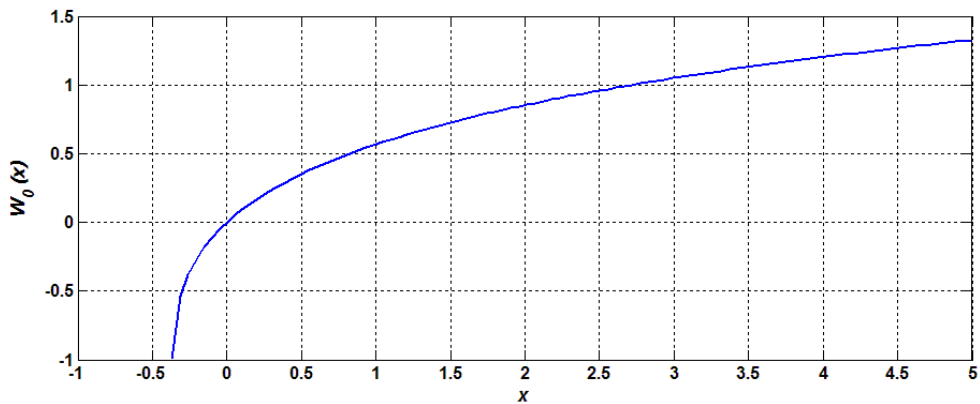

شکل ۲-۱۳. نمودار توصیف‌کننده تابع اُمگای لمبرت ریشه صفرم

## ۲-۵-۲ سیکل‌های حرکتی مرکب (چندگامی)

یک سیکل حرکتی مرکب دارای دوره تناوبی مساوی یا بزرگتر دو گام است. به این معنی که شرایط اولیـه پـس از دو گام یا بیشتر تکرار می‌شود. با این حساب برای سیکلهای مرکب دوگامی می‌توان روابط (۲-۲۹) را نوشت.

$$\begin{cases} p_{c_2} = p_{c_1} e^{-\omega T_{c_1}} - L_{c_1} \\ q_{c_2} = q_{c_1} e^{\omega T_{c_1}} - L_{c_1} \end{cases}$$
$$and$$
$$\begin{cases} p_{c_1} = p_{c_2} e^{-\omega T_{c_2}} - L_{c_2} \\ q_{c_1} = q_{c_2} e^{\omega T_{c_2}} - L_{c_2} \end{cases} \Rightarrow \begin{cases} p_{c_1} = -\dfrac{L_{c_1} e^{-\omega T_{c_2}} + L_{c_2}}{1 - e^{-\omega(T_{c_1}+T_{c_2})}} \\[2mm] q_{c_1} = \dfrac{L_{c_1} e^{\omega T_{c_2}} + L_{c_2}}{e^{\omega(T_{c_1}+T_{c_2})} - 1} \\[2mm] p_{c_2} = -\dfrac{L_{c_2} e^{-\omega T_{c_1}} + L_{c_1}}{1 - e^{-\omega(T_{c_1}+T_{c_2})}} \\[2mm] q_{c_2} = \dfrac{L_{c_2} e^{\omega T_{c_1}} + L_{c_1}}{e^{\omega(T_{c_1}+T_{c_2})} - 1} \end{cases} \qquad (2-29)$$

برای سیکلهای بزرگتر از دوگام نیز روابط مشابهی می‌توان استخراج کرد. اگرچه این سیکلها بـرای مـا جـذابیت زیادی ندارند ولی برخی حالات خاص سیکل دوگامی شیوه‌هایی را در راه رفتن توصیف می‌کند که سعی مـی‌کنیم در این قسمت به صورت مختصر به آنها بپردازیم. شکل ۲-۱۴، دو نوع سیکل دوگامی پیش‌رونده و پـس‌رونـده کـه توسط روابط (۲-۲۹) بدست می‌آیند را به تصویر می‌کشد.

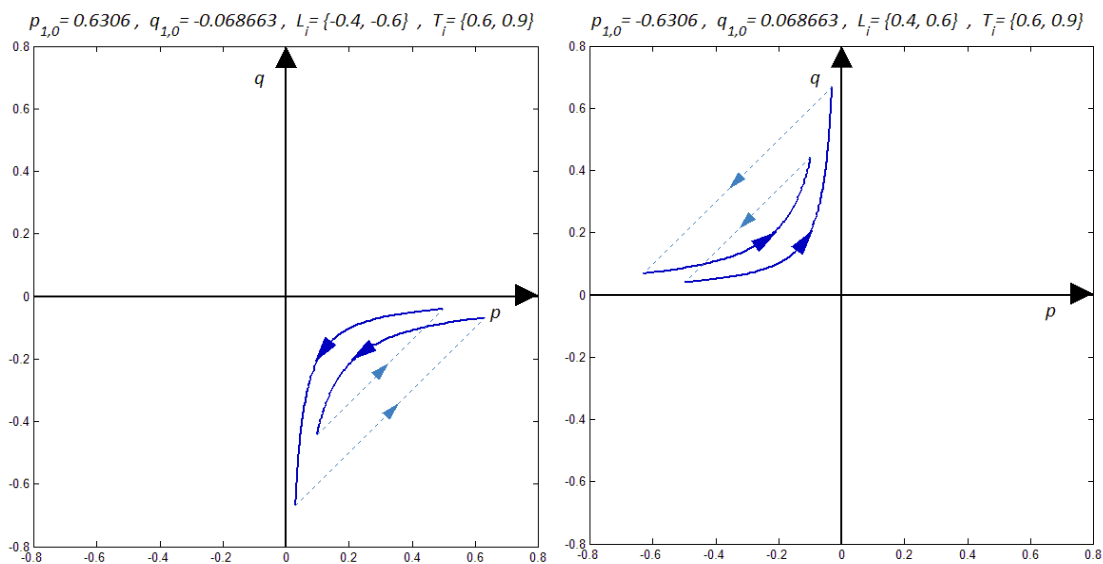

شکل ۲-۱۴. صفحه‌های فاز مسیر حرکت مولفه واگرا نسبت به مولفه همگرا برای سیکل حرکتی مرکب دوگامی پیش‌رونده (صفحه فاز سمت راست) و پس‌رونده (صفحه فاز سمت چپ)

سرعت متوسط در دو گام مختلف سیکل‌های مرکب این شکل ثابت است. می‌توان سیکل‌هـای دو گـامی تـک‌جهتی(پیش‌رونده یا پس‌رونده) دیگری را نیز ازجمله سیکل‌های دوگامی با طول گام ثابت در دوگام و یا زمان گـام ثابت در دوگام، در نظرگرفت.

علاوه بر سیکل‌های دوگامی تک‌جهتی(پیش‌رونده و پس‌رونده)، شکل ۲-۱۵ دو نمونه از سیکل دوگامی عقـب و جلو رونده در صفحه طولی را به تصویر می‌کشد. سیکل‌های موجود در این شکل نیز دارای اندازه سرعت متوسـط برابری در دوگام هستند و یکی دارای طول گام مثبت و دیگری منفی است کـه اگر انـدازه گـام مثبـت بزرگ‌تـر از اندازه گام منفی باشد، حرکت راه‌رونده با وجود عقب و جلو رفتن، در جهت گام مثبت انحراف پیـدا مـی‌کنـد و در غیراین‌صورت بالعکس. نوع خاصی از این سیکل که دارای اندازه گام برابر در دو گام عقب و جلو اسـت، منجـر بـه انحرافی به جلو و عقب نمی‌شود و حرکتی درجا را توصیف می‌کند. حالت متقارن درجای سیکل متقارن مـی‌نـامیم و در ادامه این مطلب به بررسی آن می‌پردازیم.

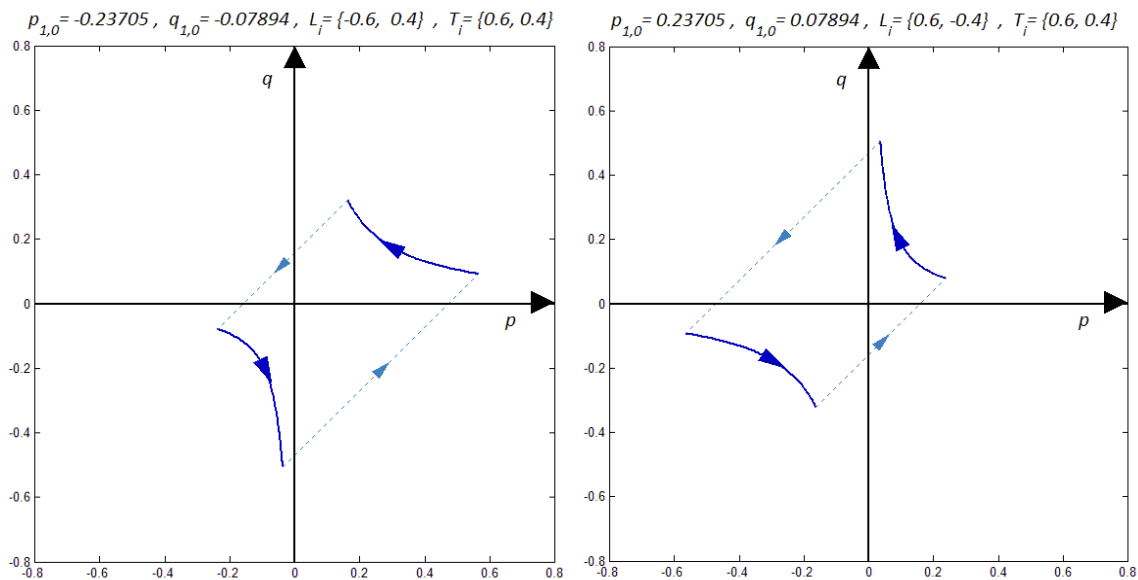

$p_{1,0} = -0.23705$ , $q_{1,0} = -0.07894$ , $L_i = \{-0.6,\ 0.4\}$ , $T_i = \{0.6,\ 0.4\}$

$p_{1,0} = 0.23705$ , $q_{1,0} = 0.07894$ , $L_i = \{0.6,\ -0.4\}$ , $T_i = \{0.6,\ 0.4\}$

شکل ۲-۱۵. صفحه‌های فاز مسیر حرکت مولفه واگرا نسبت به مولفه همگرا برای سیکل‌های حرکتی مرکب دوگامی عقب و جلو رونده دارای انحراف به جلو (صفحه فاز سمت راست) و انحراف به عقب (صفحه فاز سمت چپ)

- سیکل حرکتی درجای[1] متقارن

اگرچه حرکت در صفحه جانبی موضوع بحث این پژوهش نیست ولی اگر معادلات حرکت را در صفحه جانبی بنویسیم، با انجام فرض‌های ساده‌کننده مشابهی برای حرکت در راستای صفحه جانبی همانند آنچه برای حرکت در راستای صفحه طولی برقرار است، به معادله گام به‌گام یکسانی با آنچه برای حرکت در صفحه طولی رسیدیم، خواهیم رسید. با این مقدمه، نوع خاصی از سیکل مرکب، سیکل دوگامی جلو و عقب رونده(با راست و چپ رونده) است که رفتار انسان در صفحه جانبی هنگام راه رفتن در صفحه طولی را برای یک مدل سه‌بعدی توصیف می‌کند. این حرکت هیچ‌گونه جابجایی را در صفحه جانبی موجب نمی‌شود و در صورتی که در راستای صفحه طولی نیز حرکتی نداشته‌باشیم، حرکتی درجا را توصیف می‌کند و به همین دلیل می‌توان آن را سیکل درجا نامگذاری کرد. در این سیکل حرکتی مرکز فشار یا تکیه‌گاه بدن در بازه زمانی یک گام روی پای راست و در بازه زمانی برابری در گام بعدی روی پای چپ خواهد بود و این کار تکرار می‌شود. این سیکل حرکتی همچنین می‌تواند حرکت درجای جلو و عقب در صفحه طولی را نیز توصیف کند که سیکلی خاص محسوب می‌شود و کاربردی عام در راه‌رفتن ندارد. برای به دست آوردن ارتباط حاکم بین مشخصات گام و شرایط اولیه این سیکل حرکتی، روابط (۲-۳۰) را می‌توان نوشت.

---

[1] Idle Cycle

$$forward \; : \begin{cases} -p_c = p_c e^{-\omega T_{c_1}} - L_{c_1} \\ -q_c = q_c e^{\omega T_{c_1}} - L_{c_1} \end{cases}$$

and

$$backward \; : \begin{cases} p_c = -p_c e^{-\omega T_{c_2}} - L_{c_2} \\ q_c = -q_c e^{\omega T_{c_2}} - L_{c_2} \end{cases}$$

$$\Rightarrow \begin{cases} L_{c_1} = -L_{c_2} = (p_c + q_c) \\ T_{c_1} = T_{c_2} = \dfrac{1}{\omega} Ln\left(\dfrac{p_c}{q_c}\right) \end{cases}$$

$$(۲-۳۰)$$

$p_c = \{0.358538, -0.358538, 0.542895, -0.542895, 0.359009, -0.359009\}$

$q_c = \{0.141462, -0.141462, 0.157105, -0.157105, 0.040991, -0.040991\}$

$L_c = \{0.5, -0.5, 0.7, -0.7, 0.4, -0.4\}$ , $T_c = \{0.3, 0.3, 0.4, 0.4, 0.7, 0.7\}$

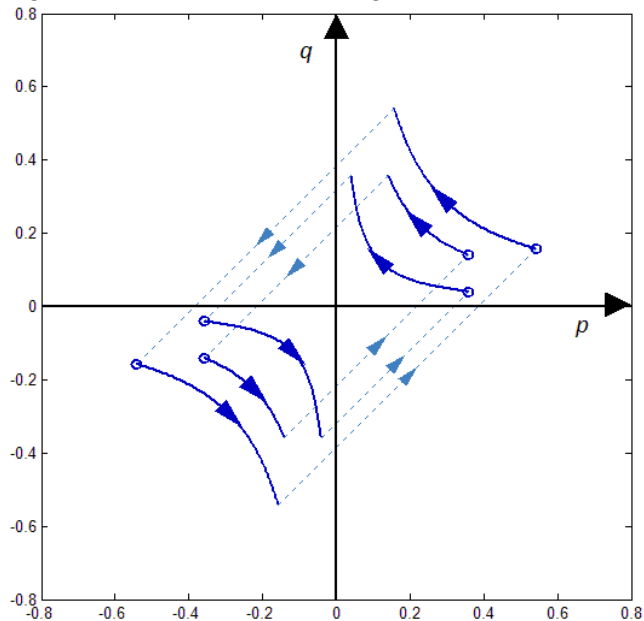

شکل ۲-۱۶. صفحه فاز مسیر حرکت مولفه واگرا نسبت به مولفه همگرا برای سه نمونه سیکل حرکتی مرکب دوگامی درجا

## ۲-۵-۳ بررسی پایداری سیکل‌های حرکتی(ساده)

اگر فرض کنیم از نقطه‌ای نزدیک و یا منطبق بر شرایط اولیه یک سیکل حرکتی ساده، شروع به برداشتن گام-هایی پیاپی با طول و زمان گام مربوط به همین سیکل حرکتی کنیم، آیا به این سیکل حرکتی نزدیک و یا از آن دور می‌شویم؟ نقطه شروع حرکت در چه محدوده‌ای از شرایط اولیه این سیکل حرکتی اگر قرار گیرد به آن سیکل نزدیک و همگرا شده و در چه محدوده‌ای اگر قرارگیرد، از سیکل دور و نسبت به آن واگرا خواهیم شد. برای این کار کافی است شرایط اولیه دلخواهی برای معادله گسسته حرکت گام به گام در نظر بگیریم و نتیجه حرکت را پس از پیمودن گام‌هایی پی‌درپی با طول و زمان گام ثابت و برابر با طول و زمان گام سیکل مورد نظر، بررسی کنیم.

رابطه (۲-۳۱) بر اساس جمع دنباله‌های حسابی و هندسی به‌دست آمده است که درستی آن به روش استقرا قابل تحقیق می‌باشد.

$$\begin{cases} p_{i+1,0} = p_{i,0}e^{-\omega T_c} - L_c \\ q_{i+1,0} = q_{i,0}e^{\omega T_c} - L_c \end{cases} \Rightarrow \begin{cases} p_{k,0} = p_{1,0} - \left(p_{1,0} + \dfrac{L_c}{1-e^{-\omega T_c}}\right)\left(1-e^{-(k-1)\omega T_c}\right) \\ q_{k,0} = q_{1,0} + \left(q_{1,0} - \dfrac{L_c}{e^{\omega T_c}-1}\right)\left(e^{(k-1)\omega T_c}-1\right) \end{cases} \tag{۲-۳۱}$$

این رابطه مشخصا نشان می‌دهد که شرایط اولیه مولفه همگرای معادله گسسته نیز همچنان برای معادله گسسته نیز همگرا است و به سمت شرایط اولیه مولفه همگرای سیکل حرکتی میل می‌کند(رابطه(۲-۳۲)).

$$\lim_{k\to\infty} p_{k,0} = \lim_{k\to\infty} p_{1,0}e^{-(k-1)\omega T_c} - \frac{L_c}{1-e^{-\omega T_c}}\left(1-e^{-(k-1)\omega T_c}\right) = -\frac{L_c}{1-e^{-\omega T_c}} = p_c \tag{۲-۳۲}$$

درحالی که شرایط اولیه مولفه واگرا تنها دارای پایداری مرزی [1] است (رابطه (۲-۳۳)) و با کوچکترین اختلافی نسبت به شرایط اولیه سیکل حرکتی، واگرا خواهد شد که به دلیل رخدادن این اختلاف در عمل، همواره رفتاری واگرا خواهد داشت و می‌توان گفت شرایط اولیه مولفه واگرا برای معادله حرکت گسسته نیز همچون مولفه همگرای معادله حرکت پیوسته واگرا می‌ماند.

$$\lim_{k\to\infty} q_{k,0} = \begin{cases} +\infty\,, & q_{1,0} > \dfrac{L_c}{e^{\omega T_c}-1} \\ q_c\,, & q_{1,0} = \dfrac{L_c}{e^{\omega T_c}-1} = q_c \\ -\infty\,, & q_{1,0} < \dfrac{L_c}{e^{\omega T_c}-1} \end{cases} \tag{۲-۳۳}$$

در این قسمت به دلیل اهمیت سیکل حرکتی ساده و همچنین به جهت خلاصه نویسی مطالب تنها به بررسی پایداری سیکل حرکتی ساده پیش‌رونده پرداختیم، با این‌حال می‌توان نشان داد برای انواع دیگر سیکل‌های حرکتی نیز، به نتایج مشابهی در مورد همگرایی و واگرایی شرایط اولیه مولفه همگرا و واگرا می‌توان رسید.

موضوع اصلی این تحقیق در واقع ارائه روشی است که بتواند حرکت راه رفتن را به سمت یکی از این سیکل‌های حرکتی ساده سوق دهد و سپس پایداری را حول آن سیکل حرکتی تضمین کند. در فصل بعدی سعی خواهیم کرد با ارائه کنترل‌کننده‌هایی، پایداری مجانبی حول این سیکل‌های حرکتی ناپایدار را تضمین کنیم.

---

[1] Marginal Stability

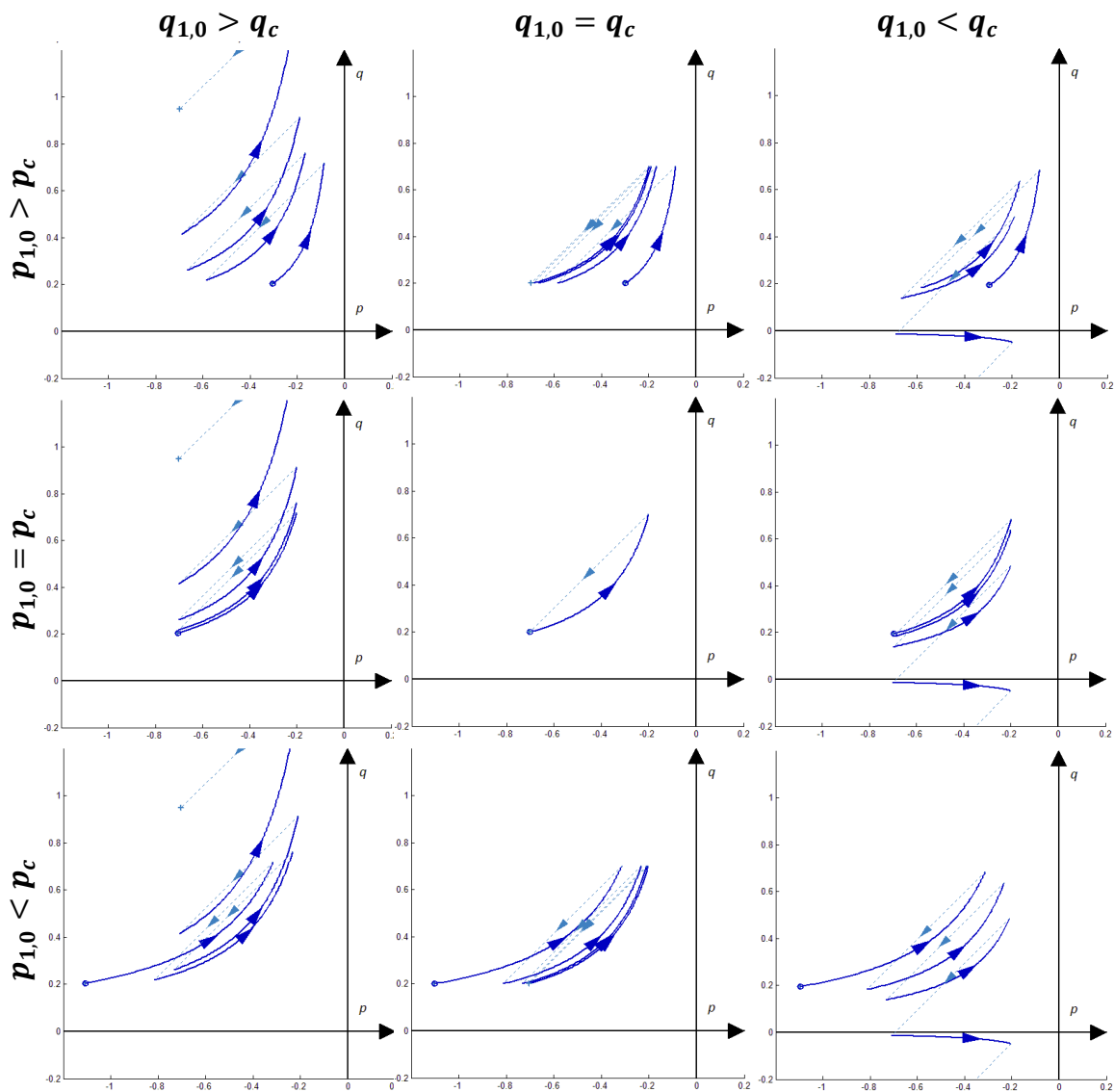

شکل ۲-۱۷. بررسی پایداری سیکل حرکتی ساده در صفحه فاز مولفه واگرا نسبت به مولفه همگرا برای شرایط اولیه متفاوت

## ۶-۲   مقایسه معادله حرکت کامل با معادله حرکت ساده‌شده راه رفتن طی یک گام

بررسی انحراف مدل حرکتی کامل از مدل ساده شده به ما کمک می‌کند که درک کامل‌تری از مسیرهای ممکن در صفحه فاز برای حرکت راه رفتن داشته باشیم. همچنین می‌توانیم تخمینی از محدوده قابل قبول متغیرهایی که در هنگام استخراج مدل ساده از آنها صرف‌نظر کردیم، داشته باشیم تا با اطمینان بیشتری به طراحی کنترل‌کننده بر اساس مدل ساده‌شده راه‌رونده بپردازیم. برای به‌دست‌آوردن معادلات حرکت پیوسته برای مدل کامل، ابتدا به انجام برخی تعاریف در روابط (۲-۳۴) و بازنویسی معادله حرکت پایه می‌پردازیم تا درانتها بتوانیم معادله حرکت

پایه را به یک سیستم دومتغیره شامل مولفه‌های همگرا و واگرایی مشابه با آنچه برای مدل ساده شده به دست آمد، تبدیل کنیم.

$$\begin{cases} (1)\ \ \omega(t) \triangleq \sqrt{\dfrac{g}{y(t)}} \\[4mm] (2)\ \ z = z_i + \Delta z_i \\ \quad\ \ , \ z_i = cte \\[4mm] (3)\ \ s \triangleq x - z_i \\[4mm] (4)\ \ f_1(t) \triangleq \dfrac{\ddot{y}(t)}{y(t)} \\[4mm] (5)\ \ f_2(t) \triangleq \dfrac{\dot{H}(t)}{M\,y(t)} \end{cases} \Rightarrow$$

$$z = x + \frac{\dot{H}/M - y\ddot{x}}{g + \ddot{y}}$$
$$\downarrow$$
$$\ddot{x} - \frac{g + \ddot{y}}{y}(x - z) = \frac{\dot{H}}{M\,y} \qquad (۳۴-۲)$$
$$\downarrow$$
$$\ddot{s}_i - \left(\omega(t)^2 + f_1(t)\right) s_i = f_2(t) - \left(\omega(t)^2 + f_1(t)\right)\Delta z_i$$

این تبدیل با نوشتن معادله حرکت در فضای حالت، رابطه (۳۵-۲)،

$$\ddot{s}_i - (\omega^2 + f_1)\, s_i = f_2 - (\omega^2 + f_1)\Delta z_i$$

$$\overset{STATE}{\underset{SPACE}{\Longrightarrow}} \begin{bmatrix} \dot{s}_i \\ \ddot{s}_i \end{bmatrix} = \begin{bmatrix} 0 & 1 \\ \omega^2 + f_1 & 0 \end{bmatrix}\begin{bmatrix} s_i \\ \dot{s}_i \end{bmatrix} + \begin{bmatrix} 0 \\ f_2 - (\omega^2 + f_1)\Delta z_i \end{bmatrix} \qquad (۳۵-۲)$$

و استفاده از نگاشت تبدیل (۱۰-۲) به صورت رابطه (۳۶-۲) انجام می‌شود که ما را به رابطه نهایی (۳۷-۲) و یا (۳۸-۲) خواهد رساند.

$$\begin{bmatrix} \dot{p}_i \\ \dot{q}_i \end{bmatrix} = T \begin{bmatrix} 0 & 1 \\ \omega^2 + f_1 & 0 \end{bmatrix} T^{-1} \begin{bmatrix} p_i \\ q_i \end{bmatrix} + T \begin{bmatrix} 0 \\ f_2 + (\omega^2 + f_1)\Delta z_i \end{bmatrix} \qquad (۳۶-۲)$$

$$\begin{bmatrix} \dot{p}_i \\ \dot{q}_i \end{bmatrix} = \begin{bmatrix} -\omega - \dfrac{f_1}{2\omega} & -\dfrac{f_1}{2\omega} \\[3mm] \dfrac{f_1}{2\omega} & \omega + \dfrac{f_1}{2\omega} \end{bmatrix}\begin{bmatrix} p_i \\ q_i \end{bmatrix} + \begin{bmatrix} -\dfrac{f_2}{\omega} + \left(\omega + \dfrac{f_1}{\omega}\right)\Delta z_i \\[3mm] \dfrac{f_2}{\omega} - \left(\omega + \dfrac{f_1}{\omega}\right)\Delta z_i \end{bmatrix} \qquad (۳۷-۲)$$

$$\begin{matrix} Continuous\ Part \\ of \\ Motion\ Equation \end{matrix} : \begin{cases} \dot{p}_i + \left(\omega + \dfrac{f_1}{2\omega}\right)p_i = -\dfrac{f_1}{2\omega}q_i - \dfrac{f_2}{\omega} + \left(\omega + \dfrac{f_1}{\omega}\right)\Delta z_i \\[4mm] \dot{q}_i - \left(\omega + \dfrac{f_1}{2\omega}\right)q_i = \ \ \dfrac{f_1}{2\omega}p_i + \dfrac{f_2}{\omega} - \left(\omega + \dfrac{f_1}{\omega}\right)\Delta z_i \end{cases} \qquad (۳۸-۲)$$

برای بررسی انحراف مورد نظر، معادله حرکت کامل (۲-۳۸) را برای توابعی که می‌توانند نامزد احتمالی $f_1$ و $f_2$ باشند، شبیه‌سازی می‌کنیم. اندازه حرکت خطی در راستای عمودی و اندازه حرکت زاویه‌ای، $M\dot{y}(t)$ و $H(t)$، با توجه به تناوبی بودن حرکت راه رفتن در حالت دائمی، توابعی هارمونیک خواهند بود که مود اول بسط فوریه آنها تابعی سینوسی با دوره تناوبی به اندازه بازه زمانی یک گام از سیکل حرکتی ساده می‌باشد. شکل ۲-۱۸ و شکل ۲-۱۹، شبیه‌سازی معادله حرکت کامل برای شرایط اولیه مشابهی با شرایط اولیه شکل ۲-۱۱ را نشان می‌دهد. در این شبیه‌سازی توابع $y(t)$ و $H(t)$ که تشکیل دهنده $f_1$ و $f_2$ هستند، به صورت جدول ۲-۴ درنظرگرفته‌شده اند. پارامترهای این توابع به گونه‌ای درنظرگرفته شده است که منحنی‌ها شبیه به منحنی‌های اندازه‌گیری از حرکت انسان در مراجع [۱۲] ، [۳۰] ، [۳۱] و [۳۲] و برای حرکت‌هایی با بیشترین دامنه ممکن باشد. به عنوان مثال، دامنه حرکت انداز حرکت زاویه‌ای، $A_H$، برابر با $0.1\,M\bar{V}h$ درنظرگرفته‌شده است که مربوط به حرکت راه‌رفتن مبالغه آمیز [^1] می‌باشد و مقدار آن برای راه‌رفتن معمولی از $0.05\,M\bar{V}h$ تجاوز نمی‌کند[۱۲]. همچنین، حالت‌های (۱) و (۳) برای حرکت‌هایی شبیه راه‌رفتن انسان و حالت‌های (۲) و (۴) برای حرکت‌هایی دقیقا عکس حرکت انسان( با ۱۸۰ درجه اختلاف فاز) و جهت بررسی حرکت‌هایی دور از انتظار درنظرگرفته شده‌اند.

در حالی که شرایط اولیه درنظرگرفته شده و طول و زمان فرود گام مربوط به سیکل حرکتی ساده است و به بسته شدن یک سیکل حرکتی ساده در شکل ۲-۱۱ برای مدل حرکتی ساده‌شده منجر شده است، برای مدل حرکتی کامل در شکل ۲-۱۸ و شکل ۲-۱۹، منجر به شرایط نهایی متفاوتی از مدل ساده می‌شود که با برداشتن گام، شرایط اولیه گام بعدی دیگر همانند شرایط اولیه گام پیموده شده نخواهد بود و منجر به باز شدن سیکل حرکتی می‌شود. با این‌حال، مسیر حرکتی مدل کامل با وجود انحراف از مدل ساده‌شده، هنوز شباهت زیادی به حرکت مدل ساده به خصوص به سیکل‌های با سرعت متوسط بالاتر را داراد که مسئله پایداری در آنها مهم تر است. بنابراین به احتمال زیاد می‌توان مدل ساده شده را مبنای خوبی برای طراحی کنترل‌کننده پایداری در فصل بعدی قرار داد که البته با شبیه‌سازی کنترل‌کننده‌های طراحی شده بر روی یک مدل فیزیکی کامل، درستی این فرض محک زده خواهد شد.

جدول ۲-۴. تعریف توابع نامزد برای مدل حرکتی کامل راه‌رفتن

| | |
|---|---|
| $y(t) = h + A_y\,sin\left(\dfrac{2\pi t}{T_c} + \phi_y\right)$<br>$H(t) = 0$<br>$\Delta z_i = 0$ | $(1):\ A_y\ =\ 0.025\ ,\quad \phi_y = -\dfrac{\pi}{2}$ |
| | $(2):\ A_y\ =\ 0.025\ ,\quad \phi_y = \dfrac{\pi}{2}$ |
| $y(t) = h$<br>$H(t) = A_H\,sin\left(\dfrac{2\pi t}{T_c} + \phi_H\right)$<br>$\Delta z_i = 0$ | $(3):\ A_H\ =\ 0.2\,MV_ch\ ,\quad \phi_H = 0$ |
| | $(4):\ A_H\ =\ 0.2\,MV_ch\ ,\quad \phi_H = \pi$ |

[^1]: Exaggerated Walking

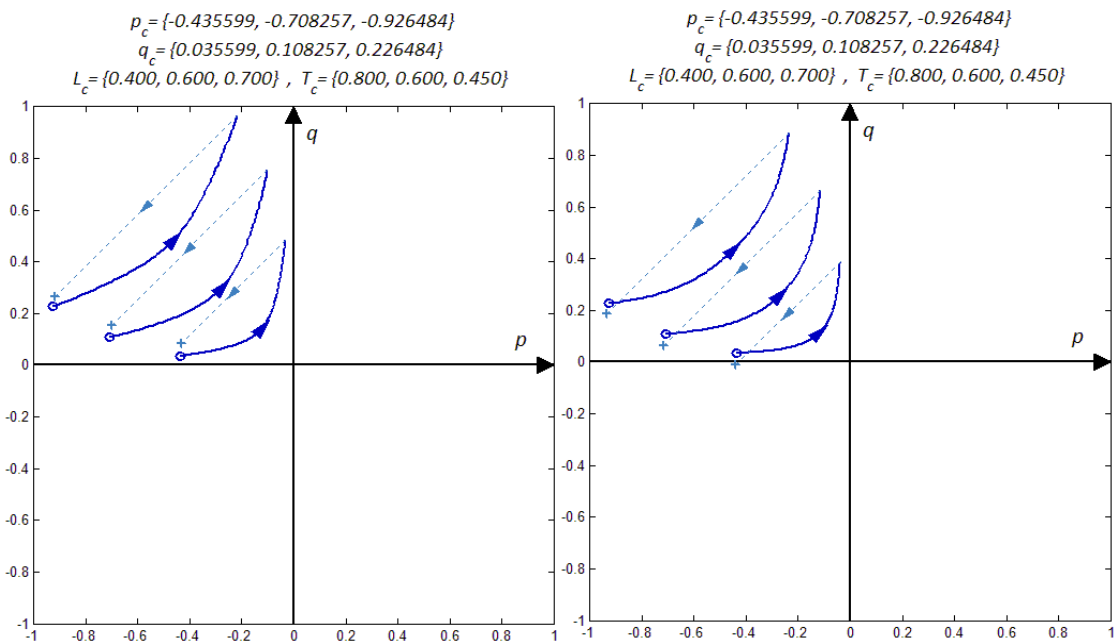

ب- سه نمونه از حرکتی پیشرونده برای توابع حالت (2)      الف- سه نمونه از حرکتی پیشرونده برای توابع حالت (1)

شکل ۲-۱۸. صفحه فاز مسیرهای حرکت مولفه واگرا نسبت به مولفه همگرا برای مدل حرکتی کامل بر روی شرایط اولیه سیکل‌های حرکتی ساده

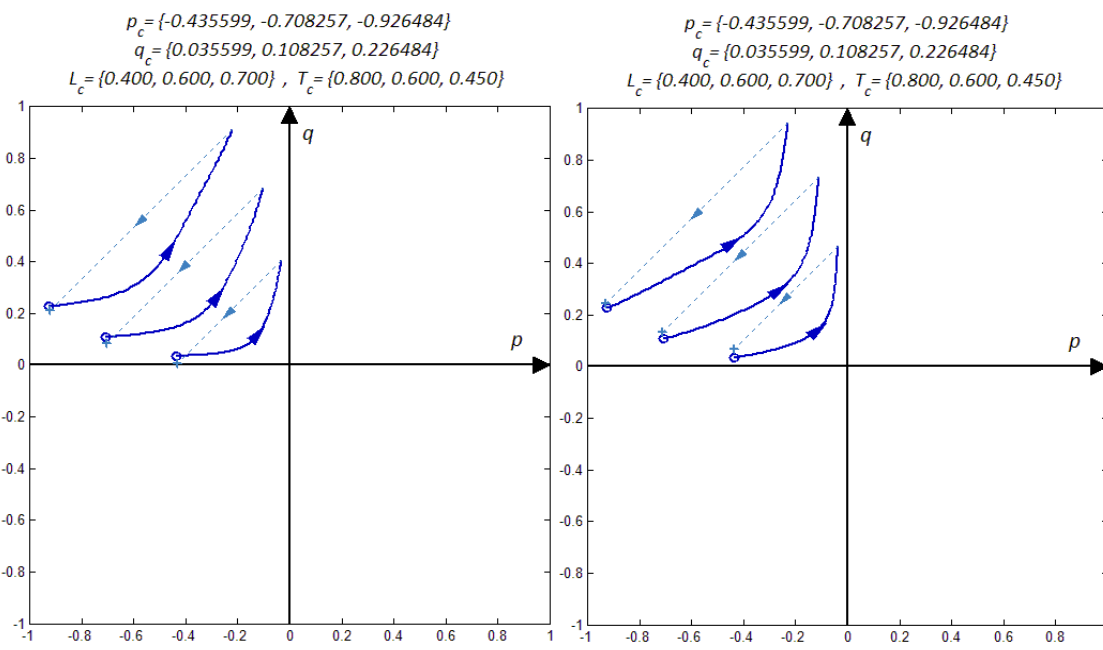

ب- سه نمونه از حرکتی پیشرونده برای توابع حالت (2)      الف- سه نمونه از حرکتی پیشرونده برای توابع حالت (3)

شکل ۲-۱۹. صفحه فاز مسیرهای حرکت مولفه واگرا نسبت به مولفه همگرا برای مدل حرکتی کامل بر روی شرایط اولیه سیکل‌های حرکتی ساده

## ۷-۲ تعریف مدل‌های ریاضی ساده‌شده(SWM) و کامل (CWM) راه‌رفتن

در این بخش سعی خواهیم کرد با نگاهی سیستمی به جمع‌بندی معادلات حرکتی که تاکنون در این فصل مطرح و مطالعه شد، بپردازیم و دو مدل ریاضی برای حرکت راه‌رفتن به نام‌های مدل ساده راه‌رفتن $SWM$ [1] و مدل کامل راه‌رفتن $CWM$ [2] ارائه دهیم.

در ابتدای فصل، با فرض برقرار بودن همیشگی تماس با زمین، دانستیم که معادله حرکت پایه (۳-۲) همواره برای هر راه‌رونده‌ای برقرار است. سپس، با اعمال فرض‌هایی نزدیک به واقعیت بر این معادله حرکت پایه، همچون ناچیز بودن حرکت در راستاهای عمودی و دورانی، و با بازتعریف برخی متغیرهای جدید به صورت نگاشتی خطی بر روی متغیرهای قبلی، مدلی ساده برای حرکت طبیعی راه‌رفتن(دینامیک صفر) به صورت معادله حرکت (۱۲-۲) استخراج شد که در آن به صورت پیش‌فرض یکی از ورودی‌ها $\Delta z_i = 0$ درنظرگرفته شد تا دینامیک صفر این مدل ساده‌شده(حرکت طبیعی [3]) به دست آید. (استفاده از این معادله حرکت ساده‌شده با ورودی $\Delta z_i = 0$ به ما کمک کرد تا معادله حرکت گام به‌گام را برای مجموع بخش پیوسته و گسسته حرکت استخراج کنیم و به مفاهیمی همچون سیکل‌های حرکتی و پایداری آن‌ها بپردازیم.)

حال اگر بخواهیم بخش پیوسته معادلات حرکت ساده‌شده راه‌رفتن را با در نظر گرفتن متغیر جابجایی فشار غیرصفر، $\Delta z_i \neq 0$، که در حالت کلی برای همه راه‌رونده‌ها (به غیر از راه‌رونده‌های دارای کف پای نقطه‌ای یا دوپای کم‌عملگر) می‌تواند غیر صفر باشد، بازنویسی کنیم، به رابطه (۳۹-۲) خواهیم رسید. این رابطه با قرار دادن $f_1 = f_2 = 0$ در معادله حرکت کامل (۳۸-۲) نیز، قابل استخراج است.

$$z = z_i + \Delta z_i \Rightarrow \ddot{x} - \omega^2(x - z_i) = -\omega^2 \Delta z_i \Rightarrow \begin{cases} \dot{p}_i + \omega p_i = \omega \Delta z_i \\ \dot{q}_i - \omega q_i = -\omega \Delta z_i \end{cases} \qquad (۳۹-۲)$$

بر این اساس، مدل ریاضی ساده‌شده راه‌رفتن(SWM) را به صورت بلوک دیاگرام شکل ۲-۲۰ معرفی می‌کنیم.

---

[1] Simplified Walking Model
[2] Complete Walking Model
[3] Natural Walking Motion

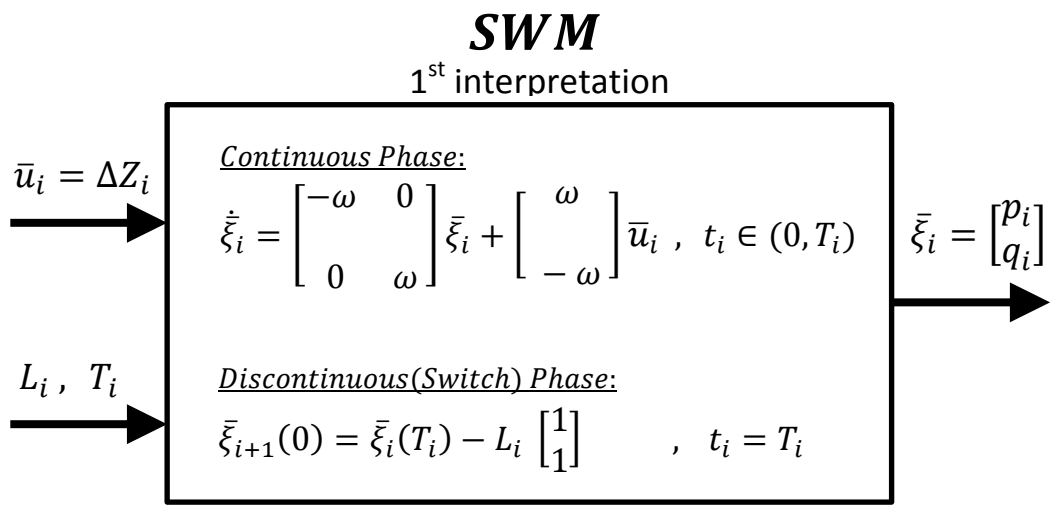

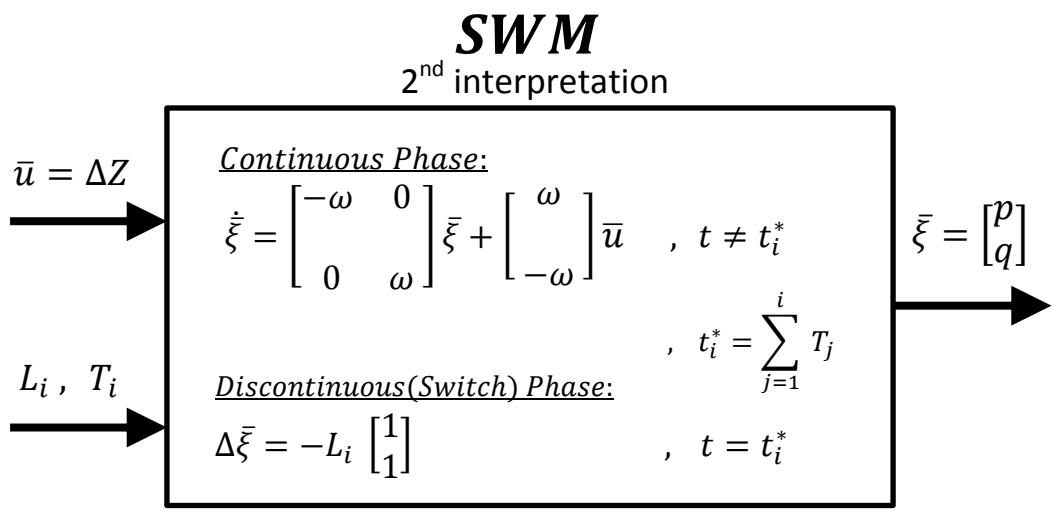

شکل ۲-۲۰. مدل سادهشده راهرفتن ( SWM: Simplified Walking Model )

حال اگر بخواهیم معادلات پیوسته حرکت کامل راهرفتن در رابطه (۲-۳۸) را درنظر بگیریم، باید توجه کنیم که فرضیاتی که در مدل ساده انجام شد از جمله قابل صرفنظر بودن حرکت در راستای دو درجه آزادی عمودی و دورانی، دیگر برای این مدل معتبر نخواهد بود و باید معادلات حرکت حاکم بر این دو درجه آزادی، معادلات رابطه (۲-۴۰)، را نیز بر این مدل بیافزاییم.

$$\begin{cases} \ddot{y} = y\, f_1 \\ \dot{H} = M\, y\, f_2 \end{cases} \qquad\qquad (\text{۲-۴۰})$$

بر این اساس، میتوان مدل ریاضی کامل راهرفتن(CWM) را به صورت بلوک دیاگرام شکل ۲-۲۱ معرفی کرد. در روابط این مدل، $\tilde{H}$ انتگرال اندازهحرکت زاویهای به صورت $\tilde{H} = \int_0^t H(\tau)\,d\tau$ میباشد که معرف وضعیت کلی قرارگیری دورانی پیکره بندی است.

۵۸

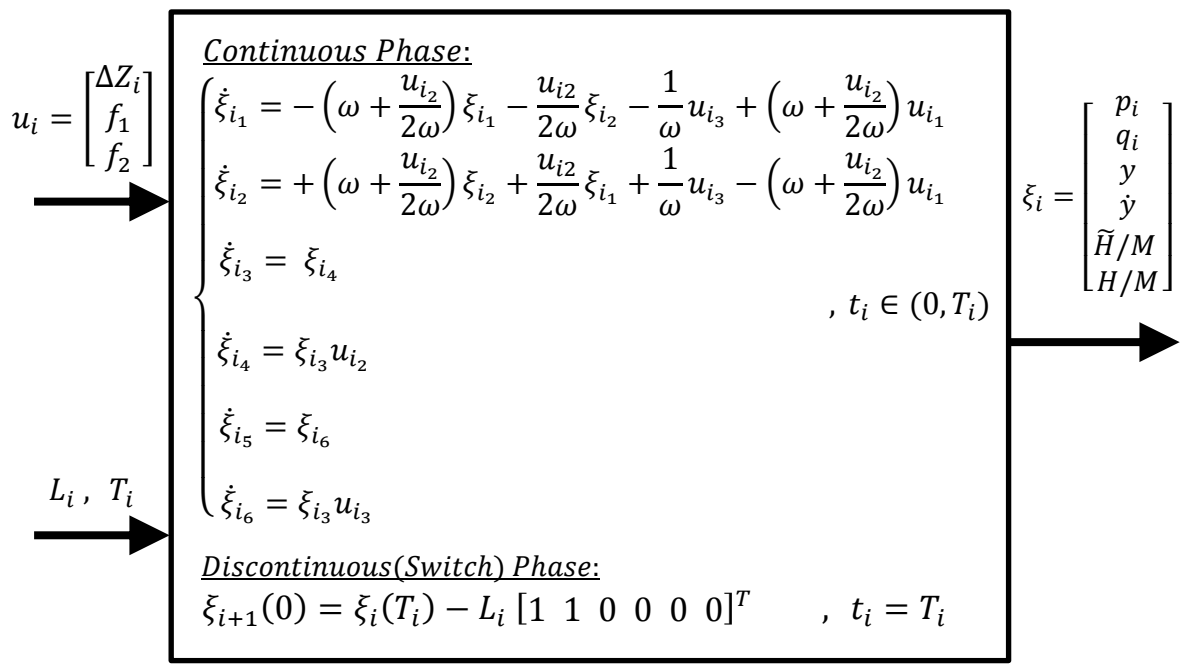

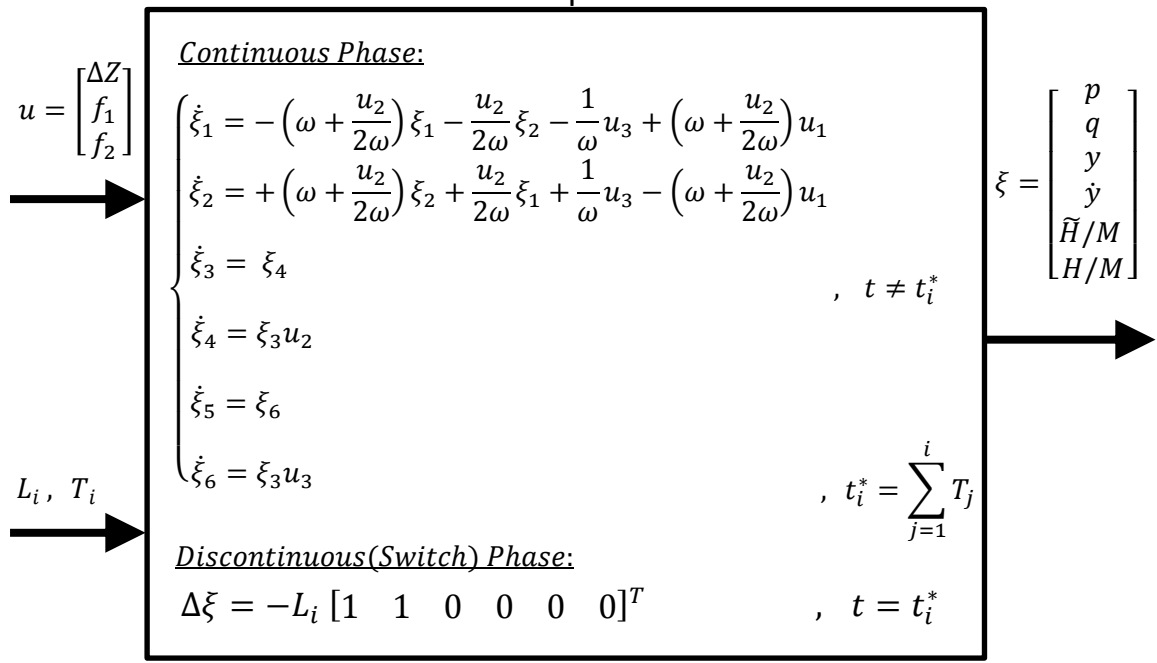

شکل ۲-۲۱. مدل کامل راه‌رفتن ( CWM: Complete Walking Model )

**فصل سوم**
**کنترل پایداری**

بررسی معادله حرکت حاکم بر راهروندهها و به ویژه دوپای راهرونده در فصل پیش، نشان داد که مـدل فیزیکـی حاکم بر آنها ، چه برای مدل ساده راهرفتن(SWM) و چه برای مدل کامل راهرفتن(CWM)، یک سیسـتم دینامیکـی تکهتکه پیوسته[1] است که نوعی سیستم دینامیکی هیبرید[2] است. گسستگی این مدل مربوط به عمل گامبرداشتن است که جزیی جداییناپذیر از حرکت راهروندههاست: برای حرکت راهرفتن در لحظه فـرود آمـدن هـر گـام محـدوده تکیهگاهی تغییر میکند و به تبع آن مرکز فشار جابجا میشود، همچنین برای حرکت دویدن اگرچه موضـوع بحـث تحقیق فعلی نیست، میتوان گفت که در هر گام برای یک بازه زمانی کوتاه نیروهـای خـارجی وارد شـونده از کـف زمین ناپدید میشوند و سپس با بازگشت این نیروها، محدوده تکیهگاهی و به تبع آن مرکز فشار جابجا میشود.

در بخشهای پیوسته این نوع سیستمها(سیستمهای دینامیکی تکهتکه پیوسته)، معادلات دیفرانسیلی حـاکم اسـت که برای بخش اصلی سیستم دینامیکی مدل کامل راهرونده(CWM)، همان معادله حرکت پیوسته (۲-۳۸) است و در بخشهای گسسته این سیستم، معادلات تفاضلی[3] حاکم است که برای سیستم دینامیکی راهرونـده، همـان معادلـه حرکت گسسته (۲-۱۷) است.

---

در این فصل، ابتدا راهبردهای کنترلی برای پایدارسازی بخش اصلی (مؤلفه‌های همگرا و واگرا) مدل کامل راه‌رونده(CWM) مطالعه خواهد شد. سپس، با مبنا قراردادن مدل ساده‌شده راه‌رفتن (SWM) و معادله حرکت گـام بـه‌گام (۲–۱۹)، که از ترکیب رابطه (۲–۱۲) برای بخش پیوسته مدل ساده‌شـده و رابطـه (۲–۱۷) بـرای بخش گسسـته استخراج شده است، دو دسته کنترل‌کننده پایداری بر اساس دو راهبرد کنترلی مختلف، با عنوان کلی پایدارسازهـای سیکل حرکتی، طراحی می‌شود و پایداری آن‌ها اثبات خواهـد شـد. در ادامـه بـا معرفـی یـک مـدل فیزیکـی کامـل (غیرخطی) و تکمیل نمودن کنترل‌کننده برای بخش‌های دیگر این مدل، به شبیه‌سازی عملکرد پایدارسازهای سیکل حرکتی بر روی این مدل می‌پردازیم. این شبیه‌سازی‌ها، برای صورت مسئله‌های مختلفی از جمله مقاومـت در برابـر اختلال(از نوع ضربه) و نیز برای شرایط اولیه دور از سیکل و با درنظرگرفتن محدودیت‌هایی بر روی حرکـت راه‌رونده انجام می‌شود تا توانایی و نارسایی کنترل‌کننده‌های طراحی شده بررسی شود.

در فصل بعدی، با ارائه روشی جدید بر نارسایی‌های پایدارسازهای سیکل حرکتی فائق خواهیم آمد و به راه‌حلی کامل برای مسئله کنترل پایداری به نام کنترل پایداری بهینه خواهیم رسید که به روش برنامه‌ریزی غیرخطـی[۱] و بـا توجه همزمان به عملکرد پایدارسازهای سیکل حرکتی در حفظ پایداری و رعایت قیود مسئله عمل می‌کند.

## ۳–۱ راهبردهای قابل تصور برای کنترل پایداری

اگر بخواهیم تنها مؤلفه‌های واگرا و همگرای مدل کامل راه‌رونده(CWM) را بدون توجه به دیگر متغیرهای ایـن مدل، تحت کنترل درآوریم و در عین حال به هر دو بخش معادلات حرکت(مربوط به مؤلفه‌هـای واگـرا و همگـرا) در بخش پیوسته آن، معادلات (۲–۳۸)، و در بخش گسسته آن، معادلات (۲–۱۷)، توجه کنیم، متوجه خـواهیم شـد که سه ورودی کنترلی $\Delta z_i$، $f_1$ و $f_2$، در بخش پیوسته و دو ورودی کنترلی $T_i$ و $L_i$، در بخش گسسـته وجـود دارد. با این حال به علت نیاز به استفاده از دو ورودی $f_1$ و $f_2$ برای کنترل حرکت راه‌رونده در راستای دو درجه آزادی حرکت عمودی (سرعت و موقعیت مرکز جرم در راستای عمودی : $y, \dot{y}$) و حرکت دورانـی (انـدازه حرکت زاویه‌ای و وضعیت قرارگیری دورانی پیکربندی : $H, \tilde{H}$)، ترجیح می‌دهیم از این ورودی‌هـا بـرای ایـن اهداف به جای استفاده در کنترل پایداری (مؤلفه‌های واگرا و همگرای حرکت) استفاده کنیم و تـأثیر آن‌هـا را در بخش پیوسته معادلات حرکت (۲–۳۸) به صورت اغتشاش درنظر بگیریم.

با این حساب در بخش پیوسته معادلات حرکت (۲–۳۸)، تنهـا، ورودی کنترلـی متغیـر جابجـایی مرکـز فشـار، $\Delta z_i$، باقی می‌ماند که استفاده از آن را با عنوان راهبرد تغییر مرکز فشار در بازه زمانی هر گام نامگذاری مـی‌کنیم و در بخش گسسته معادلات حرکت، دو ورودی کنترلی طول و زمان فرود گام، $T_i$ و $L_i$، را در دست خواهیم داشت که استفاده از آن را، راهبرد تغییر گام با برداشتن گام‌هایی با طول و زمان فـرود متغیـر مـی‌نـامیم. علاوه بـر ایـن دو،

---

راهبرد سومی نیز برای کنترل پایداری می‌تواند وجود داشته باشد که تلفیقی از این دو راهبرد است، بـا ایـن‌حـال در این تحقیق به بررسی آن نخواهیم پرداخت و مطالعه آن را برای ادامه کار پیشنهاد مـی‌کنـیم. بـا توجـه بـه نامشـخص بودن ورودی‌های $f_1$ و $f_2$ به طور کلی، و بـا علـم بـه اینکـه حرکـت در راسـتای عمـودی و دورانـی تکرارشـونده و نزدیک به صفر است، از $f_1$ و $f_2$ در بخش طراحی کنترل‌کننده پایداری صرف‌نظر مـی‌کنـیم و بنـابراین مـدل سـاده راهرونده(SWM) ، معادله حرکت (۲–۳۹)، را مبنای طراحی کنترل‌کننده پایداری در این فصل قرار می‌دهیم.

### ۳-۱-۱ کنترل پایداری حول سیکل حرکتی

همان‌طور که در فصل قبل دیدیم، اگر راهروندهای بخواهد حرکتی دائمی را تجربه کند، باید حرکتی نزدیک بـه یک سیکل حرکتی از خود بروز دهد به این معنی کـه سـیکل‌هـای حرکتـی کـه از مـدل سـاده‌شـده در فصـل قبلـی استخراج شد، معیار خوبی به عنوان یک مسیر حرکت دایمی مطلوب بـرای راهرونـده مـی‌باشـد، ولـی همـان‌طور کـه دیدیم این سیکل‌ها تنها دارای پایداری مرزی هستند و در عمل بسیار ناپایدارند(واگرایی نمایی). به همین جهـت، اگر شرایط مولفه واگرا و همگرا نزدیک به مسیر یک سیکل حرکتی باشد و بتوان کنترل‌کننده‌ای یافت کـه حرکـت راهرونده را به سمت این سیکل حرکتی همگرا کند، نوعی کنترل‌کننده پایداری خواهیم داشت کـه آن را پایدارسـاز سیکل حرکتی می‌نامیم. پایدارسازهای سیکل حرکتی که در این فصل پیشنهاد خواهد شد، از هـر دو راهبـرد ممکـن استفاده می‌کنند و به همین جهت به دو دسته تقسیم‌بندی شده‌اند.

- **جابجایی مرکز فشار جهت تعقیب مسیر در بازه زمانی هر گام**

این دسته از کنترل‌کننده‌های پیشنهادی بخش پیوسته معادلات حرکت را به نحوی تحت کنترل درمـی‌آورد کـه معادلات حرکت گسسته نیز پایدار شوند. در این دسته، یک کنترل‌کننده پیشنهاد می‌شود که بـا اسـتفاده از جابجـایی مرکز فشار در بازه زمانی هر گام سعی می‌کند حرکت راهرونده را در بخش پیوسته معـادلات حرکـت همگـرا و در نهایت حرکت کلی را به سمت یک سیکل حرکتی دلخواه همگرا کند. باید توجه داشت این کنترل‌کننده با توجه بـه الزامش در استفاده از محدوده‌ای برای اعمال تغییرات در موقعیت مرکز فشار در طـول زمـان، بـرای راهرونـده‌هـایی کاربرد دارد که دارای کف پا هستند و برای دوپاهای کم‌عملگر که دارای تماس نقطه بـا زمـین هسـتند، بـی‌کـاربرد است.

- **تغییر طول یا زمان گام و یا هر دو برای جبران انحراف از مسیر**

این دسته از کنترل‌کننده‌های پیشنهادی که در واقع بخش گسسته معادلات حرکت را تحت کنترل درمـی‌آورنـد، از نوع کنترل‌کننده‌های تحریک یک[1] محسوب می‌شوند.

---

[1] Impulsive Control

در این دسته طراحی کنترل‌کننده بر راهبرد تغییر گام استوار است و در آن سه روش برای پایدارسازی سیکل حرکتی پیشنهاد می‌شود که هر یک با محاسبه طول گام، یا زمان فرود گام، و یا هر دو، بخش گسسته معادلات حرکت را پایدار می‌سازند.

مزیت این کنترل‌کننده‌ها نسبت به دسته قبلی در این است که به جای کنترل پیوسته حرکت در طول بازه‌های زمانی هر گام، پایداری کلی حرکت را با اعمال کنترل تنها در لحظات فرود گام تضمین می‌کنند و می‌توانند خطاهای بوجود آمده دربخش پیوسته معادلات حرکت را در صورت محدود بودن جبران‌کنند. همچنین فرض صفر بودن متغیر جابجایی مرکز فشار، $\Delta z_i$، در این روش و یا به عبارت بهتر، عدم استفاده از این متغیر که پیش‌نیاز استفاده از آن داشتن کف پا است، این دسته از کنترل‌کننده‌ها را بدون هیچ تغییری برای کنترل پایداری راه‌روندهای دوپای کم‌عملگر[1] قابل استفاده می‌کند. علاوه بر آن، همانطور که خواهیم دید، توانایی این دسته از کنترل‌کننده‌ها برای پایدارسازی بازه گسترده‌تری از شرایط اولیه‌ای دلخواه، بسیار بیشتر از دسته اول است و بنابراین می‌توان گفت پایدارسازهای با راهبرد تغییر گام، به طور کلی، مقاومت بیشتری از خود در مقابل تغییرات شرایط اولیه هرگام نشان می‌دهند. در مقابل، روش ارائه شده در دسته اول، به علت اعمال کنترل در بخش پیوسته معادلات و حاضر بودن در طول بازه زمانی هرگام، دارای مزیتی نسبی به روش‌هایی است که در دسته دوم قرار می‌گیرند، چراکه نسبت به اغتشاشات خارجی از جمله ضربه که ممکن است در طول گام به راه‌رونده وارد شود، حساس‌تر است و سرعت همگرایی بیشتری دارد. به همین جهت، می‌توان گفت دسته اول در برابر اغتشاشات خارجی محدود تا حدی مقاوم‌تر است. علت واضح این امر، در نظرگرفتن انحراف از مسیر سیکل حرکتی ناشی از ضربه و سعی در برطرف کردن اثر آن در طول بازه زمانی گام و نه در انتهای گام است.

### ۳-۱-۲ برنامه ریزی غیرخطی[2] برای سوق دادن مسیر به سمت یک سیکل حرکتی دلخواه

اگر بخواهیم به محدودیت‌های راه‌رونده واقعی بپردازیم، ناگزیر از جستجو در فضای جواب‌های همگرا در این محدوده خواهیم بود. در این روش که در فصل آتی به آن می‌پردازیم، سعی خواهیم کرد با توجه به دانش بدست آمده در بخش پایدارسازهای سیکل‌های حرکتی، با تعریف تابع شاخص پایداری و کمینه‌کردن آن با در نظر گرفتن قیود حرکتی، جوابی بهینه برای هدایت مسیر حرکت به سمت یک سیکل حرکتی دلخواه و کنترل پایداری حول آن سیکل بیابیم. روش برنامه‌ریزی خطی کنترل پایداری یا کنترل پایداری بهینه، مبتنی بر راهبرد دوم کنترل پایداری است که در آن استفاده از برداشتن گام‌هایی با طول و زمان فرود متغیر برای پایدارسازی حرکت مورد توجه است.

---

## ۳-۲  پایدارسازهای سیکل حرکتی

بررسی پایداری سیکل‌های حرکتی در فصل قبل نشان داد که سیکل‌های حرکتی دارای پایداری مرزی هستند و در عمل با کوچک‌ترین اختلالی در شرایط اولیه مولفه واگرای حرکت، ناپایدار خواهد شد. از آنجا که دیدیم شرایط اولیه مولفه همگرای حرکت، پس از برداشتن گام‌های متوالی با طول و زمان مربوط به یک سیکل همواره به سـمت مقدار مطلوب خود همگرا می‌شود، در این بخش، چهار کنترل‌کننده پیشنهاد خواهد شد تا شرایط اولیه مولفه واگـرا و در نهایت کل حرکت را پایدار کند. کنترل‌کننده اول بر اساس راهبرد تغییر مرکز فشار اسـت و سـه کنترل‌کننده بعدی بر اساس راهبرد تغییر طول و زمان گام می‌باشند. در هر قسمت، پایداری، محدوده ضرایب و محدوده پایداری را بررسی خواهیم کرد. در این فصل، با وجود اینکه برای بررسی همگرایی کنترل‌کننده‌هـایی کـه بـر روی معادلـه گسسته گام به‌گام طراحی می‌شوند می‌توان از قضایای پایداری معادلات تفاضلی [1] که در مبحث کنترل دیجیتال نیـز مطرح می‌شوند، استفاده نمود، ترجیح می‌دهیم برای درک مستقیم و راحتی بیشتر از مبانی پایه‌ای ریاضی هماننـد همگرایی دنباله‌های عددی استفاده‌شده در این منابع برای اثبات پایداری استفاده کنیم [۳۳] و [۳۴].

### ۳-۲-۱  قضیه محدود ماندن شرایط اولیه مولفه همگرا پس از برداشتن گام‌های دلخواه پی‌درپی

از آنجا که تمرکز کنترل کننده‌های پیشنهادی بر کنترل مولفه واگرای حرکت می‌باشد، برای اثبات پایداری کلی حرکت در صورت اعمال کنترل‌کننده‌های پیشنهادی، نیاز به بررسی محدود ماندن شرایط اولیه مولفه همگرا در حالت کلی با حضور اختلال و پیمودن گام‌های دلخواه خواهیم داشت تا از همگرایی دینامیک داخلی حرکـت مطمئن شویم. در این قسمت به بررسی دو ویژگی شرایط اولیه مولفه همگرا، $p_{i,0}$، پس از برداشتن تعـدادی گـام دلخواه می‌پردازیم.

بخش همگرای معادله حرکت گام به‌گام در صورت وجود اختلال می‌تواند به صورت رابطه (۳-۱) بیان‌شود:

$$p_{i+1,0} = p_{i,0}\, e^{\omega T_i} - L_i + \Delta p_i \qquad\qquad (۳-۱)$$

که $\Delta p_i$ در آن، اختلال ایجاد شده است کـه مـی‌توانـد در اثـر عوامـل مختلفـی ماننـد اغتشـاش، نیـروی خـارجی و انحراف از معادله حرکت ساده باشد. متغیر $p^*_{i,0}$ توسط رابطه (۳-۲) تعریف می‌کنیم و آن‌را مولفه همگرای شاخص برای گام $i$أم می‌نامیم.

$$p^*_{i,0} \triangleq \frac{-L_i + \Delta p_i}{1 - e^{-\omega T_i}} \qquad\qquad (۳-۲)$$

---

ویژگی اول $p_{i,0}$، تعقیب مسیر حرکت متغیر $\boldsymbol{p}_{i,0}^*$ است به این معنی که $p_{i,0}$ در گام بعـدی، $p_{i+1,0}$، همـواره بـه $\boldsymbol{p}_{i,0}^*$ نزدیک‌تر خواهد شد. حال اگر مقدار متغیر $\boldsymbol{p}_{i,0}^*$ تثبیت شود و یا تغییرات آن بسیار محـدود گـردد، مـی‌تـوان نتیجه‌گیری کرد که $p_{i,0}$ به سمت $\boldsymbol{p}_{i,0}^*$ همگرا خواهدشد.

ویژگی دوم $p_{i,0}$، محدود ماندن آن در بازه مشخصی است که توسط رابطه (۳-۳) قابل بیان است.

$$\begin{cases} |L_i| < L_{max} \\ T_i > T_{min} \\ \Delta p_{min} < \Delta p_i < \Delta p_{max} \end{cases} \tag{۳-۳}$$

$$\Rightarrow \min\left(\frac{-L_{max} + \Delta p_{min}}{1 - e^{-\omega T_{min}}}\ ,\ p_{1,0}\right) \leq\ p_{i,0}\ \leq \max\left(\frac{L_{max} + \Delta p_{max}}{1 - e^{-\omega T_{min}}}\ ,\ p_{1,0}\right)$$

در این رابطه با توجه به محدودیت عملی در حرکت، $L_{max}$، کران بالای اندازه طـول گـام، $T_{min}$، کـران پـایین زمان گام‌برداشتن و $\Delta p_{min}$ و $\Delta p_{max}$، به ترتیب حداقل و حداکثر اختلال وارد شده در کل گام‌های سپری شده می‌باشند.

برای اثبات این مدعا، ابتدا با بازنویسی رابطه با توجه به تعریف $\boldsymbol{p}_{i,0}^*$ به رابطه (۴-۳) خواهیم رسید.

$$p_{i+1,0} = p_{i,0} - k_i\left(p_{i,0} - \boldsymbol{p}_{i,0}^*\right)\ ,\ 0 < k_i = e^{-\omega T_i} < 1$$

$$\Rightarrow \begin{cases} (1):\ p_{i+1,0} - p_{i,0} = -k_i\left(p_{i,0} - \boldsymbol{p}_{i,0}^*\right) \\ (2):\ p_{i+1,0} - \boldsymbol{p}_{i,0}^* = \alpha_i\left(p_{i,0} - \boldsymbol{p}_{i,0}^*\right)\ ,\ 0 < \alpha_i = 1 - k_i < 1 \end{cases} \tag{۴-۳}$$

این رابطه در واقع یک معادله تفاضلی ناهمگن[1] و بیانگر یک فیلتر دیجیتال[2] درجه اول با ثابت زمانی متغیر اسـت کـه در آن $p_{i,0}$ همواره سعی می‌کنـد $\boldsymbol{p}_{i,0}^*$ را تعقیب کند. این واقعیـت از نتـایج (1) و (2) حاصـل از ایـن معادلـه تفاضلی به صورت روابط (۵-۳) قابل بیان است که نشان می‌دهد $p_{i+1,0}$ همواره بین مقدار قبلی خود یعنـی $p_{i,0}$ و متغیر $\boldsymbol{p}_{i,0}^*$ قرارمی‌گیرد.

---

[1] Nonhomogeneous Difference Equation
[2] Digital Filter

$$Variation \atop from \ p_{i,0} \atop to \ p_{i+1,0} : \begin{cases} p_{i,0} < \pmb{p}_{i,0}^* : \begin{cases} \overset{(1)}{\Rightarrow} p_{i,0} < p_{i+1,0} \\ \overset{(2)}{\Rightarrow} p_{i+1,0} < \pmb{p}_{i,0}^* \end{cases} or \quad p_{i,0} < p_{i+1,0} < \pmb{p}_{i,0}^* \\ \\ p_{i,0} > \pmb{p}_{i,0}^* : \begin{cases} \overset{(1)}{\Rightarrow} p_{i,0} > p_{i+1,0} \\ \overset{(2)}{\Rightarrow} p_{i+1,0} > \pmb{p}_{i,0}^* \end{cases} or \quad p_{i,0} > p_{i+1,0} > \pmb{p}_{i,0}^* \end{cases} \tag{۳-۵}$$

این نتیجه بیانگر همان ویژگی اول است که شرایط اولیه مولفه همگرا همواره مقدار شاخص خود را تعقیب می‌کند، همچنین، در صورت ثابت شدن $\pmb{p}_{i,0}^*$، $p_{i,0}$ با نرخ نزولی $\alpha_i$ به سمت آن همگرا می‌شود.

ویژگی دوم یا رابطه (۳-۳) نیز، با تعمیم روابط نامساوی به‌دست‌آمده در رابطه (۳-۵) و همچنین در نظر گرفتن کران بالا و پایین $\pmb{p}_{i,0}^*$ که توسط رابطه (۳-۶) قابل بیان است، نتیجه‌گیری می‌شود.

$$\frac{-L_{max} + \Delta p_{min}}{1 - e^{-\omega T_{min}}} < \pmb{p}_{i,0}^* < \frac{L_{max} + \Delta p_{max}}{1 - e^{-\omega T_{min}}} \tag{۳-۶}$$

### ۳-۲-۲ پایدارسازی حرکت با تغییر پیوسته موقعیت مرکز فشار در طول بازه زمانی هرگام

اگر بخواهیم مرکز فشار را در طول بازه زمانی هرگام تغییر دهیم، متغیر جابجایی مرکز فشار، $\Delta z_i$، باید غیرصفر درنظر گرفته شود و نیاز به استفاده از بخش پیوسته معادله حرکت مدل ساده شده راه‌رفتن، (۲-۳۹) یا (۳-۷) داریم.

$$\begin{cases} \dot{p}_i + \omega p_i = \ \ \omega \Delta z_i \\ \\ \dot{q}_i - \omega q_i = -\omega \Delta z_i \end{cases} \tag{۳-۷}$$

اولین کنترل‌کننده پایداری از نوع پایدارساز سیکل حرکتی را به صورت رابطه (۳-۸) پیشنهاد می‌کنیم.

$$Stability \atop Controller \ (1): \begin{cases} L_i = L_c \\ T_i = T_c \\ \Delta z_i = k_i \, e_i \ , \ \ e_i = \left( q_i - q_i^{des} \right) \ , \ \ q_i^{des} = q_c e^{\omega t_i} \end{cases} \tag{۳-۸}$$

این رابطه گام‌هایی پی‌درپی با طول و زمان گامی ثابت و برابر با طول و زمان گام سیکل حرکتی مطلوب را به اجرا می‌گذارد و همزمان متغیر جابجایی مرکز فشار را متناسب با فاصله جبری خطای مولفه واگرا از مسیر مطلوب خود تغییر می‌دهد. کنترل مرکز فشار در طول بازه زمانی هر گام منجر به معادله خطایی درجه اول برای مولفه واگرا به صورت رابطه (۳-۹) خواهد شد که جواب آن به صورت تابع نمایی (۳-۱۰) خواهد بود.

$$q_i^{des} = q_c e^{\omega t_i} \Rightarrow \dot{q}_i^{des} - \omega q_i^{des} = 0$$

$$\dot{q}_i - \omega q_i = -\omega k_i (q_i - q_i^{des}) \xrightarrow[\text{from the equation}]{\text{Subtract } \dot{q}_i^{des} - \omega q_i^{des} = 0} \boldsymbol{\dot{e}_i + \omega(k_i - 1)e_i = 0} \qquad (\text{۳-۹})$$

$$e_{i,t_i} = e_{i,0}\, e^{-\omega(k_i-1)t_i} \qquad (\text{۳-۱۰})$$

در صورت بزرگتر از واحد بودن ضریب بهره کنترل‌کننده مرکز فشار، معادله خطا در هر گام به سمت صفر همگرا می‌شود ولی به دلیل متناهی بودن زمان هرگام، خطا تنها تا لحظه $T_c$ کاهش پیدا می‌کند. با وجود این، پس از برداشتن گام در نهایت منجر به کاهش فاصله شرایط اولیه مولفه واگرا از مقدار مطلوب خود در گام بعدی مطابق با رابطه (۳-۱۱) خواهد شد.

$$\boldsymbol{k_i > 1 \Rightarrow \left(q_{i,T_c} - q_c e^{\omega T_c}\right) = \alpha_i (q_{i,0} - q_c)\;,\;\; 0 < \alpha_i = e^{-\omega(k_i-1)T_c} < 1}$$

$$\begin{cases} (۲-۱۷): q_{i+1,0} = q_{i,T_c} - L_c \\ (۲-۲۳): L_c = q_c(e^{\omega T_c} - 1) \end{cases} \Rightarrow q_{i,T_c} - q_c e^{\omega T_c} = q_{i+1,0} - q_c \Rightarrow \dfrac{q_{i+1,0} - q_c}{q_{i,0} - q_c} = \alpha_i \qquad (\text{۳-۱۱})$$

این رابطه شرط همگرایی کنترل‌کننده‌های پایداری است زیرا طبق رابطه (۳-۱۲)، این شرط در نهایت، همگرایی شرایط اولیه مولفه واگرا به سمت شرایط اولیه واگرای سیکل مطلوب را اثبات می‌کند.

$$\lim_{k \to \infty} |q_{k,0} - q_c| = \lim_{i \to \infty} |q_{1,0} - q_c| \prod_{i=1}^{k-1} \left| \dfrac{q_{i+1,0} - q_c}{q_{i,0} - q_c} \right| = |q_{1,0} - q_c| \lim_{i \to \infty} \prod_{i=1}^{k-1} \alpha_i = 0 \quad (\text{۳-۱۲})$$

همچنین با استفاده از نامعادله از متوسط‌های حسابی-هندسی[1] می‌توان کران بالایی برای نرخ همگرایی مجانبی این کنترل‌کننده به صورت رابطه (۳-۱۳) مشخص کرد.

$$\begin{matrix} AM - GM\ Inequality \\ or \\ Jensen's\ Inequality \end{matrix} : 0 < \prod_{i=1}^{n} \alpha_i < \bar{\alpha}^n = \left( \dfrac{\sum_{i=1}^{n} \alpha_i}{n} \right)^n \Rightarrow \prod_{i=1}^{n} \alpha_i < e^{-n\,Ln(1/\bar{\alpha})} \qquad (\text{۳-۱۳})$$

همگرایی کلی حرکت را می‌توان این گونه نتیجه‌گیری کرد که بخش همگرای معادله حرکت (۳-۷) در واقع معرف یک فیلتر آنالوگ درجه اول با خروجی $p_{i,0}$ بر روی ورودی $\Delta z_i$ است و بنابراین $p_{i,0}$ همواره $\Delta z_i$ را تعقیب می‌کند، به همین جهت $p_{i,0}$ در هر گام حداکثر می‌تواند دارای اختلالی کراندار با کرانی برابر $\Delta p_i = \Delta z_i$ شود که بنابر ویژگی دوم شرایط اولیه مولفه همگرا، $p_{i,0}$ محدود می‌ماند. با توجه به پایداری مجانبی

---

[1] Inequality of Arithmetic and Geometric Means

و محدود ماندن $p_{i,0}$، دینامیک داخلی سیستم نیز پایدار است. بنابراین می‌توان رابطه (۳-۱۴) را نتیجه گرفت $q_{i,0}$ که پایداری کلی حرکت و همگرایی آن به سمت سیکل حرکتی مطلوب را اثبات می‌کند.

$$k_i > 1 \Rightarrow \lim_{i \to \infty} q_{i,0} = q_c \Rightarrow \lim_{i \to \infty} \Delta z_i = 0 \Rightarrow \lim_{i \to \infty} \Delta p_i = 0$$
$$\Rightarrow \lim_{i \to \infty} \boldsymbol{p}_{i,0}^* = \frac{-L_c}{1 - e^{-\omega T_c}} = p_c \Rightarrow \lim_{i \to \infty} p_{i,0} = p_c \tag{۱۴-۳}$$

با توجه به محدودیت ناحیه تکیه‌گاهی، حداکثر انحراف مرکز فشار از مرکز تکیه‌گاه همواره دارای کران بالا و پایین (فاصله مرکز پا تا نوک پا و یا تا پاشنه پا) است که منجر به محدودیت در انتخاب ضریب کنترل‌کننده به صورت رابطه (۳-۱۵) خواهد شد. بدیهی است بزرگ‌ترین ضریبی که از یک بزرگ‌تر است و در این محدوده قرار می‌گیرد، منجر به کمینه کردن $\alpha_i$ می‌شود و بهترین انتخاب خواهد بود زیرا سرعت همگرایی را به حداکثر میزان ممکن خود می‌رساند. همچنین، درصورتی که این محدوده هیچ ضریب بزرگ‌تر از یکی را در خود جا ندهد، کنترل‌کننده واگرا خواهد شد. در واقع پایداری حرکت توسط کنترل‌کننده مرکز فشار قابل تضمین است، اگر و تنها اگر، محدوده شرایط اولیه مولفه واگرا تابع رابطه (۳-۱۶) باشد.

$$\Delta z_{min} < \Delta z_i = k_i(q_{i,0} - q_c) < \Delta z_{max}$$

$$\Leftrightarrow \begin{array}{c} Gain \\ Limits \end{array} : \begin{cases} -\dfrac{\Delta z_{min}}{q_{i,0} - q_c} < k_i < \dfrac{\Delta z_{max}}{q_{i,0} - q_c} \ , \quad q_{i,0} > q_c \\[3mm] -\dfrac{\Delta z_{max}}{q_c - q_{i,0}} < k_i < \dfrac{\Delta z_{min}}{q_c - q_{i,0}} \ , \quad q_{i,0} < q_c \end{cases} \tag{۳-۱۵}$$

$$\begin{cases} k_i > 1 \\ \Delta z_{min} < k_i(q_{i,0} - q_c) < \Delta z_{max} \end{cases} \Leftrightarrow \begin{array}{c} Initial \\ Condition: q_c - \Delta z_{min} < q_{i,0} < q_c + \Delta z_{max} \\ Limits \end{array} \tag{۳-۱۶}$$

- ### نتایج شبیه‌سازی پایدارساز اول بر روی مدل ساده‌شده حرکت

شبیه‌سازی پایدارساز اول بر روی مدل ساده با شرایط اولیه‌ای نزدیک به حد بالایی شرایط اولیه مرزی، انجام شد. منظور از شرایط اولیه مرزی، شرایط اولیه‌ای است که در آن مولفه واگرا در نزدیکی حد بالایی یا پایینی رابطه (۳-۱۶) باشد. همچنین ضریب کنترلی $k_i = 5$ و محدوده متغیر جابجایی مرکز فشار $\Delta z_{max} = 0.11 \ m$ و $\Delta z_{min} = -0.11 \ m$ درنظرگرفته شده است. شکل ۳-۱، صفحه فاز مسیر حرکت مولفه‌های همگرا و واگرا را برای این شبیه‌سازی نمایش می‌دهد. نمودارهای شکل ۳-۲، سرعت و شتاب مرکز جرم و متغیر کنترلی جابجایی مرکز فشار را برای بخش پیوسته معادلات حرکت نمایش می‌دهند و بخش گسسته معادلات حرکت در شکل ۳-۳، به صورت نمودارهای از متغیرهای کنترلی طول و زمان فرود هر گام، شرایط اولیه مولفه‌های همگرا و واگرا در

ابتدای هرگام و سرعت متوسط مرکز جرم در هرگام نمایش داده شده است. سرعت متوسط حرکت مرکز جرم در هر گام، $V_{i,avg}$، توسط رابطه (۳–۱۷) محاسبه شده است.

$$V_{i,avg} = \frac{x_{i,T_i} - x_{i,0}}{T_i} = \frac{\frac{p_{i,T_i} + q_{i,T_i}}{2} - \frac{p_{i,0} + q_{i,0}}{2}}{T_i} = \frac{p_{i,0}(e^{-\omega t} - 1) + q_{i,0}(e^{\omega t} - 1)}{2\,T_i} \qquad (۳–۱۷)$$

مقادیر پارامترهای مربوط به راهرونده، سیکل حرکتی مطلوب و برخی شرایط اولیه حرکت، برای شبیه‌سازی این پایدارساز و پایدارسازهای بعدی بر روی مدل ساده‌شده راهرونده به صورت جدول ۳-۱است. مقادیر سیکل حرکتی مطلوب و پارامترهای راهرونده را نزدیک به حرکت راهرفتن یک انسان با قد متوسط درنظر می‌گیریم. به عنوان مثال، ارتفاع مرکز جرم، ۱ متر و سرعت متوسط سیکل حرکتی، ۱/۲۵ متر برثانیه برابر با ۴/۵ کیلومتر بر ساعت فرض شده است که تقریبا برابر با سرعت متوسط راهرفتن انسان است. همچنین طول قدم در سیکل حرکتی، ۵۰ سانتی‌متر و حداکثر طول هر قدم، ۷۵ سانتی متر فرض شده است.

جدول ۳-۱. پارامترهای مربوط به شبیه‌سازی مدل ساده

| پارامترهای راهرونده | |
|---|---|
| $h = 1\ m$ | ارتفاع متوسط مرکز جرم |
| $g = 9.8\ m.s^{-2}$ | شتاب جاذبه زمین |
| $L_{max} = 0.75\ m$ | حداکثر اندازه طول گام |
| $V_{max} = 3\ m.s^{-1}$ | حداکثر سرعت متوسط قابل دسترس یا حداکثر سرعت نسبی متوسط پای متحرک به بالاتنه |
| $T_0 = 0.05\ s$ | جمع کل زمان لحظات جدا شدن و فرود آوردن پا |
| **پارامترهای سیکل حرکتی مطلوب** | |
| $p_c = -0.7\ m$ | شرایط اولیه مولفه همگرا |
| $q_c = 0.2\ m$ | شرایط اولیه مولفه همگرا |
| $L_c = 0.5\ m$ | طول گام |
| $T_c = 0.4\ s$ | زمان فرود گام |
| $V_c = 1.25\ m.s^{-1}$ | سرعت متوسط حرکت |
| **شرایط اولیه حرکتی** | |
| $x_0 = x_{1,0} = -0.1\ m$ | موقعیت مرکز جرم در ابتدای حرکت |
| $z_0 = -0.2\ m$ | موقعیت پای متحرک در ابتدای حرکت |
| $z_1 = 0$ | موقعیت پای تکیه گاهی در ابتدای حرکت |

$$p_c = -0.7 \ , \ q_c = 0.2 \ , \ T_c = 0.4 \ , \ L_c = 0.5 \ , \ V_c = 1.25$$
$$p_{1,0} = -0.5 \ , \ q_{1,0} = 0.3 \ , \ x_0 = -0.1 \ , \ dx/dt_0 = 1.253$$
$$z_0 = -0.2 \ , \ z_1 = 0 \ , \ \Delta z_{min} = -0.11 \ , \ \Delta z_{max} = 0.11$$

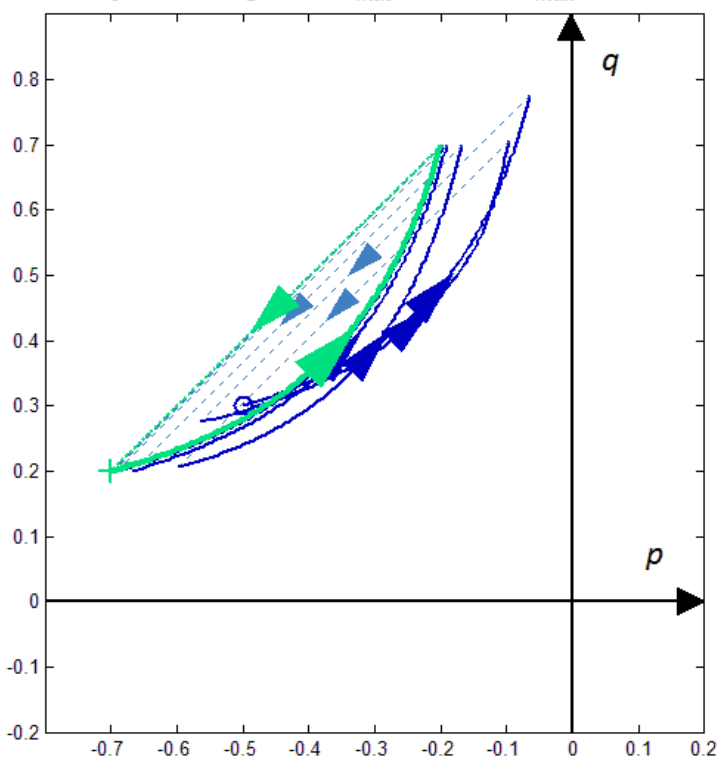

شکل ۳-۱. صفحه فاز مسیر مولفه واگرا نسبت به مولفه همگرا برای شبیه‌سازی پایدارساز اول سیکل حرکتی بر روی مدل ساده راه‌رونده با شرایط اولیه مرزی — منحنی ضخیم و کمرنگ مسیر گام آخر را نمایش می‌دهد.

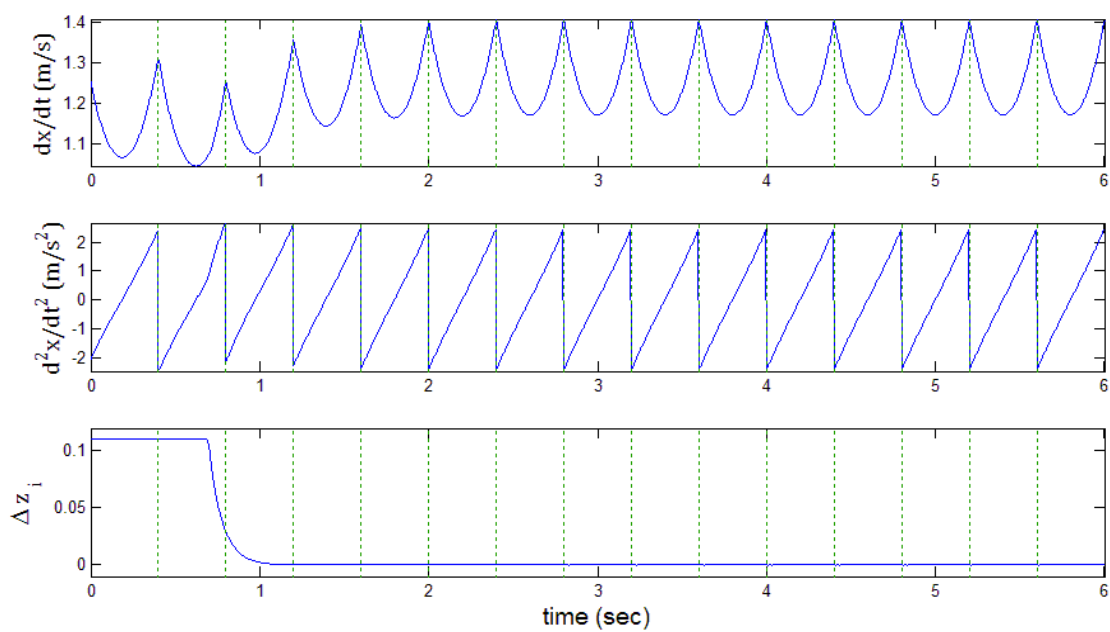

شکل ۳-۲. نمودارهای بخش پیوسته معادلات حرکت برای شبیه‌سازی پایدارساز اول سیکل حرکتی بر روی مدل ساده راه‌رونده با شرایط اولیه مرزی

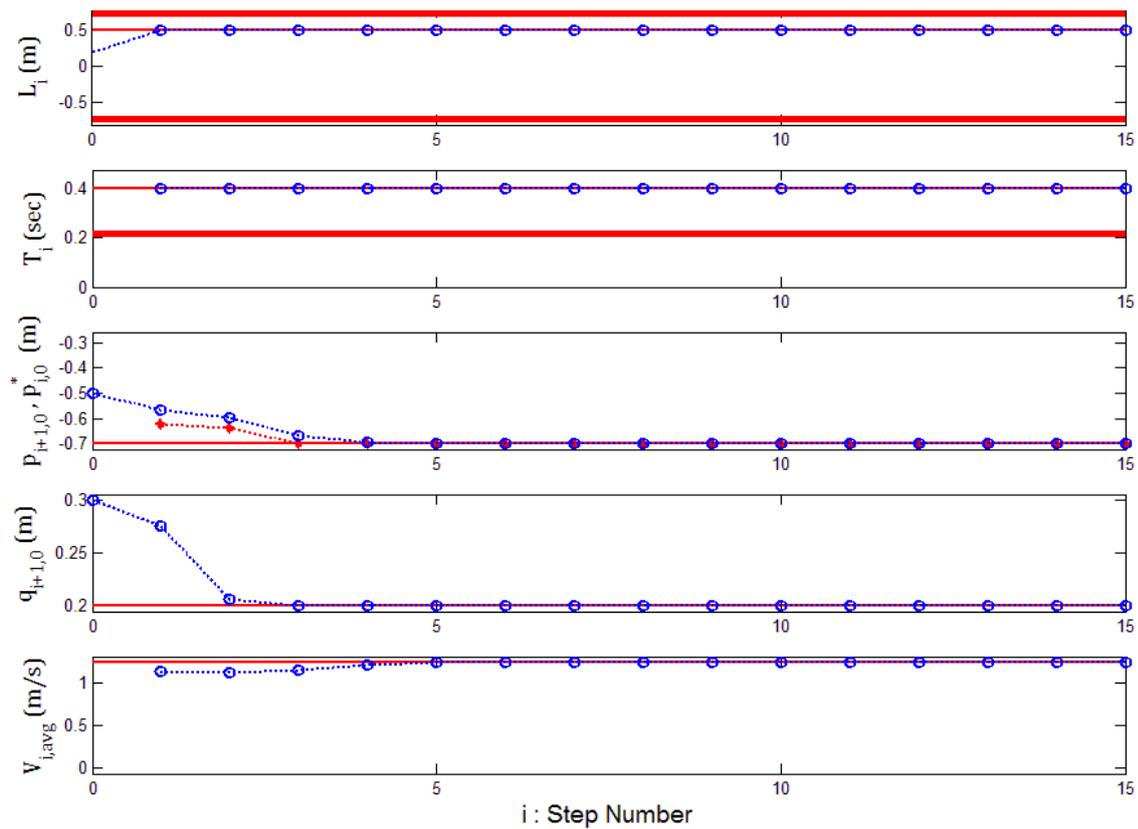

شکل ۳-۳. نمودارهای بخش گسسته معادلات حرکت برای شبیه‌سازی پایدارساز اول سیکل حرکتی بر روی مدل ساده راه‌رونده با شرایط اولیه مرزی- خطوط با ضخامت متوسط و زیاد، به ترتیب مقدار مطلوب متغیر در سیکل حرکتی و محدودیت بالا و یا پایین آن را نمایش می‌دهند.

### ۳-۲-۳ پایدارسازی حرکت با تغییر پی‌درپی طول هر گام

پایدارساز دوم را به صورت رابطه (۳-۱۸) پیشنهاد می‌کنیم. این رابطه گام‌هایی پی‌درپی دارای زمان گام سیکل مطلوب ولی با طول گامی متفاوت را به اجرا می‌گذارد که اختلاف آن از طول گام سیکل مطلوب در هر گام، متناسب با خطای مولفه واگرا در لحظه شروع گام از مقدار مطلوب خود است.

$$\begin{matrix} Stability \\ Controller \end{matrix} (2): \begin{cases} T_i = T_c \\ L_i = L_c + k_i\, e_i = q_c(e^{\omega T_c} - 1) + k_i(q_{i,0} - q_c) \quad,\quad e_i = (q_{i,0} - q_c) \\ \Delta z_i = 0 \end{cases} \quad (\text{۳-۱۸})$$

با جایگذاری این کنترل‌کننده در بخش واگرای معادله حرکت گام به‌گام، به رابطه (۳-۱۹) می‌رسیم که می‌توان آن را به صورت رابطه (۳-۲۰) بازنویسی کرد.

$$q_{i+1,0} = q_{i,0}e^{\omega T_i} - L_i = q_{i,0}e^{\omega T_c} - q_c(e^{\omega T_c} - 1) - k_i(q_{i,0} - q_c) \quad (\text{۳-۱۹})$$

$$\frac{q_{i+1,0} - q_c}{q_{i,0} - q_c} = e^{\omega T_c} - k_i \qquad (۲۰-۳)$$

با درنظرگرفتن ضریب کنترلی در بازه‌ای مشخص، شرط همگرایی شرایط اولیه مؤلفه همگرا به سمت مقدار مطلوب خود به صورت رابطه (۳-۲۱) تکرار می‌شود. اثبات همگرایی همانند کنترل‌کننده قبلی توسط رابطه (۳-۱۲) انجام می‌شود.

$$0 < e^{\omega T_c} - 1 < k_i < e^{\omega T_c} + 1 \quad \Rightarrow \quad \alpha_i = |e^{\omega T_c} - k_i| = \left|\frac{q_{i+1} - q_c}{q_{i,0} - q_c}\right| < 1 \qquad (۲۱-۳)$$

همچنین اثبات پایداری کلی حرکت، با توجه به محدود ماندن $p_{i,0}$، به صورت روابط (۳-۲۲) خواهد بود.

$$0 < e^{\omega T_c} - 1 < k_i < e^{\omega T_c} + 1 \quad \Rightarrow \lim_{i \to \infty} q_{i,0} = q_c$$

$$\lim_{i \to \infty} q_{i,0} = q_c \Rightarrow \lim_{i \to \infty} L_i = \lim_{q_i \to q_c} L_c + k_i(q_{i,0} - q_c) = L_c \qquad (۲۲-۳)$$

$$\Rightarrow \lim_{i \to \infty} \boldsymbol{p}_{\boldsymbol{i,0}}^* = \frac{-L_c}{1 - e^{-\omega T_c}} = p_c \Rightarrow \lim_{i \to \infty} p_{i,0} = p_c$$

در عمل، اندازه طول هر گام دارای کران بالاست که همان حداکثر فاصله ممکن بین دو پای تکیه‌گاهی است. این محدودیت منجر به محدودیت در انتخاب ضریب کنترل‌کننده به صورت رابطه (۳-۲۳) خواهد شد. برای هرگام، بدیهی است ضریبی که منجر به کمینه کردن $\alpha_i$ می‌شود و در این محدوده قرارمی‌گیرد، بهترین انتخاب خواهد بود زیرا سرعت همگرایی را به حداکثر میزان ممکن خود می‌رساند. همچنین، درصورتی‌ یافت نشدن ضریبی در این محدوده که $\alpha_i$ را کوچک‌تر از واحد کند، کنترل‌کننده واگرا خواهد شد. در واقع پایداری حرکت توسط این کنترل‌کننده تضمین خواهد شد، اگر و تنها اگر، محدوده شرایط اولیه مؤلفه واگرا تابع رابطه (۳-۲۴) باشد.

$$|L_i| < \boldsymbol{L_{max}} \iff -L_{max} < L_c + k_i(q_{i,0} - q_c) < L_{max}$$

$$\iff \begin{cases} -\dfrac{\boldsymbol{L_{max}} + L_c}{q_{i,0} - q_c} < k_i < \dfrac{\boldsymbol{L_{max}} - L_c}{q_{i,0} - q_c}, & q_{i,0} > q_c \\[4mm] -\dfrac{\boldsymbol{L_{max}} - L_c}{q_c - q_{i,0}} < k_i < \dfrac{\boldsymbol{L_{max}} + L_c}{q_c - q_{i,0}}, & q_{i,0} \leq q_c \end{cases} \qquad (۲۳-۳)$$

$$\iff \begin{array}{c} Gain \\ Limits \end{array} : -\frac{\boldsymbol{L_{max}}}{|q_{i,0} - q_c|} - \frac{L_c}{q_{i,0} - q_c} < k_i < \frac{\boldsymbol{L_{max}}}{|q_{i,0} - q_c|} - \frac{L_c}{q_{i,0} - q_c}$$

$$q_c - \frac{L_{max} + L_c}{e^{\omega T_c} - 1} < q_{i,0} < q_c + \frac{L_{max} - L_c}{e^{\omega T_c} - 1}$$

$$\begin{cases} 0 < \ e^{\omega T_c} - 1 < k_i < e^{\omega T_c} + 1 \\ \\ -L_{max} < L_c + k_i(q_{i,0} - q_c) < L_{max} \end{cases} \Leftrightarrow \quad q_c = L_c/(e^{\omega T_c} - 1) \quad \downarrow$$

$$\begin{array}{c} Initial \\ Condition: \\ Limits \end{array} - \frac{L_{max}}{e^{\omega T_c} - 1} < q_{i,0} < \frac{L_{max}}{e^{\omega T_c} - 1} \qquad (۲۴-۳)$$

• **نتایج شبیه‌سازی پایدارساز دوم بر روی مدل ساده‌شده حرکت**

شبیه‌سازی پایدارساز دوم بر روی مدل ساده با شرایط اولیه‌ی نزدیک به حد بالایی شرایط اولیه مرزی، انجام شد. منظور از شرایط اولیه مرزی، شرایط اولیه‌ای است که در آن مولفه واگرا در نزدیکی حد بالایی یا پایینی رابطه (۳–۲۴) باشد. همچنین، برای انتخاب ضریب کنترل‌کننده $k_i$ در هر گام، بهترین گزینه ممکن برای بالاترین سرعت همگرایی($\min \alpha_i$) و با توجه به محدوده مجاز این ضریب، رابطه (۳–۲۳)، انجام شده است. شکل ۳-۴، صفحه فاز مسیر حرکت مولفه‌های همگرا و واگرا را برای این شبیه‌سازی نمایش می‌دهد. نمودارهای شکل ۳-۵، سرعت و شتاب مرکز جرم و متغیر کنترلی جابجایی مرکز فشار را برای بخش پیوسته معادلات حرکت نمایش می‌دهند و بخش گسسته معادلات حرکت در شکل ۳-۶، به صورت نمودارهای از متغیرهای کنترلی طول و زمان فرود هر گام، شرایط اولیه مولفه‌های همگرا و واگرا در ابتدای گام و سرعت متوسط مرکز جرم در هر گام نمایش داده شده است.

$p_c = -0.7 , q_c = 0.2 , T_c = 0.4 , L_c = 0.5 , V_c = 1.25$
$p_{1,0} = -0.49 , \ q_{1,0} = 0.29 , x_0 = -0.1 , dx/dt_0 = 1.222$
$z_0 = -0.2 , \ z_1 = 0$

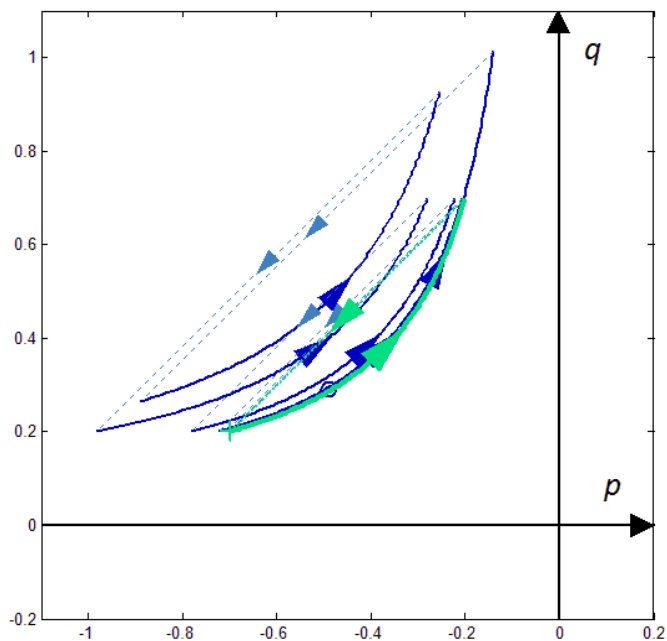

شکل ۳-۴. صفحه فاز مسیر حرکت مولفه واگرا نسبت به مولفه همگرا برای شبیه‌سازی پایدارساز دوم سیکل حرکتی بر روی مدل ساده راه‌رونده با شرایط اولیه مرزی – منحنی ضخیم و کمرنگ مسیر گام آخر را نمایش می‌دهد.

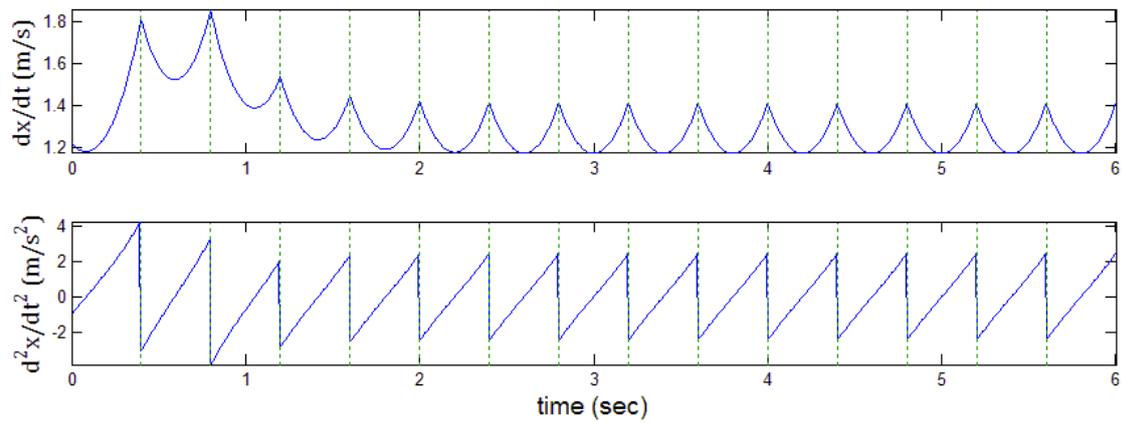

شکل ۳-۵. نمودارهای بخش پیوسته معادلات حرکت برای شبیه‌سازی پایدارساز دوم سیکل حرکتی بر روی مدل ساده راه‌رونده با شرایط اولیه مرزی

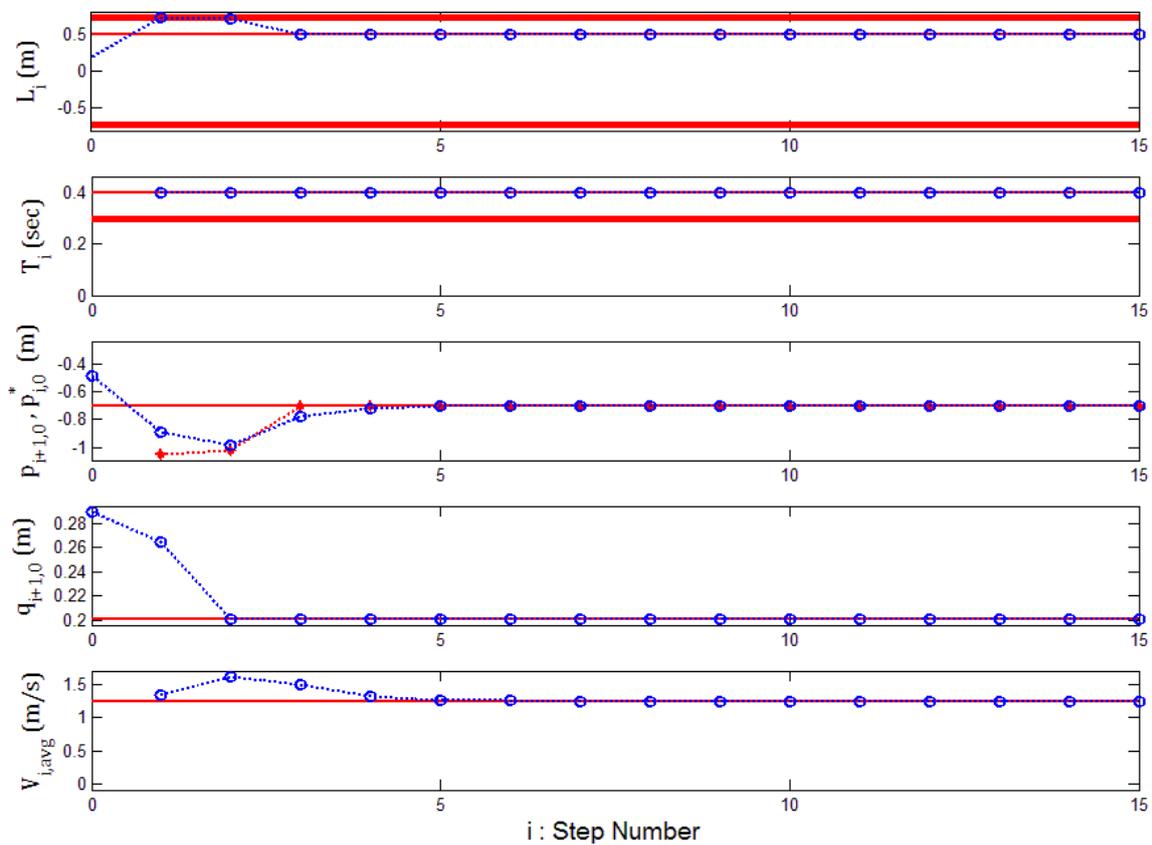

شکل ۳-۶. نمودارهای بخش گسسته معادلات حرکت برای شبیه‌سازی پایدارساز دوم سیکل حرکتی بر روی مدل ساده راه‌رونده با شرایط اولیه مرزی– خطوط با ضخامت متوسط و زیاد، به ترتیب مقدار مطلوب متغیر در سیکل حرکتی و محدودیت بالا ویا پایین آن را نمایش می‌دهند.

### ۴-۲-۳ پایدارسازی حرکت با تغییر پی‌درپی زمان فرود هرگام

پایدارساز سوم را به صورت رابطه (۳-۲۵) پیشنهاد می‌کنیم. این رابطه گام‌هایی پی‌درپی دارای طول گام سیکل مطلوب ولی با زمان گامی متفاوت را به اجرا می‌گذارد که اختلاف آن از زمان گام سیکل مطلوب در هر گام، تابعی از خطای مولفه واگرا در لحظه شروع گام از مقدار مطلوب خود است.

$$
\text{Stability}\atop \text{Controller (3)}: \begin{cases} T_i = T_c + \dfrac{1}{\omega} Ln\left(1 - \dfrac{k_i\, e_i}{(q_c + e_i)e^{\omega T_c}}\right) \\ L_i = L_c \\ \Delta z_i = 0 \end{cases} \quad ,\ \ e_i = \left(q_{i,0} - q_c\right) \qquad (۳-۲۵)
$$

با جایگذاری این کنترل‌کننده در بخش واگرای معادله حرکت گام به گام، به رابطه (۳-۲۶) می‌رسیم که همان رابطه نهایی (۳-۱۹) را تولید خواهد کرد. به همین علت، اثبات همگرایی شرایط اولیه مولفه واگرا به سمت مقدار مطلوب خود، دقیقا مشابه با کنترل‌کننده قبلی قابل بیان است.

$$
\begin{aligned}
q_{i+1,0} &= q_{i,0}e^{\omega T_i} - L_i = q_{i,0}e^{\omega T_c + Ln\left(1 - \frac{k_i\left(q_{i,0}-q_c\right)}{q_{i,0}e^{\omega T_c}}\right)} - L_c \\
&= q_{i,0}e^{\omega T_c} - q_c(e^{\omega T_c} - 1) - k_i\left(q_{i,0} - q_c\right)
\end{aligned} \qquad (۳-۲۶)
$$

در عمل، زمان هر گام به دلیل محدودیت سرعت گام برداشتن دارای کران پایین است که وابسته به طول گام است ($T_{min} = T_{min}(L_c) = \dfrac{L_c}{V_{max}} + T_0$). این محدودیت منجر به محدودیت در انتخاب ضریب کنترل‌کننده به صورت رابطه (۳-۲۷) خواهد شد. برای هرگام، بدیهی است ضریبی که منجر به کمینه کردن $\alpha_i$ می‌شود، بهترین انتخاب خواهد بود زیرا سرعت همگرایی را به حداکثر میزان ممکن خود می‌رساند. همچنین، درصورتی که ضریبی در این محدوده یافت نشود که $\alpha_i$ را کوچکتر از واحد کند، کنترل‌کننده واگرا خواهد شد. در واقع پایداری حرکت توسط این کنترل‌کننده تضمین خواهد شد، اگر و تنها اگر، محدوده شرایط اولیه مولفه واگرا تابع رابطه (۳-۲۸) باشد.

$$
T_i > T_{min} \Leftrightarrow\ 1 - \frac{k_i\, err_i}{(q_c + err_i)e^{\omega T_c}} > e^{\omega(T_{min}-T_c)} \Leftrightarrow\ k_i\frac{q_{i,0} - q_c}{q_{i,0}} < e^{\omega T_c} - e^{\omega T_{min}}
$$

$$
\Leftrightarrow\ \text{Gain}\atop\text{Limits}: \begin{cases} k_i < \ \left|\dfrac{q_{i,0}}{q_{i,0}-q_c}\right|\left(e^{\omega T_c} - e^{\omega T_{min}}\right), & \dfrac{q_c}{q_{i,0}} < 1 \\[2mm] k_i > -\left|\dfrac{q_{i,0}}{q_{i,0}-q_c}\right|\left(e^{\omega T_c} - e^{\omega T_{min}}\right), & \dfrac{q_c}{q_{i,0}} > 1 \end{cases} \qquad (۳-۲۷)
$$

$$\begin{cases} 0 < \ e^{\omega T_c} - 1 < k_i < e^{\omega T_c} + 1 \\ k_i \dfrac{q_{i,0} - q_c}{q_{i,0}} < e^{\omega T_c} - e^{\omega T_{min}} \end{cases} \Leftrightarrow \begin{matrix} Initial \\ Condition: \end{matrix} \begin{cases} 0 < q_{i,0} < \dfrac{L_c}{e^{\omega T_{min}} - 1} \ , \quad q_c > 0 \\ \dfrac{L_c}{e^{\omega T_{min}} - 1} < q_{i,0} < 0 \ , \quad q_c < 0 \end{cases} \qquad (۲۸-۳)$$

- **نتایج شبیه‌سازی پایدارساز سوم بر روی مدل ساده‌شده حرکت**

شبیه‌سازی پایدارساز سوم بر روی مدل ساده با شرایط اولیه‌ای نزدیک به حد بالایی شرایط اولیه مـرزی، انجـام شد. منظور از شرایط اولیه مرزی، شرایط اولیه‌ای است که در آن مولفه واگرا در نزدیکی حد بالایی یا پـایینی رابطـه (۲۸-۳) باشد. همچنین، برای انتخاب ضریب کنترل‌کننده $k_i$ در هرگام، بهترین گزینه ممکن برای بـالاترین سـرعت همگرایی($\min \alpha_i$) و با توجه به محدوده مجاز این ضریب، رابطه (۲۷-۳)، انجام شـده اسـت. شکل ۳-۷، صـفحه فاز مسیر حرکت مولفه‌های همگرا و واگرا را برای این شبیه‌سازی نمایش می‌دهد. نمودارهای شـکل ۳-۸، سـرعت و شتاب مرکز جرم و متغیر کنترلی جابجایی مرکز فشار را برای بخش پیوسته معادلات حرکـت نمـایش می‌دهنـد و بخش گسسته معادلات حرکت در شکل ۳-۹، به صورت نمودارهای از متغیرهای کنترلی طول و زمان فرود هر گام، شرایط اولیه مولفه‌های همگرا و واگرا در ابتدای هرگام و سرعت متوسط مرکز جرم را نمایش داده شده است.

$$p_c = -0.7 \ , \ q_c = 0.2 \ , \ T_c = 0.4 \ , \ L_c = 0.5 \ , \ V_c = 1.25$$
$$p_{1,0} = -0.705 \ , \ q_{1,0} = 0.505 \ , \ x_0 = -0.1 \ , \ dx/dt_0 = 1.895$$
$$z_0 = -0.2 \ , \ z_1 = 0$$

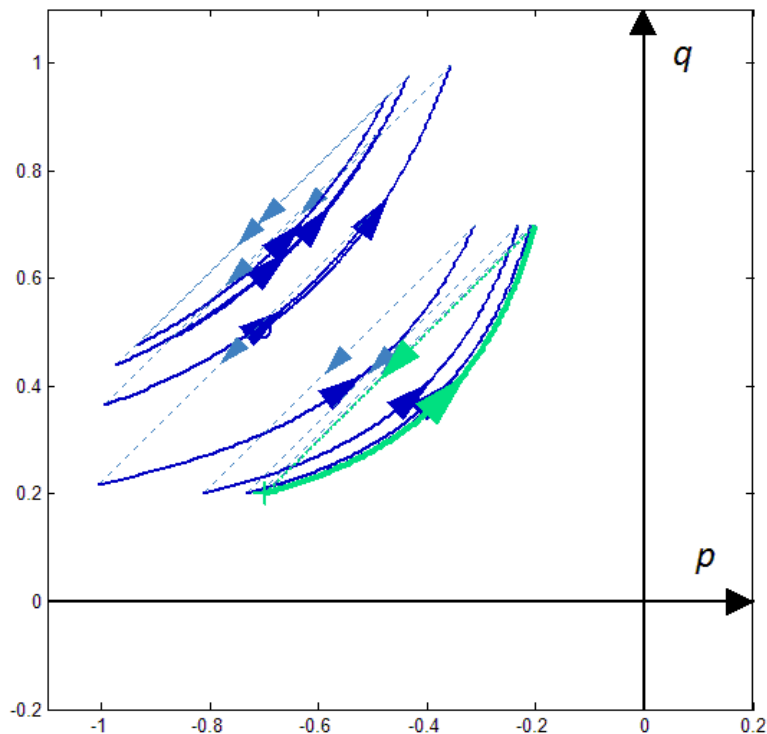

شکل ۳-۷. صفحه فاز مسیر حرکت مولفه واگرا نسبت به مولفه همگرا برای شبیه‌سازی پایدارساز سوم سیکل حرکتی بر روی مدل ساده راه‌رونده با شرایط اولیه مرزی - منحنی ضخیم و کمرنگ مسیر گام آخر را نمایش می‌دهد.

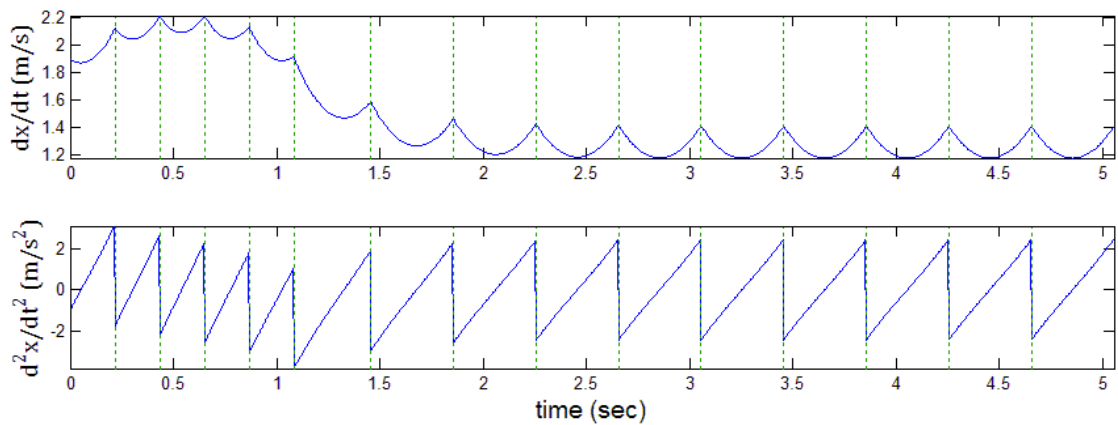

شکل ۳-۸ نمودارهای بخش پیوسته معادلات حرکت برای شبیه‌سازی پایدارساز سوم سیکل حرکتی بر روی مدل ساده راهرونده با شرایط اولیه مرزی

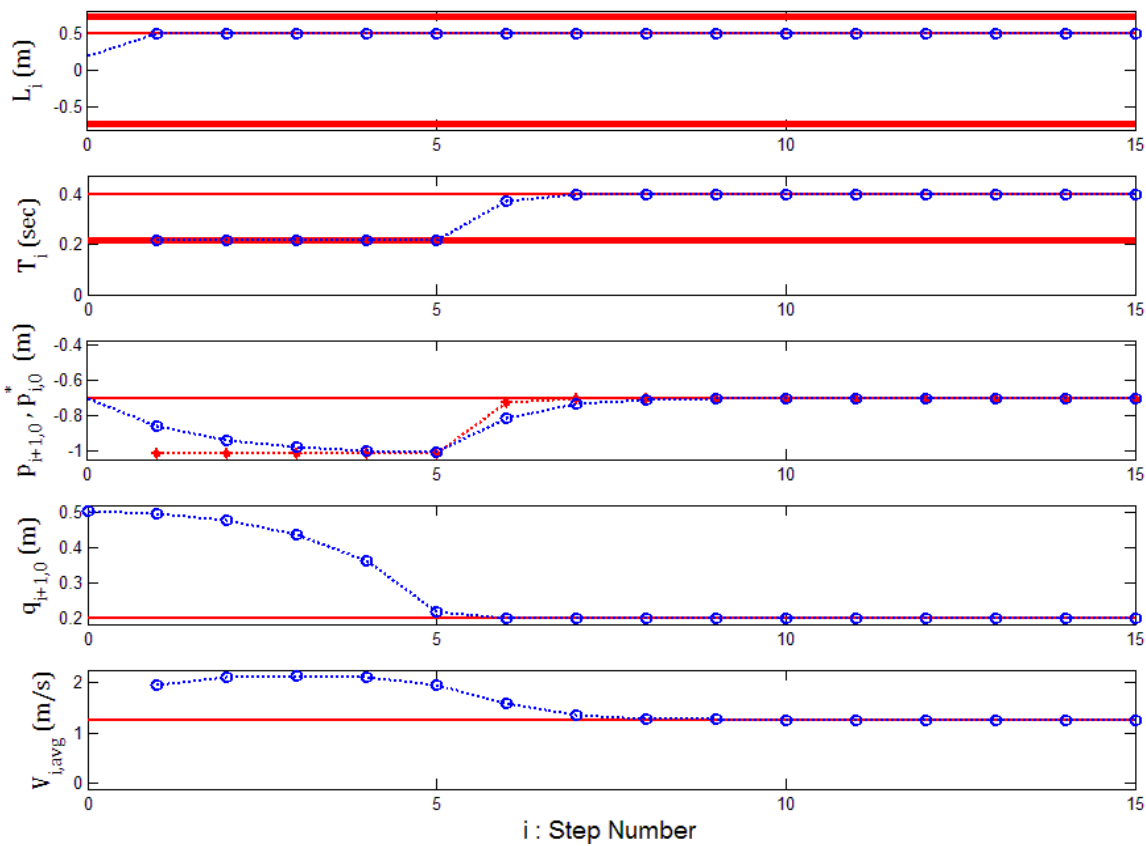

شکل ۳-۹. نمودارهای بخش گسسته معادلات حرکت برای شبیه‌سازی پایدارساز اول سیکل حرکتی بر روی مدل ساده راهرونده با شرایط اولیه مرزی− خطوط با ضخامت متوسط و زیاد، به ترتیب مقدار مطلوب متغیر در سیکل حرکتی و محدودیت بالا و/یا پایین آن را نمایش می‌دهند.

### ۵-۲-۳ پایدارسازی حرکت با تغییر پی‌درپی طول و زمان فرود هرگام

پایدارساز سوم را به صورت رابطه (۲۵-۳) پیشنهاد می‌کنیم. این رابطه گام‌هایی پی‌درپی دارای طـول و زمـانی متفاوت با مشخصات گام سیکل مطلوب را به‌اجرا می‌گذارد. اختلاف طول هر گام از طول گام سیکل مطلوب در هر گام، متناسب با خطای مولفه واگرا در لحظه شروع گام از مقدار مطلوب خود است و اختلاف زمـان هـر گـام از زمان گام سیکل مطلوب در هر گام، تابعی از این خطا می‌باشد.

$$Stability\ Controller\ (4): \begin{cases} T_i = T_c + \dfrac{1}{\omega}Ln\left(1 - \dfrac{k_{i,T}\ e_i}{(q_c + e_i)e^{\omega T_c}}\right) \\ L_i = L_c + k_{i,L}\ e_i = q_c(e^{\omega T_c} - 1) + k_{i,L}(q_{i,0} - q_c)\ , e_i = (q_{i,0} - q_c) \\ \Delta z_i = 0 \end{cases} \quad (۲۹-۳)$$

با جایگذاری این کنترل‌کننده در بخش واگرای معادله حرکت گام به گام، بـه رابطـه نهـایی (۳۰-۳) مـی‌رسـیم. اثبات همگرایی شرایط اولیه مولفه واگرا به سمت مقدار مطلوب خود، مشابه با کنترل‌کننـده‌هـای قبلـی اسـت بـا ایـن تفاوت که $(k_{i,L} + k_{i,T})$ جایگزین $k_i$ در روابط قبلی خواهد است.

$$\frac{q_{i+1,0} - q_c}{q_{i,0} - q_c} = e^{\omega T_c} - (k_{i,L} + k_{i,T})$$

$$0 < e^{\omega T_c} - 1 < k_{i,L} + k_{i,T} < e^{\omega T_c} + 1 \Rightarrow \quad \alpha_i = \left|e^{\omega T_c} - k_{i,L} - k_{i,T}\right| = \left|\frac{q_{i+1} - q_c}{q_{i,0} - q_c}\right| < 1 \quad (۳۰-۳)$$

در عمل، زمان هر گام به دلیل محدویت سرعت گام برداشتن دارای کـران پـایین و انـدازه طـول هـر گـام دارای کران بالاست. با توجه به تاثیر طول گام در کران پایین زمـان گـام، ایـن کـران را بـرای بیشـترین مقـدار طـول گـام فرض می‌کنیم تا برای طول گام‌های کوچکتر نیز معتبر بماند($T_{min} = T_{min}(L_{max}) = \dfrac{L_{max}}{V_{max}} + T_0$). ایـن محدودیت‌ها منجر به محدودیت در انتخاب ضرایب کنترل‌کننده به صورت رابطه (۳۱-۳) خواهد شد. برای هرگام، بدیهی است ضریب یا ضرایبی در این محدوده که منجر به کمینه کردن $\alpha_i$ می‌شـود، بهتـر انتخـاب خواهنـد بـود زیرا سرعت همگرایی را به حداکثر میزان ممکن خود می‌رساند. همچنین، درصورتی کـه ضـریبی در ایـن محـدوده یافت نشود که $\alpha_i$ را کوچکتر از واحد کند، کنترل‌کننده واگرا خواهد شـد، در واقـع پایـداری حرکـت توسـط ایـن کنترل‌کننده تضمین خواهد شد، اگر و تنها اگر، محدوده شرایط اولیه مولفه واگرا تابع رابطه (۳۲-۳) باشد.

$$
Gains\ Limits : \begin{cases} (1)\quad 0\ <\ e^{\omega T_c} - 1 < k_{i,L} + k_{i,T} < e^{\omega T_c} + 1 \\[2mm] (2)\quad -\dfrac{\boldsymbol{L_{max}}}{|q_{i,0} - q_c|} - \dfrac{L_c}{q_{i,0} - q_c} < k_{i,L} < \dfrac{\boldsymbol{L_{max}}}{|q_{i,0} - q_c|} - \dfrac{L_c}{q_{i,0} - q_c} \\[4mm] (3)\quad \begin{cases} k_{i,T}\ <\ \left|\dfrac{q_{i,0}}{q_{i,0} - q_c}\right| \left(e^{\omega T_c} - e^{\omega T_{min}}\right), \qquad \dfrac{q_c}{q_{i,0}} < 1 \\[3mm] k_{i,T}\ >\ -\left|\dfrac{q_{i,0}}{q_{i,0} - q_c}\right| \left(e^{\omega T_c} - e^{\omega T_{min}}\right), \qquad \dfrac{q_c}{q_{i,0}} > 1 \end{cases} \end{cases} \tag{۳-۳۱}
$$

$$
\begin{cases} 0 < e^{\omega T_c} - 1 < k_{i,L} + k_{i,T} < e^{\omega T_c} + 1 \\[2mm] -\boldsymbol{L_{max}} < L_c + k_{i,L}(q_{i,0} - q_c) < \boldsymbol{L_{max}} \\[2mm] k_{i,T}\dfrac{q_{i,0} - q_c}{q_{i,0}} < e^{\omega T_c} - e^{\omega T_{min}} \end{cases} \Leftrightarrow \begin{matrix} Initial \\ Condition \\ Limits \end{matrix} : \dfrac{-\boldsymbol{L_{max}}}{e^{\omega T_{min}} - 1} < q_{i,0} < \dfrac{\boldsymbol{L_{max}}}{e^{\omega T_{min}} - 1} \tag{۳-۳۲}
$$

- **نتایج شبیه‌سازی پایدارساز چهارم بر روی مدل ساده‌شده حرکت**

شبیه‌سازی پایدارساز سوم بر روی مدل ساده با شرایط اولیه‌ای نزدیک به حد بالایی شرایط اولیه مرزی، انجام شد. منظور از شرایط اولیه مرزی، شرایط اولیه‌ای است که در آن مؤلفه واگرا در نزدیکی حد بالایی یا پایینی رابطه (۳-۳۲) باشد. . همچنین، برای انتخاب ضرایب کنترل‌کننده $k_{i,L}$ و $k_{i,T}$ در هرگام، بهترین گزینه ممکن برای بالاترین سرعت همگرایی ($\min \alpha_i$) و با توجه به محدوده مجاز این ضرایب، رابطه (۳-۳۱)، انجام شده است. شکل ۳-۱۰، صفحه فاز مسیر حرکت مؤلفه‌های همگرا و واگرا را برای این شبیه‌سازی نمایش می‌دهد. نمودارهای شکل ۳-۱۱، سرعت و شتاب مرکز جرم و متغیر کنترلی جابجایی مرکز فشار را برای بخش پیوسته معادلات حرکت نمایش می‌دهند و بخش گسسته معادلات حرکت در شکل ۳-۱۲، به صورت نمودارهایی از متغیرهای کنترلی طول و زمان فرود هر گام، شرایط اولیه مؤلفه‌های همگرا و واگرا در ابتدای هرگام و سرعت متوسط مرکز جرم در هرگام نمایش داده شده است.

$$p_c = -0.7 \; , \; q_c = 0.2 \; , \; T_c = 0.4 \; , \; L_c = 0.5 \; , \; V_c = 1.25$$
$$p_{1,0} = -0.67 \; , \; q_{1,0} = 0.47 \; , \; x_0 = -0.1 \; , \; dx/dt_0 = 1.786$$
$$z_0 = -0.2 \; , \; z_1 = 0$$

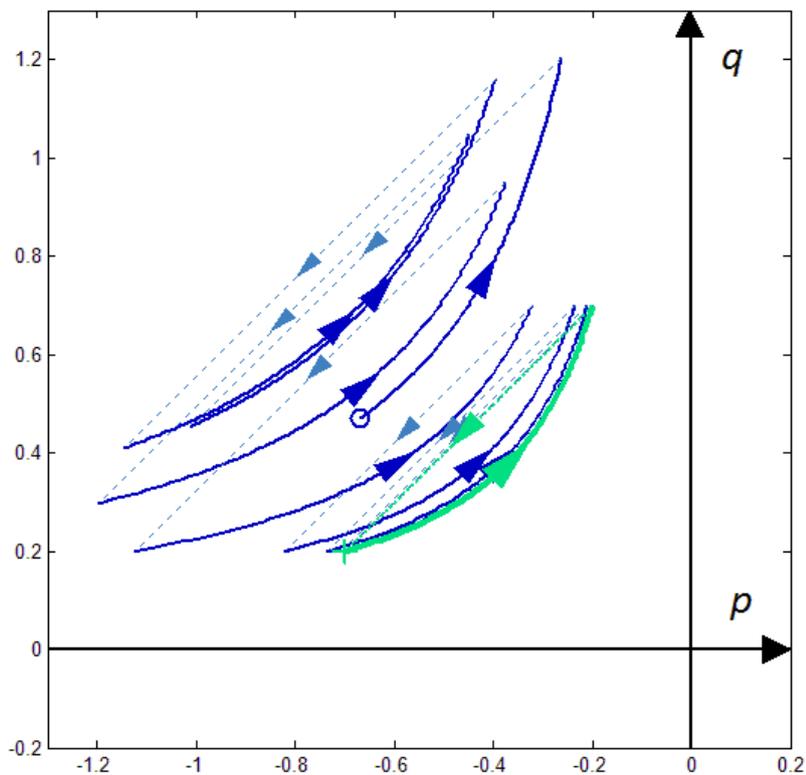

شکل ۳-۱۰. صفحه فاز مسیر حرکت مولفه واگرا نسبت به مولفه همگرا برای شبیه‌سازی پایدارساز چهارم سیکل حرکتی بر روی مدل ساده راه‌رونده با شرایط اولیه مرزی ─ منحنی ضخیم و کم‌رنگ مسیر گام آخر را نمایش می‌دهد.

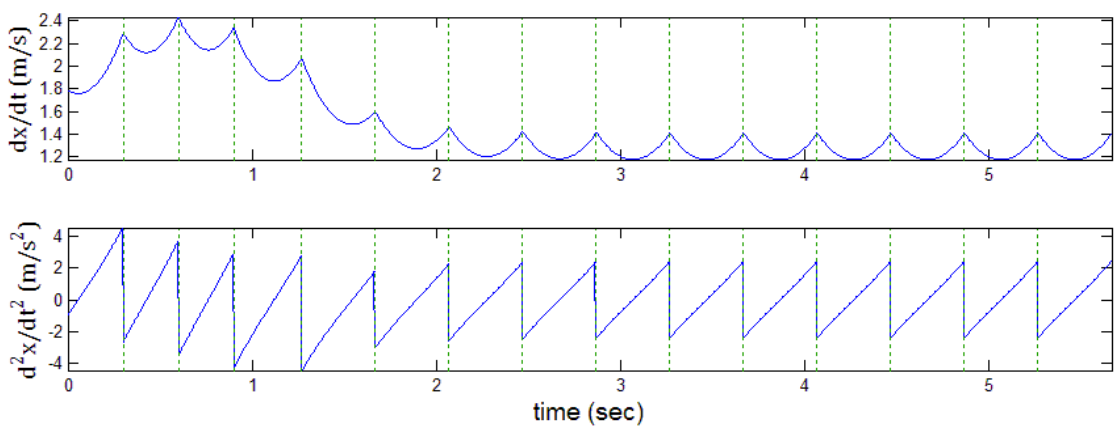

شکل ۳-۱۱. نمودارهای بخش پیوسته حرکت برای شبیه‌سازی پایدارساز چهارم سیکل حرکتی بر روی مدل ساده راه‌رونده با شرایط اولیه مرزی

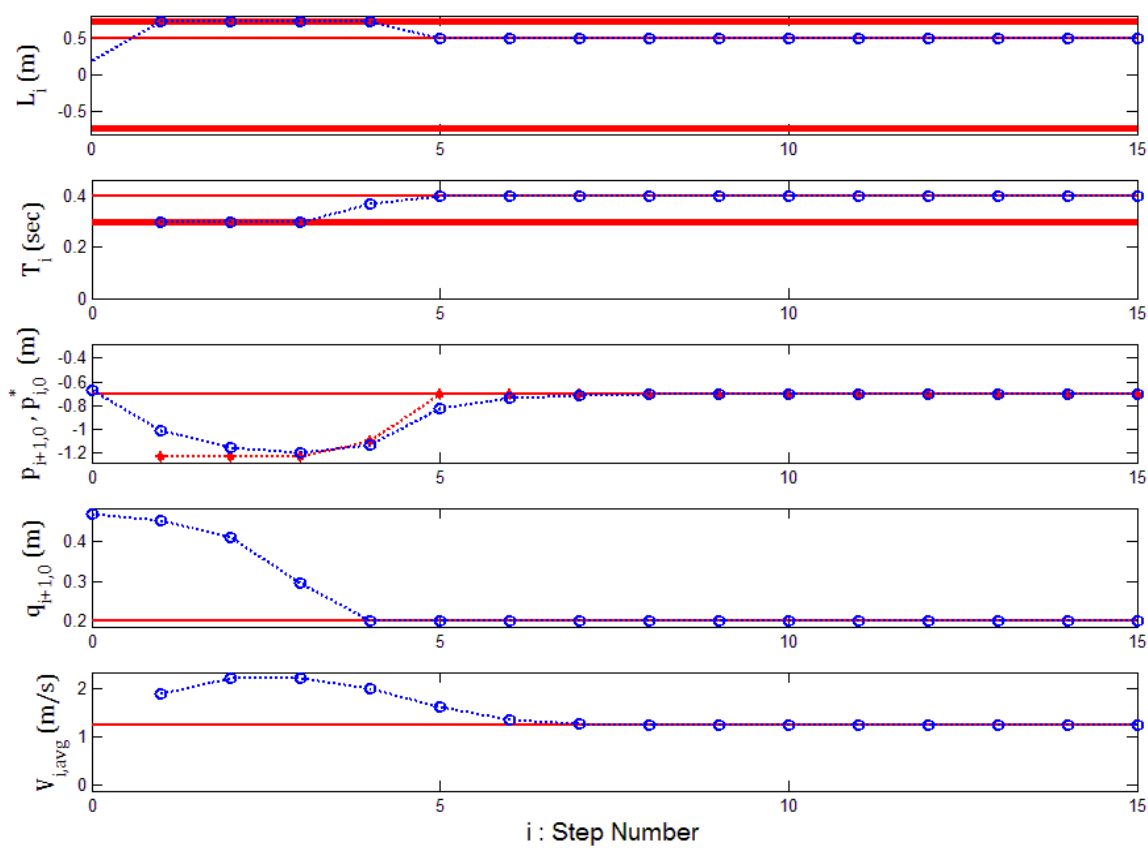

شکل ۳-۱۲. نمودارهای بخش گسسته معادلات حرکت برای شبیه‌سازی پایدارساز چهارم سیکل حرکتی بر روی مدل ساده راه‌رونده با
شرایط اولیه مرزی– خطوط با ضخامت متوسط و زیاد، به ترتیب مقدار مطلوب متغیر در سیکل حرکتی و محدودیت بالا ویا پایین آن را
نمایش می‌دهند.

## ۳-۳  شبیه‌سازی پایدارسازهای سیکل حرکتی بر روی مدل فیزیکی کامل یک راه‌رونده

### ۳-۳-۱  مدل فیزیکی کامل یک راه‌رونده با بالاتنه صلب و پاهایی با جرم قابل صرف‌نظر

مدل انتخاب‌شده برای شبیه‌سازی‌ها، یک مدل تک جرمی صلب در قسمت بدنه یا بالاتنه است کـه حرکـت در
هر سه درجه‌آزادی افقی، عمودی و  دورانی را در صفحه نمایندگی می‌کند. انتخاب ایـن مـدل از چنـد جهـت قابـل
توجه است. اول اینکه حرکت بالاتنه این مدل به علت تک جرمی بودن، دقیقا کمیت‌هـای انـدازه حرکـت خطـی در
راستای افقی و عمودی و همچنین اندازه حرکت زاویه‌ای را به نمایش می‌گذارد و به همین جهت کمیت‌های سیستم
به راحتی قابل مشاهده و ادراک است. همچنین با فرض قرار گرفتن مفصل ران بـر روی مرکـز جـرم بالاتنـه، نوشـتن
روابط قید مربوط به محدودیت‌های حرکتی در قسمت بعدی آسان خواهد بود. علاوه بر این، کنترل‌کننده‌ای کـه در

ادامه برای این مدل پیشنهاد خواهدشد، می‌تواند بر روی ربات‌هایی واقعی مانند ربات فلامینگو[1] دانشگاه M.I.T [۳۵] و یا ربات‌های مشابهی که با پیش‌فرض‌های این مدل همخوانی دارند، بـا اسـتفاده از روش معکـوس ژاکـوبین پیـاده-سازی شود(شکل ۳-۱۳).

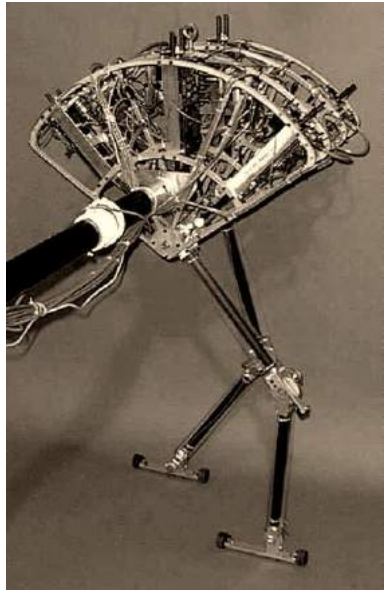

شکل ۳-۱۳. ربات فلامینگو دانشگاه M.I.T [۳۵]

معادلات حرکت این مدل فیزیکی که شمایی از حرکت و دیاگرام آزاد نیرویی آن در شکل ۳-۱۴ نمایش داده شده است، مطابق روابط (۳۳-۳) می‌باشد. بالانویس $control$ در این روابط، مربوط به نیروها و گشتاور اعمـالی از طرف کنترل‌کننده (به واسطه اعمال گشتاور داخلی بر عضوها که برآینـد نیرویـی آن برابـر بـا نیروهـای واکنشـی وارده از کف زمین است) است و به مرکز جرم بالاتنه وارد می‌شود. همچنین بالانویس $disturbance$، مربـوط به برآیند نیروها و گشتاور خارجی اختلالی وارده از سوی محیط اطراف (به استثنای نیروی وزن و نیروهـای وارده از کف زمین) می‌باشد و می‌تواند ناشی از هر گونه اختلالی ازجمله ضربه به بالاتنه و یا نیروها و گشتاورهای اغتشاشـی باشند که نقطه اثر آن‌ها نیز، مرکز جرم بالاتنه فرض شده‌است.

پیش‌فرض این مدل فیزیکی، وجود دائمی حداقل یک نقطه از بدن(پای) راهرونده با زمین به نحوی اسـت که علاوه بر نیروی عمودی، نیروی افقی مورد نیاز برای اعمال نیروهـای کنترلـی را بـدون خـارج شـدن از مخـروط اصطکاک($|GRF_X/GRF_Y| \cong \left| F_x^{control}/F_y^{control} \right| < \mu$) تامین می‌کند. ربات در لحظه شروع حرکـت حائز این شرایط فرض می‌شود و بدیهی است در صورت اعمال نیروهایی که این شرایط را رعایـت کننـد، معـادلات حرکت ارائه شده، مدل فیزیکی حاضر را به درستی نمایندگی می‌کنند. از آنجا کـه مسـئله اصـلی بـرای مـا، کنتـرل پایداری حرکت در سطح اندازه حرکت است، این پیش‌فرض‌ها خللی به کلیت مسئله وارد نمی‌کند.

---

[1] Flamingo

$$\begin{cases} \ddot{x} = \dfrac{\dot{L}_X}{M} = \dfrac{F_x}{M} = \dfrac{F_x^{control} + F_x^{disturbance}}{M} \\[3mm] \ddot{y} = \dfrac{\dot{L}_Y}{M} = \dfrac{F_y - M\,g}{M} = \dfrac{F_y^{control} + F_y^{disturbance} - M\,g}{M} \\[3mm] \ddot{\theta} = \dfrac{\dot{H}_{G_Z}}{I} = \dfrac{T_z}{I} = \dfrac{T_z^{control} + T_z^{disturbance}}{I} \end{cases} \qquad (\text{۳–۳۳})$$

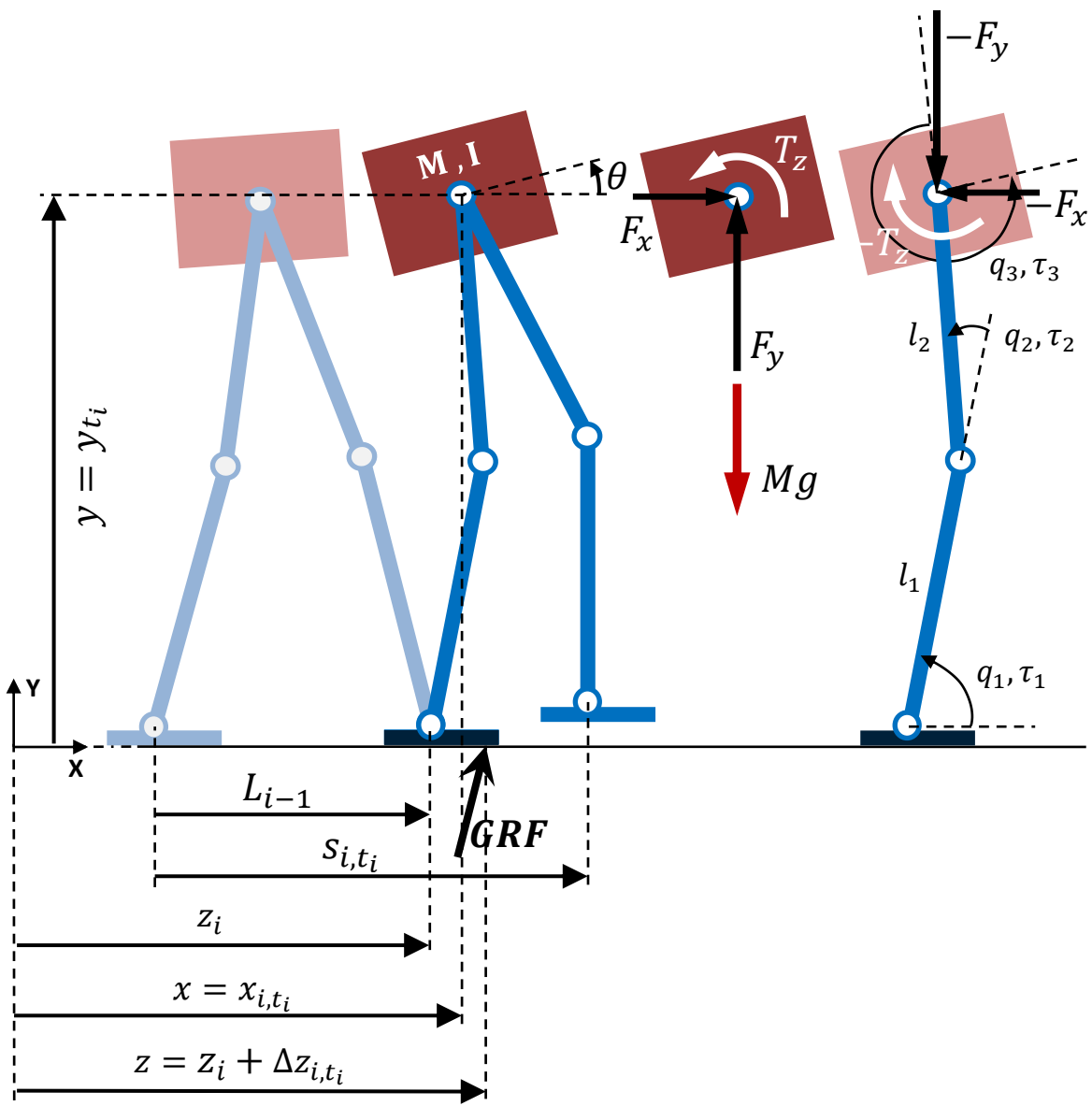

شکل ۳-۱۴. شمایی از حرکت و دیاگرام آزاد نیرویی برای مدل فیزیکی راهروندهای با بالاتنه صلب و پاهایی با جرم قابلصرفنظر

**۲-۳-۳ کنترل اندازه حرکت خطی عمودی و زاویه‌ای و تامین اندازه حرکت خطی افقی حرکت طبیعی**

تا به این قسمت، همواره فرض بر این بود که حرکت راه‌رونده در راستای عمودی(اندازه حرکت خطی عمودی) و در راستای دورانی(اندازه حرکت زاویه‌ای) به دلیل عدم نیاز به تغییردادن این متغیرها در انجام وظیفه راه‌رفتن و همچنین با توجه به اطلاعات ضبط شده از حرکت راه‌رونده‌هایی همچون انسان، همواره محدود می‌ماند. امـا برای محدود نگه داشتن این دو کمیت نیاز به کنترل‌کننده‌ای داریم که حرکت آنها را حول صفر و یا حول مسیری تناوبی با دامنه محدود تنظیم کند. بخش‌های (1) و (2) کنترل‌کننده اندازه‌حرکت در رابطه (۳-۳۴) می‌توانند این وظیفه را به خوبی انجام دهند. با جایگذاری ایـن دو بخش از کنترل‌کننـده انـدازه‌حرکت در دو بخش مربوطه معادلات حرکت مدل فیزیکی، به راحتی می‌توان دید که تابع خطای هر یک، فیلتری درجه دوم بر روی اغتشاشـات خارجی خواهد بود که در صورت محدود ماندن اغتشاشات خارجی، فاصله آن‌ها نزدیک به صفر محدود می‌ماند.

بخش (3) کنترل‌کننده، در واقع به دنبال کنترل کمیت مشخصی نیست و تنها تلاش دارد تا با توجه بـه موقعیت مطلوب مرکز فشار و با استفاده از معکوس معادله حرکت پایه، نیروی مورد نیاز بـرای انجام حرکت طبیعـی مـورد انتظار از راه‌رونده در راستای افق را تامین کند. موقعیت مطلوب مرکز فشار، حاصـل اجـرای طـول گام‌هـای قبلـی درکنترل مرکز تکیه‌گاه، $z_i$، و احتمالا محاسبات جاری مقدار مطلوب متغیر جابجایی مرکز فشار، $\Delta z_i^{des}$، است کـه هر دو توسط کنترل‌کننده پایداری انجام می‌شود. در صورت به وجود آمدن خطایی در حرکت افقی راه‌رونده، ایـن وظیفه کنترل‌کننده پایداری و نه کنترل‌کننده حاضر است که به وسیله تغییر مـداوم $\Delta z_i^{des}$ بـه عنوان ورودی ایـن قسمت و یا به وسیله محاسبه و اجرای گامی با طول و زمان فرود مناسب، $L_i$ و $T_i$، کـه منجر بـه $z_i$ جدیـد خواهـد شد، به جبران این خطا و همگرایی حرکت افقی به سمت سیکل حرکتی مطلوب بپردازد.

$$(1): \ F_y^{control} = M\left(g + \ddot{y}_{des} + k_{v1}(\dot{y}_{des} - \dot{y}) + k_{p1}(y_{des} - y)\right)$$

$$(2): \ T_z^{control} = \dot{H}_{des} + k_{v2}(H_{des} - H) + k_{p2}\int_{\tau=0}^{t}(H_{des} - H)\,d\tau$$

$$= I\left(\ddot{\theta}_{des} + k_{v2}(\dot{\theta}_{des} - \dot{\theta}) + k_{p2}(\theta_{des} - \theta)\right)$$

$$(3): \ F_x^{control} = \frac{T_z^{control} + (x - z_{des})\,F_y^{control}}{y} \ , \quad z_{des} = z_i + \Delta z_i^{des}$$

$$(۳-۳۴)$$

**۳-۳-۳ محاسبه گشتاور مفاصل برای راه‌رونده واقعی با استفاده از روش معکوس ژاکوبین**

از آنجا که پاها بدون جرم یا با جرمی قابل صرف‌نظر درنظرگرفته‌شده‌است، گشتاور مورد نیاز برای حرکت پای متحرک ناچیز است و نیروها و گشتاوری که از کل پای متحرک به بالاتنه وارد می‌شود، یـا کـاملا قابل‌صـرف‌نظر خواهد بود و یا می‌تواند به عنوان نیرو و گشتاور خارجی به عنوان اختلال در مـدل فیزیکی درنظرگرفته‌شـود. ولی برای پای تکیه‌گاهی که وظیفه حمل بالاتنه در راستای افقی و کنترل حرکت آن در راستای عمـودی و دورانـی را دارد، می‌توان با استفاده از ترانهاده ژاکوبین، گشتاورهای مورد نیاز در مفاصل را محاسبه کرد.

اگر رابطه موقعیت و زاویه بالاتنه با زوایای سه مفصل پای متحرک را بنویسیم، مـاتریس ژاکـوبینی بـه صـورت روابط(۳–۳۵) استخراج خواهد شد.

$$\begin{bmatrix} x \\ y \\ \theta \end{bmatrix} = \begin{bmatrix} l_1 \cos(q_1) + l_2 \cos(q_1 + q_2) + z_i \\ l_1 \sin(q_1) + l_2 \sin(q_1 + q_2) \\ q_1 + q_2 + q_3 \end{bmatrix}$$

$$\Rightarrow \begin{bmatrix} \dot{x} \\ \dot{y} \\ \dot{\theta} \end{bmatrix} = J \begin{bmatrix} \dot{q}_1 \\ \dot{q}_2 \\ \dot{q}_3 \end{bmatrix} \quad , \quad J = \begin{bmatrix} -l_1 \sin(q_1) - l_2 \sin(q_1 + q_2) & -l_2 \sin(q_1 + q_2) & 0 \\ l_1 \cos(q_1) + l_2 \cos(q_1 + q_2) & l_2 \cos(q_1 + q_2) & 0 \\ 1 & 1 & 1 \end{bmatrix}$$

(۳–۳۵)

بر اساس روابط کار مجازی، گشتاور مورد نیاز در مفاصل برای تامین نیروها و گشتاور کنترلی بر روی بالاتنه که باید در نقطه انتهایی مفصل بالایی(ران) اثرکنند، برابر با ضرب مـاتریس ترانهـاده ژاکـوبین در بـردار متشـکل از ایـن نیروها و گشتاور با ترتیبی متناظر خواهد بود که با جایگـذاری نیروهـا و گشـتاور کنترلـی رابطـه (۳–۳۴) در آن، بـه روابط نهایی (۳–۳۶) خواهیم رسید. اگر به گشتاور مفصل اول(مچ پا) توجه کنیم، مـی‌بینیم کـه درصـورت صـفر بـودن متغیر جابه‌جایی مرکز فشار، $\Delta z_i^{des}$، مقدار آن صفر خواهد بود که در آن صورت از داشتن کف پا بی‌نیـاز خـواهیم بود و می‌توان از این کنترل‌کننده برای دوپاهای کم‌عملگر نیز استفاده نمود. بـه غیـر از روش اول در پایدارسـازی‌هـای سیکل حرکتی در مابقی روش‌های پایدارساز سیکل حرکتی و نیز روش کنترل پایـداری بهینـه کـه در فصـل بعـدی بحث خواهدشد، این متغیر صفر فرض می‌شود و بنابراین همه روش‌های کنترل پایداری بـه غیـر از روش اول، قابـل کاربرد برای راه‌روندهای دارای تماس کف پای نقطه‌ای نیز می‌باشند. اگرچه این نتیجه(صفر بودن گشتاور مفصل مچ پا) با توجه به روابط ژاکوبینی که برای این مدل خاص نوشته شده‌است، بـه‌دسـت آمـد، ولـی در حالـت کلـی نیـز می‌توان با نوشتن معادلات نیوتن برای عضو کف پا به نتیجه یکسانی رسید.

$$\begin{bmatrix} \tau_1^{control} \\ \tau_2^{control} \\ \tau_3^{control} \end{bmatrix} = J^T \begin{bmatrix} F_x^{control} \\ F_y^{control} \\ T_z^{control} \end{bmatrix}$$

$$(\text{۳۶}-\text{۳})$$

$$\Rightarrow \begin{cases} \tau_1^{control} = \Delta z_i^{des} \, F_y^{control} \\ \tau_2^{control} = -l_2 \sin(q_1 + q_2) \, F_x^{control} + l_2 \cos(q_1 + q_2) \, F_y^{control} + T_z^{control} \\ \tau_3^{control} = T_z^{control} \end{cases}$$

### ۴-۳-۳ محدودیت‌های حرکتی ربات ناشی از پیکره‌بندی و محدودیت سرعت نسبی متوسط پای متحرک

به غیر از محدودیت حرکتی حداکثر طول گام، رابطه (۳–۳۷)، که ناشی از طول عضوهای ساق و ران پا است، طول این اعضا منجر به محدودیت‌های حرکتی دیگری نیز در هنگام فرود آوردن پا در یک نقطه می‌شوند که از آن جمله می‌توان از محدودیت فاصله افقی بالاتنه تا پای تکیه‌گاهی فعلی، $|D_{i,1}| > \frac{L_{max}}{2}$، و همچنین فاصله افقی بالاتنه تا پای تکیه‌گاهی بعدی، $|D_{i,2}| > \frac{L_{max}}{2}$، نام برد. این محدودیت‌ها همانطور که در شکل ۳-۱۵ نمایش‌داده شده‌اند، برای مدل فیزیکی این مسئله به صورت روابط (۳–۳۸) و (۳–۳۹) قابل بیان هستند.

$$L_{max} = 2\sqrt{(l_1 + l_2)^2 - h^2} \qquad\qquad (\text{۳۷}-\text{۳})$$

$$|D_{i,1}| < \frac{L_{max}}{2} \ , \quad D_{i,1} = x_{i,T_i} - z_i = \frac{p_{i,T_i} + q_{i,T_i}}{2}$$
$$or \ \ D_{i,1} = x_{i+1,0} - z_i = (x_{i+1,0} - z_{i+1}) + (z_{i+1} - z_i) = \frac{p_{i+1,0} + q_{i+1,0}}{2} + L \qquad (\text{۳۸}-\text{۳})$$

$$|D_{i,2}| < \frac{L_{max}}{2} \ , \quad D_{i,2} = z_{i+1} - x_{i,T_i} = (z_{i+1} - z_i) - (x_{i,T_i} - z_i) = L - \frac{p_{i,T_i} + q_{i,T_i}}{2}$$
$$or \ \ D_{i,2} = z_{i+1} - x_{i+1,0} = -(x_{i+1,0} - z_{i+1}) = -\frac{p_{i+1,0} + q_{i+1,0}}{2} \qquad (\text{۳۹}-\text{۳})$$

محدودیت دیگر مسئله ما، محدودیت حداکثر سرعت نسبی پای متحرک نسبت به بالاتنه است که تاکنون از آن به عنوان کران پایین زمان نام می‌بردیم، رابطه این قید به صورت رابطه (۳–۴۰) برای مدل فیزیکی مسئله مـا خواهـد بود.

$$|\bar{V}_{i,swing}| < V_{max} \ ,$$
$$\bar{V}_{i,swing} = \frac{(s_{i,T_i} - s_{i,0}) - (x_{i,T_i} - x_{i,0})}{T - T_0} = \frac{(L_{i-1} + L_i) - \left(\frac{p_{i,T_i} + q_{i,T_i}}{2} - \frac{p_{i,0} + q_{i,0}}{2}\right)}{T - T_0} \qquad (\text{۴۰}-\text{۳})$$

در ابتدای حرکت گام جدید(فعلی)، یک تخمین ساده برای محاسبه این کمیت‌های مقید بر اساس مدل ساده‌شده راهرونده برای گامی با طول و زمان فرود مشخص، $L_i$ و $T_i$، به صورت روابط (۳-۴۱) خواهند بود.

$$D_{i,1} = \frac{p_{i,0}e^{-\omega T_i} + q_{i,0}e^{\omega T_i}}{2}$$

$$D_2 = L_i - \frac{p_{i,0}e^{-\omega T_i} + q_{i,0}e^{\omega T_i}}{2}$$

$$\bar{V}_{swing} = \frac{L_{i-1} + L_i - \frac{(p_{i,0}e^{-\omega T_i} + q_{i,0}e^{\omega T_i}) - (p_{i,0} + q_{i,0})}{2}}{T_i - T_0}$$

(۳-۴۱)

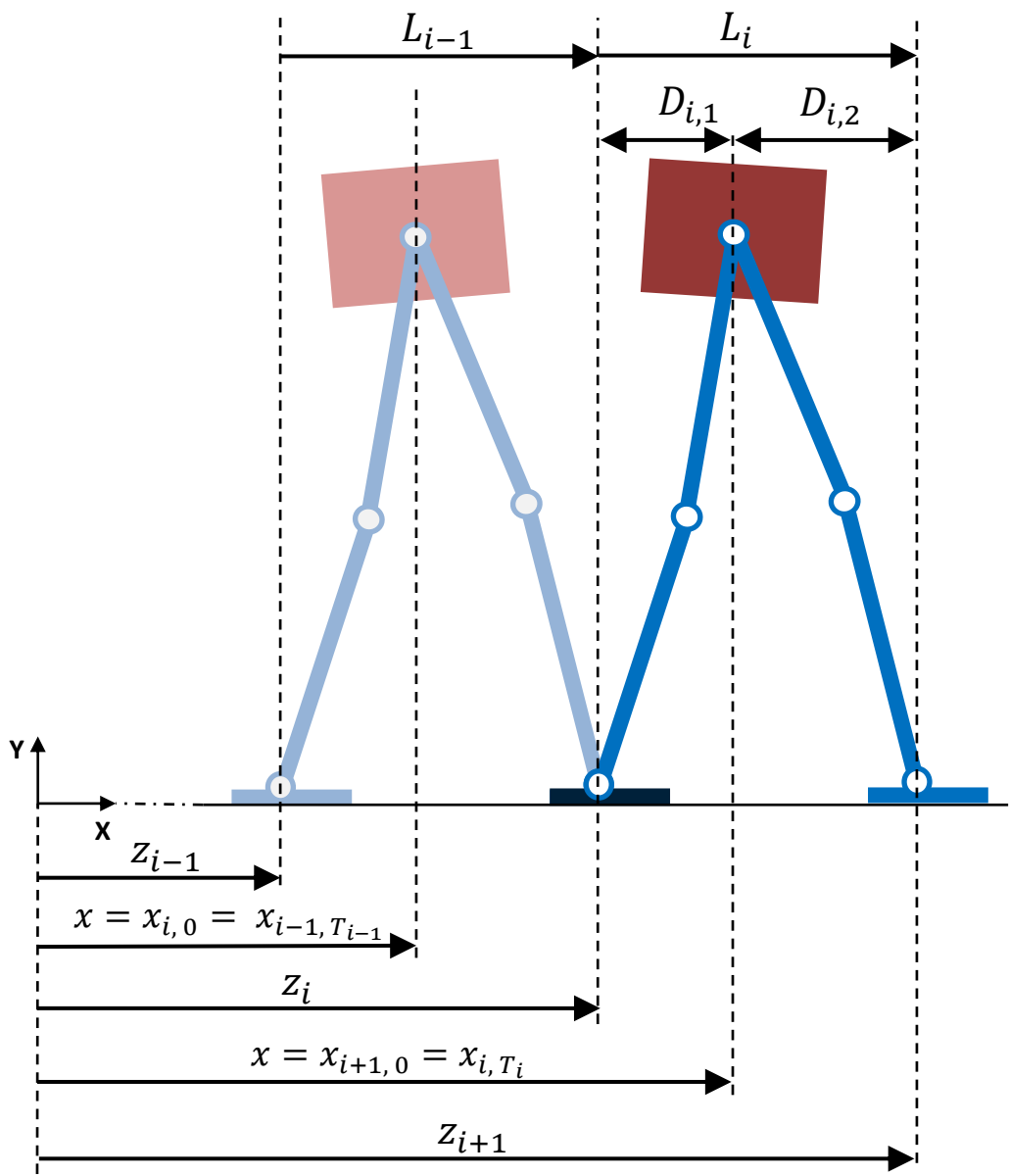

شکل ۳-۱۵. شمایی از حرکت مدل فیزیکی راهرونده با بالاتنه صلب و پاهایی با جرم قابل‌صرف‌نظر در ابتدا و انتهای یک گام

### ۵-۳-۳ بررسی نارسایی‌های پایدارسازهای سیکل‌های حرکتی با شبیه‌سازی بر روی مدل فیزیکی کامل

برای شبیه‌سازی کنترل‌کننده پیشنهادی در دو سطح کنترل پایداری و کنترل اندازه حرکت بـر روی مـدل کامـل، ابتدا یک شرایط معیار برای مسئله به نام شرایط مسئله محک[1] درنظرمی‌گیریم و سپس توانایی و نارسایی هـر یـک از کنترل‌کننده‌های پایداری را بر روی مدل فیزیکی کامل راه‌رونده با درنظرگرفتن شرایط مسئله محک، بررسی مـی‌کنیم.

همچنین دو پایدارساز دوم و‌چهارم سیکل حرکتی از میان پایدارسازهای پیشنهادی سیکل حرکتی، بـا توجـه بـه محدوده شرایط اولیه قابل‌کنترل‌شان، توانایی هدایت حرکت از یک شرایط اولیه دور از سیکل بـه سـمت سیکل را دارند و می‌توانند راه‌رونده را از یک شرایط اولیه مربوط به حرکتی پس‌رونده به سمت یـک سـیکل حرکتـی پیش‌رونده حرکت دهند که از آن با عنوان توانایی هدایت یا سوق دادن حرکت از هـر شـرایط اولیـه دلخـواه بـه سـمت سیکل حرکتی مطلوب یاد می‌کنیم (اگرچه این دو کنترل‌کننده الزامـی بـه رعایـت کامـل قیـود مسـئله ازجملـه قیـود موجود بر روی $D_{i,1}$ و $D_{i,2}$ ندارند.). به همین جهت و همچنین برای مقایسه آنها با روش کنترل پایـداری بهینـه کـه در فصل بعدی مطرح خواهد شد، علاوه بر مسئله محک، در ادامه همین قسمت به شبیه‌سازی و بررسی عملکـرد ایـن دو پایدارساز برای هدایت حرکت از یک شرایط اولیه پس‌رونده به سمت یک سیکل حرکتی پیش‌رونده خـواهیم پرداخت.

همان‌طور که خواهیم دید، اگرچه برخی از پایدارسازهای سیکل‌های حرکتی مـی‌تواننـد شـرایط اولیـه دور از سیکل را نیز به سمت سیکل همگرا کنند ولی با توجه به درنظر نگرفتن رابطه قیود در این کنترل‌کننـده‌هـا، الزامـی در رعایت قیود ندارند. اگرچه می‌توان اصلاحی بر روی این کنترل‌کننده‌ها انجام داد تا همه قیـود را رعایـت کننـد ولـی ترجیح می‌دهیم در فصل بعدی با توجه به درکی که از شاخص پایداری پیدا کرده‌ایم، نوعی کنترل پایداری بـر پایـه بهینه‌سازی غیرخطی طراحی کنیم که همزمان شاخص پایداری را بهینه و قیود را رعایت خواهد کرد.

● **شرایط مسئله محک**

در این مسئله، مقادیر سیکل حرکتی مطلوب و پارامترهای راه‌رونده را نزدیک به حرکت راه‌رفتن یـک انسـان بـا قد متوسط درنظر می‌گیریم. به عنوان مثال، ارتفاع مرکز جرم، ۱ متـر و سـرعت متوسـط سـیکل حرکتـی، ۱/۲۵ متـر برثانیه برابر با ۴/۵ کیلومتر بر ساعت فرض شده است که تقریبا برابر با سرعت متوسط راه‌رفتن انسان اسـت. همچنـین طول قدم در سیکل حرکتی، ۵۰ سانتی‌متر و حداکثر طول هر قدم، ۷۵ سانتی متر فرض شده است.

شبیه‌سازی در نزدیکی حد بالایی شرایط اولیه مرزی مربوط هر کنترل‌کننده شروع می‌شـود (بـا توجـه بـه روابـط مربوط به محدوده شرایط اولیه قابل کنترل هر پایدارساز) و تا ۲۰ گام پیش خواهد رفت که در میانه گام ۷، ضـربه‌ای با نسبتی مشخص از یک ضربه معیار در هر سه بخش اندازه حرکت به بالاتنه راه‌رونده وارد مـی‌شـود. جهـت ضـربه معیار برای ایجاد حداکثر ناپایداری ممکن، در راستای اندازه حرکت خطی، رو به جلو وپایین بـا زاویـه ۴۵ زیـر افـق

---

[1] Benchmark

(در راستای افقی روبه‌جلو، در جهت عمودی رو به زمین) و در راستای اندازه حرکت زاویه‌ای، ساعتگرد می‌باشد. برای هر روش کنترلی، حداکثر ضربه‌ای که قابل تحمل است، وارد می‌شود و نسبت آن به صورت درصدی از ضربه معیار ضبط می‌گردد. اگرچه برای رسیدن به حداکثر ضربه قابل تحمل، شبیه‌سازی های زیادی انجام شد، تنها نتایج نهایی برای حداکثر ضربه قابل تحمل معیار قضاوت ما خواهد بود.

پارامترهای راه‌رونده، سیکل حرکتی مطلوب، وضعیت قرارگیری پاها وبالاتنه در لحظه شروع و پارامترهای محیط از جمله اندازه ضربه معیار برای این مسئله محک به صورت جدول ۳-۲ است.

جدول ۳-۲. پارامترهای مربوط به شبیه‌سازی مدل فیزیکی کامل و مسئله محک

| پارامترهای راه‌رونده | |
|---|---|
| ارتفاع متوسط مرکز جرم | $h = 1\ m$ |
| شتاب جاذبه زمین | $g = 9.8\ m.s^{-2}$ |
| حداکثر اندازه طول گام | $L_{max} = 0.75\ m$ |
| حداکثر سرعت متوسط قابل دسترس | $V_{max} = 3\ m.s^{-1}$ |
| جمع کل زمان لحظات جدا شدن و فرود آوردن پا | $T_0 = 0.05\ s$ |
| جرم بالاتنه (جرم کل) | $M = 50\ kg$ |
| ممان اینرسی بالاتنه | $I = 4\ kg.m^2$ |
| مسیر مطلوب حرکت در راستای عمودی | $y_{des} = h + A_y \sin(2\pi t_i/T_c + \varphi_y)$ $A_y = 0.025\ m,\ \varphi_y = -\pi/2$ |
| مسیر مطلوب حرکت در راستای دورانی | $H_{des} = I\dot{\theta}_{des} = A_H \sin(2\pi t_i/T_c + \varphi_H)$ $A_H = 0.05MV_cH\ N.m.s,\ \varphi_H = 0$ |
| پارامترهای ضربه معیار | |
| ضربه در راستای اندازه حرکت افقی | $\Delta L_x = 10\ N.s$ |
| ضربه در راستای اندازه حرکت عمودی | $\Delta L_y = -10\ N.s$ |
| ضربه در راستای اندازه حرکت زاویه‌ای | $\Delta H_z = -10\ N.m.s$ |
| زمان شروع و مدت زمان وارد شدن ضربه | $i = 7,\ t_i = 0.15\ s,\ \Delta t = 0.05\ s$ |
| پارامترهای سیکل حرکتی مطلوب | |
| شرایط اولیه مولفه همگرا | $p_c = -0.7\ m$ |
| شرایط اولیه مولفه همگرا | $q_c = 0.2\ m$ |
| طول گام | $L_c = 0.5\ m$ |
| زمان فرود گام | $T_c = 0.4\ s$ |
| سرعت متوسط حرکت | $V_c = 1.25\ m.s^{-1}$ |

| شرایط اولیه حرکتی | |
|---|---|
| موقعیت مرکز جرم در ابتدای حرکت | $x_0 = x_{1,0} = -0.1\ m$ |
| موقعیت پای متحرک در ابتدای حرکت | $z_0 = -0.2\ m$ |
| موقعیت پای تکیه گاهی در ابتدای حرکت | $z_1 = 0$ |
| ارتفاع مرکز جرم در ابتدای حرکت | $y_0 = 0.95\ m$ |
| زاویه بالاتنه در ابتدای حرکت | $\theta_0 = -10° = -0.175\ rad$ |

- **شبیه سازی پایدارساز اول سیکل حرکتی بر روی مدل کامل با شرایط مسئله محک**

نتایج به صورت مسیر حرکت مؤلفه واگرا به همگرا در صفحه فاز شکل ۱۶-۳ نمایش داده شده است. برای بخش پیوسته معادلات حرکت، نمودارهای شکل ۱۷-۳، متغیرهای سیستم از جمله سرعت و شتاب حرکت در راستای افقی ($dx/dt$ و $d^2x/dt^2$)، موقعیت عمودی ($y$) و زاویه بالاتنه ($\theta$) را به همراه متغیرهای کنترلی نیروهای افقی و عمودی و گشتاور بالاتنه ($F_x^{control}$، $F_x^{control}$ و $T_z^{control}$) و همچنین متغیر جابجایی مرکز فشار ($\Delta z_i^{des}$)، را نمایش می‌دهند. همچنین ضریب اصطکاک مورد نیاز ($\mu_{required}$) برای تامین نیروهای کنترلی این حرکت، به صورت نموداری بر حسب زمان مشخص شده‌است. برای بخش گسسته معادلات حرکت، نمودارهای شکل ۱۸-۳، متغیرهای کنترلی از جمله طول و زمان فرود هرگام ($L_i$ و $T_i$) با کران‌های بالا و پایین هریک ($L_{max}$ و $T_{min}$) را به همراه متغیرهای سیستم از جمله شرایط اولیه مؤلفه‌های واگرا ($q_{i+1,0}$) و همگرا ($p_{i+1,0}$) با مقدار شاخص آن ($p_i^*$) و نیز سرعت متوسط حرکت نسبی پای متحرک نسبت به بالاتنه ($V_{i,swing}$) با کران بالای آن ($V_{max}$) و سرعت متوسط مرکز جرم در طول هر گام ($V_{i,avg}$)، نمایش داده شده است. همچنین متغیرهای $D_{i,1}$ و $D_{i,2}$ به همراه حداکثر مقدار مجازشان ($L_{max}/2$) در این نمودارها مشخص شده است.

پایدارساز اول با شروع از شرایط اولیه مرزی پس از سه گام به پایداری کامل می‌رسد و با وارد شدن ضربه در گام هفتم نیز، پس از سه گام دوباره به پایداری کامل می‌رسد و در عین حال قیود مسئله بر روی $D_{i,2}$، $D_{i,1}$ و $V_{i,swing}$ را رعایت می‌کنند. پایدارساز اول با وجود ناتوانی تئوریک از کنترل شرایط اولیه دور از سیکل ولی توانست ضربه‌ای با شدت ۹۰٪ ضربه معیار را تحمل کند و با سرعت همگرایی زیادی(طی سه گام) به پایداری کامل برسد. علت این امر، حاضر بودن این کنترل‌کننده پایداری در کل بازه زمانی گام‌ها است که می‌تواند بلافاصله به جبران عوامل اختلالی بپردازد.

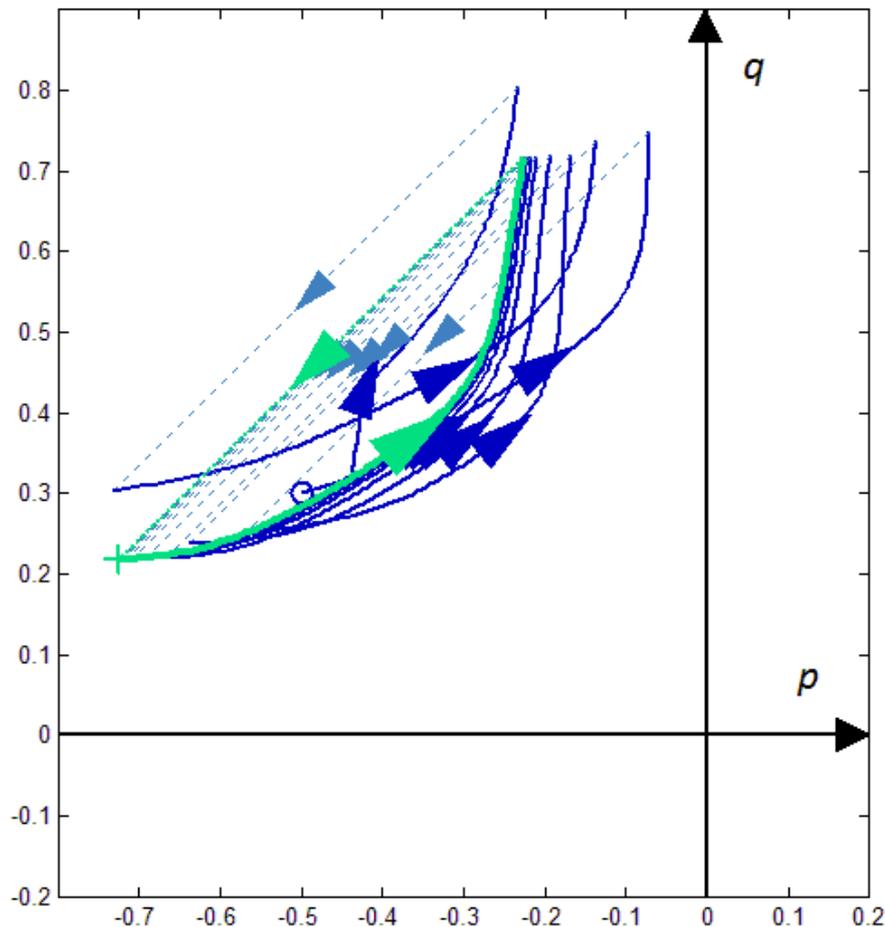

$p_c = -0.7$ , $q_c = 0.2$ , $T_c = 0.4$ , $L_c = 0.5$ , $V_c = 1.25$
$p_{1,0} = -0.5$ , $q_{1,0} = 0.3$ , $x_0 = -0.1$ , $dx/dt_0 = 1.253$
$z_0 = -0.2$ , $z_1 = 0$ , $\Delta z_{min} = -0.11$ , $\Delta z_{max} = 0.11$

شکل ۳-۱۶. صفحه فاز مسیر حرکت مولفه واگرا نسبت به مولفه همگرا برای شبیه‌سازی پایدارساز اول سیکل حرکتی بر روی مدل کامل راه‌رونده با شرایط مسئله محک (۹۰٪ ضربه) — منحنی ضخیم و کم‌رنگ مسیر گام آخر را نمایش می‌دهد.

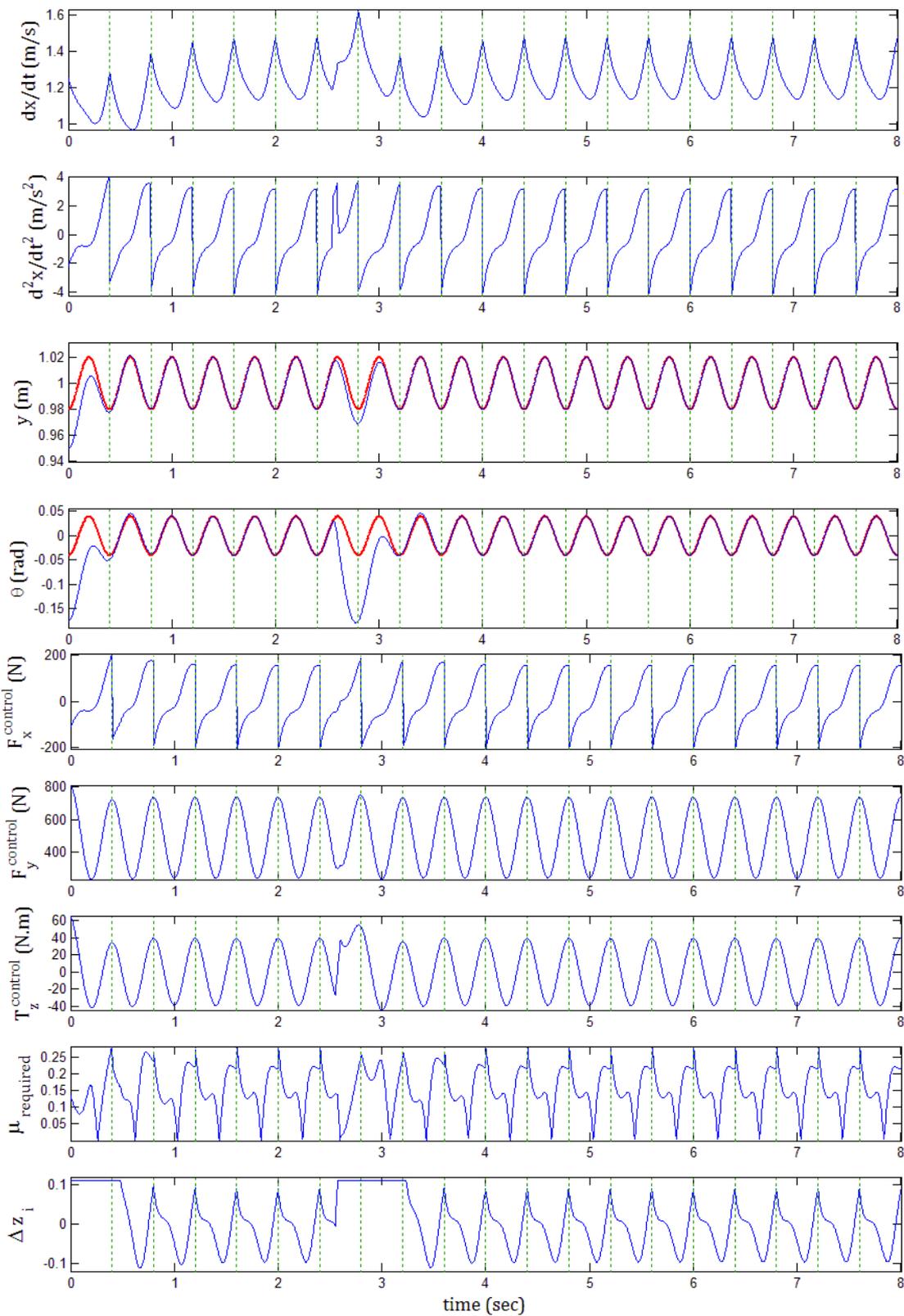

شکل ۳-۱۷. نمودارهای بخش پیوسته معادلات حرکت برای شبیه‌سازی پایدارساز اول سیکل حرکتی بر روی مدل کامل راه‌رونده با
شرایط مسئله محکک (۹۰٪ ضربه) — خطوط ضخیم، مسیر مطلوب متغیر را نمایش می‌دهند.

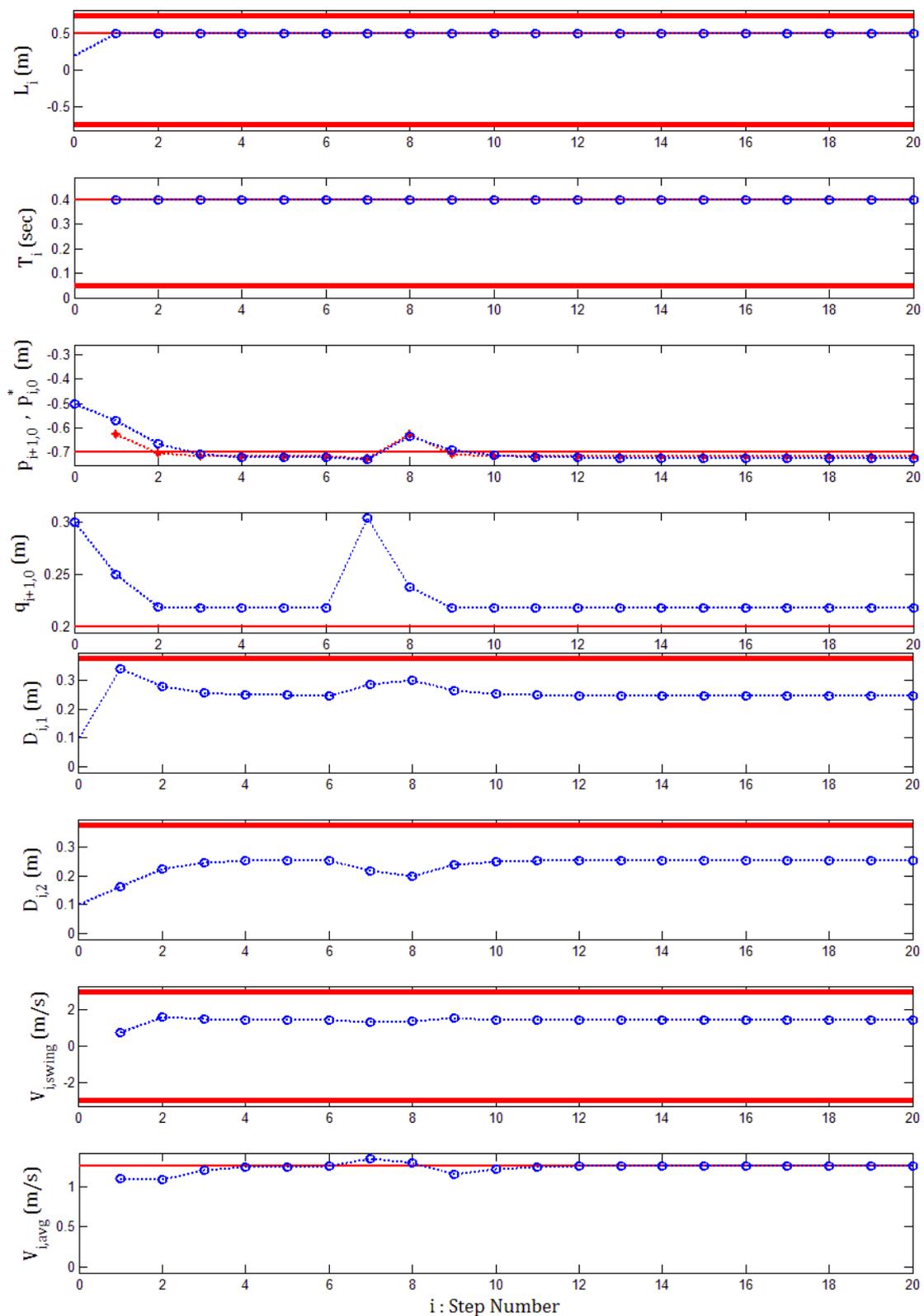

شکل ۳-۱۸. نمودارهای بخش گسسته معادلات حرکت برای شبیه‌سازی پایدارساز اول سیکل حرکتی بر روی مدل کامل راهرونده با شرایط مسئله محک (۹۰٪ ضربه) – خطوط با ضخامت متوسط و زیاد، به ترتیب مقدار مطلوب متغیر در سیکل حرکتی و محدودیت بالا ویا پایین آن را نمایش می‌دهند.

- **شبیه سازی پایدارساز دوم سیکل حرکتی بر روی مدل کامل با شرایط مسئله محک**

نتایج به صورت مسیر حرکت مؤلفه واگرا به همگرا در صفحه فاز شکل ۱۹-۳ نمایش داده شده است. برای بخش پیوسته معادلات حرکت، نمودارهای شکل ۲۰-۳، متغیرهای سیستم از جمله سرعت و شتاب حرکت در راستای افقی( $dx/dt$ و $d^2x/dt^2$ ) ، موقعیت عمودی( $y$ ) و زاویه بالاتنه( $\theta$ ) را به همراه متغیرهای کنترلی نیروهای افقی و عمودی و گشتاور بالاتنه( $T_z^{control}$ و $F_x^{control}$ ، $F_x^{control}$ )، نمایش می‌دهند. همچنین ضریب اصطکاک مورد نیاز( $\mu_{required}$ ) برای تامین نیروهای کنترلی این حرکت، به صورت نموداری بر حسب زمان مشخص شده‌است. برای بخش گسسته معادلات حرکت، نمودارهای شکل ۲۱-۳، متغیرهای کنترلی از جمله طول و زمان فرود هرگام( $L_i$ و $T_i$ ) با کران‌های بالا و پایین هریک( $L_{max}$ و $T_{min}$ ) را به همراه متغیرهای سیستم از جمله شرایط اولیه مؤلفه‌های واگرا( $q_{i+1,0}$ ) و همگرا ( $p_{i+1,0}$ ) با مقدار شاخص آن( $p_i^*$ ) و نیز سرعت متوسط حرکت نسبی پای متحرک نسبت به بالاتنه( $V_{i,swing}$ ) با کران بالای آن( $V_{max}$ ) و سرعت متوسط مرکز جرم در طول هر گام( $V_{i,avg}$ )، نمایش داده شده است. همچنین متغیرهای قید $D_{i,1}$ و $D_{i,2}$ به همراه حداکثر مقدار مجازشان ( $L_{max}/2$ ) در این نمودارها مشخص شده است.

پایدارساز دوم با شروع از شرایط اولیه مرزی پس از پنج گام به پایداری کامل می‌رسد و با وارد شدن ضربه در گام هفتم نیز، پس از شش گام دوباره به پایداری کامل می‌رسد ولی دو قید مسئله بر روی $D_{i,1}$ ، $D_{i,2}$ در هر دو مرحله ابتدای حرکت و پس از دریافت ضربه تا حدی رعایت نشده است. پایدارساز دوم با وجود توانایی کنترل شرایط اولیه نسبتا دور از سیکل تنها توانست ضربه‌ای با شدت ۵۰٪ ضربه معیار را تحمل کند و علاوه بر آن قیود مسئله را برای شرایط مرزی و حداکثر ضربه قابل تحمل خود به درستی رعایت نمی‌کند. اگرچه برای شرایط اولیه‌ای نزدیک به سیکل حرکتی و ضرباتی کوچک، این قیود رعایت می‌شوند و این روش می‌تواند برای پایداری راه‌رونده به کار رود ولی برای شرایط اولیه دور و یا ضربات با شدت زیاد، این کنترل‌کننده توانایی بالایی ندارد.

$$p_c = -0.7 \ , \ q_c = 0.2 \ , \ T_c = 0.4 \ , \ L_c = 0.5 \ , \ V_c = 1.25$$
$$p_{1,0} = -0.49 \ , \ q_{1,0} = 0.29 \ , \ x_0 = -0.1 \ , \ dx/dt_0 = 1.222$$
$$z_0 = -0.2 \ , \ z_1 = 0$$

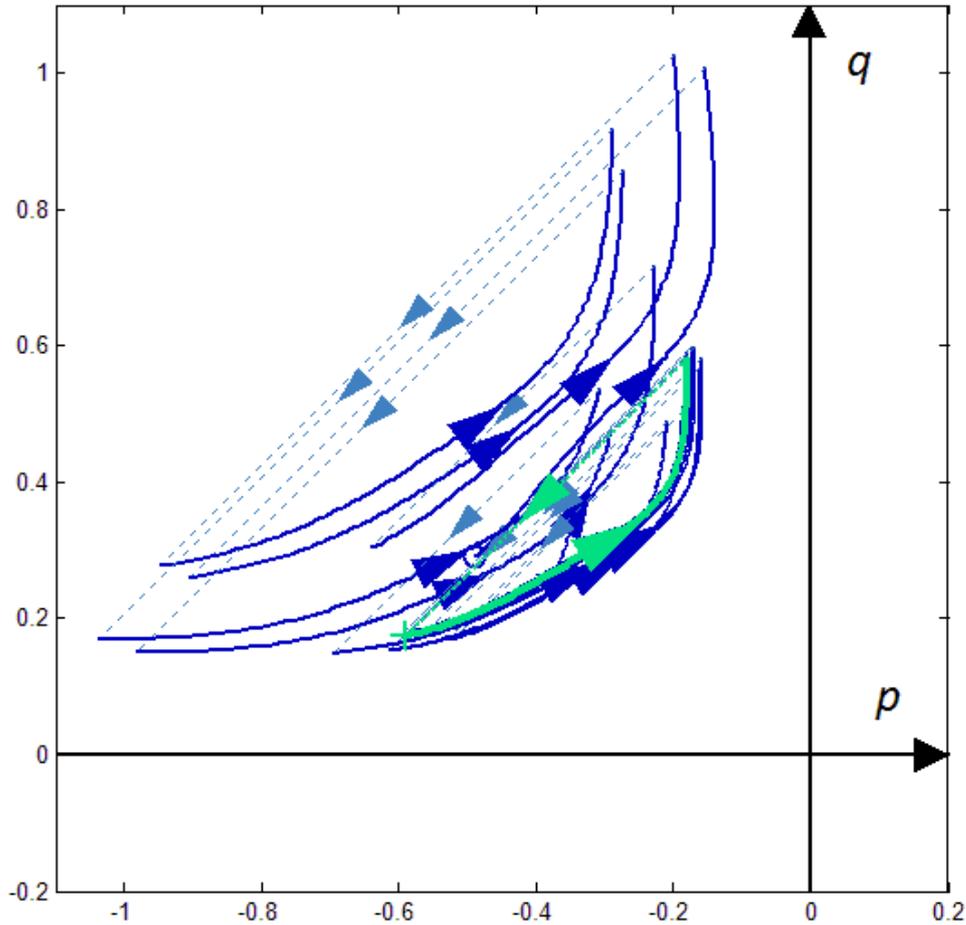

شکل ۳-۱۹. صفحه فاز مسیر حرکت مولفه واگرا نسبت به مولفه همگرا برای شبیه‌سازی پایدارساز دوم سیکل حرکتی بر روی مدل

فیزیکی با شرایط مسئله محک (۵۰٪ ضربه) − منحنی ضخیم و کمرنگ مسیر گام آخر را نمایش می‌دهد.

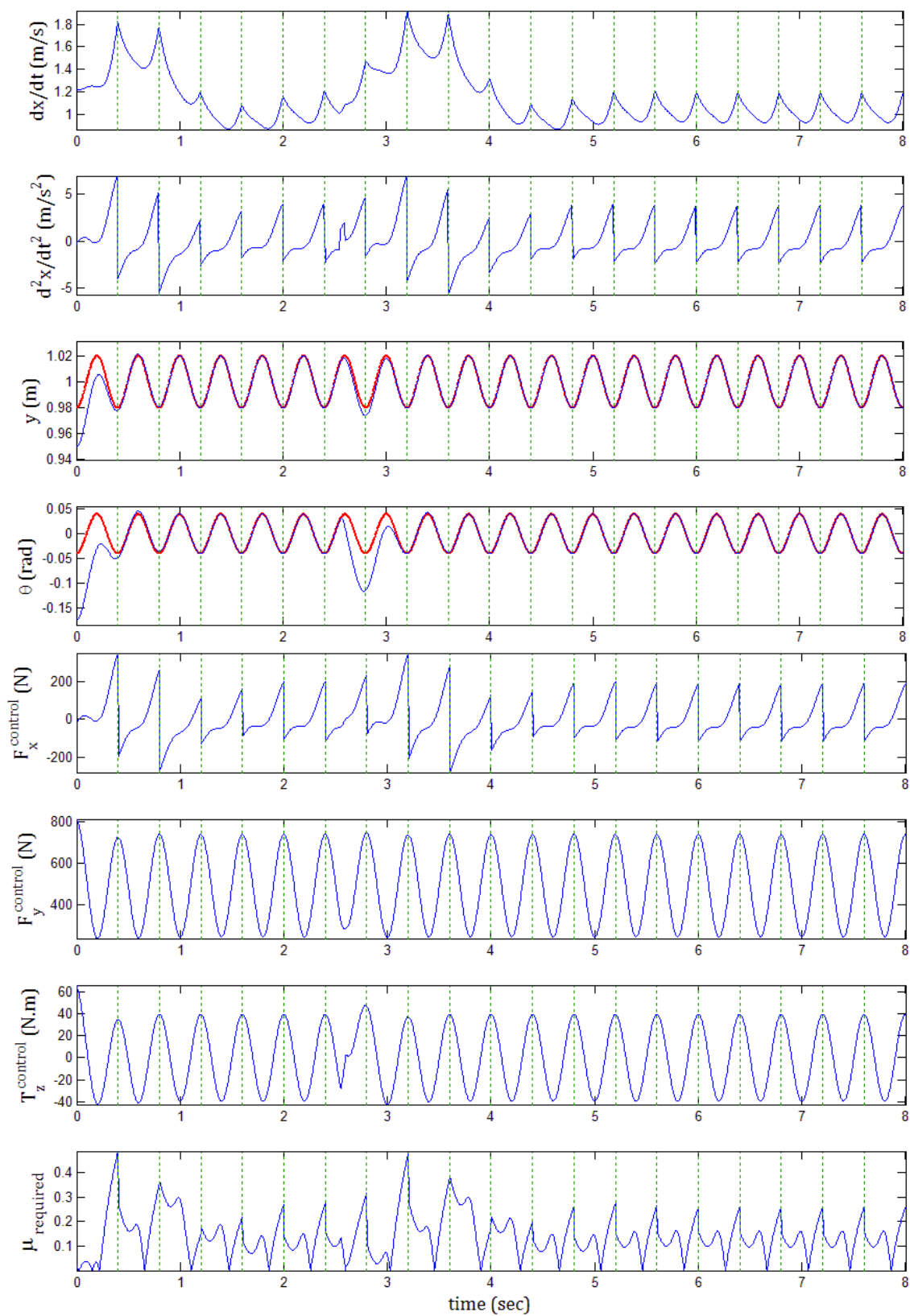

شکل ۳-۲۰. نمودارهای بخش پیوسته معادلات حرکت برای شبیه‌سازی پایدارساز دوم سیکل حرکتی بر روی مدل کامل راه‌رونده با

شرایط مسئله محک (۵۰٪ ضربه) — خطوط ضخیم، مسیر مطلوب متغیر را نمایش می‌دهند.

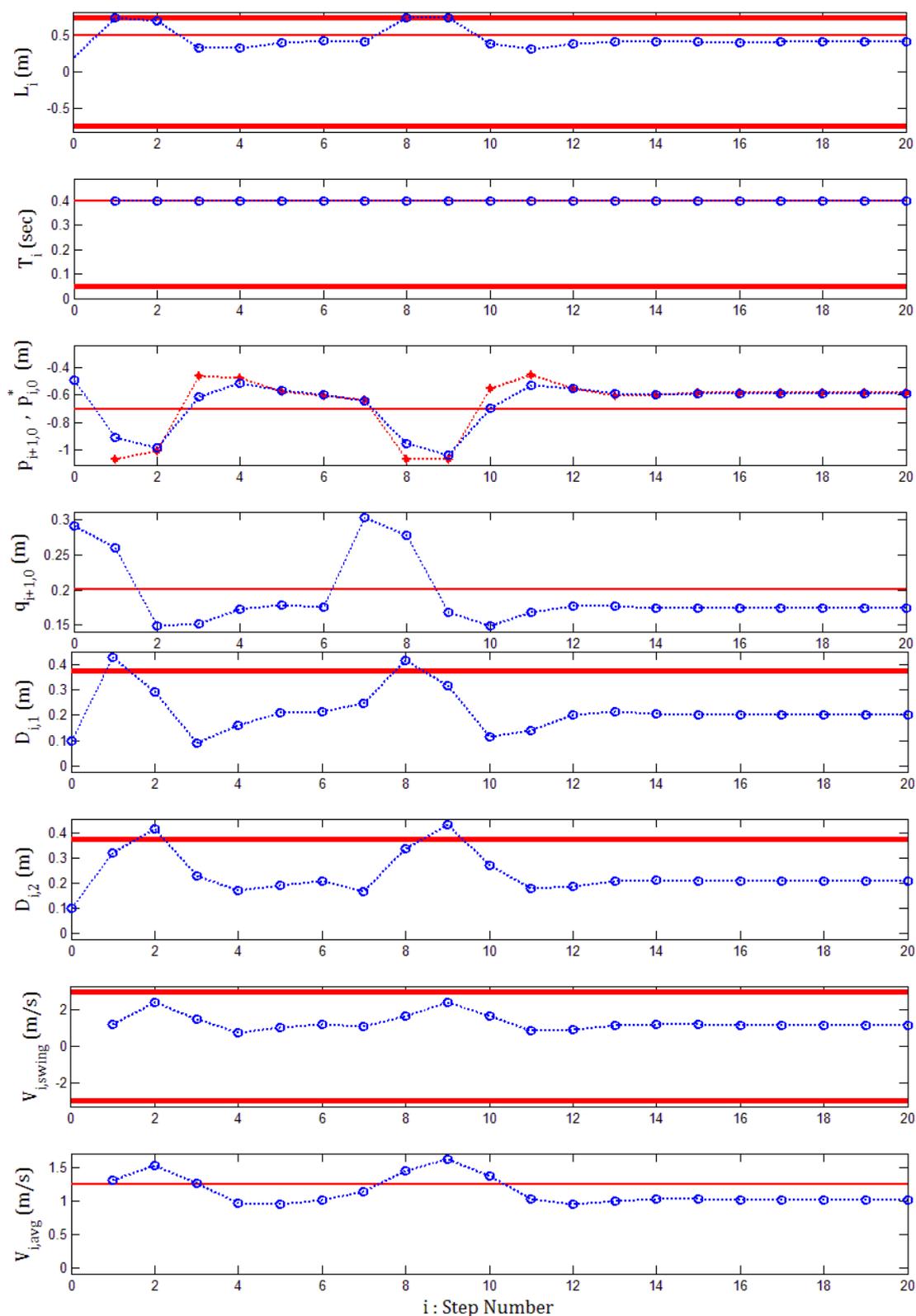

شکل ۳-۲۱. نمودارهای بخش گسسته معادلات حرکت برای شیه‌سازی پایدارساز دوم سیکل حرکتی بر روی مدل کامل راه‌رونده با

شرایط مسئله محک (۵۰٪ ضربه) – خطوط با ضخامت متوسط و زیاد، به ترتیب مقدار مطلوب متغیر در سیکل حرکتی و محدودیت بالا

ویا پایین آن را نمایش می‌دهند.

- **شبیه سازی پایدارساز سوم سیکل حرکتی بر روی مدل کامل با شرایط مسئله محک**

نتایج به صورت مسیر حرکت مؤلفه واگرا به همگرا در صفحه فاز شکل ۲۲-۳ نمایش داده شده است. برای بخش پیوسته معادلات حرکت، نمودارهای شکل ۲۳-۳، متغیرهای سیستم از جمله سرعت و شتاب حرکت در راستای افقی( $dx/dt$ و $d^2x/dt^2$ )، موقعیت عمودی( $y$ ) و زاویه بالاتنه( $\theta$ ) را به همراه متغیرهای کنترلی نیروهای افقی و عمودی و گشتاور بالاتنه( $T_z^{control}$ و $F_x^{control}$ ، $F_x^{control}$ )، نمایش می‌دهند. همچنین ضریب اصطکاک مورد نیاز( $\mu_{required}$ ) برای تأمین نیروهای کنترلی این حرکت، به صورت نموداری بر حسب زمان مشخص شده‌است. برای بخش گسسته معادلات حرکت، نمودارهای شکل ۲۴-۳، متغیرهای کنترلی از جمله طول و زمان فرود هر گام( $L_i$ و $T_i$ ) با کران‌های بالا و پایین هریک( $L_{max}$ و $T_{min}$ ) را به همراه متغیرهای سیستم از جمله شرایط اولیه مؤلفه‌های واگرا( $q_{i+1,0}$ ) و همگرا ( $p_{i+1,0}$ ) با مقدار شاخص آن( $p_i^*$ ) و نیز سرعت متوسط حرکت نسبی پای متحرک نسبت به بالاتنه( $V_{i,swing}$ ) با کران بالای آن( $V_{max}$ ) و سرعت متوسط مرکز جرم در طول هر گام( $V_{i,avg}$ )، نمایش داده شده است. همچنین متغیرهای قید $D_{i,1}$ و $D_{i,2}$ به همراه حداکثر مقدار مجازیشان ( $L_{max}/2$ ) در این نمودارها مشخص شده است.

پایدارساز سوم با شروع از شرایط اولیه مرزی نسبتا دور از سیکل پس از پنج گام به پایداری کامل می‌رسد و با وارد شدن ضربه در گام هفتم نیز، پس از پنج گام دوباره به پایداری کامل می‌رسد و در عین حال دو قید مسئله را بر روی $D_{i,1}$ ، $D_{i,2}$ در هر دو مرحله ابتدای حرکت و پس از دریافت ضربه کاملا رعایت کرده است، با این وجود تا حدی از قید سرعت بر روی $V_{i,swing}$ تجاوز می‌کند. پایدارساز سوم هم توانایی بالایی در کنترل شرایط اولیه نسبتا دور از سیکل دارد و همزمان توانست ضربه‌ای با شدت ۱۳۰٪ ضربه معیار را تحمل کند و در عین حال قیود مسئله را برای شرایط مرزی و حداکثر ضربه قابل تحمل خود را تا حد زیادی رعایت می‌کند. اما همان‌طور که در مباحث تئوری دیدیم، این کنترل‌کننده ناتوان از هدایت راه‌رونده از شرایط اولیه پس‌رونده به سمت سیکلی پیش‌رونده است که به دلیل آن جهت یکطرفه گام‌ها در این پایدارساز همواره رو به جلو فرض شده‌است. با این‌حال با توجه به رعایت قیود، به نظر می‌رسد که عملکرد آن برای حرکت حول سیکل‌های پیش‌رونده دارای سرعت متوسط بالا و همچنین مقاوت آن در برابر ضربات شدید، مناسب باشد و میتواند بر روی راه‌رونده‌های واقعی به‌کار گرفته شود.

$$p_c = -0.7 \ , \ q_c = 0.2 \ , \ T_c = 0.4 \ , \ L_c = 0.5 \ , \ V_c = 1.25$$
$$p_{1,0} = -0.705 \ , \ q_{1,0} = 0.505 \ , \ x_0 = -0.1 \ , \ dx/dt_0 = 1.895$$
$$z_0 = -0.2 \ , \ z_1 = 0$$

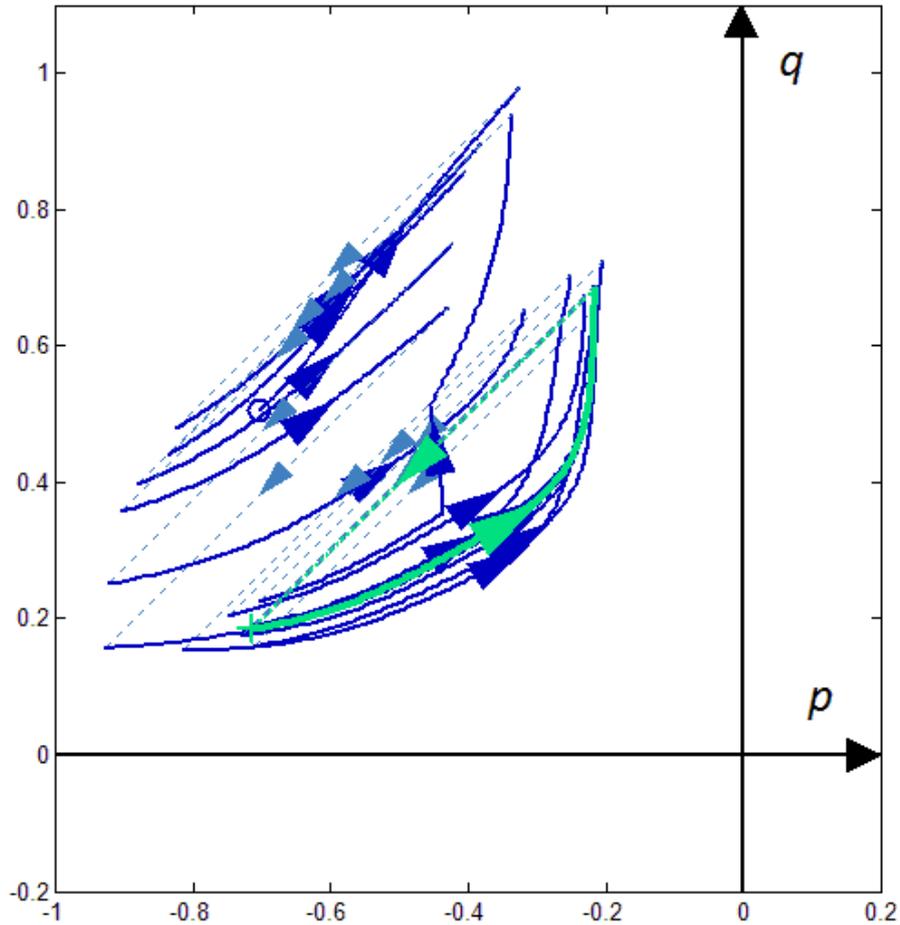

شکل ۳-۲۲. صفحه فاز مسیر حرکت مولفه واگرا نسبت به مولفه همگرا برای شبیه‌سازی پایدارساز سوم سیکل حرکتی بر روی مدل کامل راه‌رونده با شرایط مسئله محک (۱۳۰٪ ضربه) — منحنی ضخیم و کمرنگ مسیر گام آخر را نمایش می‌دهد.

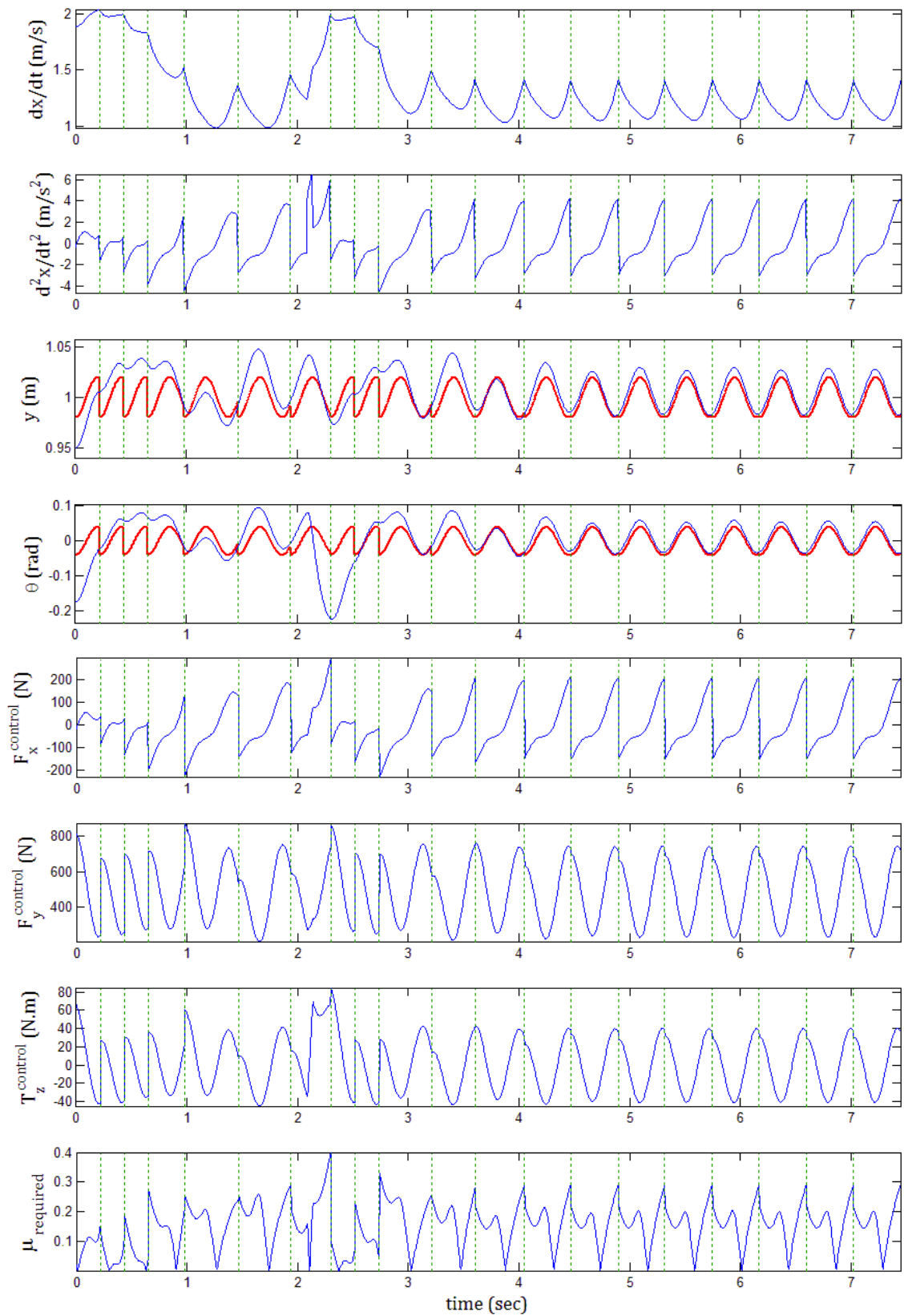

شکل ۳-۲۳. نمودارهای بخش پیوسته معادلات حرکت برای شبیه‌سازی پایدارساز سوم سیکل حرکتی بر روی مدل کامل راه‌رونده با شرایط مسئله محک (۱۳۰٪ ضربه) — خطوط ضخیم، مسیر مطلوب متغیر را نمایش می‌دهند.

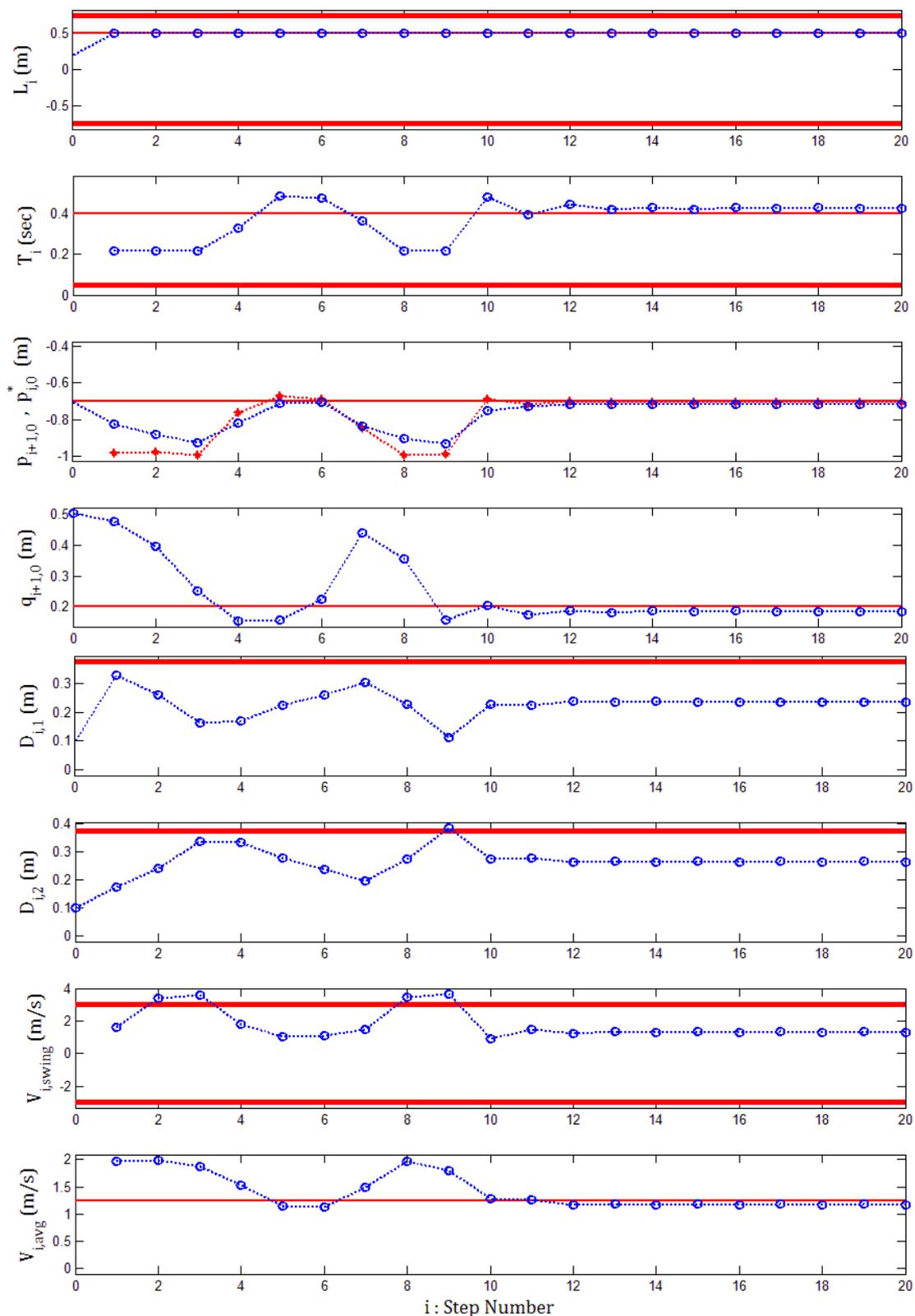

شکل ۳-۲۴. نمودارهای بخش گسسته معادلات حرکت برای شبیه‌سازی پایدارساز سوم سیکل حرکتی بر روی مدل کامل راه‌رونده با شرایط مسئله محک (۱۳۰٪ ضربه) – خطوط با ضخامت متوسط و زیاد، به ترتیب مقدار مطلوب متغیر در سیکل حرکتی و محدودیت بالا ویا پایین آن را نمایش می‌دهند.

- **شبیه سازی پایدارساز چهارم سیکل حرکتی بر روی مدل کامل با شرایط مسئله محک**

نتایج به صورت مسیر حرکت مولفه واگرا به همگرا در صفحه فاز شکل ۳-۲۵ نمایش داده شده است. برای بخش پیوسته معادلات حرکت، نمودارهای شکل ۳-۲۶، متغیرهای سیستم از جمله سرعت و شتاب حرکت در راستای افقی($dx/dt$ و $d^2x/dt^2$)، موقعیت عمودی($y$) و زاویه بالاتنه($\theta$) را به همراه متغیرهای کنترلی نیروهای افقی و عمودی و گشتاور بالاتنه($F_x^{control}$، $F_x^{control}$ و $T_z^{control}$)، نمایش میدهند. همچنین ضریب اصطکاک مورد نیاز($\mu_{required}$) برای تامین نیروهای کنترلی این حرکت، به صورت نموداری بر حسب زمان مشخص شده است. برای بخش گسسته معادلات حرکت، نمودارهای شکل ۳-۲۷، متغیرهای کنترلی از جمله طول و زمان فرود هر گام($L_i$ و $T_i$) با کرانهای بالا و پایین هریک($L_{max}$ و $T_{min}$) را به همراه متغیرهای سیستم از جمله شرایط اولیه مولفههای واگرا($q_{i+1,0}$) و همگرا($p_{i+1,0}$) با مقدار شاخص آن($p_i^*$) و نیز سرعت متوسط حرکت نسبی پای متحرک نسبت به بالاتنه($V_{i,swing}$) با کران بالای آن($V_{max}$) و سرعت متوسط مرکز جرم در طول هر گام($V_{i,avg}$)، نمایش داده شده است. همچنین متغیرهای قید $D_{i,1}$ و $D_{i,2}$ به همراه حداکثر مقدار مجازشان($L_{max}/2$) در این نمودارها مشخص شده است.

پایدارساز چهارم با شروع از شرایط اولیه مرزی پس از شش گام به پایداری کامل میرسد و با وارد شدن ضربه در گام هفتم نیز، پس از نه گام دوباره به پایداری کامل میرسد ولی سه قید مسئله بر روی $D_{i,1}$، $D_{i,2}$ و $V_{i,swing}$ در هر دو مرحله ابتدای حرکت و پس از دریافت ضربه تا حدی رعایت نشده است. پایدارساز چهارم با وجود توانایی کنترل شرایط اولیه نسبتا دور از سیکل و توانایی تحمل ضربهای با شدت ۱۴۰٪ ضربه معیار، قیود مسئله را برای شرایط اولیه مرزی و حداکثر ضربه قابل تحمل خود به درستی رعایت نمیکند. اگرچه برای شرایط اولیهای نزدیکتر به سیکل حرکتی و ضرباتی متوسط، این قیود رعایت میشوند و این روش میتواند برای پایداری راه‌رونده واقعی به کار رود ولی برای شرایط اولیه دور و یا ضربات با شدت زیاد به دلیل رعایت نکردن قیود، کارایی بالایی ندارد.

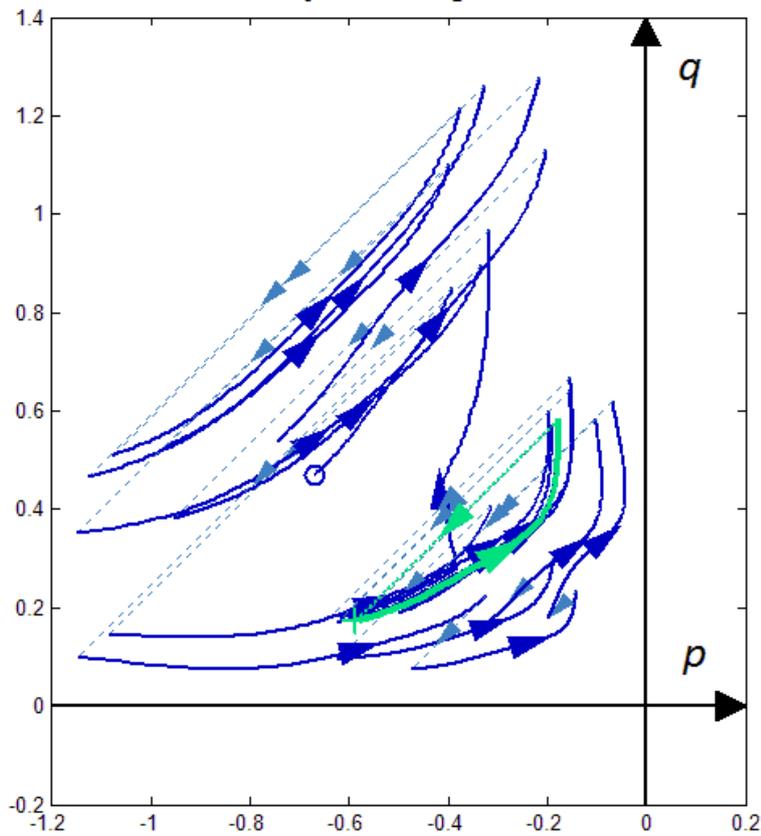

شکل ۳-۲۵. صفحه فاز مسیر حرکت مولفه واگرا نسبت به مولفه همگرا برای شبیه‌سازی پایدارساز چهارم سیکل حرکتی بر روی مدل کامل راه‌رونده با شرایط مسئله محکک (۱۴۰٪ ضربه) — منحنی ضخیم و کمرنگ مسیر گام آخر را نمایش می‌دهد.

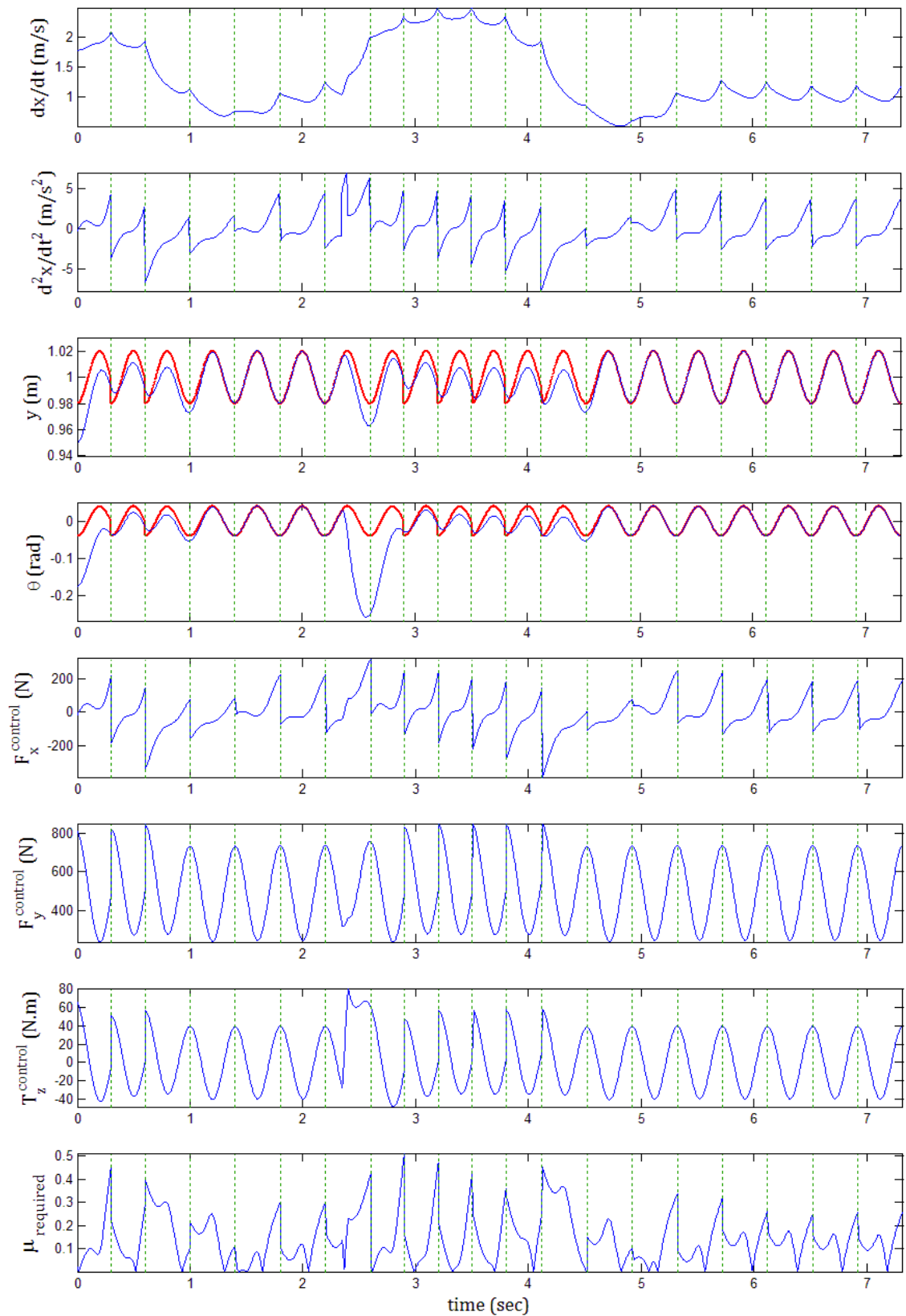

شکل ۳-۲۶. نمودارهای بخش پیوسته معادلات حرکت برای شبیه‌سازی پایدارساز چهارم سیکل حرکتی بر روی مدل کامل راه‌رونده با

شرایط مسئله محک (۱۳۰٪ ضربه) — خطوط ضخیم، مسیر مطلوب متغیر را نمایش می‌دهند.

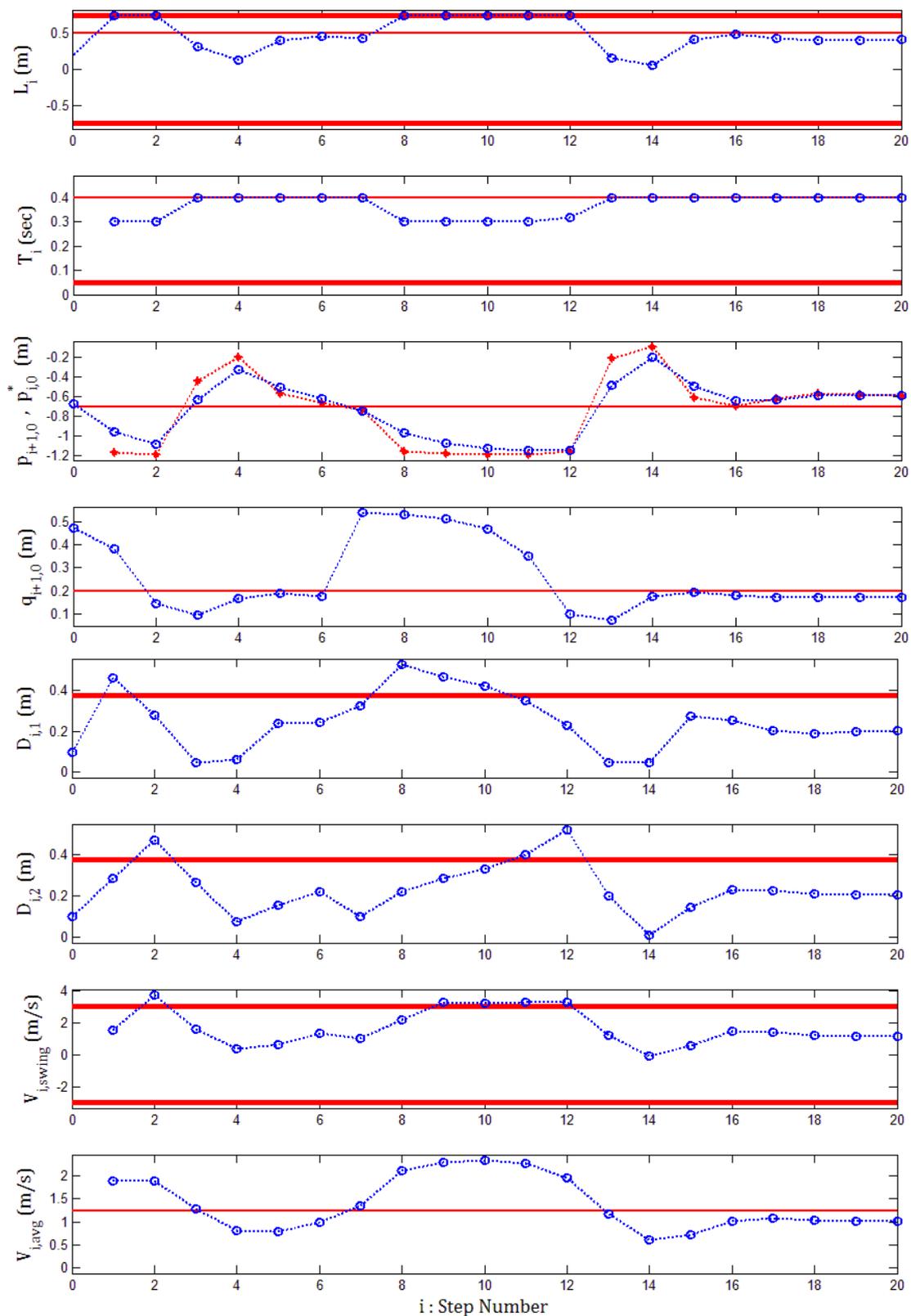

شکل ۳-۲۷. نمودارهای بخش گسسته معادلات حرکت برای شبیه‌سازی پایدارساز چهارم سیکل حرکتی بر روی مدل کامل راه‌رونده با شرایط مسئله محک (۱۴۰٪ ضربه) – خطوط با ضخامت متوسط و زیاد، به ترتیب مقدار مطلوب متغیر در سیکل حرکتی و محدودیت بالا و/یا پایین آن را نمایش می‌دهند.

- **شبیه سازی پایدارساز دوم سیکل حرکتی بر روی مدل کامل با شرایط اولیه پس‌رونده دور از سیکل**

نتایج به صورت مسیر حرکت مولفه واگرا به همگرا در صفحه فاز شـکل ۲۸-۳ نمـایش داده شـده اسـت. بـرای بخش پیوسته معادلات حرکت، نمودارهای شـکل ۲۹-۳، متغیرهـای سیسـتم از جملـه سـرعت و شـتاب حرکـت در راستای افقی( $dx/dt$ و $d^2x/dt^2$ ) ، موقعیت عمودی( $y$ ) و زاویه بالاتنه( $\theta$ ) را به همراه متغیرهـای کنترلـی نیروهای افقی و عمـودی و گشـتاور بالاتنـه( $F_x^{control}$ ، $F_x^{control}$ و $T_z^{control}$ )، نمـایش مـی‌دهنـد. همچنـین ضریب اصطکاک مورد نیاز( $\mu_{required}$ ) برای تامین نیروهای کنترلی این حرکت، به صورت نمـوداری بـر حسـب زمان مشخص شده‌است. برای بخش گسسته معادلات حرکت، نمودارهای شکل ۳۰-۳، متغیرهـای کنترلـی از جملـه طول و زمان فرود هرگام( $L_i$ و $T_i$ ) با کران‌های بـالا و پـایین هریـک( $L_{max}$ و $T_{min}$ ) را به همراه متغیرهـای سیستم از جمله شرایط اولیه مولفه‌های واگرا( $q_{i+1,0}$ ) و همگرا ( $p_{i+1,0}$ ) بـا مقـدار شـاخص آن( $p_i^*$ ) و نـیز سرعت متوسط حرکت نسبی پای متحرک نسبت به بالاتنه( $V_{i,swing}$ ) با کران بالای آن( $V_{max}$ ) و سرعت متوسـط مرکز جرم در طول هـر گـام( $V_{i,avg}$ )، نمـایش داده شـده اسـت. همچنـین متغیرهـای قیـد $D_{i,1}$ و $D_{i,2}$ بـه همـراه حداکثر مقدار مجازشان ( $L_{max}/2$ ) در این نمودارها مشخص شده است.

پایدارساز دوم با شروع از شرایط اولیه پس‌رونده دور از سیکل، طی هفت گام بـه پایـداری کامـل حـول سـیکل پیش‌رونده مطلوب می‌رسد ولی دو قید مسئله بر روی $D_{i,1}$، $D_{i,2}$، تا حد زیادی رعایت نشده است. پایدارسـاز دوم با وجود توانایی کنترل شرایط اولیه پس‌رونده دور از سیکل، قیود مسئله را به درستی رعایت نمـی‌کنـد و بنـابراین از لحاظ عملی این روش برای هدایت حرکت از یک شرایط اولیه پس‌رونده دور به سمت سیکل حرکتی پیش‌رونده، کارایی ندارد.

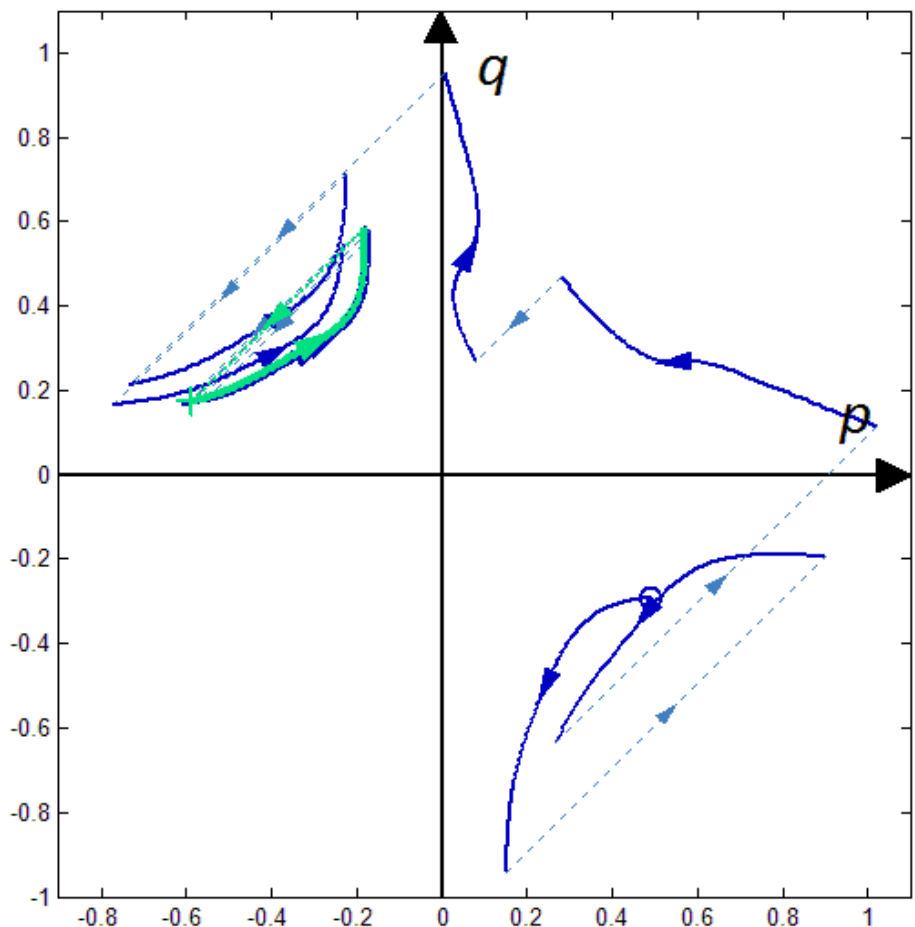

شکل ۳-۲۸. صفحه فاز مسیر حرکت مولفه واگرا نسبت به مولفه همگرا برای شبیه‌سازی پایدارساز دوم سیکل حرکتی بر روی مدل کامل راه‌رونده با شرایط اولیه دور از سیکل(پس‌رونده) — منحنی ضخیم و کمرنگ مسیر گام آخر را نمایش می‌دهد.

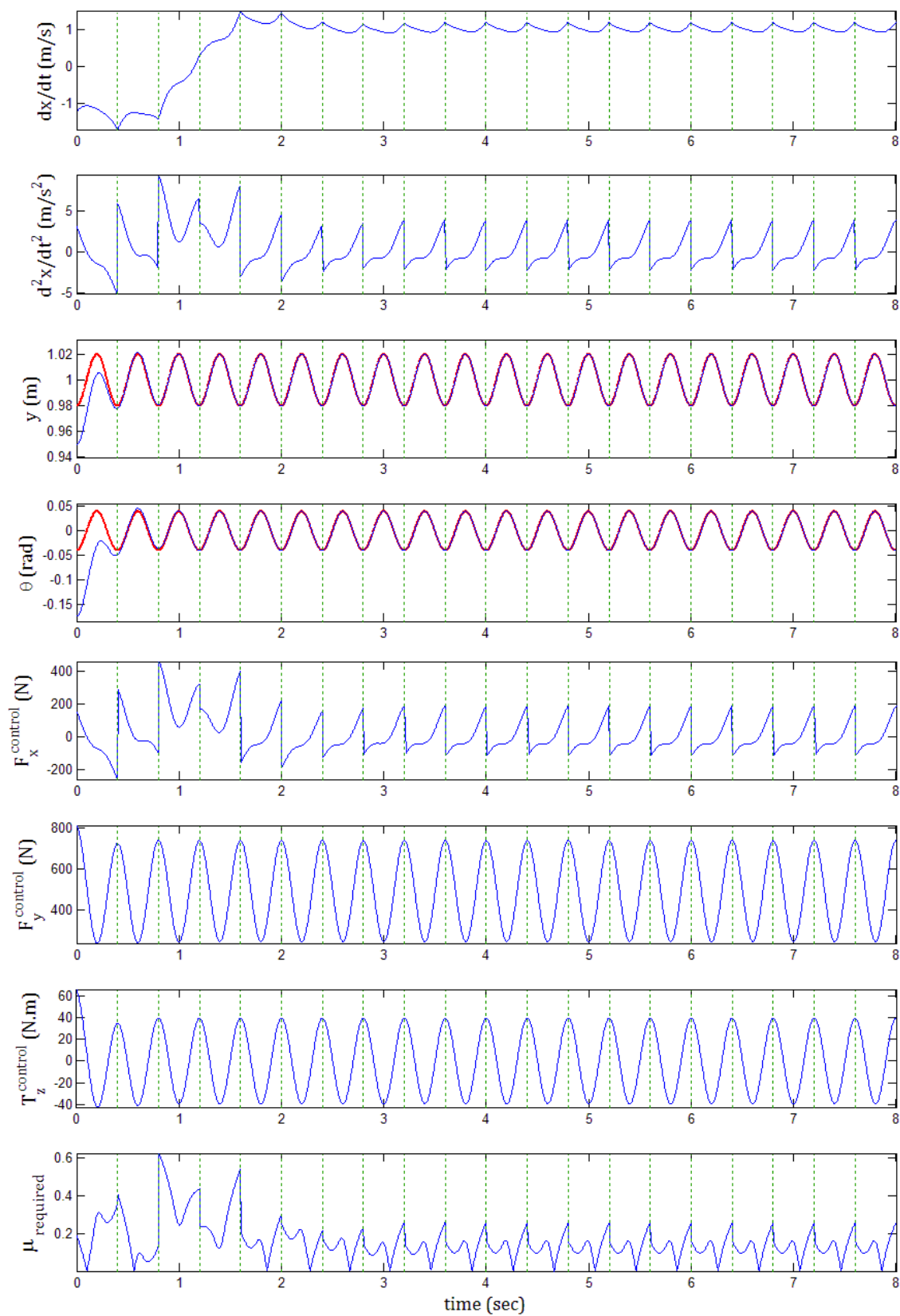

شکل ۳-۲۹. نمودارهای بخش پیوسته معادلات حرکت برای شبیه‌سازی پایدارساز دوم سیکل حرکتی بر روی مدل کامل راه‌رونده با
شرایط اولیه دور از سیکل(پس‌رونده) — خطوط ضخیم، مسیر مطلوب متغیر را نمایش می‌دهند.

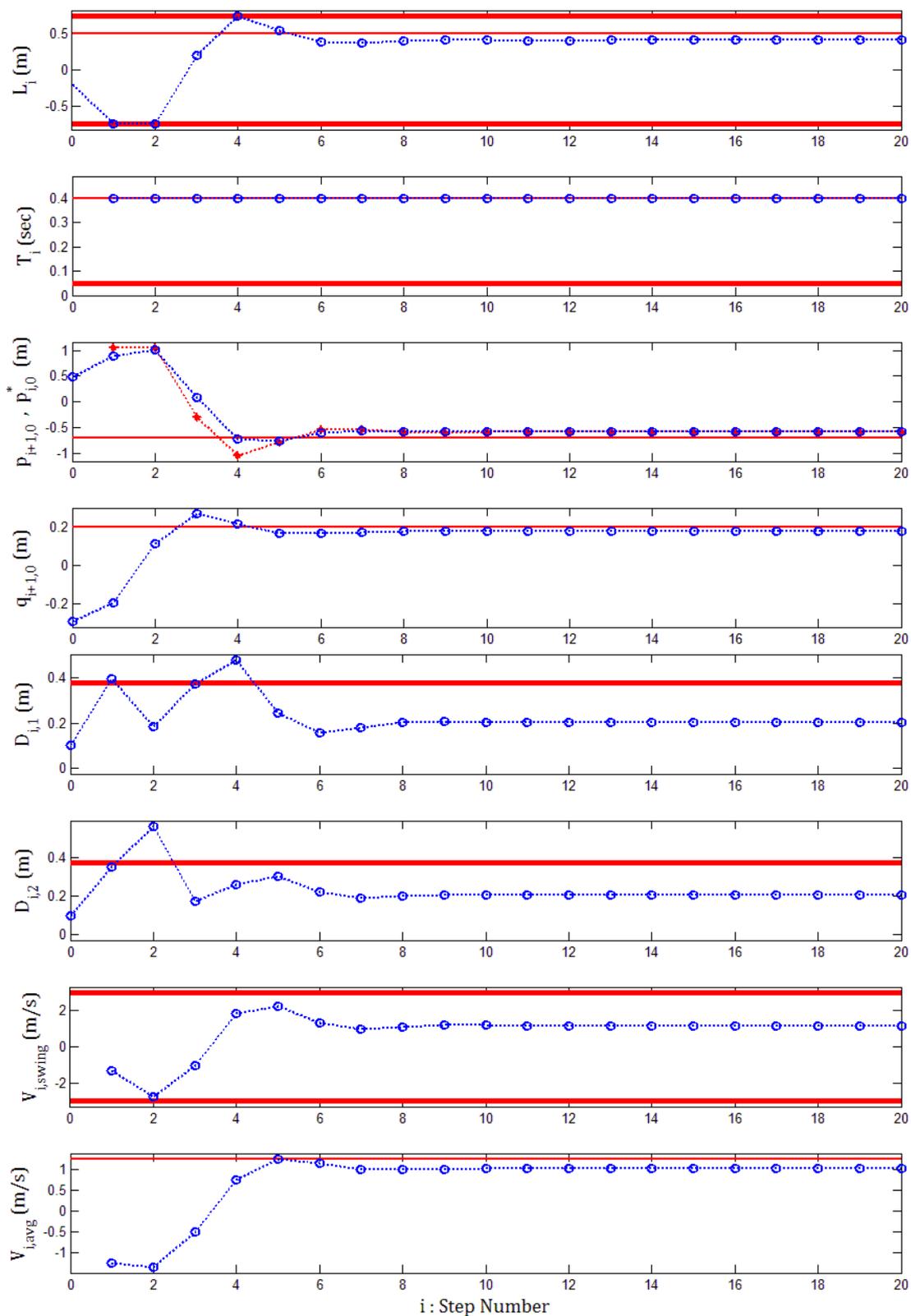

شکل ۳-۳۰. نمودارهای بخش گسسته معادلات حرکت برای شبیه‌سازی پایدارساز دوم سیکل حرکتی بر روی مدل کامل راهرونده با شرایط اولیه دور از سیکل(پس‌رونده) — خطوط با ضخامت متوسط و زیاد، به ترتیب مقدار مطلوب متغیر در سیکل حرکتی و محدودیت بالا و یا پایین آن را نمایش می‌دهند.

• **شبیه سازی پایدارساز چهارم سیکل حرکتی بر روی مدل کامل با شرایط اولیه پس‌رونده دور از سیکل**

نتایج به صورت مسیر حرکت مولفه حرکت واگرا به همگرا در صفحه فاز شکل ۳۱-۳ نمایش داده شده است. برای بخش پیوسته معادلات حرکت، نمودارهای شکل ۳۲-۳، متغیرهای سیستم از جمله سرعت و شتاب حرکت در راستای افقی( $dx/dt$ و $d^2x/dt^2$ )، موقعیت عمودی( $y$ ) و زاویه بالاتنه( $\theta$ ) را به همراه متغیرهای کنترلی نیروهای افقی و عمودی و گشتاور بالاتنه( $T_z^{control}$ و $F_x^{control}$، $F_x^{control}$ )، نمایش می‌دهند. همچنین ضریب اصطکاک مورد نیاز( $\mu_{required}$ ) برای تامین نیروهای کنترلی این حرکت، به صورت نموداری بر حسب زمان مشخص شده‌است. برای بخش گسسته معادلات حرکت، نمودارهای شکل ۳۳-۳، متغیرهای کنترلی از جمله طول و زمان فرود هرگام( $L_i$ و $T_i$ ) با کران‌های بالا و پایین هریک( $L_{max}$ و $T_{min}$ ) را به همراه متغیرهای سیستم از جمله شرایط اولیه مولفه‌های واگرا( $q_{i+1,0}$ ) و همگرا( $p_{i+1,0}$ ) با مقدار شاخص آن( $p_i^*$ ) و نیز سرعت متوسط حرکت نسبی پای متحرک نسبت به بالاتنه( $V_{i,swing}$ ) با کران بالای آن( $V_{max}$ ) و سرعت متوسط مرکز جرم در طول هر گام( $V_{i,avg}$ )، نمایش داده شده است. همچنین متغیرهای قید $D_{i,1}$ و $D_{i,2}$ به همراه حداکثر مقدار مجازشان ( $L_{max}/2$ ) در این نمودارها مشخص شده است.

پایدارساز چهارم با شروع از شرایط اولیه پس‌رونده دور از سیکل، طی شش گام به پایداری کامل حول سیکل پیش‌رونده مطلوب می‌رسد ولی دو قید مسئله بر روی $D_{i,1}$، $D_{i,2}$، تا حد زیادی رعایت نشده است. بنابراین پایدارساز چهارم نیز، با وجود توانایی کنترل شرایط اولیه پس‌رونده دور از سیکل، قیود مسئله را به درستی رعایت نمی‌کند و از لحاظ عملی این روش برای هدایت حرکت از یک شرایط اولیه پس‌رونده دور به سمت سیکل حرکتی پیش‌رونده، کارایی ندارد.

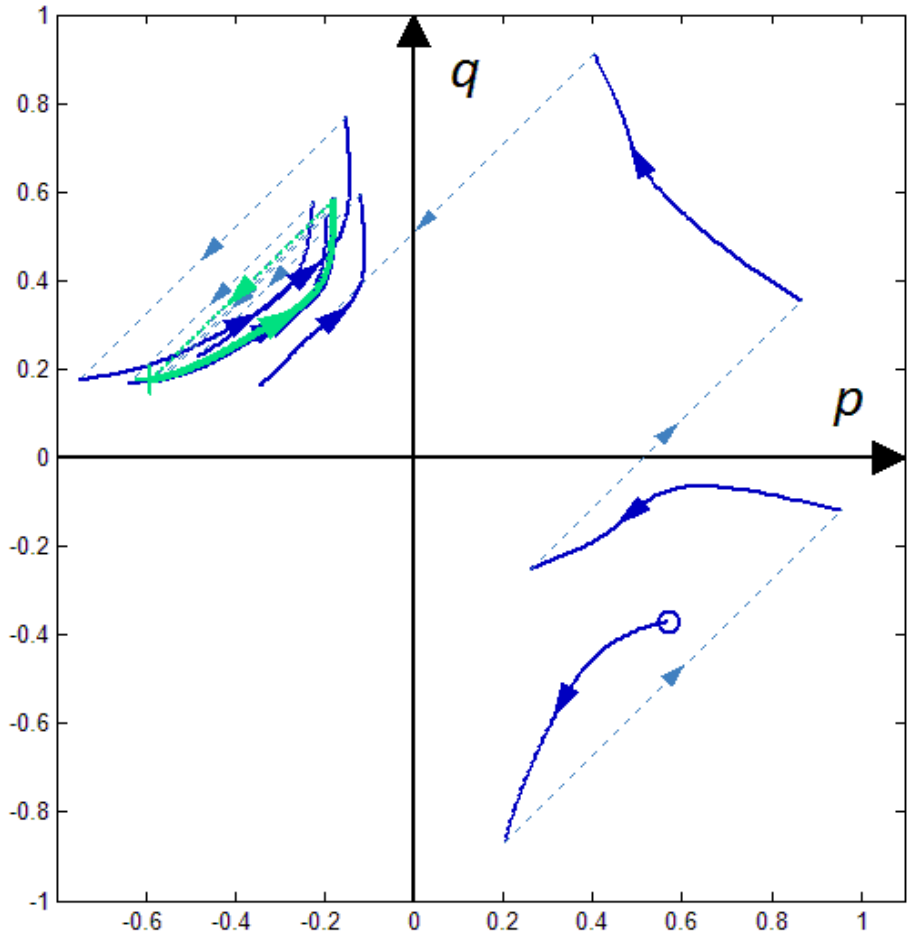

$p_c = -0.7$ , $q_c = 0.2$ , $T_c = 0.4$ , $L_c = 0.5$ , $V_c = 1.25$
$p_{1,0} = 0.57$ , $q_{1,0} = -0.37$ , $x_0 = 0.1$ , $dx/dt_0 = -1.473$
$z_0 = 0.2$ , $z_1 = 0$

شکل ۳-۳۱. صفحه فاز مسیر حرکت مولفه واگرا نسبت به مولفه همگرا برای شبیه‌سازی پایدارساز چهارم سیکل حرکتی بر روی مدل کامل راه‌رونده با شرایط اولیه دور از سیکل(پس‌رونده) − منحنی ضخیم و کم‌رنگ مسیر گام آخر را نمایش می‌دهد.

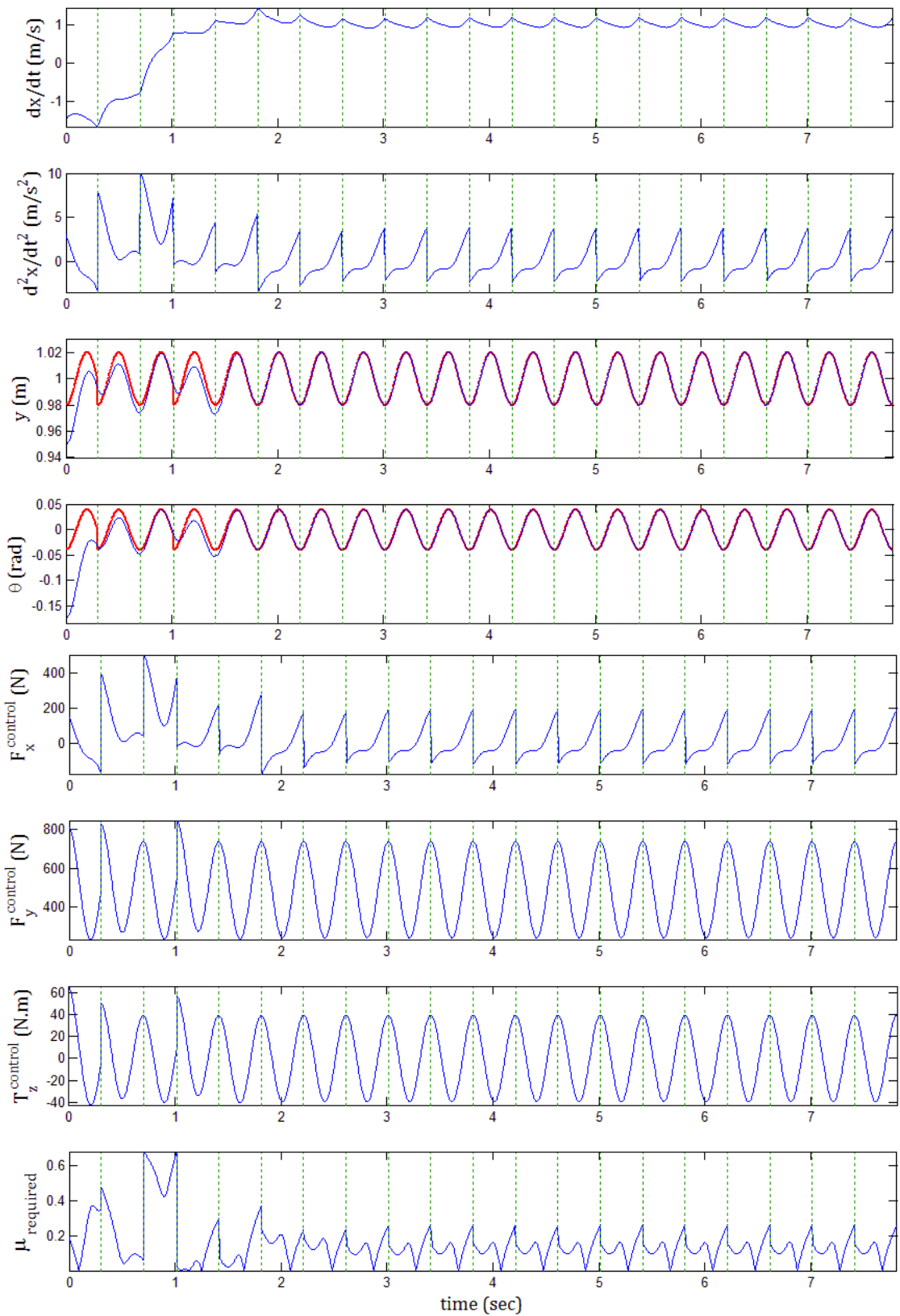

شکل ۳-۳۲. نمودارهای بخش پیوسته معادلات حرکت برای شبیه‌سازی پایدارساز سیکل حرکتی چهارم سیکل کامل راه‌رونده با شرایط اولیه دور از سیکل (پس‌رونده) − خطوط ضخیم، مسیر مطلوب متغیر را نمایش می‌دهند.

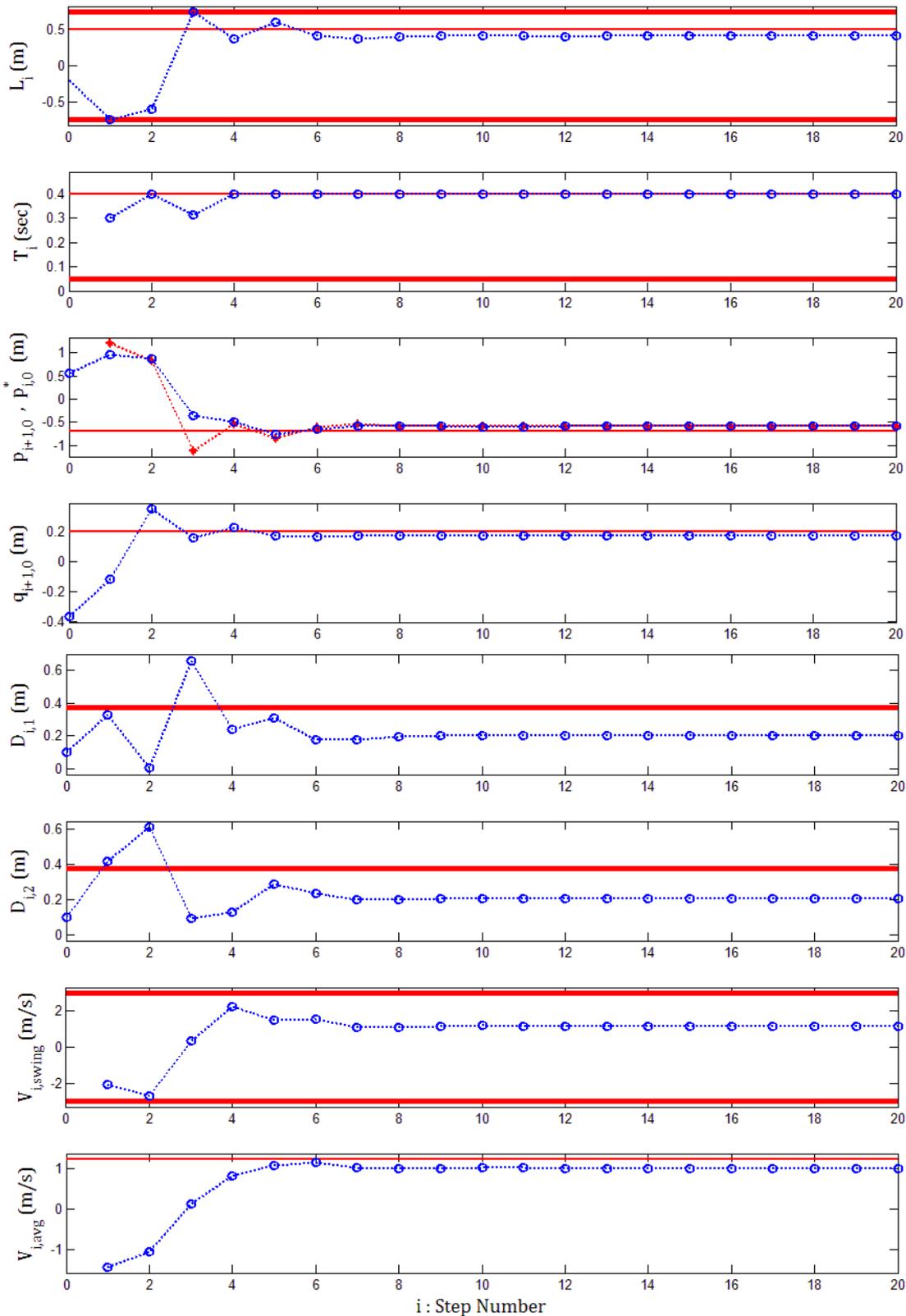

شکل ۳-۳۳. نمودارهای بخش گسسته معادلات حرکت برای شبیه‌سازی پایدارساز چهارم سیکل حرکتی بر روی مدل کامل راه‌رونده با شرایط اولیه دور از سیکل(پس‌رونده) − خطوط با ضخامت متوسط و زیاد، به ترتیب مقدار مطلوب متغیر در سیکل حرکتی و محدودیت بالا و یا پایین آن را نمایش می‌دهند.

**فصل چهارم**
**کنترل‌کننده پایداری بهینه**

در فصل قبل با معرفی پایدارسازهای سیکل حرکتی، روشی موثر برای هـدایت حرکـت بـه سـمت یـک سیکل حرکتی مطلوب یافتیم ولی با معرفی مدل فیزیکی دارای قیود واقعی و انجام شبیه‌سازی بر روی آن متوجه شـدیم کـه پایدارسازها الزامی به رعایت این قیود ندارند و با وجود اینکه برای پایدارکردن شرایط اولیـه نزدیـک بـه سیکل، بـه خوبی عمل می‌کنند، در عمل برای شرایط اولیه‌های مرزی دور از سیکل، نمی‌توانند بر روی راه‌رونده واقعی پیـاده‌سازی شوند. در این فصل سعی می‌کنیم به راه‌حلی دست یابیم که این مشکل را کاملا برطرف کند.

اگر بخواهیم به محدودیت‌های راه‌رونده واقعی بپردازیم، ناگزیر از جستجو در فضای جواب‌های همگـرا در ایـن محدوده خواهیم بود. در این فصل سعی خواهیم کرد با مبنا قراردادن دانش بهدسـت آمـده در بخـش پایدارسـازهای سیکل‌های حرکتی فصل گذشته، تابع شاخص پایداری را تعریف کنیم و سپس با کمینه‌کردن همزمان آن و رعایـت قیود حرکتی، جوابی بهینه برای هدایت مسیر حرکت به سمت یک سیکل حرکتی دلخـواه و کنتـرل پایـداری حـول آن سیکل بیابیم. روش برنامه‌ریزی خطی پایداری یا کنترل پایداری بهینه، مبتنی بر راهبرد دوم کنترل پایداری اسـت که در آن استفاده از برداشتن گام‌هایی با طول و زمان فرود متغیر برای پایدارسازی حرکت مورد توجه است.

## ۴-۱ مقدمه‌ای بر روش کنترل پایداری با استفاده از برنامه‌ریزی غیرخطی

اگر برای سادگی بیشتر، نمادهای شرایط اولیه ، طول و زمان فرود گام جدیدی(فعلی) را که درصدد برداشتن آن هستیم بدون اندیس گام فرض کنیم و به ترتیب به صورت $(p_0, q_0)$ ، $L$ و $T$ نمایش دهیم و برای شرایط اولیه گام آتی از اندیس $1+$ به صورت $(p_{+1,0}, q_{+1,0})$ استفاده کنیم، معادله حرکت گام به گام به صورت رابطه (۴-۱) قابل بازنویسی است. حال اگر بخواهیم شرایط اولیه گام آتی به مقادیر مطلوبی برسد، سیستم معادلات درجه دومی بر حسب دو کمیت طول و زمان فرود گام جدید، $L$ و $T$ ، به صورت رابطه (۴-۲) نتیجه‌گیری می‌شود که در صورت وجود جوابی حقیقی، یک یا دو دسته جواب به صورت رابطه (۴-۳) برای آن قابل تصور است.

$$\begin{cases} p_{+1,0} = p_0 e^{-\omega T} - L \\ q_{+1,0} = q_0 e^{\omega T} - L \end{cases} \tag{۴-۱}$$

$$\begin{cases} p_{+1,0} = p_{+1,0}^{des} \\ q_{+1,0} = q_{+1,0}^{des} \end{cases} \Rightarrow \begin{cases} q_0 (e^{\omega T})^2 - (q_{+1,0}^{des} - p_{+1,0}^{des}) e^{\omega T} - p_{f,0} = 0 \\ L^2 + (q_{+1,0}^{des} + p_{+1,0}^{des}) L + p_{+1,0}^{des} q_{+1,0}^{des} - p_0 q_0 = 0 \end{cases} \tag{۴-۲}$$

$$Answer(1) \begin{cases} e^{\omega T_{(1)}} = \dfrac{(q_{+1,0}^{des} - p_{+1,0}^{des}) - \sqrt{\Delta}}{2q_0} \\ L_{(1)} = \dfrac{-(q_{+1,0}^{des} + p_{+1,0}^{des}) - \sqrt{\Delta}}{2} \end{cases} , Answer(2) \begin{cases} e^{\omega T_{(2)}} = \dfrac{(q_{+1,0}^{des} - p_{+1,0}^{des}) + \sqrt{\Delta}}{2q_0} \\ L_{(2)} = \dfrac{-(q_{+1,0}^{des} + p_{+1,0}^{des}) + \sqrt{\Delta}}{2} \end{cases} \tag{۴-۳}$$

$$for\ both\ aswers\ (1), (2): \quad \Delta = (q_{+1,0}^{des} - p_{+1,0}^{des})^2 + 4 p_0 q_0$$

بهترین انتخاب برای شرایط اولیه مطلوب گام بعدی، شرایط اولیه سیکل حرکتی مطلوبی است که تصمیم داریم حرکت را به سمت آن سوق داده و پایداری را حول آن برقرار کنیم. با این‌حال در صورت نزدیک نبودن شرایط اولیه گام فعلی(جدید) حرکت نسبت به شرایط اولیه سیکل حرکتی، یافتن جواب برای این انتخاب مطلوب، به هیچ وجه قابل تضمین نیست(رابطه (۴-۴))، علاوه بر آن، در صورت وجود احتمالی جواب، جواب‌ها باید قیود مسئله ازجمله کران بالای طول گام و سرعت را نیز ارضاء کنند که احتمال یافتن جوابی مطلوب را کمتر می‌کند.

$$\begin{cases} p_{+1,0}^{des} = p_c \\ q_{+1,0}^{des} = q_c \end{cases} \overset{?}{\Rightarrow} (T, L) \tag{۴-۴}$$

انتخاب دیگری برای شرایط اولیه مطلوب گام بعدی، شرایط اولیه‌ای است که به سمت شرایط اولیه سیکل حرکتی مطلوب همگراست. رابطه (۴-۵)، یک معادله مشخصه تفاضلی درجه اول برای حرکت بر روی مسیری همگرا به سمت شرایط اولیه سیکل حرکتی مطلوب معرفی می‌کند. شرط همگرایی این رابطه قرار گرفتن $K_{1,i}$ و $K_{2,i}$ در بازه $[0 , 2]$ است. می‌توان به صورت عددی نشان داد که همواره برای محدوده مشخصی از شرایط اولیه، می‌توان دنباله‌ای از $\{K_{1,0}, K_{1,2}, ... , K_{1,i}\}$ و $\{K_{2,0}, K_{2,2}, ... , K_{2,i}\}$ در بازه $[0 , 2]$ را یافت که منجر به دنباله‌ای از $\{p_{1,0}^{des}, p_{2,0}^{des}, ... , p_{n,0}^{des} = p_c\}$ و $\{q_{1,0}^{des}, q_{2,0}^{des}, ... , q_{n,0}^{des} = q_c\}$ می‌شوند به نحوی که بازای هر جفت $(p_{i,0}^{des}, q_{i,0}^{des})$ در هر گام، سیستم معادلات(۴-۳) همواره دارای جواب است و قیود مسئله نیز رعایت می‌شوند.

$$\begin{cases} p_{i+1,0}^{des} = p_{i,0}^{des} + K_{1,i}(p_c - p_{i,0}^{des}) \\ q_{i+1,0}^{des} = q_{i,0}^{des} + K_{2,i}(q_c - q_{i,0}^{des}) \end{cases} , \qquad p_{1,0}^{des} = p_{1,0} \ , \ q_{1,0}^{des} = q_{1,0} \qquad (۴-۵)$$

$$q_{i+1,0}^{des} - q_c = (1 - K_{2,i})(q_{i,0}^{des} - q_c) \ \Rightarrow \ \frac{q_{i+1,0}^{des} - q_c}{q_{i,0}^{des} - q_c} = (1 - K_{2,i})$$

$$q_{i+1,0}^{des} - q_c = (q_{1,0}^{des} - q_c) \prod_{j=1}^{i} \frac{q_{i+1,0}^{des} - q_c}{q_{i,0}^{des} - q_c} = (q_{1,0}^{des} - q_c) \prod_{j=1}^{i} (1 - K_{2,j}) \qquad (۴-۶)$$

$$0 < K_{2,j} < 2 \ \Rightarrow \ -1 < 1 - K_{2,j} < 1 \ \Rightarrow \ |1 - K_{2,j}| < 1$$

$$\lim_{i \to \infty} |q_{i,0}^{des} - q_c| = |q_{1,0}^{des} - q_c| \lim_{i \to \infty} \prod_{j=1}^{i-1} |1 - K_{2,j}| = 0 \ \Rightarrow \ \lim_{i \to \infty} q_{i,0}^{des} = q_c \qquad (۴-۷)$$

این روش، راهکاری کلی برای برنامه‌ریزی غیرخطی سوق دادن حرکت از هر شرایط اولیه دلخواه به سمت شرایط اولیه سیکل حرکتی مطلوب را نشان می‌دهد. اگرچه این فرمول‌بندی برای یافتن جوابی مناسب برای مسئله، قابل کاربرد است ولی دارای کاستی‌هایی است که نمی‌توان از آن برای حل مسئله در حالت کلی استفاده نمود. ازجمله اینکه، روشی غیرمستقیم محسوب می‌شود که به جای محاسبه مستقیم متغیرهای مسئله، $L$ و $T$ ، ما را درگیر بررسی وجود جواب و حل معادله درجه دوم (۴-۳) می‌کند که برای استفاده از روش‌های بهینه‌سازی عددی نامناسب است. همچنین الزام پیروی از معادله مشخصه پیشنهادی (۴-۵)، محدودیتی بر مسئله وارد می‌کند که می‌تواند ناحیه کنترل‌پذیر شرایط اولیه را محدود کند. به همین دلیل در بخش بعدی سعی می‌کنیم تابع شاخصی برای حل بهینه مسئله در یک گام بسازیم که در صورت کمینه شدن، هم پایداری را بهینه کند و همزمان همه قیود مسئله را ارضا کند، سپس، با استفاده از بهینه‌سازی غیرخطی به حل این مسئله کنترل‌کننده پایداری بهینه و شبیه‌سازی آن بر روی مدل فیزیکی کامل راه‌رونده می‌پردازیم.

## ۴-۲  طرح مسئله کنترل پایداری با استفاده از روش بهینه‌سازی بر روی طول و زمان فرود گام

همانطور که در بخش پایدارسازهای سیکل حرکتی در فصل گذشته دیدیم، شاخص اصلی برای حرکت به سمت یک سیکل حرکتی، همگرا کردن شرایط اولیه مولفه واگرا، $q_{i,0}$، به سمت شرایط اولیه مولفه واگرای سیکل حرکتی مطلوب، $q_c$، است. با توجه به ویژگی اول در قضیه محدود ماندن شرایط اولیه مولفه همگرا، $p_{i,0}$، که در ابتدای فصل بررسی شد، این مولفه با برداشتن گام‌های دلخواه متوالی، همواره شرایط اولیه شاخص مولفه همگرا، $p_i^*$، را تعقیب می‌کند و نیز با توجه به ویژگی دوم این مولفه، $p_{i,0}$، همواره محدود می‌ماند. بنابراین برای سوق‌دادن حرکت راه‌رونده از هر شرایط اولیه دلخواه به سمت یک سیکل حرکتی مطلوب بدون توجه به شرایط اولیه مولفه همگرا، تنها کافی است در هر گام تابع شاخص فاصله بین شرایط اولیه مولفه واگرا در گام بعدی، $q_{+1,0}$، و شرایط اولیه مولفه واگرای سیکل حرکتی مطلوب، $q_c$، را با توجه به قیود مسئله کمینه‌سازی کنیم. اگر بخواهیم مسئله سوق‌دادن حرکت راه‌رونده از هر شرایط اولیه دلخواه به سمت یک سیکل حرکتی مطلوب را با توجه به قید-های موقعیتی و سرعتی به صورت یک مسئله بهینه‌سازی فرمول‌بندی کنیم، رابطه (۴-۸)، صورت اولیه آن خواهد بود که به بهینه‌کردن شاخص پایداری و رعایت قیود را برای یک گام درنظر می‌گیرد.

$$\min_{L,\,T} \; |q_{+1,0} - q_c| \;, \quad Subject \; to:$$

$$|L| \;<\; L_{max}$$

$$|D_1| \;<\; \frac{L_{max}}{2}$$

$$|D_2| \;<\; \frac{L_{max}}{2} \qquad\qquad (۴-۸)$$

$$|\bar{V}_{swing}| \;<\; V_{max}$$

اگرچه این فرمول‌بندی به همگرایی به معنی همگرایی $q_{i,0}$ به سمت $q_c$ و محدود ماندن $p_{i,0}$ منجر خواهد شد ولی هیچ تضمینی برای همگرایی $p_{i,0}$ به سمت $p_c$ و جلوگیری از نوسان کردن نخواهد داد. اگر بار دیگر به ویژگی اول شرایط اولیه مولفه همگرا توجه کنیم که همواره مقدار شاخص خود، $p_i^*$، را تعقیب می‌کند و با توجه به اینکه شرایط اولیه مولفه همگرا $p_i^*$ تابعی از طول و زمان فرود گام است، کمینه‌کردن فاصله طول و زمان فرود هر گام، $L$ و $T$، با طول و زمان فرود گام سیکل حرکتی مطلوب، $L_c$ و $T_c$، هنگامی که مسیر حرکت به اندازه کافی نزدیک به مسیر حرکتی سیکل مطلوب شده است، می‌تواند به پایدار شدن شرایط اولیه مولفه همگرا، $p_i^*$، و به دنبال آن شرایط اولیه مولفه همگرا، $p_{i,0}$، و درنهایت کل حرکت بیانجامد و از نوسان طول و زمان گام‌ها جلوگیری کند. با این حساب اگر بخواهیم فرمول‌بندی مسئله بهینه‌سازی برای سوق‌دادن حرکت راه-رونده از هر شرایط اولیه دلخواه به سمت یک سیکل حرکتی مطلوب را با توجه به این دو شاخص جدید اصلاح کنیم، رابطه آن به صورت (۴-۹) خواهد بود که ضرایب این دو شاخص جدید از رابطه (۴-۱۰) پیروی می‌کنند.

همچنین برداشتن قدم‌هایی کوتاه‌تر از نصف طول کف پا برای راه‌رونده نامطلوب است، زیرا تغییر مرکز فشار برای فواصلی کوتاه‌تر از این مقدار، می‌تواند بدون برداشتن گام در مدت زمانی نزدیک به صفر در کف پای تکیه‌گاهی فعلی صورت بپذیرد. قید دوم رابطه (۴-۱۰)، بیانگر این خواسته است که درنظرگرفتن یا نگرفتن آن اختیاری خواهد بود.

$$\min_{L\,,\,T} r_1|q_{+1,0} - q_c| + r_2|L - L_c| + r_3|T - T_c| \;,\; Subject\ to:$$

$$|L| \;<\; L_{max}$$

$$|L| \;>\; L_{min}\ (Optional)$$

$$|D_1| \;<\; \frac{L_{max}}{2}$$

$$|D_2| \;<\; \frac{L_{max}}{2}$$

$$|\bar{V}_{swing}| \;<\; V_{max}$$

$$(۴-۹)$$

$$Step\ Stabilizer \atop Coefficients : \begin{cases} r_2 = r_3 > 0 & if\ |q_0 - q_c| < \epsilon\ and\ |p_0 - p_c| < E \\ r_2 = r_3 = 0 & if\ |q_0 - q_c| > \epsilon\ or\ |p_0 - p_c| > E \end{cases}$$

$$(۴-۱۰)$$

اگرچه روش‌های مختلفی برای حل عددی این مسئله بهینه‌سازی مقید[1] وجود دارد ولی در این تحقیق به جای استفاده از روش‌های مستقیم مسئله، راهکاری جایگزین به نام روش تاوان[2] را استفاده خواهیم کرد که با گنجاندن قیود مسئله در یک تابع شاخص ترکیبی، مسئله را با روش‌های بهینه‌سازی غیرمقید[3] حل می‌کند. مزیت این راهکار در مقایسه با استفاده از روش‌های مقید، سادگی در پیاده‌سازی، پایین‌آمدن هزینه محاسباتی و از همه مهم‌تر تضمین سرعت همگرایی بالا در برخی از روش‌های بهینه‌سازی غیرمقید است که در این تحقیق از آن استفاده خواهد شد. این راهکار به علت سرعت محاسباتی بالا، برای استفاده بر روی راه‌رونده واقعی نیز قابل کاربرد می‌باشد.

---

[1] Constrained Optimization
[2] Penalty Method
[3] Unconstrained Optimization

### ۴-۳  تعریف تابع شاخص پایداری ترکیبی  برای استفاده از روش‌های بهینه‌سازی غیرمقید

اگر بخواهیم یک مسئله بهینه‌سازی مقید با قیود نامساوی را به مسئله‌ای غیرمقید تبدیل کنیم(رابطه (۴-۱۱))، یکی از روش‌های کاربردی، شیوه تاوان[1] است[۳۶]. در این روابط، $g_i$ ها، توابع شاخصی هستند که هدف اصلی مسئله بهینه سازی، جستجو در فضای برداری $X$ برای نزدیک ساختن مقدار این توابع شاخص به مقدار بهینه خودشان، $g_{opt}$ ها، است.

همچنین در این روابط، $h_j$ ها، توابع قیدی هستند که مقدار این توابع با توجه به جهت‌شان، $dir_j$ ها، نباید از مقدار مجاز آنها، $h_{j,ext}$ ها، فراتر برود. $h_j$ ها، درواقع، محدوده فضای برداری $X$ را برای انجام عملیات جستجو و یافتن جواب مسئله، $X^*$، تعیین می‌کنند.

$$\min_{X} \sum_{i=1}^{n} \left| g_i(X) - g_{opt} \right|, \quad i = 1,2,..,n,$$
$$Subject\ to:\quad dir_j * h_j(X) > h_{j,ext}, \quad j = 1,2,..,m, \quad dir_j = 1 \ or\ -1$$

(۴-۱۱)

$$\min_{X} U(X), \quad U(X) = U\left(g_1(X), g_2(X), ..., g_n(X), \ h_1(X), h_2(X), ..., h_m(X)\right)$$

در مسئله حاضر، $X = \begin{bmatrix} T & L \end{bmatrix}^{T}$ می‌باشد. برای رسیدن به تابع شاخصی که کمینه‌سازی آن علاوه بر اینکه منجر به بهینه‌سازی شاخص‌ها می‌شود به رعایت قیود نامساوی مسئله نیز بیانجامد، آن را به صورت ترکیبی از دو نوع توابع هدف[2] و تاوان[3] به صورت رابطه (۴-۱۲) تعریف می‌کنیم که $C$ در آن ضریب اثر تاوان است.

وجود متغیر زمان در این تابع، برای درنظر داشتن جامعیت مسئله ذکر شده است، زیرا پارامترهای توابع هدف و توابع قید می‌توانند تابعیتی از زمان نیز داشته باشند(به عنوان مثال استفاده از شرایط اولیه تخمینی به جای شرایط اولیه گام فعلی، تابعیتی از زمان را وارد مسئله می‌کند.). با این وجود اما، به علت انجام‌شدن کل عملیات بهینه‌سازی در یک لحظه کوتاه از زمان، بهینه‌سازی بدون توجه به متغیر زمان انجام می‌شود و در روابط مربوط به بهینه‌سازی تأثیری ندارد.

$$U(T,L,t) = \sum_{i=1}^{n} R_i \, GoalFunction_i\,(T,L,t) + C \sum_{j=1}^{m} PenaltyFunction_j\,(T,L,t) \quad \text{(۴-۱۲)}$$

---

شاخص‌های بهینه سازی را به صورت توابع هدف مرکب رابطه (۴–۱۳) تعریف می‌کنیم که تابعی درجه دوم بر روی فاصله تابع شاخص تا مقدار بهینه خود است. مقدار این تابع درجه دوم هنگامی که تابع شاخص بر روی مقدار بهینه خود است، صفر می‌باشد و با دور شدن تابع شاخص از مقدار بهینه خود، به صورت درجه دوم زیاد می‌شـود تـا در فاصله‌ای از مقدار بهینه به نام حاشیه بیشینه، $\Delta g_{i,max}$ ، برابر واحد گردد و با فرا رفتن از این حاشیه مقـدار آن از واحد تجاوز می‌کند. شکل ۴-۱ ، بیانگر تعریف این تابع هدف مرکب است.

$$GoalFunction_i\,(T,L,t) \quad = GF\big(g_i(T,L,t)\,,\,g_{i,opt}\,,\,\Delta g_{i,max}\big) = GF(T,L,t,i)$$

$$GF\big(g_i(T,L,t)\,,\,g_{i,opt}\,,\,\Delta g_{i.max}\big) = \frac{1}{2}\,\frac{\big(g_i(T,L,t) - g_{i,opt}\big)^2}{\Delta g_{i,max}{}^2}$$

$$(۴–۱۳)$$

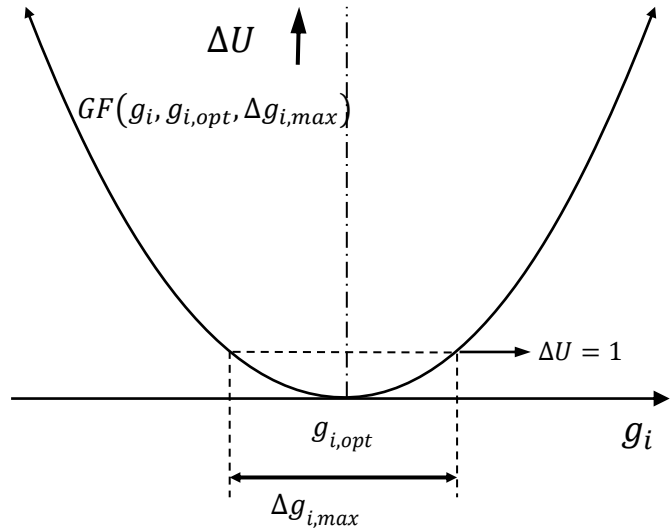

شکل ۴-۱. تعریف توابع هدف به صورتی تابعی مربعی بر روی فاصله هر تابع شاخص از مقدار بهینه خود

قیود مسئله را نیز به صورت توابع تاوان مرکب رابطه (۴–۱۴) تعریف می‌کنیم که تابعی درجه دوم و یک‌طرفه بر روی میزان فرارفتن تابع قید از مقدار مجاز حـداکثر یـا حـداقل خـود اسـت. مقـدار ایـن تـابع درجه دوم، در فاصله کوچکی قبل از رسیدن تابع مجاز خود به مقدار مجاز خود به نام حاشیه کمینه، $\Delta h_{j,min}$ ، صفر است و هنگامی که تابع قیـد به روی مقدار مجاز خود می‌رسد، برابر واحد می‌گردد و با فرارفتن تابع قید از این مقدار مجاز، با نرخی مربعی زیـاد می‌شود. شکل ۴-۲ بیانگر تعریف این تابع تاوان مرکب است.

$$PenaltyFunction_j(T,L,t) = PF\big(h_j(T,L,t),h_{j,ext},\Delta h_{j,min},dir_j\big) = PF(T,L,t,j)$$

$$PF\big(h_j(T,L,t),h_{j,ext}\,,\Delta h_{j,min}\,,dir_j\big) = \frac{1}{2}\max\left(0\,,\,1 + dir_j\,\frac{h_j(T,L,t) - h_{j,ext}}{\Delta h_{j,min}}\right)^2$$

$$(۴–۱۴)$$

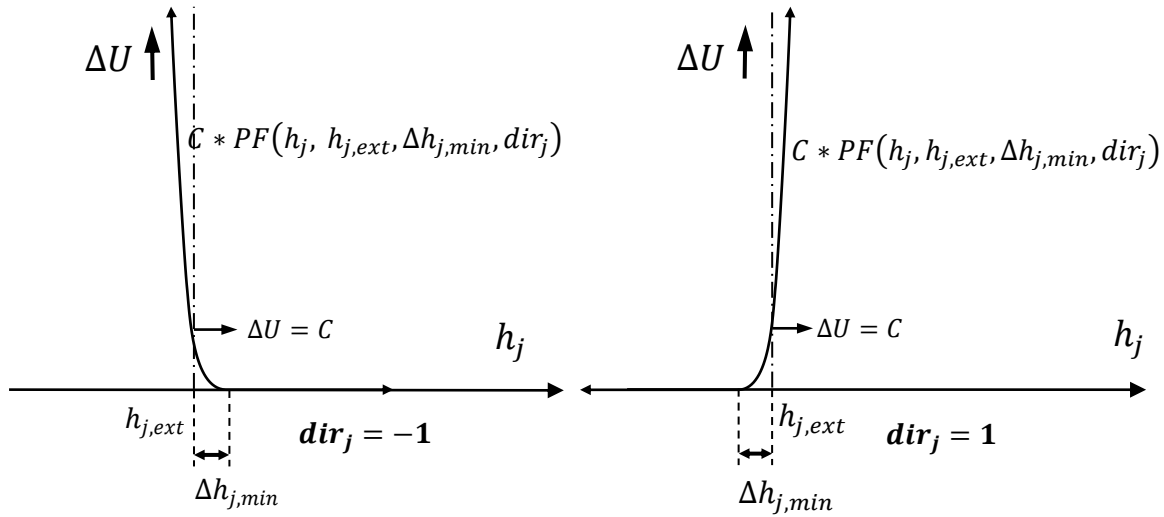

شکل ۴-۲. تعریف توابع تاوان به صورتی تابعی مربعی و یکطرفه بر روی فرارفتن هر تابع قید از مقدار مجاز کمینه یا بیشینه خود

برای استفاده از روش‌های بهینه‌سازی عددی در مسئله حاضر، نیاز به مشتق‌های اول و دوم تابع شاخص ترکیبی خواهیم داشت که با استفاده از رابطه مشتق‌گیری زنجیره‌ای، از روابط (۴-۱۵)، (۴-۱۶)، (۴-۱۷) و (۴-۱۸) محاسبه می‌شوند که $x$ و $y$ در این روابط، هر یک می‌تواند با $L$ و یا $T$ جایگزین شود.

$$\frac{\partial U}{\partial x} = \frac{\partial}{\partial X}\left(\sum_{i=1}^{n} R_i \, GoalFunction_i(T,L,t) + C\sum_{j=1}^{m} PenaltyFunction_j(T,L,t)\right)$$

$$\frac{\partial U}{\partial x} = \sum_{i=1}^{n} R_i \, GF_i{'}(T,L,t,i)\frac{\partial g_i}{\partial x} + C\sum_{j=1}^{m} PF_j{'}(T,L,t,j)\frac{\partial h_j}{\partial x}$$

$$(۱۵-۴)$$

$$\frac{\partial^2 U}{\partial x \partial y} = \sum_{i=1}^{n} GF_i{'}(T,L,t,i)\frac{\partial^2 g_i}{\partial x \partial y} + GF_i{''}(T,L,t,i)\left(\frac{\partial g_i}{\partial x}\right)\left(\frac{\partial g_i}{\partial y}\right)$$

$$+ C\sum_{j=1}^{m} PF_j{'}(T,L,t,j)\frac{\partial^2 h_j}{\partial x \partial y} + PF_j{''}(T,L,t,j)\left(\frac{\partial h_j}{\partial x}\right)\left(\frac{\partial h_j}{\partial y}\right)$$

$$(۱۶-۴)$$

$$GF'\big(g_i(T,L,t)\,,\,g_{i,opt}\,,\,\Delta g_{i.max}\big) = \frac{\big(g_i(T,L,t) - g_{i,opt}\big)}{\Delta g_{i,max}{}^2}$$

$$GF''\big(g_i(T,L,t)\,,\,g_{i,opt}\,,\,\Delta g_{i.max}\big) = \frac{1}{\Delta g_{i,max}{}^2}$$

$$(۱۷-۴)$$

$$PF'\big(h_j(T,L,t)\,,\,h_{j,ext}\,,\,\Delta h_{j,min}\,,\,dir_j\big)$$
$$= \frac{dir_j}{\Delta h_{j,min}} \max\left(0\,,\,1 + dir_j\,\frac{h_i(T,L,t) - h_{j,ext}}{\Delta h_{j,min}}\right)$$

$$PF''\big(h_j(T,L,t)\,,\,h_{j,ext}\,,\,\Delta h_{j,min}\,,\,dir_j\big)$$
$$= \frac{1}{2}\left(\frac{dir_j}{\Delta h_{j,min}}\right)^2\left(1 + \text{sign}\left(1 + dir_j\,\frac{h_j(T,L,t) - h_{j,ext}}{\Delta h_{j,min}}\right)\right)$$

(۴–۱۸)

با به‌دست‌آمدن مولفه‌های مشتق اول و دوم تابع، بردار گرادیان و ماتریس هسیان برای مسئله ما به صورت روابط (۴–۱۹) محاسبه می‌شوند که در ادامه از آن استفاده خواهیم کرد.

$$\nabla U(T,L) = \begin{bmatrix} \dfrac{\partial U}{\partial T} & \dfrac{\partial U}{\partial L} \end{bmatrix}^T \quad,\quad \nabla^2 U(T,L) = \begin{bmatrix} \dfrac{\partial^2 U}{\partial T^2} & \dfrac{\partial U}{\partial T \partial L} \\[2ex] \dfrac{\partial^2 U}{\partial T \partial L} & \dfrac{\partial^2 U}{\partial L^2} \end{bmatrix}$$

(۴–۱۹)

## ۴-۴ روش‌های عددی بهینه‌سازی غیرخطی غیر مقید

فرض کنیم تابع حقیقی اسکالر $U$ بر فضای حقیقی برداری $X$ عمل می‌کند. موضوع اصلی بهینه‌سازی غیرمقید، یافتن $X = X^*$ خواهد بود که به ازای آن $U(X^*)$ دارای کمترین مقـدار در بـرد خـود باشـد. رابطـه (۴–۲۰) یـک مسئله کلی غیرمقید را نمایش می‌دهد. روش‌های عددی سعی می‌کنند از طریق یک الگوریتم تکرارشونده و با انجام تعداد تکرار متناهی از هر نقطه دلخواه به این نقطه بهینه برسند. هزینه محاسـباتی و سـرعت همگرایـی دو معیـار مهـم برای سنجش عملکرد این روش‌هاست. در ادامه دو روش اصلی در این مبحث را شرح می‌دهیم و از روش نیوتـون برای حل مسئله با توجه به مزایای آن استفاده خواهیم کرد. در این قسمت به بررسی روش‌های عـددی بهینه‌سـازی غیرخطی غیرمقید بر اساس مرجع [۳۷] خواهیم پرداخت.

$$\underset{X}{\text{Min}}\,U(X)\,,\quad U(X) \in \mathbf{R}\,,\quad X \in \mathbf{R}^n$$

(۴–۲۰)

## ۴-۴- ۱    روش بیشترین شیب نزولی  یا  کاهش گرادیانی[1]

این روش، در واقع بهینه‌سازی مرتبه اول است که بر اساس تخمین درجه اول بسط تیلور تابع بـه دسـت مـی‌آیـد. اگر مقدار تابع را در همسایگی نقطه‌ای از دامنه تابع که اکنون در آن قرارگرفته‌ایم را با بسط تیلـور درجـه اول برابـر بدانیم، به ازای حرکتی در فضای دامنه با گامی کوچک و در خلاف جهت بردار گرادیان تابع در این نقطه، مـی‌تـوان انتظار داشت که مقدار تابع بر اساس رابطه (۴-۲۱) کاهش پیدا کند. بدیهی است، صحت این روش، بـه کوتـاه بـودن گام‌های جستجو بستگی دارد زیرا بسط درجه اول تیلور تخمین مناسبی بـرای گـام‌هـای کوتـاه در توابـع غیرخطـی است. اگرچه استفاده از روش جستجوی خطی که در ادامه شرح داده می‌شود، عملکرد روش بیشترین شـیب نزولـی را تا حدی بهبود می‌بخشد ولی مشخصات کلی این روش مزیت در هزینه محاسباتی پایین و عیب در سرعت همگرایـی پایین است که میزان سرعت همگرایی آن قابل محاسبه یا اثبات نیست و به همین جهت برای مقاصد کنترلی کـه نیـاز به سرعت محاسباتی بالا دارد، مناسب نیست. شکل ۴-۳، اجرای ایـن الگـوریتم بـر روی یـک تـابع مربعـی نمونـه در فضای دوبعدی را نشان می‌دهد.

$$U(X + \Delta X) = U(X) + \nabla U(X)^T \, \Delta X$$

$$\Delta X = -\alpha \, \nabla U(X) \Rightarrow U(X + \Delta X) = U(X) - \alpha \|\nabla U(X)\|^2 < U(X) \; , \; 0 < \alpha \ll 1 \quad \text{(۴-۲۱)}$$

$$X_{k+1} = X_k - \alpha_k \, \nabla U_k$$

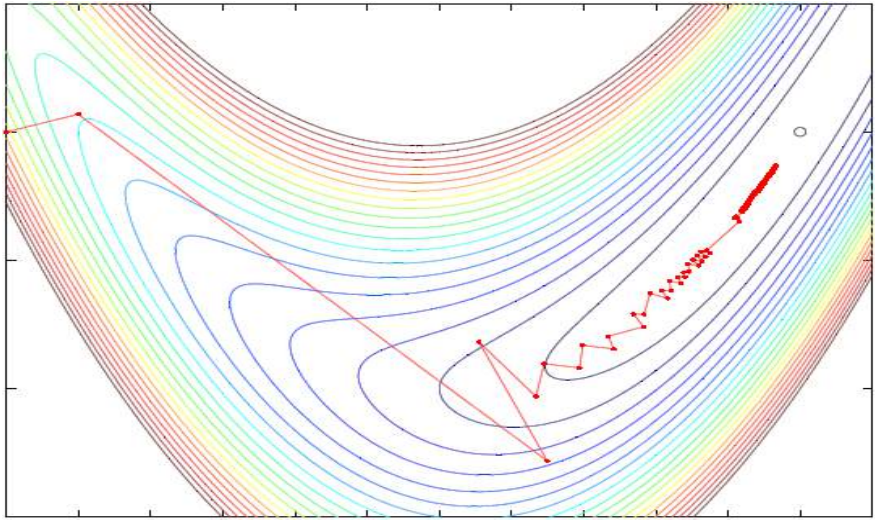

شکل ۴-۳. نمایش گرافیکی الگوریتم بهینه‌سازی غیرمقید بیشترین شیب نزولی بر روی یک تابع مربعی نمونه در فضای دوبعدی

---

[1] Steepest Descent or Gradient Descent

### ۴-۴-۲ روش نیوتن[1]

این روش، در واقع بهینه‌سازی مرتبه دوم است که بر اساس تخمین درجه دوم بسط تیلور تابع به دست می‌آید. اگر مقدار تابع را در همسایگی نقطه‌ای از دامنه تابع که اکنون در آن قرارگرفته‌ایم را با بسط تیلور درجه دوم برابر بدانیم، تغییرات بردار گرادیان به صورت رابطه (۴-۲۲) خواهد بود. صفر شدن بردار گرادیان به معنی، قرارداشتن تابع در نقطه حداقل (و یا حداکثر) خود خواهد بود، بنابراین با صفر قرار دادن بردار گرادیان تخمینی در همسایگی نقطه‌ای که هم‌اکنون در آن قرار گرفته‌ایم، گامی دارای جهت و طول مشخص به دست می‌آید که با برداشتن آن گام در نقطه کنونی (رابطه (۴-۲۳) )، می‌توان انتظار داشت که مقدار گرادیان تابع بر اساس الگوریتم تکرارشونده رابطه (۴-۲۲) تقریبا برابر با صفر شود و یا به صفر نزدیک شود. در عمل، به دلیل تخمینی بودن این روابط، همواره ضریبی مثبت از این گام که کوچکتر و یا برابر واحد باشد، به صورت رابطه (۴-۲۴) استفاده می‌شود.

$$U(X + \Delta X) = U(X) + \nabla U(X)^T \, \Delta X + \, \Delta X^T \, \nabla^2 U(X) \, \Delta X$$
$$\nabla U(X + \Delta X) = \nabla U(X) + \nabla^2 U(X) \, \Delta X$$
$$\nabla U(X + \Delta X) = 0 \quad \Rightarrow \quad \Delta X = -\nabla^2 U(X)^{-1} \, \nabla U(X)$$

(۴-۲۲)

$$X_{k+1} = X_k - \nabla^2 U(X_k)^{-1} \, \nabla U(k)$$

(۴-۲۳)

$$X_{k+1} = X_k - \alpha_k \, \nabla^2 U(X_k)^{-1} \, \nabla U(X_k) \; , \; 0 < \alpha_k \leq 1$$

(۴-۲۴)

این روش نسبت به روش قبلی دارای این مزیت است که گام‌هایی به نسبت بزرگتر را می‌تواند بپیماید زیرا مبتنی بر بسط درجه دوم تیلور است که تخمینی غیرخطی از توابع غیرخطی است. همچنین استفاده از روش جستجوی خطی که در ادامه شرح داده می‌شود، عملکرد روش نیوتن را تقویت می‌کند. مشخصات کلی این روش عیب در هزینه محاسباتی نسبتا بالا و مزیت در سرعت همگرایی بسیار بالا است که نرخ همگرایی مربعی آن قابل محاسبه و اثبات است و به همین جهت گزینه‌ای بسیار مناسب برای مقاصد کنترلی است. شکل ۴-۴، اجرای این الگوریتم بر روی یک تابع مربعی در فضای دوبعدی را نشان می‌دهد. همچنین استفاده از این روش با استفاده از این رابطه اولیه، در برخی مسائل منجر به بروز اشکالی در جهت حرکت می‌شود که به جای حرکت در جهت نزولی، منجر به حرکتی صعودی می‌شود. این عیب با استفاده از اصلاح‌کننده جهت که در ادامه شرح داده‌خواهد شد، قابل رفع است.

---

[1] Newton's Method in optimization

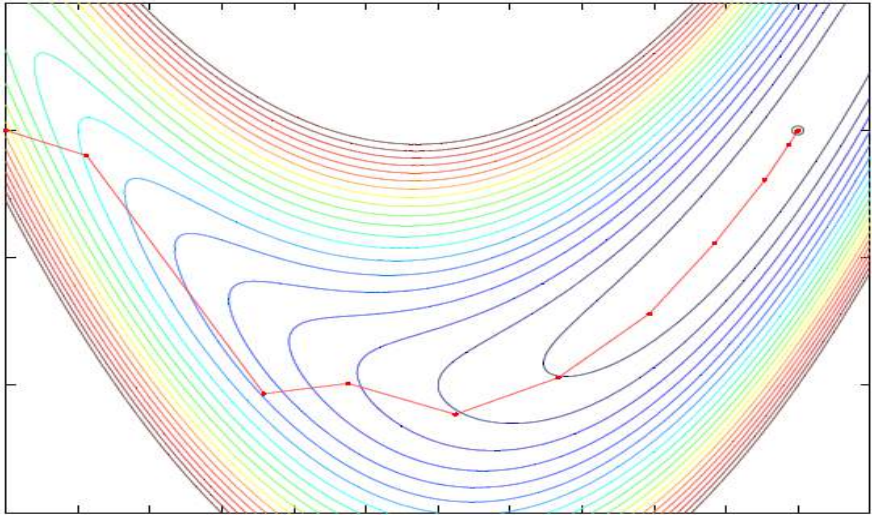

شکل ۴-۴. نمایش گرافیکی الگوریتم بهینه‌سازی غیرمقید نیوتن بر روی یک تابع مربعی نمونه در فضای دوبعدی

## ۴-۴-۳ جستجوی خطی[1] در جهت حرکت[2]

پس از یافتن جهت حرکت توسط هر یک از دو روشی که از آن ذکر شد، محاسبه طول گامی که می‌تواند مقدار تابع را در این جهت کمینه کند نیز می‌تواند به بالابردن سرعت همگرایی کمک کند. اگرچه روش نیوتن، خود طول گامی با ضریب واحد را توصیه می‌کند ولی به علت تخمینی بودن روابطی که این روش‌ها بر آن پایه‌ریزی شده‌اند، ترجیح می‌دهیم برای هر دو روش، جستجویی خطی در جهت حرکت برای یافتن بهترین طول گام انجام دهیم. اگر جهت حرکت، $P_k$، را بر هر یک از این روش‌ها به صورت روابط (۴–۲۵) درنظربگیریم، مسئله جستجوی خطی با تعریف تابع $\varphi(\alpha)$، به صورت رابطه(۴–۲۶) قابل بیان است. این مسئله در شکل ۴-۵ نمایش داده شده‌است.

$Steepest\ Descent\ search\ direction:\ \boldsymbol{P_k} = -\boldsymbol{\nabla U_k}$
$Newton's\ Method\ search\ direction:\ \boldsymbol{P_k} = -\boldsymbol{\nabla^2 U_k}^{-1}\boldsymbol{\nabla U_k}$   (۴–۲۵)
$\boldsymbol{X_{k+1}} = \boldsymbol{X_k} + \alpha_k \boldsymbol{P_k}$

$\varphi(\alpha) = U(\boldsymbol{X_k} + \alpha \boldsymbol{P_k}) \Rightarrow\ Line\ Search:\ \underset{\alpha}{Min}\ \varphi(\alpha)$   (۴–۲۶)

---

[1] Line Search
[2] Search Direction

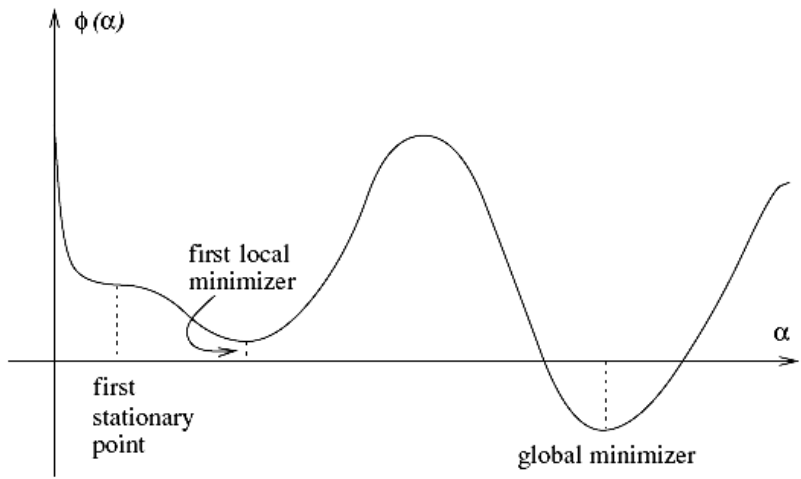

شکل ۴-۵. تابع جستجوی غیرخطی به عنوان تابعی از پارامتر ضریب طول گام

- **شرایط ولف[1]**

در مسئله جستجوی خطی، شرایط ولف، شرایط حداقلی را بیان می‌کند کـه در صـورت رعایـت آن، همگرایـی مسئله در روش بیشترین شیب نزولی و نرخ همگرایی مربعی در روش نیوتن تضمین خواهد شد. اگر تـابعی خطـی بـر حسب $\alpha$ به صورت رابطه $l(\alpha) = U(X_k) + c_1\alpha\, \nabla U_k^T P_k$ (کـه $c_1$ در آن، یـک ضـریب کوچـک و در عمل نزدیک به $10^{-4}$ است،) را درنظربگیریم، شرط اول از دو شرط ولف بیان مـی‌کنـد کـه بایـد طـول گـام را در بازه‌ای یافت که $\varphi(\alpha)$ در آن کوچکتر یا مساوی $l(\alpha)$ باشد تا کاهشـی کـافی در مقـدار تـابع اتفـاق بیافتـد. ایـن شرط به صورت نامعادله (۱) رابطه (۴-۲۷) قابل بیان است و عملکرد آن به صورت نمـادین در شـکل ۴-۶ نمـایش داده شده است. رعایت این شرط به تنهایی ممکن است منجر به برداشتن گام‌هایی کوتاه شود که سـرعت همگرایـی را کاهش می‌دهد و مطلوب نیست.

نامعادله (۲) در این رابطه، شرط دوم ولف یا شرط خمیدگی است که برداشتن طول گامی در بازه‌هایی که ایـن شرط را ارضا بکند، حرکت گرادیان تابع در جهت مثبت را تضمین خواهد کرد که در نتیجـه آن، از برداشـتن گـام‌های کوتاه جلوگیری خواهد شد. عملکرد نمادین شرط دوم در شکل ۴-۷ و عملکرد همزمان هردو شرط ولـف در شکل ۴-۸ نمایش داده شده‌است.

$$
\begin{array}{l} Wolf \\ Conditions \\ 0 < c_1 < c_2 < 1 \end{array} :
\begin{cases}
(1)\ sufficient\ decrease:\ U(X_k + \alpha_k P_k) \le\ U(X_k) + c_1\alpha_k \nabla U_k^T P_k \\
(2)\ curvature\ condition:\ \nabla U(X_k + \alpha_k P_k)^T P_k \ge c_2 \nabla U_k^T P_k
\end{cases}
\quad (\text{۴}-\text{۲۷})
$$

---

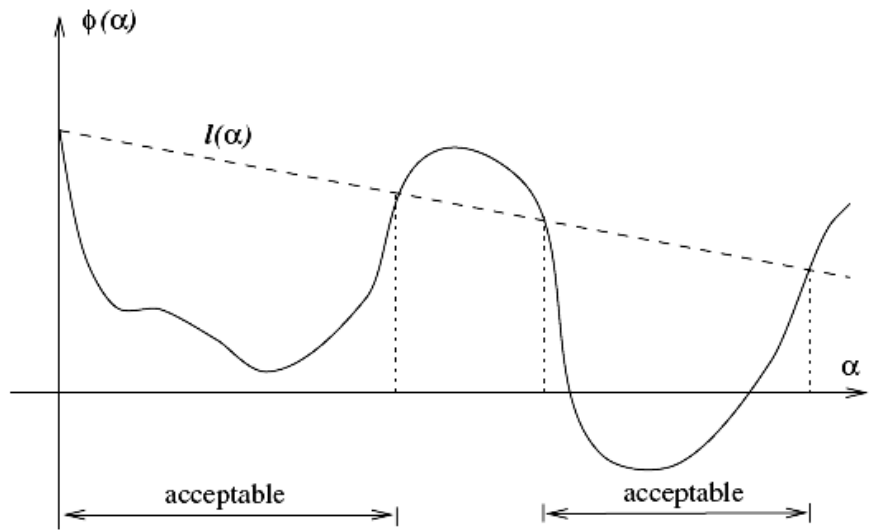

شکل ۴-۶. عملکرد نمادین شرط اول ولف (شرط کاهش کافی)

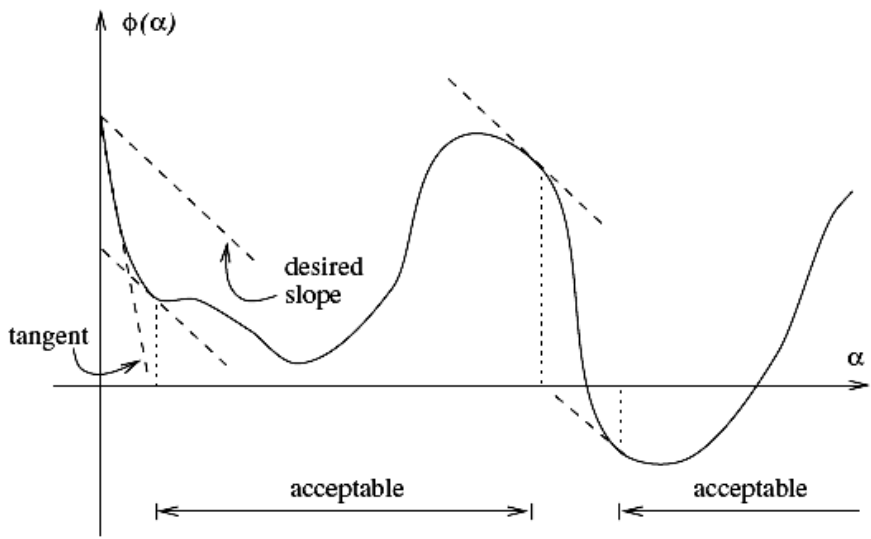

شکل ۴-۷. عملکرد نمادین شرط دوم ولف (شرط خمیدگی)

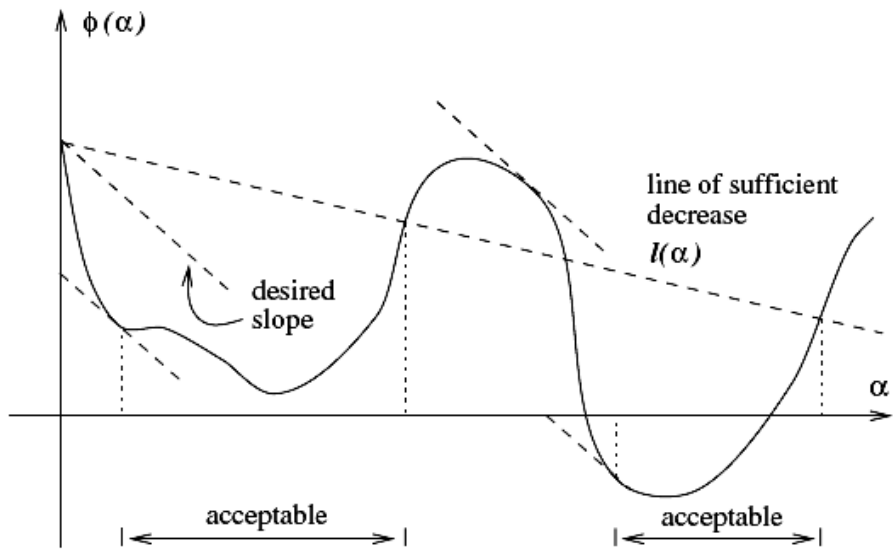

شکل ۴-۸ عملکرد نمادین همزمان شرط اول ودوم ولف (شرط کاهش کافی و شرط خمیدگی)

یک شرط جایگزین محدودکننده‌تر که می‌تواند سرعت همگرایی را افزایش بدهد، شرایط سخت‌گیرانه ولف در رابطه (۴-۲۸) است که تفاوت آن با شروط معمول ولف در بیانی سخت‌گیرانه‌تر از شرط دوم یا شرط خمیدگی است. برداشتن گام در بازه‌هایی که این شرط را ارضا بکند، نزدیک شدن گرادیان تابع به سمت صفر را تضمین می‌کند که همان خواسته اصلی در بهینه‌سازی است. با این‌حال، به علت کافی بودن شرایط معمول ولف برای تضمین همگرایی و همچنین به دلیل ترس از کم‌شدن احتمال یافتن جواب و یا بالابردن زمان محاسبات، این شروط جایگزین، کمتر مورد استفاده قرار می‌گیرند.

$$
\begin{matrix}
Strong \\
Wolf \\
Conditions \\
0 < c_1 < c_2 < 1
\end{matrix}
: \begin{cases}
(1)\ sufficient\ decrease: U(X_k + \alpha_k P_k) \leq\ U(X_k) + c_1 \alpha_k \nabla U_k^T P_k \\
(2)\ curvature\ condition: |\nabla U(X_k + \alpha_k P_k)^T P_k| \leq c_2 |\nabla U_k^T P_k|
\end{cases} \quad (۴-۲۸)
$$

– **الگوریتم جستجو به عقب[1] برای روش نیوتون**

استفاده از این الگوریتم جستجوی خطی که با شرط اول ولف کار می‌کند، در روش نیوتن می‌تواند ما را از بررسی شرط دوم بی‌نیاز کند. از آنجا که روش نیوتن خود گامی با ضریب واحد را پیشنهاد می‌کند، جستجو بر روی تابع به صورت برگشت از ضریب واحد به سمت صفر، همزمان هم کاهش تابع و هم بزرگ بودن طول گام را

---

[1] Backtracking Algorithm

تضمین می‌کند و در عمل شیوه‌ای بسیار مفید برای جستجوی خطی در روش نیوتن است. این الگوریتم ساده به صورت الگوریتم رابطه (۴—۲۹) بیان می‌شود.

***Backtracking Algorithm*** (*to oblige wolf conditions for Newton's Method*)

*Choose* $\alpha$ *in* $0 \ll \alpha \le 1$ *e.g.* $\alpha \leftarrow 1$
*Choose* $\rho$ *in* $0 \ll \rho < 1$ *e.g.* $\rho \leftarrow 0.9$
*Choose* $c_1$ *in* $0 < c_1 \ll 1$ *e.g.* $c_1 \leftarrow 10^{-4}$
***repeat*** *until* $U(X_k + \alpha P_k) \le U(X_k) + c_1 \alpha \, \nabla U_k^T P_k$ (۴—۲۹)
    $\alpha \leftarrow \rho \, \alpha$
***end***(*repeat*)
*terminate with* $\alpha_k = \alpha$

- **زاویه بین جهت حرکت با جهت حرکت دارای بیشترین شیب نزولی [1] برای روش نیوتون**

همانطور که در روش نیوتون مطرح شد، اشکال این روش این است که می‌تواند منجر به حرکت در جهت صعودی شود، زیرا این روش تمایلی یکسان برای حرکت به سمت نقاطی دارای گرادیان صفر ازجمله نقاط کمینه، بیشینه یا زین اسبی [2] دارد که البته تنها می‌تواند به سمت نقاط بیشینه یا کمینه همگرا شود. همواره با مقایسه زاویه جهت حرکت الگوریتم با جهت معکوس بردار گرادیان که همان جهت بیشترین شیب نزولی است، شرط حرکت نزولی را می‌توان به صورت رابطه (۴—۳۰) نوشت. اگر همواره زاویه مابین جهت حرکت با این جهت حاده بماند، آنگاه می‌توان اطمینان داشت که حرکت صعودی نخواهد بود.

*Optimization Method search direction*: $P_k$
*Steepest Descent search direction*: $-\nabla U_k$ (۴—۳۰)
$$\cos(\theta_k) = \frac{-\nabla U_k^T P_k}{\|\nabla U_k\| \, \|P_k\|} \quad , \quad -90 < \theta_k < 90 \quad or \quad \cos(\theta_k) > 0$$

به عنوان مثال برای روش نیوتون با توجه به رابطه (۴—۳۱)، اگر معکوس یا خود ماتریس هسیان [3] مثبت معین باشد، همواره حرکت نزولی نتیجه می‌شود، در غیر این صورت، نیاز است تا جهت حرکت این روش اصلاح شود.

---

[1] Steepest Descent Direction
[2] Saddle Point
[3] Hessian Matrix

$$Newton's\ Method\ search\ direction:\ \boldsymbol{P_k} = -\boldsymbol{\nabla}^2\boldsymbol{U_k}^{-1}\ \boldsymbol{\nabla U_k}$$

$$\boldsymbol{\nabla}^2\boldsymbol{U_k} > 0 \Rightarrow \cos(\theta_k) = \frac{-\boldsymbol{\nabla U_k}^T(-\boldsymbol{\nabla}^2\boldsymbol{U_k}^{-1}\ \boldsymbol{\nabla U_k})}{\|\boldsymbol{\nabla U_k}\|\ \|\boldsymbol{P_k}\|} > 0 \qquad (\text{۴}-\text{۳۱})$$

### ۴-۴-۴ روش اصلاح‌شده نیوتون[1] برای تضمین حرکت در جهت نزولی

اگر بخواهیم از نزولی بودن حرکت در روش نیوتن مطمئن شویم باید در صورت منفی‌شـدن مـاتریس هسیان، $\nabla^2 U(X_k)$، ماتریسی مثبت معین، $B_k$، که نزدیک به آن است را جایگزینش کنیم. یکـ از این روش‌هـا، روش اصلاح مقادیر ویژه است که توسط رابطه (۴–۳۲) بیان می‌شود. روش‌های جایگزین دیگری نیز بـرای ایـن اصلاح وجود دارد که در مسئله ما به علت کوچک بودن حجم ماتریس، نیازی به استفاده از آن نیست. همچنین بـه جـای اصلاح ماتریس هسیان، روش‌های دیگری با نام کلی روش‌های شبه‌نیـوتنی[2] ازجملـه $SR1$، $BFGS$ و $DFP$، نیـز وجود دارند که این ماتریس را با استفاده از یک دنباله ریاضی و با ماتریسی مثبت معین، تخمین می‌زنند.

$$P_i = -B_k^{-1}\ \nabla U_k$$
$$X_{k+1} = X_k + \alpha_k P_k$$
$$\nabla^2 U(X_k)_{\,n\times n} = Q\ \Lambda\ Q^T = \sum_{i=1}^{n} \lambda_i\, q_i q_i^{\,T} \qquad (\text{۴}-\text{۳۲})$$
$$\Rightarrow B_k = \sum_{i=1}^{n} \max(\lambda_i, \lambda_{min})\ q_i q_i^{\,T}, \quad \lambda_{min} > 0\ e.g.\ \lambda_{min} = 10^{-8}$$

### ۴-۵ شبیه‌سازی کنترل‌کننده پایداری بهینه بر روی مدل فیزیکی ربات راه‌رونده

#### ۴-۵-۱ بازنویسی توابع شاخص و توابع قید مسئله برای حل به شیوه تاوان

برای حل مسئله به شیوه تاوان، می‌توان توابع هدف و توابع قید مسئله را به همراه پارامترهای هریک به ترتیـب بـه صورت روابط (۴–۳۳) و (۴–۳۴) بازنویسی کرد.

$$g_1 = q_{+1,0} = q_0 e^{\omega T} - L\ ,\quad g_{1,opt} = q_c \qquad (\text{۴}-\text{۳۳})$$

---

$$g_2 = T \quad , \quad g_{2,opt} = T_c$$
$$g_3 = L \quad , \quad g_{3,opt} = L_c$$

$$h_1 = L^2 \quad , \quad h_{1,ext} = L_{max}{}^2 \quad , \quad dir_1 = -1$$

$$h_2 = L^2 \quad , \quad h_{2,ext} = L_{min}{}^2 \quad , \quad dir_2 = 1$$

$$h_3 = D_1(L,T)^2 = \left(\frac{p_0 e^{-\omega T} + q_0 e^{\omega T}}{2}\right)^2 \quad , \quad h_{3,ext} = \left(\frac{L_{max}}{2}\right)^2 \quad , \quad dir_3 = -1$$

$$h_4 = D_2(L,T)^2 = \left(L - \frac{p_0 e^{-\omega T} + q_0 e^{\omega T}}{2}\right)^2 \quad , \quad h_{4,ext} = \left(\frac{L_{max}}{2}\right)^2 \quad , \quad dir_4 = -1 \qquad \text{(۴-۳۴)}$$

$$h_5 = \bar{V}_{swing}(L,T)^2 = \left(\frac{L_{-1} + L - \dfrac{(p_0 e^{-\omega T} + q_0 e^{\omega T}) - (p_0 + q_0)}{2}}{T - T_0}\right)^2$$

$$, \quad h_{5,ext} = V_{max}{}^2 \quad , \quad dir_5 = -1$$

## ۴-۵-۲ نتایج شبیه‌سازی کنترل‌کننده پایداری بهینه بر روی مدل فیزیکی کامل راه‌رونده

با توجه به نارسایی پایدارسازهای حرکتی در ناتوانی از همگرا نمودن شرایط اولیه دور از سیکل و عـدم رعایـت کامل قیدهای مسئله برای شرایط مسئله محک، همچنین عدم رعایت قیود برای مسئله هدایت حرکت از شرایط اولیـه پس‌رونده دور از سیکل به سمت یک سیکل حرکتی پیش‌رونده، انتظار داریم کنترل پایداری بهینه بتوانـد بـر ایـن نارسایی‌ها فائق آید. به همین جهت شبیه‌سازی این کنترل‌کننده را یک بار برای شرایط مسئله محک و یک بار بـرای سوق‌دادن حرکت پس‌رونده به سمت سیکل پیش‌رونده انجام می‌دهیم و نتایج به دست‌آمده را بررسی خواهیم کرد.

● **شبیه سازی کنترل‌کننده پایداری بهینه بر روی مدل کامل با شرایط مسئله محک**

نتایج به صورت مسیر حرکت مؤلفه واگرا به همگرا به صفحه فاز شکل ۴-۹ نمایش داده شده است. برای بخش پیوسته معادلات حرکت، نمودارهای شکل ۴-۱۰، متغیرهـای سیسـتم از جملـه سـرعت و شـتاب حرکـت در راستای افقی( $dx/dt$ و $d^2x/dt^2$ )، موقعیت عمودی( $y$ ) و زاویه بالاتنه( $\theta$ )را به همراه متغیرهـای کنترلـی نیروهای افقی و عمودی و گشتاور بالاتنه( $F_x^{control}$ ، $F_y^{control}$ و $T_z^{control}$ )، نمـایش مـی‌دهند. همچنـین ضریب اصطکاک مورد نیاز( $\mu_{required}$ )برای تامین نیروهای کنترلی این حرکت، به صورت نمـوداری بـر حسـب زمان مشخص شده‌است. برای بخش گسسته معادلات حرکت، نمودارهای شکل ۴-۱۱، متغیرهـای کنترلـی از جملـه طول و زمان فرود هرگام( $L_i$ و $T_i$ )با کـران‌هـای بـالا و پـایین هریـک( $L_{max}$ و $T_{min}$ )را بـه همـراه متغیرهـای

سیستم از جمله شرایط اولیه مولفه‌های واگـرا($q_{i+1,0}$) و همگـرا ($p_{i+1,0}$) بـا مقـدار شـاخص آن($p_i^*$) و نیـز سرعت متوسط حرکت نسبی پای متحرک نسبت به بالاتنه($V_{i,swing}$) با کران بالای آن($V_{max}$) و سرعت متوسط مرکز جرم در طول هـر گـام($V_{i,avg}$)، نمـایش داده شـده است. همچنین متغیرهـای قیـد $D_{i,1}$ و $D_{i,2}$ بـه همـراه حداکثر مقدار مجازشان ($L_{max}/2$) در این نمودارها مشخص شده است.

کنترل‌کننده پایداری بهینه، با شروع از شرایط اولیه مرزی نسبتا دور از سیکل پس از پنج گام بـه پایـداری کامـل می‌رسد و با وارد شدن ضربه درگام هفتم نیز، پس از هشت گام دوباره به پایداری کامل می‌رسد و در عین حال همه قیود مسئله را از جمله دو قید موقعیت بر روی $D_{i,1}$، $D_{i,2}$ و قید سرعت بر روی $V_{i,swing}$ در هر دو مرحله ابتـدای حرکت و پس از دریافت ضربه کاملا رعایت کرده است. روش کنترل‌کننده پایداری بهینـه، هـم توانایی بـالایی در کنترل شرایط اولیه نسبتا دور از سیکل دارد و همزمان می‌تواند ضربه‌ای با شدت ۱۲۰٪ ضربه معیـار را تحمـل کنـد و در عین حال تمامی قیود مسئله را برای شرایط اولیه مرزی و حداکثر ضربه قابل تحمل خـود بـه طـور کامـل رعایـت می‌کند. به نظر می‌رسدکه عملکرد این روش برای حرکت حول سیکل‌های پیش‌رونده با شرایط اولیه دور از سیکل و همچنین مقاوت آن در برابر ضربات شدید، بسیار مناسب باشد و گزینه ایده‌آلی بـرای کنترل پایداری راهروندههای واقعی به‌حساب می‌آید.

$$p_c = -0.7 \; , \; q_c = 0.2 \; , \; T_c = 0.4 \; , \; L_c = 0.5 \; , \; V_c = 1.25$$
$$p_{1,0} = -0.67 \; , \; q_{1,0} = 0.47 \; , \; x_0 = -0.1 \; , \; dx/dt_0 = 1.786$$
$$z_0 = -0.2 \; , \; z_1 = 0$$

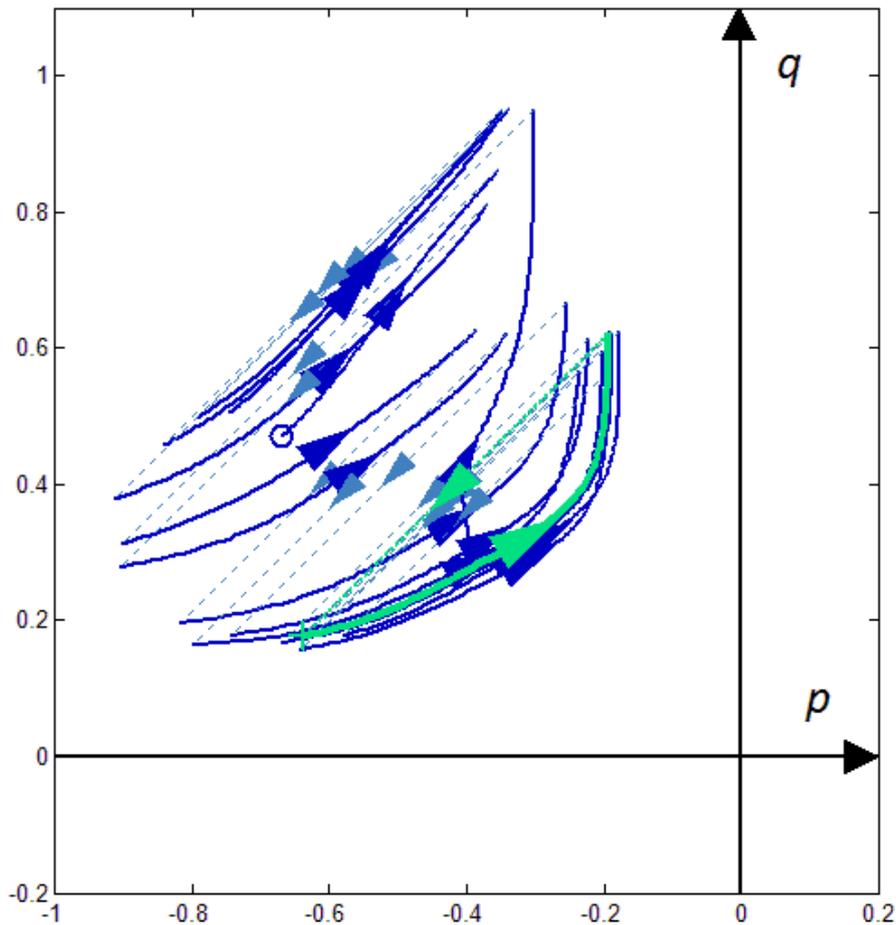

شکل ۴-۹. صفحه فاز مسیر حرکت مولفه واگرا نسبت به مولفه همگرا برای شبیه‌سازی کنترل‌کننده پایداری بهینه بر روی مدل کامل راه-
رونده با شرایط مسئله محکک (۱۲۰٪ ضربه) — منحنی ضخیم و کمرنگ مسیر گام آخر را نمایش می‌دهد.

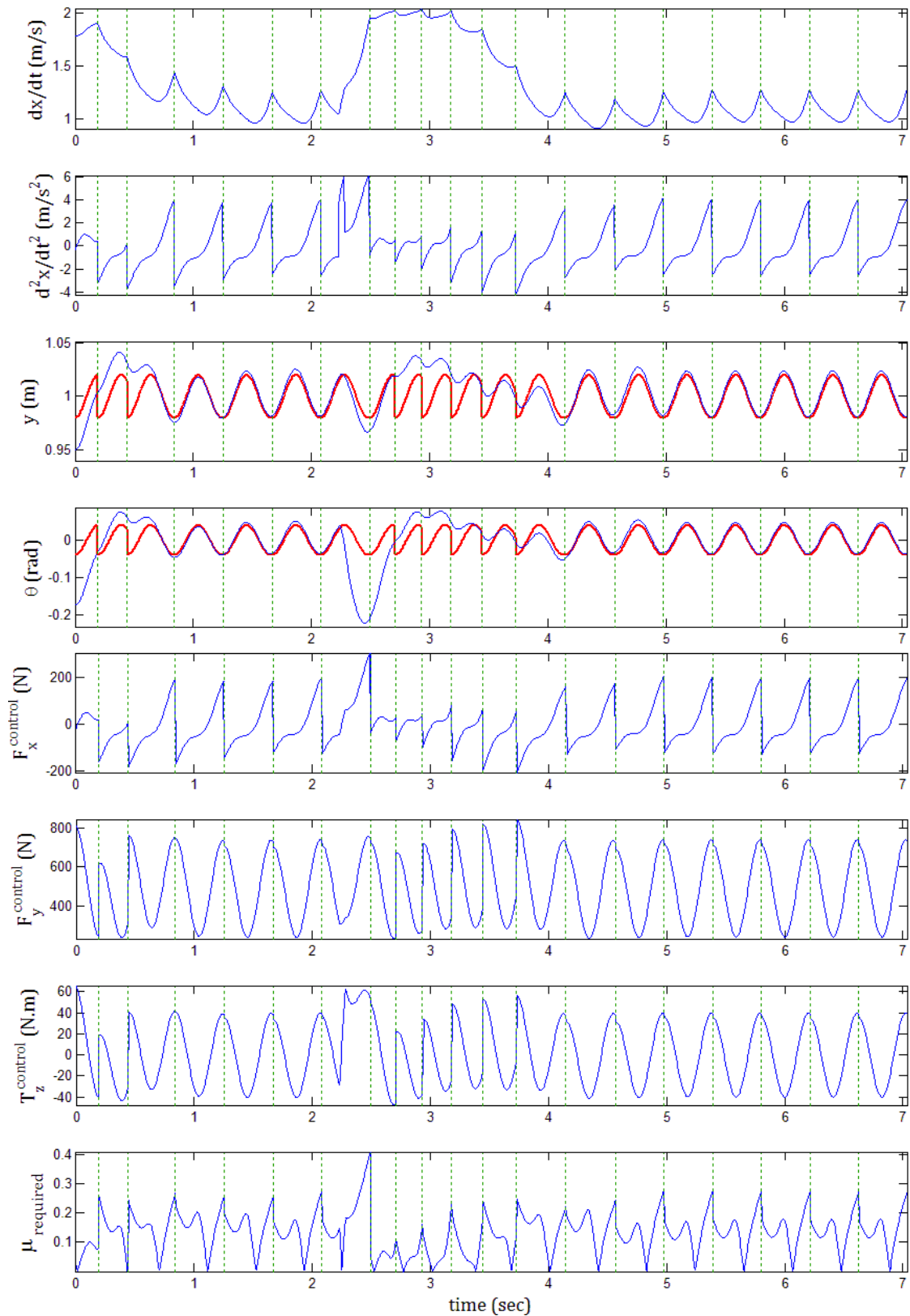

شکل ۴-۱۰. نمودارهای بخش پیوسته معادلات حرکت برای شبیه‌سازی کنترل‌کننده پایداری بهینه بر روی مدل کامل راهرونده با شرایط

مسئله محک (۱۲۰٪ ضربه) — خطوط ضخیم، مسیر مطلوب متغیر را نمایش می‌دهند.

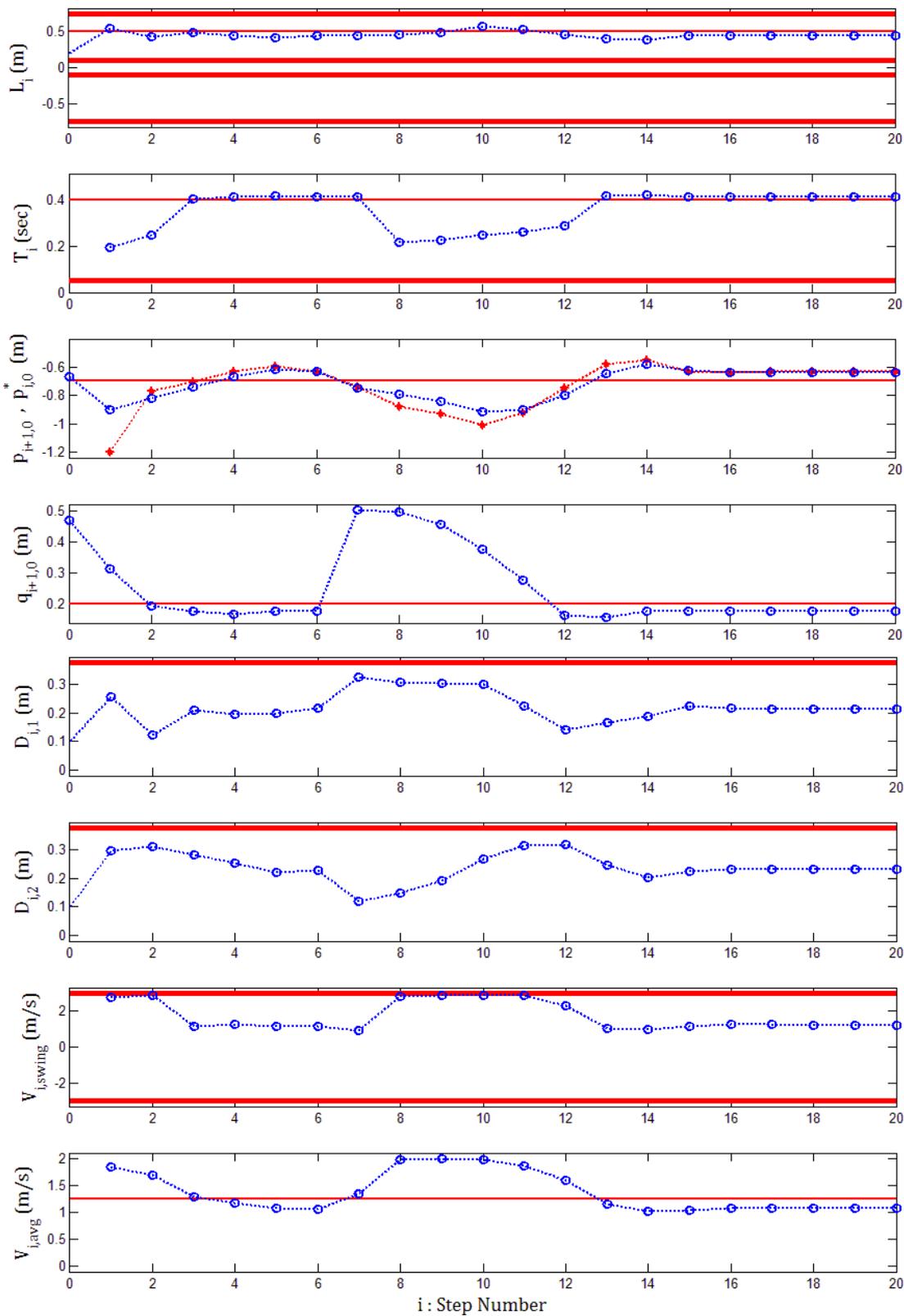

شکل ۴-۱۱. نمودارهای بخش گسسته معادلات حرکت برای شبیه‌سازی کنترل‌کننده پایداری بهینه بر روی مدل کامل راه‌رونده با شرایط مسئله محک (۱۲۰٪ ضربه) – خطوط با ضخامت متوسط و زیاد به ترتیب مقدار مطلوب متغیر در سیکل حرکتی و محدودیت بالا و یا پایین آن را نمایش می‌دهند.

• **شبیه‌سازی کنترل‌کننده پایداری بهینه بر روی مدل کامل با شرایط اولیه پس‌رونده دور از سیکل**

نتایج به صورت مسیر حرکت مؤلفه واگرا به همگرا در صفحه فاز شکل ۴-۹ نمایش داده شده است. برای بخش پیوسته معادلات حرکت، نمودارهای شکل ۴-۱۰، متغیرهای سیستم از جمله سرعت و شتاب حرکت در راستای افقی( $dx/dt$ و $d^2x/dt^2$ )، موقعیت عمودی( $y$ ) و زاویه بالاتنه( $\theta$ ) را به همراه متغیرهای کنترلی نیروهای افقی و عمودی و گشتاور بالاتنه( $T_z^{control}$ و $F_x^{control}$ ، $F_x^{control}$ )، نمایش می‌دهند. همچنین ضریب اصطکاک مورد نیاز( $\mu_{required}$ ) برای تأمین نیروهای کنترلی این حرکت، به صورت نموداری بر حسب زمان مشخص شده‌است. برای بخش گسسته معادلات حرکت، نمودارهای شکل ۴-۱۱، متغیرهای کنترلی از جمله طول و زمان فرود هر گام( $L_i$ و $T_i$ ) با کران‌های بالا و پایین هریک( $T_{min}$ و $L_{max}$ ) را به همراه متغیرهای سیستم از جمله شرایط اولیه مؤلفه‌های واگرا( $q_{i+1,0}$ ) و همگرا ( $p_{i+1,0}$ ) با مقدار شاخص آن( $p_i^*$ ) و نیز سرعت متوسط حرکت نسبی پای متحرک نسبت به بالاتنه( $V_{i,swing}$ ) با کران بالای آن( $V_{max}$ ) و سرعت متوسط مرکز جرم در طول هر گام( $V_{i,avg}$ )، نمایش داده شده است. همچنین متغیرهای قید $D_{i,1}$ و $D_{i,2}$ به همراه حداکثر مقدار مجازشان ( $L_{max}/2$ ) در این نمودارها مشخص شده است.

کنترل‌کننده پایداری بهینه، با شروع از شرایط اولیه پس‌رونده دور از سیکل، پس از نه گام به پایداری کامل حول سیکل مطلوب پیش‌رونده می‌رسد و در عین حال همه قیود مسئله را از جمله دو قید موقعیت بر روی $D_{i,1}$ ، $D_{i,2}$ و قید سرعت بر روی $V_{i,swing}$ به طور کامل رعایت کرده است. به نظر می‌رسد که عملکرد این روش برای هدایت حرکت از شرایط اولیه پس‌رونده به سمت یک سیکل پیش‌رونده و کنترل پایداری حول آن بسیار مناسب باشد و گزینه ایده‌آلی برای هدایت حرکت و کنترل پایداری راه‌روندهای واقعی حتی برای شرایط اولیه ناگواری که دور از سیکل حرکتی مطلوب قرار دارند، به حساب می‌آید.

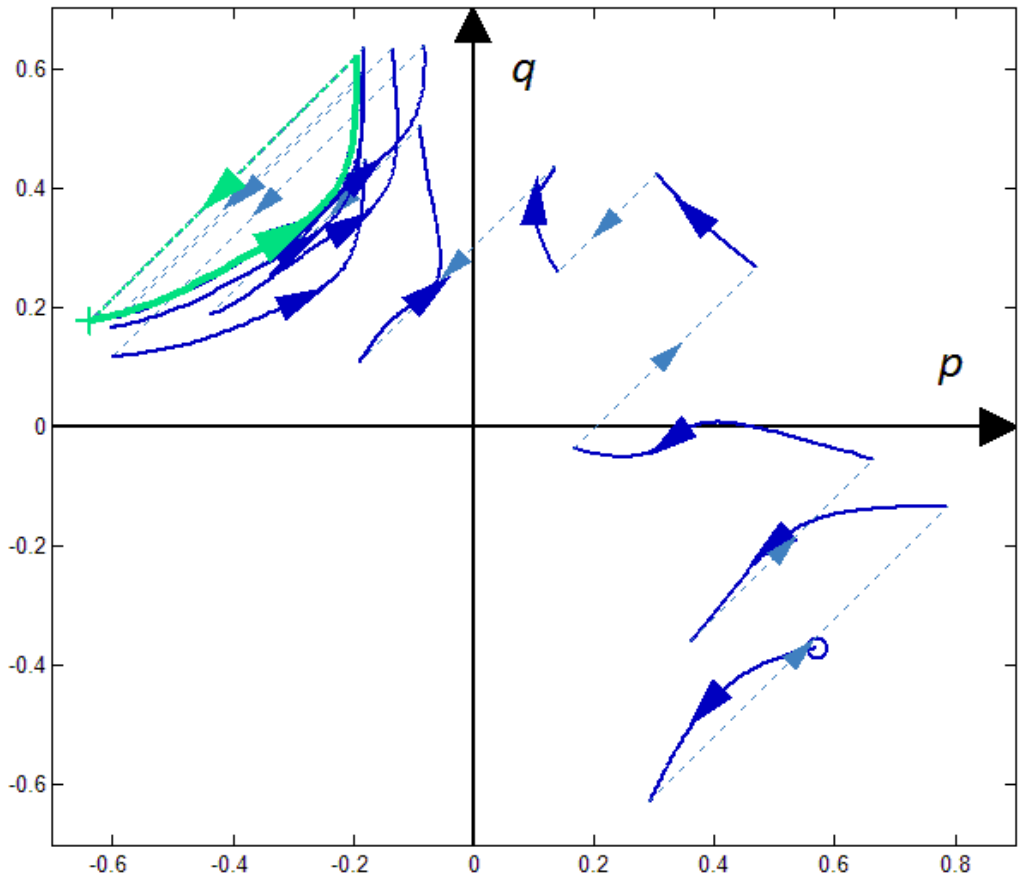

$p_c = -0.7$ , $q_c = 0.2$ , $T_c = 0.4$ , $L_c = 0.5$ , $V_c = 1.25$

$p_{1,0} = 0.57$ , $q_{1,0} = -0.37$ , $x_0 = 0.1$ , $dx/dt_0 = -1.473$

$z_0 = 0.2$ , $z_1 = 0$

شکل ۴-۱۲. صفحه فاز مسیر حرکت مولفه واگرا نسبت به مولفه همگرا برای شبیه‌سازی کنترل‌کننده پایداری بهینه بر روی مدل کامل راه‌رونده با شرایط اولیه دور از سیکل(پس‌رونده) — منحنی ضخیم و کم‌رنگ مسیر گام آخر را نمایش می‌دهد.

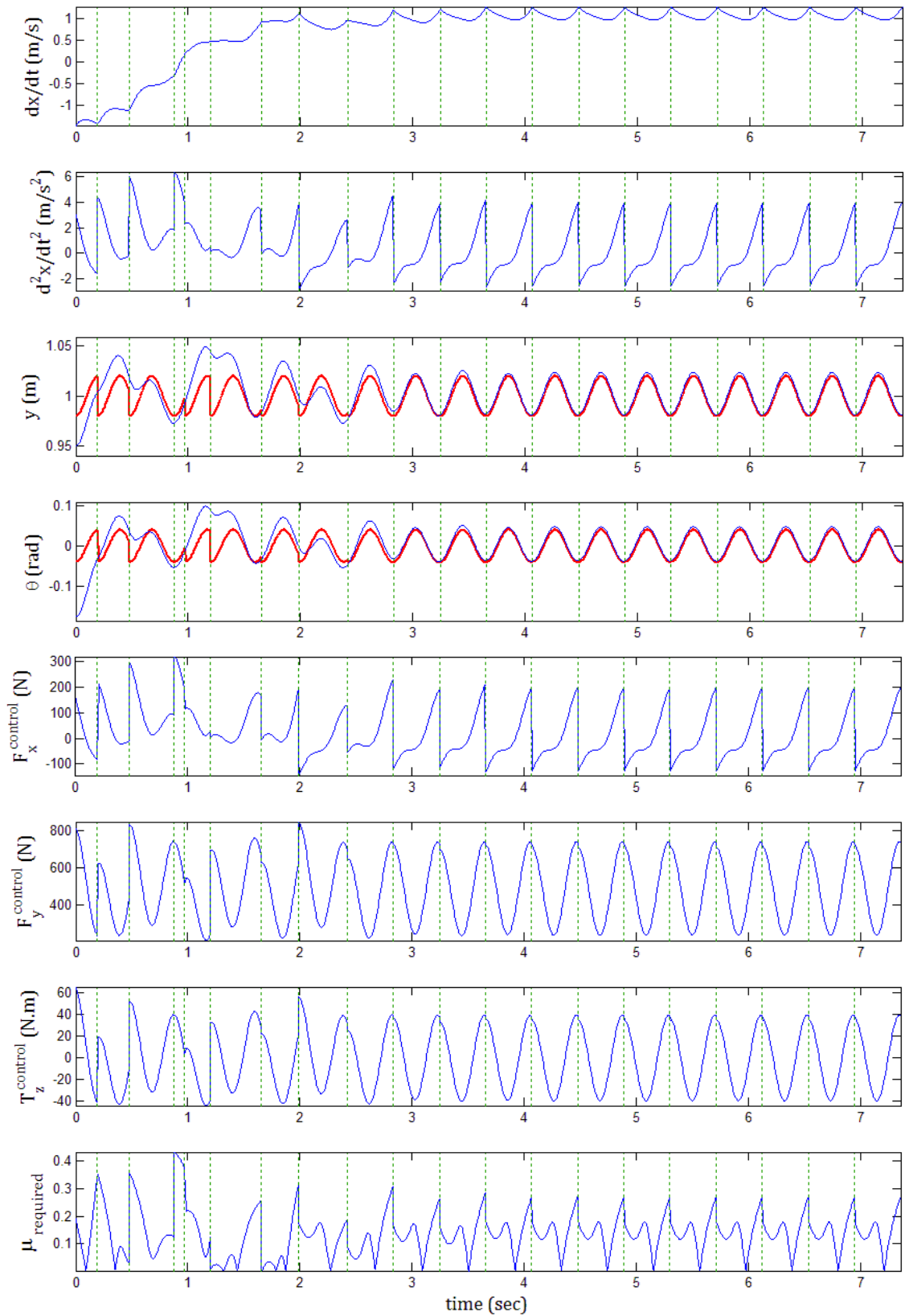

شکل ۴-۱۳. نمودارهای بخش پیوسته معادلات حرکت برای شبیه‌سازی کنترل‌کننده پایداری بهینه بر روی مدل کامل راه‌رونده با شرایط اولیه دور از سیکل(پس‌رونده) − خطوط ضخیم، مسیر مطلوب متغیر را نمایش می‌دهند.

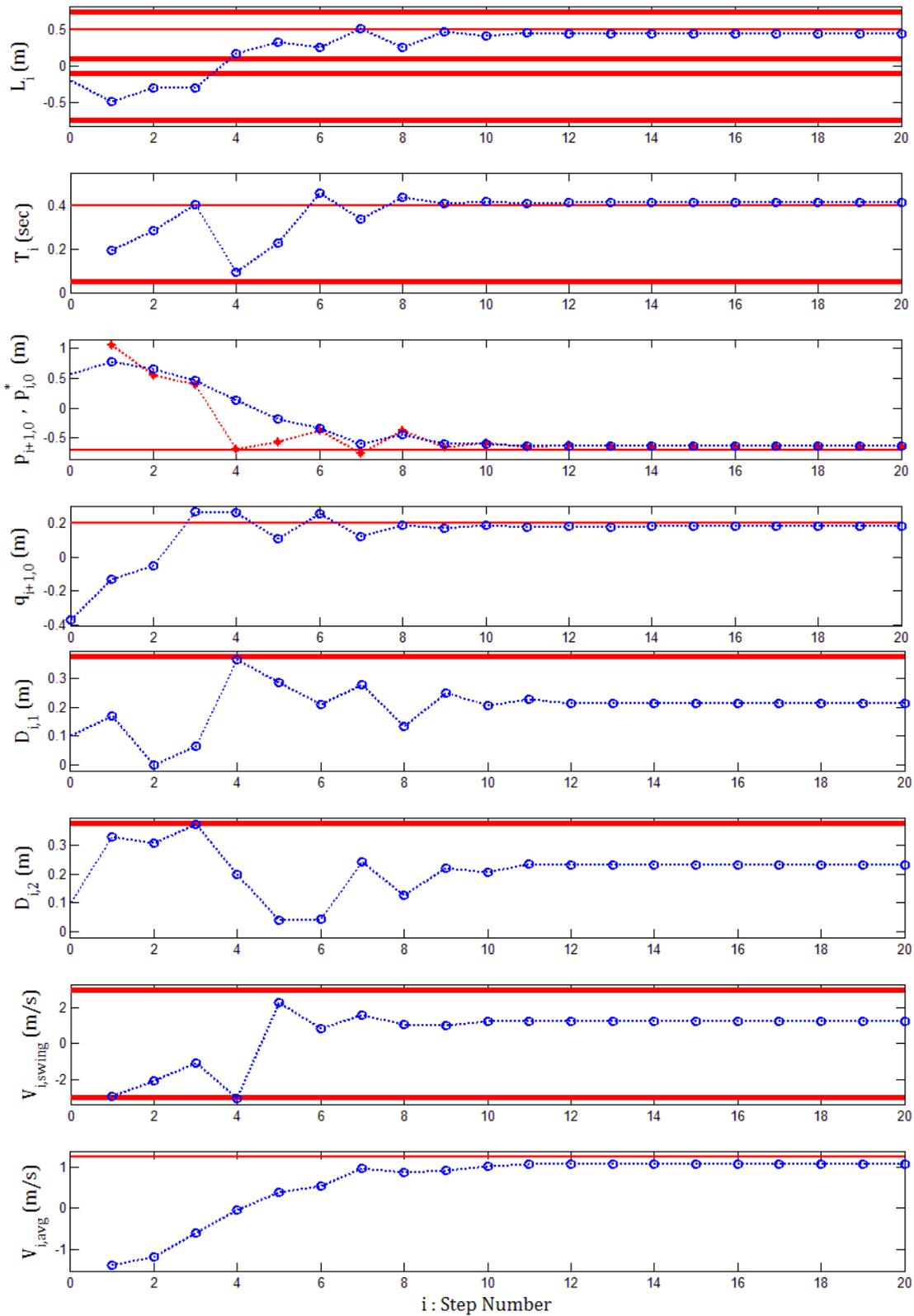

شکل ۴-۱۴. نمودارهای بخش گسسته معادلات حرکت برای شبیه‌سازی کنترل‌کننده پایداری بهینه بر روی مدل کامل راه‌رونده با شرایط اولیه دور از سیکل(پس‌رونده) – خطوط با ضخامت متوسط و زیاد، به ترتیب مقدار مطلوب متغیر در سیکل حرکتی و محدودیت بالا و/یا پایین آن را نمایش می‌دهند.

**فصل پنجم**
**جمع بندی و پیشنهادات**

در این فصل، ابتدا به جمع‌بندی از مفاهیم پایداری، کنترل پایداری، مباحث تئوری دینامیک حرکـت راه‌رفتن و نتایج شبیه‌سازی خواهیم پرداخت، و سپس نـوآوری‌هـای پژوهش حاضـر را برخـواهیم شـمرد. همچنـین، در پایان پیشنهاداتی برای ادامه کار ارائه خواهیم داد.

## ۵-۱ جمع‌بندی و بررسی نتایج

این پژوهش با بررسی مفهوم پایداری حرکت راه‌روندها در فصل اول آغاز شد و در پایان توانستیم بـه یـک راه حل جامع برای کنترل پایداری راه‌روندها دست بیابیم.

راه‌حل جدید، تفاوتی بنیادین با شیوه‌های متداول کنترل پایداری دارد که پایداری را معـادل بـا پایـداری وضعـی حرکت می‌دانند. به جای کنترل کمیت‌های جزئی حرکت همچون زوایای مفاصل راه‌رونده بـر روی یـک مسیر ازپیش طراحی شده دارای پایداری وضعی، راه حل جدید، پایداری کلی حرکت را با پایدارسازی برخی و کنترل برخی دیگر از کمیت‌هایی کلی از جمله اندازه حرکت‌های خطی و زاویه‌ای کل بدن به صورت بـه‌هنگـام تضمین می‌کند.

برای شروع کار، ابتدا در فصل دوم با نوشتن معادلات حرکت نیوتنی در فضای اندازه حرکت کل بـدن و اعمـال شرط برقراری تماس، معادله پایه حرکت راه رفتن استخراج شـد. سپس، بـا انجـام فرضـیاتی نزدیـک بـه واقعیت معادلات حرکت ساده‌سازی شد و برای دو بخش پیوسته حرکت در بازه زمانی یک گام و بخش گسسته حرکت در لحظه قبل و بعد از فرود گام نوشته شد تا معادله حرکت گام به‌گام به دست آید. در ادامه این فصل، بـا اعمـال شـرط تکرار شرایط اولیه بر روی معادله گام به‌گام، سیکل‌های حرکتـی سـاده و مرکـب استخراج شـد کـه ایـن سیکل‌هـا حرکت دائمی یک راه‌رونده را تعریف می‌کنند. با بررسی پایداری این سیکل‌ها مشخص شد که پایداری آنها امرزی است و به دلیل وجود کوچکترین اختلالی در عمل، به صورت نمایی و با نرخ رشد بالایی ناپایدار هستند و به همین جهت نیاز، ناگزیر از پایدارسازی آنها برای ادامه حرکت راه‌رونده خواهیم بود. سپس، معادلات کامـل حرکـت راه-رونده بدون در نظرگرفتن فرضیات ساده کننده، استخراج شد و اختلاف مسیر حرکت آن نسبت به مدل ساده‌شده بـا شبیه‌سازی برای بدترین شرایط ممکن بررسی گردید. در پایان فصل دوم، با جمع‌بندی معادلات حرکت راه‌رونـده، دو مدل ریاضی به نام‌های مدل ساده‌شده راه‌رفتن(SWM) و مدل کامل راه‌رفتن(CWM) بـه صـورتی دقیـق وجـامع معرفی گردید.

بـا توجـه بـه تعریـف سیکل حرکتی سـاده در فصـل دوم، ایـن سیکل الگـویی سـاده‌شـده بـرای حرکـت راه-رفتن(معمولی) دائمی است. بنابراین توانایی همگرایی حرکت به سمت یک سیکل حرکتی و یافتن الگوریتمی بـرای این کار به ترتیب معیاری برای پایداری و روشی برای کنترل پایداری حرکت خواهند بـود. بـا بررسـی مـدل کامـل راه‌رفتن(CWM)، دانستیم که کنترل پایداری راه‌رونده(پایدارسازی مولفه‌های همگرا و واگرای حرکت حول سیکل حرکتی مطلوب)، ترجیحا باید از طریق دو راهبرد کلی تغییر پیوسته موقعیت مرکز فشار در طول حرکت و یـا تغییـر گام(تغییر طول و زمان فرود گام) انجام پذیرد و به همین جهت و همچنین با فرض قابل صرف‌نظر دانستن اختلاف دو مدل ساده‌شده و کامل، مدل ساده شده راه رفتن(SWM)، مبنایی برای طراحی کنترل پایداری قرارگرفت.

در همان ابتدای فصل سوم با مطرح کردن قضیه محدود ماندن شرایط اولیه مولفه همگـرا در صـورت برداشـتن گام‌هایی پی‌درپی با طول و زمان فرود دلخواه، از پایداری دینامیک داخلی راه‌رونـده اطمینـان خـاطر پیـدا کـردیم و توجه خود را تنها به پایدارسازی مولفه واگرا برای سوق دادن آن به سمت یـک سـیکل حرکتـی مطلـوب، معطـوف ساختیم که نتیجه آن طراحی و اثبات پایداری چهار پایدارساز سیکل حرکتی در فصـل سـوم و ارائـه روش کنتـرل پایداری بهینه در فصل چهارم شد.

در مجموع، پنج کنترل‌کننده پایداری در فصل‌های سوم و چهارم پیشـنهاد شـد و عملکـرد آن‌هـا، هـم از لحـاظ تئوری و هم از لحاظ عملی با استفاده از شبیه‌سازی بر روی یک مدل فیزیکی کامل راه‌رونده، بررسـی شـد. بـه نظر می‌آید کامل‌ترین کنترل‌کننده پایداری در میان چهار پایدارساز سیکل حرکتی و روش کنترل پایداری بهینـه، گزینـه آخر، کنترل پایداری بهینه، باشد که توانایی رعایت همه قیود مسئله و ایجاد حداکثر پایداری ممکن در برابر شـرایط اولیه و ضربه خارجی را دارا می‌باشد. با وجود اینکه برای تضمین پایداری حداکثری، بهتـرین گزینـه، روش کنتـرل-کننده پایداری بهینه خواهد بود، اما، چهار پایدارساز سیکل حرکتی نیز هر کدام با توجه بـه خصوصـیات ویـژه‌شـان درجای خود می‌توانند برای راه‌رونده هایی که می‌خواهند و یا مجبور هستند کـه بـه یـک پایـداری حـداقلی بسـنده

کنند، مورد استفاده قرار بگیرند. به عنوان مثال برای راه‌روندهای که می‌خواهد همواره گام‌هایی با طـول ثابـت داشـته باشد پایدارساز اول و سوم سیکل حرکتی، گزینه‌های خوبی خواهند بود.

به نظر می‌رسد، نگاه جدید به مقوله کنترل پایداری راه‌روندهها نگاهی درست و کامل‌تر نسبت بـه کنتـرل‌کننـده‌-های متداول پایداری باشدچراکه می‌تواند فضای بزرگتری از متغیرهای سیستم را نسبت به کنتـرل‌کننـدههـای متـداول پایداری به عنوان ناحیه کنترل‌پذیر خود پوشش دهد. این شیوه جدید کنترل‌کننده پایداری براسـاس پایـداری کلـی حرکت برای رسیدن به یک سیکل حرکتی کار می‌کند و دیگر توجهی به معیارهای محدودکننده پایـداری وضـعی ندارد. اگرچه توانایی این روش کنترل پایداری تنها با شبیه‌سازی آن بر روی یک مـدل فیزیکـی سـاده بررسـی شـد، ولی کلیت مسئله برای مدلهای فیزیکی دیگر یکسان است و برای هر یک از آن‌ها نیـز مـی‌تـوان راه‌حلـی مشـابه بـا فرمول‌بندی یکسانی در بخش کنترل پایداری و فرمول‌بندی جدیدی در بخش کنترل اندازه حرکت یافت.

## ۵-۲ نوآوری‌ها و دستاوردها

در این پژوهش، دو مدل دینامیکی برای حرکت راه‌رونده دوپا برای اولین‌بار معرفی شد که یکی توصیفی کامـل از حرکت راه‌رفتن و دیگری توصیفی سادهشده از حرکت راه‌رفتن ارائه می‌دهد. بر مبنای مدل سادهشده، مفهومی به نام سیکل‌های حرکتی که حرکت‌های طبیعی و تکرارشونده راه‌رونده دوپا را نمایندگی می‌کند، بسط ریاضـی داده-شد و انواع سیکل‌های حرکتی ساده و مرکب را برای اولین بار برای حرکت راه‌رفتن دوپا معرفی گردید. همچنین بـا ارائه تعبیری جدید از پایداری را معادل توانایی هدایت حرکت راه‌رونده به سمت یـک سـیکل حرکتـی مطلوب و کنترل حول آن می‌داند، کنترل‌کننده‌هایی برای پایـداری بـه نـام پایدارسـازهـای سـیکل حرکتـی طراحـی شدکه توانستند از لحاظ تئوری مدل سادهشده را پایدار می‌سازند. همچنین در شبیه‌سازی‌ها، نشان داده شـد کـه ایـن پایدارسازهای سیکل حرکتی می‌توانند مدل کامل را نیز به خوبی پایدار سازند. در پایان، علاوه بر این پایدارسـازها و با درنظرگرفتن عملکرد آنها در پایدارسازی، نوعی کنترل‌کننده پایداری بهینه با استفاده از برنامه‌ریزی غیرخطی ارائه شد که می‌تواند مدل کامل را با رعایت تمامی قیود واقعی مسئله به خوبی پایدار سـازد. نـوآوری‌هـای ایـن تحقیـق را می‌توان به صورت خلاصه در قالب موارد زیر بیان نمود:

- ارائه دو مدل دینامیکی ساده و کامل برای حرکت راه‌رفتن

- ارائه الگویی از حرکت طبیعی و تکرارشونده از زاه‌رفتن به نام سیکل حرکتی(ساده و مرکب)

- معرفی پایدارسازهای سیکل حرکتی

- معرفی کنترل‌کننده پایداری بهینه برپایه برنامه‌ریزی غیرخطی

## ۵-۳  پیشنهادات

برای ادامه و گسترش این تحقیق پیشنهادات زیر ارائه می‌شود:

–  **استفاده از روش پایداری بهینه برای بیش از یک گام** : بهتر است به جای بهینه‌سازی شاخص پایداری پس از یک گام، شاخص پایداری برای یک افق چندگامی پس از دو، سه یا چهار گام و با استفاده از متغیرهای طول و زمان فرود این گام‌ها برنامه‌ریزی شود به نحوی که قیود در همه این گام‌ها رعایت گردد. برای این کار، با توجه به افزایش حجم متغیرهای مسئله، روش‌های برنامه‌ریزی خطی دنباله‌وار مانند برنامه‌ریزی مربعی دنباله‌وار[1] پیشنهاد می‌شود که هزینه محاساتی را کاهش، سرعت همگرایی را افزایش و زمان محاسبات را کاهش می‌دهد.

–  **استخراج روابط نگاشت برای تبدیل نرخ اندازه حرکت کلی به شتاب مفاصل یک راه‌رونده چندعضوی دارای جرم گسترده(هفت عضوی)**: مشتق روابط اندازه حرکت و سرعت مفاصل در کار انجام شده توسط کاجیتا  می‌تواند مبنای این کار قرار گیرد[7].

–  **طراحی کنترل‌کننده پایداری در صفحه جانبی و کنترل همزمان پایداری در دو صفحه طولی و جانبی به ترتیب حول سیکل حرکتی ساده پیش‌رونده و سیکل درجای متقارن برای یک ربات سه‌بعدی**: بررسی محدودیت برهم‌نهی زمان فرود گام برای پایدارسازی دو سیکل مختلف در حرکت صفحه طولی و صفحه عرضی، می‌تواند راه‌حلی اولیه برای مسئله باشد.

–  **افزودن شرط حفظ نیروهای تکیه گاهی در مخروط اصطکاک برای جلوگیری از سُرخوردن راه‌رونده:** با توجه به این واقعیت که نیروی تکیه‌گاهی افقی به تقریب نسبتا خوب تابعی خطی از فاصله مرکزجرم تا مرکز فشار است، اعمال محدودیت بیشتر بر روی طول گام با کوتاه کردن طول گام‌ها در محیط‌های لغزنده می‌تواند راهکاری عملی برای تضمین سُر نخوردن راه رونده باشد، همچنین بالا بردن حد پایین برای شتاب عمودی مرکز جرم در بازه‌های زمانی ابتدا و انتهای هر گام که اندازه نیروی افقی بیشینه است، می‌ تواند از کاهش نیروی تکیه گاهی عمودی (که متناسب با مجموع شتاب عمودی و شتاب جاذبه زمین است) جلوگیری کند و نیروی تکیه گاهی را حتی در صورت افزایش نیروی تکیه‌گاهی افقی در مخروط اصطکاک نگه دارد.

–  **حرکت در سطوح شیب‌دار:** الگوریتم حاضر می‌تواند برای حرکت در سطوح شیب‌دار با افزودن شرط مخروط اصطکاک به قیود مسئله و افزایش متوسط ارتفاع در هرگام با نرخی ثابت به‌کارگرفته شود. برای حرکت در سطوح شیب‌دار مثبت و منفی نیازی به ایجاد شتاب عمودی دائمی مثبت و یا منفی وجود ندارد و تنها سرعت ثابت مثبت یا منفی مورد نیاز است و بنابراین تاثیر حرکت در شیب در معادلات حرکت و

---

[1] Sequential Quadratic Programming(SQP)

کنترل پایداری ظاهر نمی‌شود، با این وجود در صورت کاهش یا افزایش ناخواسته سرعت عمودی که در حرکت بر روی سطوح شیبدار محتمل است، ممکن است نیاز به اعمال کوتاه مدت یک شتاب عمودی ضروری($\ddot{y}_{necessary}$) پیدا کنیم که در ادامه برای این کار راهکاری پیشنهاد خواهد شد.

–  **حرکت بر روی پله‌ها:** الگوریتم حاضر می‌تواند برای حرکت بر روی پله نیز همچون حرکت در سطوح شیبدار استفاده شود با این تفاوت که حرکت تنها می‌تواند حول آن دسته از سیکل‌های حرکتی کنترل شود که طول گام آنها تقریبا برابر با طول افقی بین دو پله است. همچنین باید توجه داشت با توجه به محدودیت فضای پلکان، استفاده از راهبرد تغییر پارامتر زمان فرود گام به جای تغییر پارامتر طول گام برای این حالت مؤثرتر خواهد بود.

–  **طراحی کنترل کننده بر اساس مدل کامل راه‌رونده(CWM):** احتمالا می‌توان با استفاده از تئوری‌های کنترل‌کننده‌های چند ورودی-چند خروجی غیرخطی کنترل‌کننده‌ای طراحی کرد که بتواند مدل کامل راه‌رونده را در برابر ضربه مقاوم کند و یا حرکت آن را به سمت یک سیکل حرکتی هدایت کند. همچنین به دلیل دوبخشی بودن معادلات حرکت(بخش پیوسته و گسسته) می‌توان از کنترل‌کننده‌های تحریک آنی [1] برای کنترل سیستم دینامیکی تکه‌تکه پیوسته [2] راه‌رونده استفاده نمود.

–  **مطالعه پایدارسازی حرکت دویدن با افزودن بخش سوم به معادلات حرکت:** اگر بین دو بخش پیوسته و گسسته حرکت، بخش پیوسته سومی به معادله حرکت پایه راه‌رونده اضافه کنیم که در آن، نرخ تمامی اندازه‌حرکت‌های کلی خطی افقی و زاویه‌ای صفر است و نرخ اندازه حرکت عمودی برابر با گرانش زمین است(شرط عدم تماس با زمین)، احتمالا می‌توان معادله حرکتی گام به گام برای حرکت دویدن استخراج نمود و با تعریف سیکل‌های دویدن به پایدارسازی آن پرداخت.

–  **ارائه مدلی برای مسیر برداشتن گام به صورت به‌هنگام و ارائه کنترل‌پایداری بهینه به‌هنگام که می‌تواند طول و زمان فرود گام را در هر لحظه تغییر داده و آن را تعقیب کند:** این روش می‌تواند پایداری را افزایش دهد، زیرا انحراف حرکت از مدل ساده‌شده را در هر لحظه درنظر می‌گیرد.

–  **افزودن نرخ متوسط اندازه حرکت خطی عمودی ضروری($\ddot{y}_{necessary}$) و نرخ متوسط اندازه حرکت زاویه‌ای ضروری($\dot{H}_{necessary}$) به معادلات حرکت برای جلوگیری از افتادن یا واژگون شدن:** اگر دریافت ضربه در راستای عمودی و دورانی چنان زیاد باشد که کنترل‌کننده عادی نتواند حرکت را

---

[1] Impulsive Control
[2] Piecewise-Continuous/Discontinuous/Impulsive Dynamical Systems

پایدارسازی مجدد کند، بهتر است یک نرخ متوسط مورد نیاز برای بازگشت اندازه حرکت ربات به مقدار مطلوب خود محاسبه و اعمال گردد و اثر آن در کنترل‌کننده پایداری لحاظ شود.

–  **اصلاح و تقویت پایدارساز اول سیکل حرکتی بر اساس راهبرد تغییر پیوسته مرکز فشار:** می‌توان برای هدایت حرکت به سمت یک سیکل حرکتی مطلوب در دوردست، به جای تلاش برای هدایت بلافاصله حرکت به سمت آن که در عمل به وسیله این راهبرد غیرممکن است، حرکت را به سمت یک سیکل حرکتی میانی همگرا کرد و این سیکل حرکتی میانی را به آرامی به سمت سیکل حرکتی مطلوب حرکت داد.

–  **تلفیق همزمان دو راهبرد تغییر پیوسته متغیرجابجایی مرکز فشار،** $\Delta z_i$**، و راهبرد تغییر پی‌درپی طول و زمان گام،** $L_i$ **و** $T_i$ **، برای افزایش قدرت پایدارسازی:** این راهبرد تلفیقی به احتمال زیاد، به هم‌افزایی قدرت پایدارسازی دو راهبرد مجزا می‌انجامد.